\documentclass{article}

\usepackage[preprint]{neurips_2025}

\usepackage[utf8]{inputenc} % allow utf-8 input
\usepackage[T1]{fontenc}    % use 8-bit T1 fonts
\usepackage[colorlinks=true, linkcolor=blue, citecolor=blue, urlcolor=blue]{hyperref}   % hyperlinks
\usepackage{url}            % simple URL typesetting
\usepackage{booktabs}       % professional-quality tables
\usepackage{amsfonts}       % blackboard math symbols
\usepackage{amsmath}
\usepackage{amssymb}
\usepackage{mathtools}
\usepackage{amsthm}
\usepackage{nicefrac}       % compact symbols for 1/2, etc.
\usepackage{microtype}      % microtypography
\usepackage{xcolor}         % colors
\newcommand{\paren}[1]{\left( #1 \right)}

\newcommand{\R}[1]{\mathbb{R}^{#1}}

\newcommand{\gauss}[2]{\mathcal{N}\paren{#1, #2}}

\usepackage{graphicx}
\usepackage{wrapfig}
\usepackage{subcaption}
\usepackage{color}
\usepackage{multirow}
\usepackage[toc,page]{appendix}
\usepackage{placeins}

\usepackage[most]{tcolorbox}
\tcbset{
  myhypobox/.style={
    colback=gray!10,
    colframe=gray!50,
    boxrule=0.4pt,
    arc=2pt,
    left=4pt,
    right=4pt,
    top=2pt,
    bottom=2pt,
    boxsep=4pt,
    enhanced
  }
}

\title{Decomposing Prediction Mechanisms \\ for In-Context Recall}

\author{%
  Sultan Daniels$^{\dagger}$ \quad Dylan Davis$^{\dagger}$ \quad Dhruv Gautam$^{\dagger}$ \quad Wentinn Liao$^{\ddagger}$ \\
  \textbf{Gireeja Ranade}$^{\dagger}$ \quad \textbf{Anant Sahai}$^{\dagger}$ \\ 
  $^{\dagger}$University of California, Berkeley \quad $^{\ddagger}$University of Pennsylvania \\
  \texttt{\{sultan\_daniels, dylanjd, dhruvgautam, gireeja, asahai\}@berkeley.edu}\\
  \texttt{wenliao@seas.upenn.edu}  \\
}

\begin{document}

\maketitle

\begin{abstract}
We introduce a new family of toy problems that combine features of linear-regression-style continuous in-context learning (ICL) with discrete associative recall. 
We pretrain transformer models on sample traces from this toy, specifically symbolically-labeled interleaved state observations from randomly drawn linear deterministic dynamical systems. We study if the transformer models can recall the state of a sequence previously seen in its context when prompted to do so with the corresponding in-context label. 
Taking a closer look at this task, it becomes clear that the model must perform two functions: (1) identify which system's state should be recalled and apply that system to its last seen state, and (2) continuing to apply the correct system to predict the subsequent states.
Training dynamics reveal that the first capability emerges well into a model's training. Surprisingly, the second capability, of continuing the prediction of a resumed sequence, develops much earlier.

% But surprisingly, well before this recall ability has emerged, a closely related task -- predicting the second token in a resumed sequence given the first token -- shows clear evidence of seemingly recall-related behavior.  

Via out-of-distribution experiments, and a mechanistic analysis on model weights via edge pruning, we find that next-token prediction for this toy problem involves at least two separate mechanisms. One mechanism uses the discrete symbolic labels to do the associative recall required to predict the start of a resumption of a previously seen sequence. The second mechanism, which is largely agnostic to the discrete symbolic labels, performs a ``Bayesian-style'' prediction based on the previous token and the context. These two mechanisms have different learning dynamics. 

To confirm that this multi-mechanism (manifesting as separate phase transitions) phenomenon is not just an artifact of our toy setting, we used OLMo training checkpoints on an ICL translation task to see a similar phenomenon: a decisive gap in the emergence of first-task-token performance vs second-task-token performance.

\end{abstract}

\section{Introduction}

%Outline

%GPT-3 shows LLMs exhibit in context learning.
%There have been a series of toy problems studied to understand in context learning, and these had shed light on it

%In context learnign is viewed as an enabler of many emergenct abilities. One interesting phenomenon is emergence that is critical for many reasons.

%in context leanring has been studied in both discrete language like tasks, and continuous models, like garg et al. 

%In this work, we make a new toy problem that combines aspects from different threads --- there is a discrete task that makes it possible to study recall, where our data consists of labelled interleaved segments of systems, where we use sybolic discrete labels to identify which segments belong to the same underlying dynamics.  
%- from the continuous work we borrow linear regression, which allows the thing in context to be at the level of meaning (i.e. hidden A matrix)

%What does this let us do? 

The release of GPT-3~\cite{brown2020language} demonstrated the power of Large Language Models' (LLMs) ability to do in-context learning (ICL). 
Since then, there has been significant progress in understanding ICL for language models themselves~\cite{olsson2022context,akyurek2024context,xie2021explanation,lin2024dualoperatingmodesincontext,wei2023largerlanguagemodelsincontext,yin2025attentionheadsmatterincontext,wies2023learnabilityincontextlearning,pan2023incontextlearninglearnsincontext,min2022rethinkingroledemonstrationsmakes}. 
There has also been work that focuses on understanding ICL for simpler toy problems~\cite{garg2022can,rajaraman2024transformers,edelman2024evolution,singh2025strategycoopetitionexplainsemergence,raventos2024pretraining,DuOymakTrans,nichani2024understandingfactualrecalltransformers}. Toys (e.g., linear regression~\cite{garg2022can, raventos2024pretraining, huang2025task}) allow us to study the learned ICL behavior of deep neural networks in settings where optimal strategies are known, allowing complex prediction mechanisms to be disentangled. In this paper, we build on previous work to create a new toy problem involving interleaved vector-valued time-series. 
%\begin{itemize}
    %\item 
    
    We start with underlying time-series that come from the evolution of random deterministic linear systems and thus these time-series play the role of noise-free least-squares problems in \cite{garg2022can} --- each consecutive time-series observation is defined by its underlying deterministic linear system (defined by an unknown matrix, just as in linear regression). This continuous-state problem has a naturally continuous error metric: mean-squared-error. As in \cite{garg2022can}, ICL here implicitly involves identifying the underlying system from observations of its evolution.  
    %\item %The interleaving is done with discretely labeled symbolic ``punctuation\footnote{Perhaps these could also be thought of as a kind of `pronoun' in how they refer to something in context.}'' tokens that unambiguously demarcate different segments of the context as belonging to different time-series.
    
    Segments of random length from different time-series are interleaved with ``symbolic punctuation labels,%\footnote{Similar to how the correct pronoun in a sentence changes based on what the object of the pronoun is, the punctuation label in our setup changes based on the context.}
    '' tokens that unambiguously demarcate different segments as belonging to different time-series.
    These discrete symbolic labels and the fact that they can occur repeatedly introduces a dimension of recall similar to multi-query-associative-recall (MQAR)~\cite{arora2023zoology}. %Because %the underlying conceptual recall involves the details of particular time-series, 
    However, successful recall is not simply a matter of copying a particular surface-level value from the context. Instead, the corresponding task (predicting the next observation in this particular sequence) must be done.
%    \todo{expand MQAR} 
    %\end{itemize}

% that represents a more ccompound task --- the model must first recall which task it is being asked to perform, and then recall how to perform that task. This setup has two key features: (1) the task itself has a discrete nature, which makes it possible to study associative recall~\cite{arora2023zoology}, and (2) the natural metric for the problem is continuous (mean-squared error). Specifically, we create sequences that consist of interleaved segments (which we call payload tokens\todo{(?)}) from distinct finite-dimensional Gauss-Markov process/system. 

% We can think of the toy problem as involving ``observation sequences'' (i.e. the time-series) that are demarcated by symbolic punctuation labels. % payload 
The discrete symbolic nature (in the sense of \cite{wei2023symbol}) of these punctuation labels means that their meanings must be learned in context to be able to complete associative recall --- \textit{they cannot be memorized as the associations change for every new instance of the problem}.
Similarly, the details of each distinct time-series itself must also be learned in context. 
We restrict attention to noiseless time-series defined by orthogonal matrices --- consequently, once we have seen enough information in the observation sequence segments for a specific time-series, in principle, perfect prediction accuracy is possible.  

% Each payload segment in the interleaved sequence is indicated by discrete symbolic labels (one can think of these are parentheses) to identify which segments belong to the same underlying process. Our setup is such that symbolic labels must be learned in-context to be able to complete the associative recall task --- they cannot be memorized as they change for every new instance of the problem. We consider the noiseless orthogonal setting, such that once the transition matrix for system is identified, all future points can be predicted with zero error. 

% Thus, to successfully complete the task the model must: (1) identify which sequence a particular segment is drawn from and (2) perform inference to predict future values in that segment.

\subsection{Contributions}
\label{sec:contributions}

In this work, we combine the discrete spirit of associative recall with continuous Markov time-series prediction \citep{DuOymakTrans, li2023transformers} to create a new toy problem (interleaved labeled time-series prediction) with natural continuous performance metrics and LLM-style pretraining with next-output prediction. 

We find clear evidence of the emergence of associative recall during training in our toy problem. The compound nature of the problem allows us to see how different abilities emerge sequentially. Specifically, the ability to continue predictions for a resumed sequence after a state has been observed develops first, and the ability to identify sequences from just their corresponding symbolic label emerges later. 

Given this, we %use the toy problem to 
explore two natural hypotheses for mechanisms by which recall is actually performed:

% Specifically, we consider three hypotheses for task performance: 

\begin{tcolorbox}[myhypobox]
\textbf{H1: Label-based recall.} The model uses in-context learning of the association of symbolic labels to time-series, and then performs inference based on recalling the queried time-series and continuing its evolution.
\end{tcolorbox}
\begin{tcolorbox}[myhypobox]
\textbf{H2: Observation-based Bayesian recall.} The model ignores the symbolic labels. The noise-free nature of the toy problem means that once we see an observation, we can figure out which previously seen time-series it could have come from. Then, we can do Bayesian prediction \citep{xie2021explanation,mackay1992practical, müller2024transformersbayesianinference} based on previous observations. % payloads for future predictions
\end{tcolorbox}
% \begin{itemize}
%     \item H1: \textbf{Label-based recall:} The model uses in-context learning of the association of symbolic-labels to time series, and then performs inference based on recalling the queried time series and continuing its evolution.
%     \item H2: \textbf{Observation-based Bayesian recall:} The model ignores the symbolic-labels. The noise-free nature of the toy problem means that once we see a payload, we can figure out which time-series it could have come from. Then, we can do Bayesian prediction \citep{xie2021explanation,mackay1992practical, müller2024transformersbayesianinference} based on the payload tokens to predict the next tokens.
% %    \item H3: \textbf{Partial Label-based recall:} The symbolic label is only used to decide if the system has been seen before or not. If it is an unsee
% % \todo{ NOT SURE WHERE THIS FITS   use the symbolic info to do the first thing, but when you don't have to, you do the baysiean thing. }
% \end{itemize}

H1 corresponds to (what we think is) the natural human approach to the task: recall the relevant information and use it. 
However, {\bf we find that H1 and H2 are both false as complete explanations.} 
Instead, both are true simultaneously! H1 is used for predicting the first token after a particular time-series is being resumed.
To get this to be better than guessing, there is no choice other than using the information in the symbolic label. But for the second token and beyond in a resumed sequence, the information in the observation allows a variant of H2 to work. 

% and the model instead performs what we term as a ``lazy Bayesian" approach. H3 is also false but \todo{FILL IN HERE}. 

% ---  H1: Do recall based on discrete symbolic label --- we falsify this
% ---  H2: Ignore symbolic labels and focus only on payload tokens. Do Bayesian based-on token history. Infer sequence from local history.
%H2 alone is not correct, lazy Bayesian. 
% ---  H3: Paren is only used to decide if the system has been seen before or not. 

%All are false

% ---  for the first token, no way of succeeding in a bayseian approach, where we must use the 

% ---  clean original plots

We observe further that there is a difference between the emergence of the ability to learn to predict the first versus second token after a symbolic label, even though information-theoretically, they both require recalling the in-context-learned nature of that specific time-series. And somewhat surprisingly, the successful use of the symbolic label (which feels conceptually easier for a human) occurs {\em after} the model has clearly learned to do some approximate version of the more Bayesian H2 for those tokens on which H2 is a viable strategy. (We verify this using out-of-distribution experiments in Section \ref{sec:out_of_dist_experiments}.) 

A further edge-pruning based investigation shows that the circuits for predicting the first and second tokens are completely distinct. This leads to the following conjecture:

\begin{tcolorbox}[myhypobox]
\textbf{C3: Transformers use multiple mechanisms for a single multi-token task.} Distinct mechanisms are used to initiate a new episode of an ICL-specified task (i.e., predict the first token), versus continuing that task (i.e., predict the second token).
\end{tcolorbox}

% \begin{itemize}
%    \item C3: \textbf{Multi-mechanism for a single multi-token task:} Distinct mechanisms are used to initiate a new episode of an ICL-specified task (i.e. predict the first token), versus continuing that task (i.e. predict the second token.)
% \end{itemize}

Given this novel conjecture emerging from our toy problem, we seek to confirm this phenomenon in the natural language setting of LLMs. 
% We find that the observations from this toy problem suggest experiments in LLMs where we can replicate the findings of the toy problem. 
By modifying one of the classic LLM emergence experiments ~\cite{wei2023largerlanguagemodelsincontext,wei2023symbol} and using OLMo checkpoints \cite{olmo20242olmo2furious}, we confirm that this conjecture holds for an NLP task --- i.e. even before the emergence of successful initiation of an ICL-specified task, models can successfully continue performing that task. % Our focus here is on the ICL-learning associating the task to a specific 

\section{Setup}\label{sec:setup}
\newcommand{\xv}[0]{\mathbf{x}}
\newcommand{\yv}[0]{\mathbf{y}}
\newcommand{\wv}[0]{\mathbf{w}}
\newcommand{\vv}[0]{\mathbf{v}}
Consider predicting the continuous-state of an {\em unknown} linear dynamical system. We focus\footnote{In the Appendix, we provide results for the identity system family which has dynamics that are even simpler than the orthogonal system family.
For the identity systems, the initial state is $\xv_0 \sim \gauss{0}{\frac{1}{5}I} \in \R{5}$. Now, the state updates as 
\begin{align}
\xv_{i+1} = \xv_i.
\end{align}
This trivial process of copying a constant is perfectly predictable after one realization is observed.} on the orthogonally evolved system family \citep{pmlr-v235-sander24a}, where the system is defined by $U \in \R{5\times 5}$, a random orthogonal matrix. 
Each $U$ is generated by the algorithm presented in \cite{mezzadri2006generate}, which ensures a uniform sampling over all $\R{5\times 5}$ orthogonal matrices. The initial state is $\xv_0 \sim \gauss{0}{\frac{1}{5}I}$, with state updates:
\begin{align}
\xv_{i+1} &= U\xv_i = U^{i+1}\xv_0.
\end{align}
The system state is in-principle perfectly predictable, but only after six positions in the sequence are observed by solving for 
\begin{align}
U = \begin{bmatrix}\xv_1 & \xv_2 & \xv_3 & \xv_4 & \xv_5 \end{bmatrix}\begin{bmatrix}\xv_0 & \xv_1 & \xv_2 & \xv_3 & \xv_4 \end{bmatrix}^{-1}.\label{eqn:perfect_solve}
\end{align}

\subsection{Optimal Pseudoinverse Predictor}
Following from \eqref{eqn:perfect_solve}, given the state observations $\{\xv_0, \dots, \xv_i\}$, an optimal predictor for this problem computes $\widehat{\xv}_{i+1} = \widehat{U}\xv_i$, where 
\begin{align}
    \widehat{U} = \begin{bmatrix}\xv_1 & \dots & \xv_i \end{bmatrix}\begin{bmatrix}\xv_0 & \dots & \xv_{i-1} \end{bmatrix}^{\dagger},\label{eqn:pseudoinv_pred}
\end{align}
and $X^{\dagger}$ denotes the Moore-Penrose pseudoinverse of $X$. Essentially, this baseline only makes non-zero errors on the first, second, third, fourth, fifth, and sixth entry in any sequence --- it gets everything else perfectly correct.

% prediction error of the model by computing the average squared-error of a predictor that

% This perfect baseline also correctly unbraids the interleaved system so it knows exactly where it is in which sequence. 

\subsection{Data generation and training}
\label{sec:datagen}

\FloatBarrier
\begin{figure}[ptbh]
    \centering
    \includegraphics[width=\linewidth]{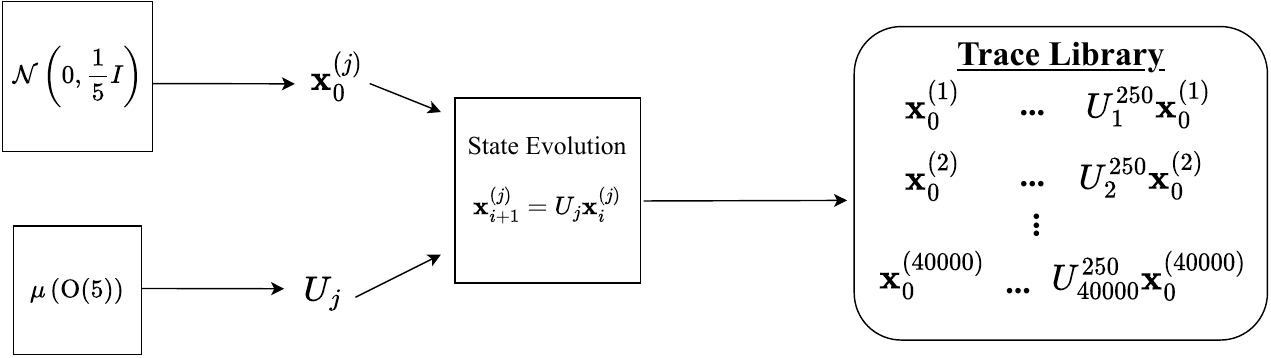}
    \caption{The generation of a train or test library of sequences.}\label{fig:library_gen}
\end{figure}
\paragraph{Generating a library of training sequences:} Depicted visually in Fig.~\ref{fig:library_gen}, we first compile a training library by the following method:

\begin{enumerate}
    \item Generate 40000 orthogonal matrices i.i.d. uniformly over all orthogonal matrices $U_1, \dots, U_{40000} \overset{iid}{\sim} \mu\left(\mathrm{O}(5)\right)$, where $\mu\left(\mathrm{O}(5)\right)$ is the Haar measure over orthogonal matrices in $\R{5\times 5}$. \cite{mezzadri2006generate}
    \item Generate 40000 i.i.d. initial states that will correspond to each training system $\xv_0^{(1)}, \dots, \xv_0^{(40000)} \overset{iid}{\sim} \gauss{0}{\frac{1}{5}I}$.
    \item Roll out the states to get observation sequences that are each 251 entries long, and compile the sequences as our training library.
\end{enumerate}

% that consists of one 251-entry-long observation sequence from each of 40,000 randomly-drawn training systems.  

\begin{figure}[tbph]
    \centering
    % Row 1: Distributions
    \begin{subfigure}[b]{0.459\linewidth}
        \centering
        \includegraphics[width=\linewidth]{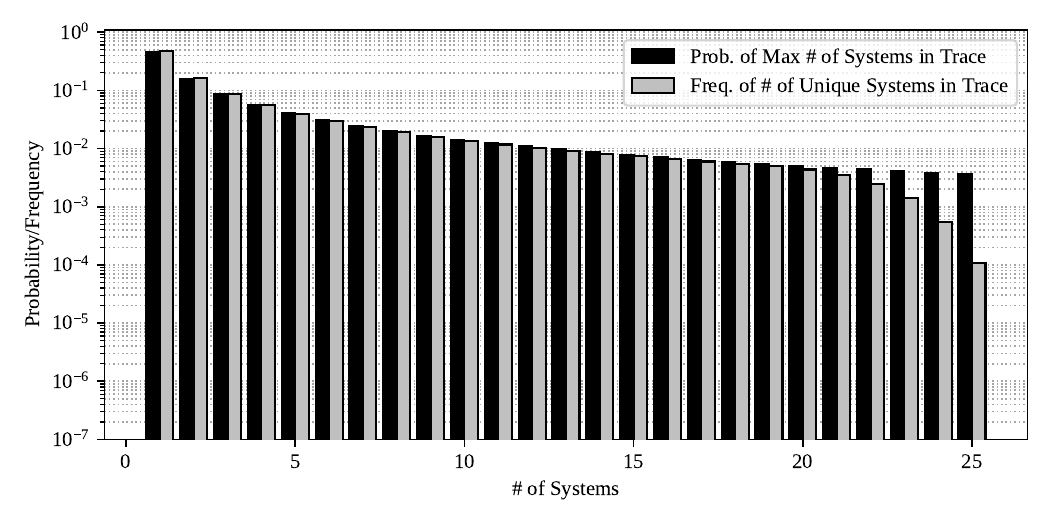}
        \caption{}
        \label{fig:zipf}
    \end{subfigure}
    \hfill
    \begin{subfigure}[b]{0.459\linewidth}
        \centering
        \includegraphics[width=\linewidth]{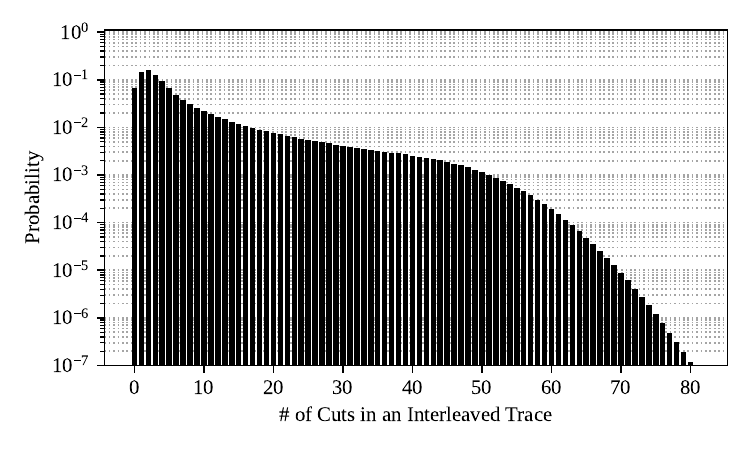}
        \caption{}
        \label{fig:num_cuts_dist}
    \end{subfigure}

    % Row 2: CCDFs
    \vspace{0.5em}
    \begin{subfigure}[b]{0.459\linewidth}
        \centering
        \includegraphics[width=\linewidth]{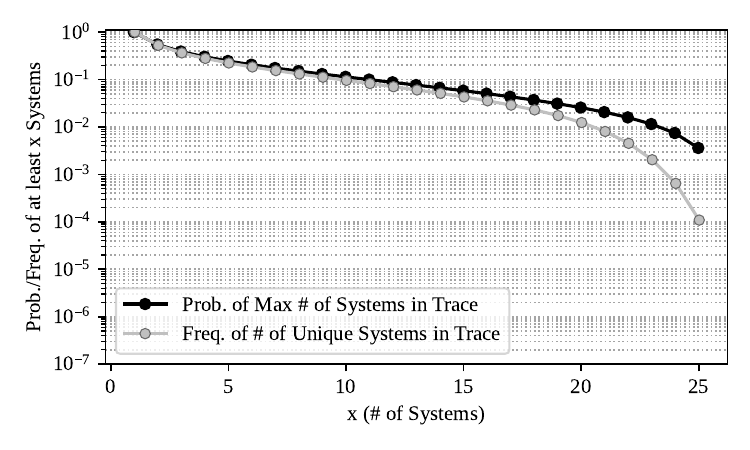}
        \caption{}
        \label{fig:zipf_ccdf}
    \end{subfigure}
    \hfill
    \begin{subfigure}[b]{0.459\linewidth}
        \centering
        \includegraphics[width=\linewidth]{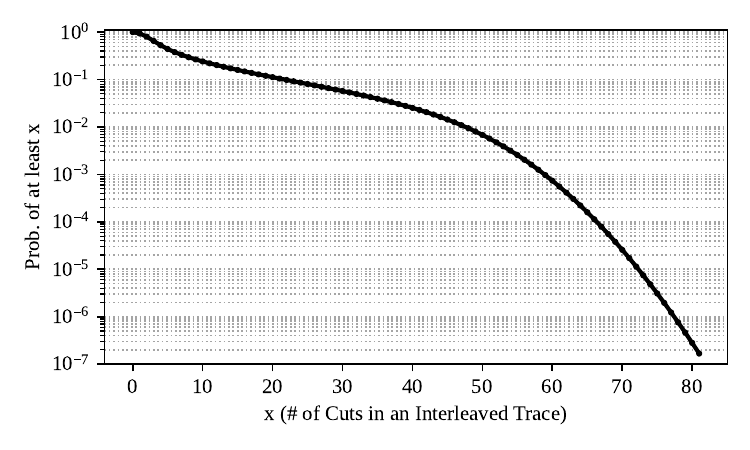}
        \caption{}
        \label{fig:num_cuts_ccdf}
    \end{subfigure}

    % Row 3: Frequencies
    \vspace{0.5em}
    \begin{subfigure}[b]{0.495\linewidth}
        \centering
        \includegraphics[width=\linewidth]{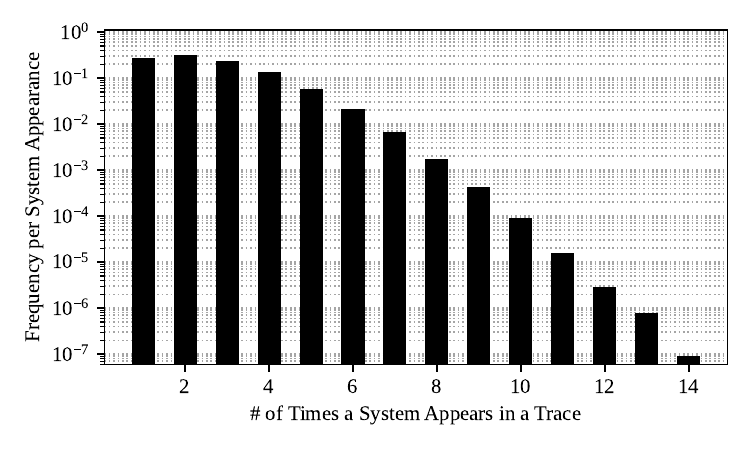}
        \caption{}
        \label{fig:sys_appears_dist}
    \end{subfigure}
    \hfill
    \begin{subfigure}[b]{0.495\linewidth}
        \centering
        \includegraphics[width=\linewidth]{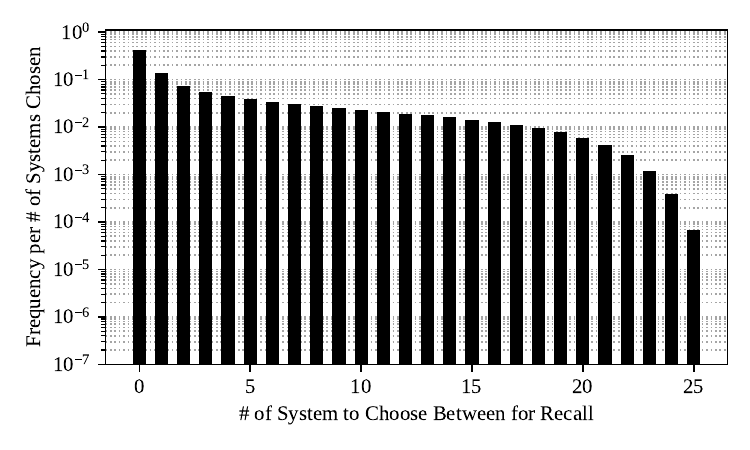}
        \caption{}
        \label{fig:sys_choice_recall}
    \end{subfigure}

    \caption{Distributions and complementary cumulative distribution functions (CCDFs) used in data generation --- Fig.~\ref{fig:zipf} is the $\mathrm{Zipf}(1.5, 25)$ distribution for the maximum number of systems per trace in black and the number of unique systems per trace for $1\times 10^7$ traces in silver. Fig.~\ref{fig:zipf_ccdf} shows the CCDFs for these distributions. Fig.~\ref{fig:num_cuts_dist} shows the PMF and Fig.~\ref{fig:num_cuts_ccdf} shows the CCDF of the number of cuts per trace. Fig.~\ref{fig:sys_appears_dist} shows the frequency of the number of times a system will appear in the same trace per system appearance. Lastly, Fig.~\ref{fig:sys_choice_recall} shows the frequency of the number of previously seen systems a predictor must choose between to recall in a trace per system appearance. For example, if system 0 is chosen then system 1, then system 0, then the model must choose between two systems to recall.}
    \label{fig:data_gen_distributions}
\end{figure}

\FloatBarrier
\paragraph{Cutting and interleaving training sequences}To form a training trace, we interleave segments of observation sequences from the %system family's 
library into a context window of length 251, by this process:
\begin{enumerate}
    \item Insert the start symbol at index 0.
    \item Sample the maximum number of systems in the trace $N$ from a $\mathrm{Zipf}(1.5,25)$ distribution depicted graphically\footnote{The Zipf distribution was chosen for its ubiquity in nature \cite{BakP.Per19471996Hnw:} and natural language \cite{SchutzeHinrich1999IP}, along with recent work pointing to its importance in modern neural networks \cite{michaud2024quantization}.} in Fig.~\ref{fig:zipf}. This means that no more than 25 systems will ever appear in a training trace. 
    \item Choose $N$ of the 40,000 systems in the training library uniformly at random without replacement.
    \item Randomly assign to each of the $N$ systems a pair of symbolic open and close labels for this training example.
    \item Sample the number of cuts $C \sim \mathrm{Poisson}(2N)$ to be made in the trace. This means that there will be $C+1$ trace segments in the trace.\footnote{On average, each training trace has $2N + 1$ segments to ensure that the trained model has seen ample interruptions and continuations of systems.}
    \item Place the $C$ cuts uniformly at random with replacement within the context window.
    \item For each segment created by the cuts, in order, uniformly at random choose one of the $N$ systems with replacement.
    \item At the cut at the beginning of the segment, the open label for this segment's system is inserted.\footnote{Since the open and close labels occupy two indices in the context window, there are three special cases that can occur: (1) If two cuts are sampled to be on top of each other, then the first of the two cuts that were sampled is ignored; (2) If the two cuts are sampled to occupy adjacent indices, then only the close label for the system corresponding to first of the two cuts is inserted, effectively making that index meaningless as close labels are masked; (3) If the two cuts are sampled so that there is only one index between them, then the open label for the system corresponding to the first of the two cuts is inserted and is immediately followed by the close label for that system, effectively making both indices meaningless due to the masking of the labels. Note that the distributions shown in Fig.~\ref{fig:data_gen_distributions} do not account for these rare special cases.}
    \item For the system chosen, check if it has appeared in a previous segment of the trace. If not, insert the system's segment from the training trace library starting at index 0. If this system has appeared in a previous segment, insert the system's segment from the training trace library starting at the index that corresponds to the continuation of the previous segment for this system.
    \item At one index before the next cut, insert the close label for this segment's system.
\end{enumerate}

\begin{figure}[ptbh]
    \centering
    \includegraphics[width=\linewidth]{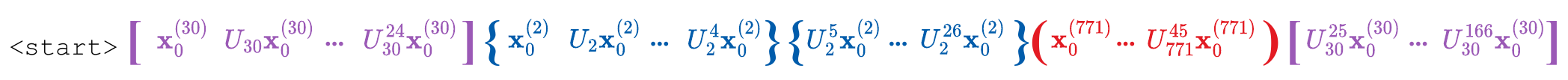}
    \caption{Example of a 251-element-long interleaved training example.}\label{fig:train_trace_example}
\end{figure}

Note that within a single training example, segments of a particular system always start with the same open token and always end with its corresponding close token. These random assignments are redrawn at the beginning of the interleaving process for each training example; therefore, \textit{the same system can have different symbolic open and close labels when it appears in different training examples}. See Fig.~\ref{fig:train_trace_example} for a diagram of an interleaved training example.

Given this randomized procedure for generating interleaved training examples, we can analyze the training distribution to better understand how frequently a model must recall a system, or sees many systems in a trace.
% \begin{figure}[tbph]
%     \centering
%     \begin{subfigure}[b]{0.45\linewidth}
%         \centering
%         \includegraphics[width=\linewidth]{diagrams/zipfian.pdf}
%         \caption{}
%         \label{fig:zipf}
%     \end{subfigure}
%     \begin{subfigure}[b]{0.45\linewidth}
%         \centering
%         \includegraphics[width=\linewidth]{diagrams/unique_systems_dist.pdf}
%         \caption{}
%         \label{fig:unique_sys_dist}
%     \end{subfigure}
%     \begin{subfigure}[b]{0.45\linewidth}
%         \centering
%         \includegraphics[width=\linewidth]{diagrams/num_cuts_dist.pdf}
%         \caption{}
%         \label{fig:num_cuts_dist}
%     \end{subfigure}
%     \caption{Distributions used in data generation: \ref{fig:zipf} the $\mathrm{Zipf}(1.5, 25)$ distribution for the maximum number of systems per trace, \ref{fig:num_cuts_dist} the frequency of the number of cuts per trace for $1\times 10^7$ trials, \ref{fig:unique_sys_dist} the frequency of the number of unique systems per trace for $1\times 10^7$ trials.}
%     \label{fig:data_gen_distributions}
% \end{figure}

In Fig.~\ref{fig:data_gen_distributions}, we show relevant distributions that are derived from the randomized interleaving procedure in this section. Figs.~\ref{fig:zipf} and \ref{fig:zipf_ccdf} show the $\mathrm{Zipf}(1.5, 25)$ distribution for the maximum number of systems per trace in black and the number of unique systems per trace in silver. The frequency of number of unique systems per trace follows closely the $\mathrm{Zipf}(1.5, 25)$ distribution for the smaller quantities of systems, but diminishes quicker for larger numbers of systems, due to the coupon-collecting phenomenon of picking the same system multiple times in a trace. The PMF for the number of cuts made in an interleaved trace is shown in Fig.~\ref{fig:num_cuts_dist}, while the CCDF for this quantity is given in Fig.~\ref{fig:num_cuts_ccdf}. Fig.~\ref{fig:sys_appears_dist} shows the frequency of how many times a system appears in a training trace. If the same system appears more than once, then the model must perform recall. Therefore, Fig.~\ref{fig:sys_appears_dist} gives an idea of how often the model must recall a system during training. Finally, Fig.~\ref{fig:sys_choice_recall} provides the frequency of the number of previously seen systems in the training trace that are candidates to be continued when a predictor is tasked to recall a system. The value zero on the x-axis of this figure means that the model is seeing a system for the first time in a training example and has no need to recall. Later, in Section \ref{sec:needle_in_a_hay_test_setup}, we construct tests for the associative recall ability of the trained model for different numbers of candidate systems to be continued in the trace. Fig.~\ref{fig:sys_choice_recall} shows that for 19 candidate systems, the largest number of candidate systems that we tested on, the model has been presented with this situation less than $1\%$ of the time during training.

\begin{figure}[ptbh]
    \centering
    \includegraphics[width=\linewidth]{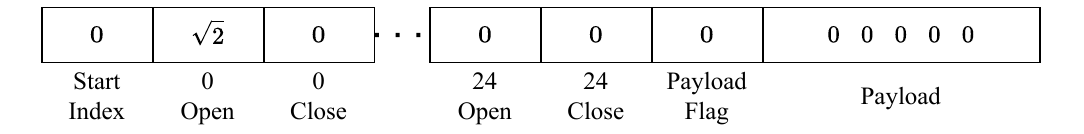}
    \caption{The one-hot encoding of an open symbolic label. In this example, the system corresponding to this label is assigned to be ``system 0.''}\label{fig:open_label}
\end{figure}

\FloatBarrier
\paragraph{Input structure and embedding}The input dimension of our models is 57.
There are 50 dimensions for encoding paired symbolic open and close labels, a dimension for the start symbol, a dimension for the payload flag and 5 dimensions to hold the 5-dimensional observation vectors. The special symbols are one-hot encoded vectors; see Fig.~\ref{fig:open_label} for an example of the open symbol.
For the observation sequence between the SPLs, the 5-dimensional state vectors are inserted into the payload portion of the input vector, the payload flag is set to 1, and the rest of the input vector dimensions are zeroed out. 

\begin{figure}[ptbh]
    \centering
    \includegraphics[width=\linewidth]{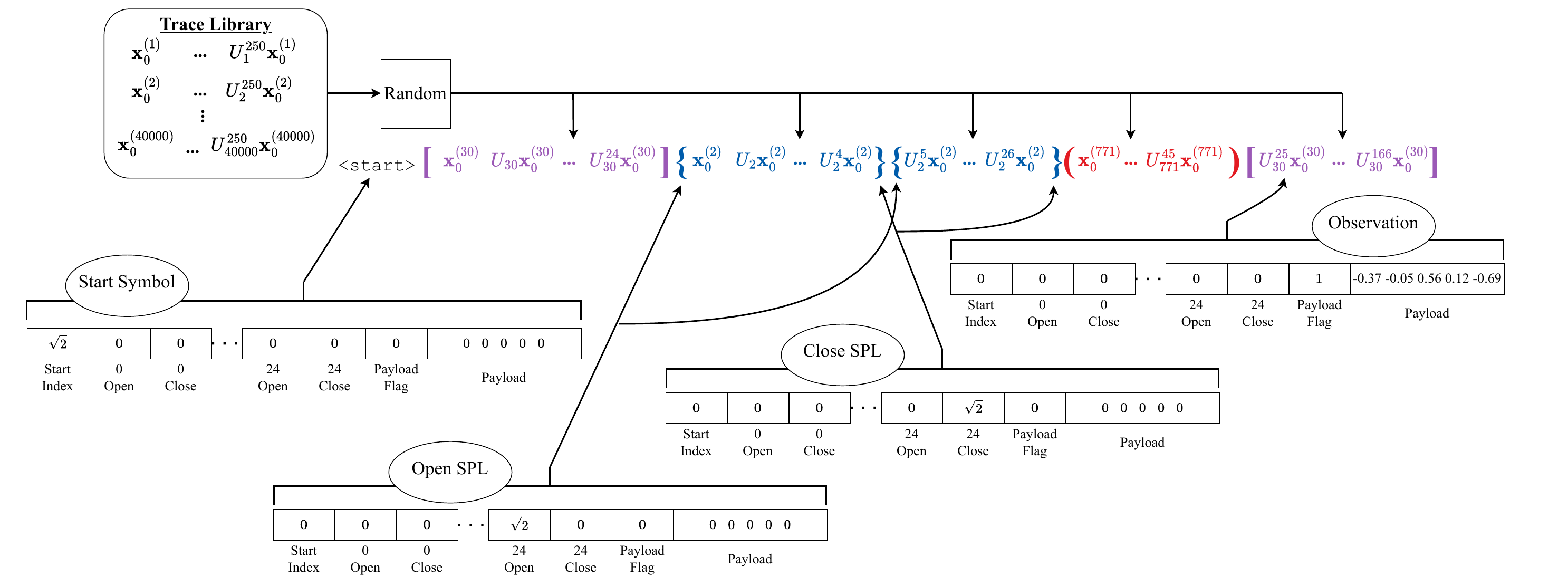}
    \caption{Generating a training example --- % An example of a special token input that signifies the beginning of a trace segment from a system that is labeled --- just for this context --- as `1'. There are 25 such distinctly labeled systems possible within a context. 
    Notice in this example the continuation from the first segment to the last (system $U_{30}$), and from the 2nd segment to the 3rd (system $U_2$). The ``parentheses" (symbolic punctuating labels) are encoded as special tokens as shown.}\label{fig:full_training_diagram}
\end{figure}

The entire randomized procedure of generating interleaved training traces is depicted in Fig.~\ref{fig:full_training_diagram} along with the structure of the inputs into the model.

\FloatBarrier
\paragraph{Model and embedding}

Building off of the codebase in \cite{DuOymakTrans}, which was influenced by \cite{garg2022can}, we train a $2.42$M parameter GPT-2 style transformer to perform this task. Our model\footnote{These parameter counts and model dimensions are for our ``medium'' model. Three other models of different sizes were also trained, and their model dimensions are given table \ref{tab:model_size_training_params} in Appendix \ref{sec:model_size_effects}.} has hidden dimension 128, 12 layers, and 8 heads. Our model's input embedding is $128\times 57$ dimensional. The model's output layer is $5 \times 128$ to ensure that the model makes 5-dimensional predictions. The input and output layers are untied \cite{DuOymakTrans}.

\paragraph{Training and Hyperparameters} \label{sec:hyperparam}
New interleaved training examples are generated for each training iteration and our GPT-2-style model was trained for next token prediction on these training traces.\footnote{The model sees newly interleaved training examples at each iteration, but the training traces that are interleaved into the training example are drawn from the fixed training library of 40,000 sequences. Therefore, the model undergoes single-epoch training where the information within a training example might have appeared in many other training examples.}
The loss for all SPLs on the output were zeroed out. A model that successfully recalls the state of a system seen previously in its context should make predictions after the open token that perform as it would have if the relevant sequence had continued on without interruption. 

Following the choice made in \cite{DuOymakTrans}, we trained our model with a weight decay of $1 \times 10^{-2}$. We used a batch size of 512, a learning rate of $\approx 1.58\times10^{-5}$, and trained on a single NVIDIA L40S GPU with 48GB of RAM. A single training run takes around 5 days. We used the AdamW Optimizer \citep{loshchilov2019decoupledweightdecayregularization} and trained using mean-squared error loss.

\section{Results}\label{sec:results}
This toy problem has two qualitatively different capabilities that a model can acquire during training: in-context learning a linear system as it is seen (ICL), and recalling what has been in-context learned about a previously seen system (associative recall). Within the ICL ability, again, there are qualitatively two sub-abilities: ICL for the first system that is seen, and ICL for the subsequent new systems that are seen. The second sub-ability requires a model to learn to restart its in-context learning for a new system. 

\subsection{Test Setup}\label{sec:test_setup}

\subsubsection{Uninterleaved sequence test}\label{sec:basictest_setup}
To test the model's ICL ability for the first system that is seen (Section \ref{sec:first_sys_learn}), we generate 100 held-out systems and 1000 different held-out initial states\footnote{We generate 1000 initial states for each system to narrow down the quartiles in the squared-error curves.} by the same method described in Section \ref{sec:datagen} to form our testing library. We then evaluate the model on the uninterleaved traces from this testing library.

\FloatBarrier
\subsubsection{Needle-in-a-haystack test}\label{sec:needle_in_a_hay_test_setup}

\begin{figure}[tbph]
        \centering
        \includegraphics[width=0.9\linewidth]{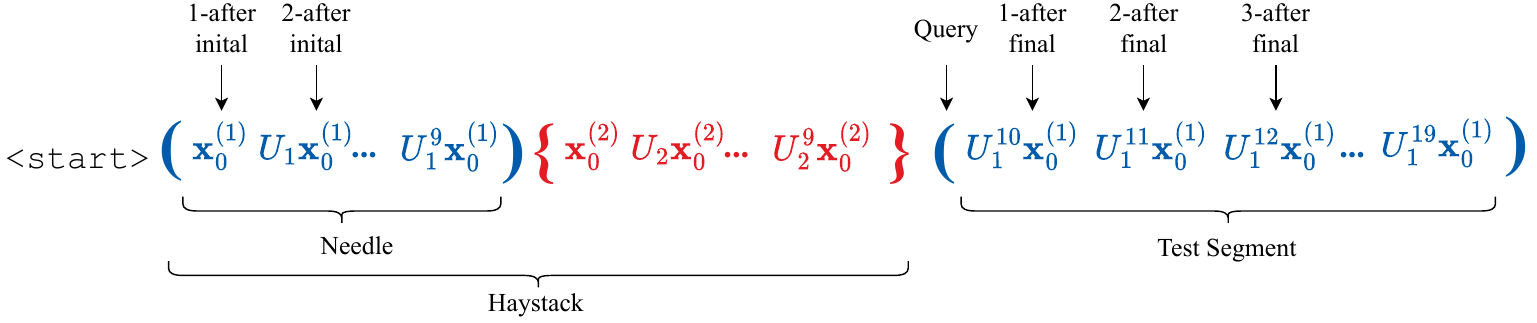}
        \caption{Needle-in-a-haystack test example. (A two system haystack.)}
        \label{fig:needle_in_haystack_diagram}
\end{figure}

 To evaluate the model's ability to restart ICL on a new system (Section \ref{sec:restart_sys_main_paper}) and recall a previously seen system (Section \ref{sec:assoc_recall_results}), we generate a series of structured ``needle-in-a-haystack'' test traces through interleaving the traces in the testing library generated in Section \ref{sec:basictest_setup}. A single ``needle-in-a-haystack'' trace is generated by the following procedure:
 \begin{enumerate}
    \item Choose $N \in [1,19]$ to be the number of distinct systems in a test trace.
    \item For each of the $N$ systems, insert a segment of 10 state observations starting from index 0 from the testing library into the ``needle-in-a-haystack'' test trace. Each of these segments are individually punctuated with a unique open and close symbol pair. We call this portion of the trace the ``haystack''.\label{insert_haystack_segs}
    \item Append a query open symbol to the test trace that signifies which system in the haystack will be be continued. The segment that will be continued is called the ``needle''.
    \item Append 10 state observations from the continuation of the system in the haystack corresponding to the query open symbol. This portion is called the ``test segment''.
 \end{enumerate}
See Fig.~\ref{fig:needle_in_haystack_diagram} for a diagram of a test trace for $N=2$ systems in the haystack and system $U_1$ as the needle.

For the full ``needle-in-a-haystack'' test dataset, we would like to ensure that we test on the same systems for the different values of $N$, while having diverse systems in our dataset so that our results are statistically meaningful. To achieve this, we test on 50 ``needle-in-a-haystack'' trace configurations. A trace configuration is a specific ordering of systems from the testing library in the positions of the haystack. For the first needle-in-a-haystack trace configuration that we generate, we place a segment from ``system 0'' from the testing library into the first position in the haystack, and fill the rest of the haystack positions consecutively until the $N^{\text{th}}$ position is filled with a segment from ``system $N - 1$''. For the next trace configuration, the first position in the haystack is filled with a segment from ``system 1'' and the rest of the haystack positions are filled consecutively until the $N^{\text{th}}$ position is filled with a segment from ``system $N$''. This pattern continues until the last trace configuration. In our case, we tested on 50 trace configurations, meaning the haystack of the last trace configuration started with ``system 49'' and ended with ``system $48+N$''.  Each trace configuration is populated with 1000 different initial states for each system. For the results in the main paper, the test segment is a continuation of the segment in the first position of the haystack. For results where segments in other haystack positions are continued in the test segment see Section \ref{sec:needle_last_seg_context} in the Appendix.

\begin{figure}[tbph]
    \centering
    \includegraphics[width=0.25\linewidth]{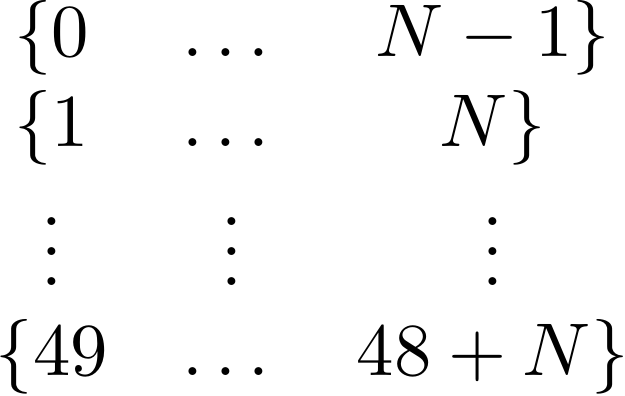}
    \caption{When testing on 50 needle-in-a-haystack trace configurations, the order of system indices from the testing library in a haystack of size $N$ for each needle-in-a-haystack test trace is given above. For each system, 1000 sequences are interleaved to build a testing dataset of shape $50 \times 1000 \times (12N + 1) \times 5$. The shape of the second to last axis is due to the start token and haystack segments being 10 context indices long plus two indices for the symbolic open and close labels. The last axis is 5-dimensional since every system has 5-dimensional observations.}\label{fig:needle_avg_indices}
\end{figure}

\subsection{In-context learning system dynamics}\label{sec:icl}
We now present results on how well the transformer-based model can in-context learn a linear system's dynamics. Section \ref{sec:first_sys_learn} shows the results for learning about the first system that is seen, while Section \ref{sec:restart_sys_main_paper} shows the results for learning about subsequent systems that are seen and the ability to restart ICL.

\subsubsection{Can a model learn the first system in-context? Yes} \label{sec:first_sys_learn}

\begin{figure}[tbph]
    \centering
    %Subfigure 1
    \begin{subfigure}{0.495\linewidth}
        \centering
        \includegraphics[width=\linewidth]{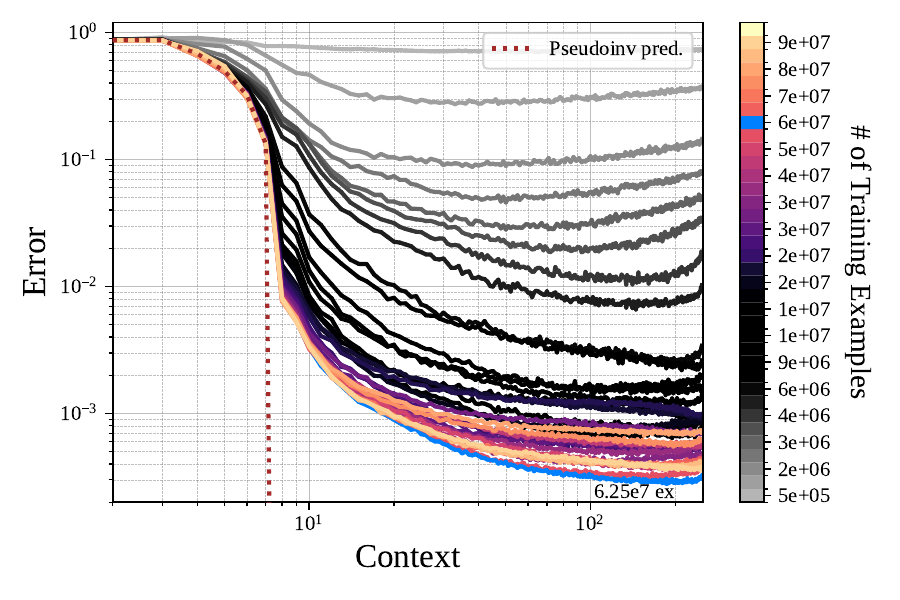}
        \caption{}  \label{fig:zero_cut_context}
    \end{subfigure}
    \hfill
   % Subfigure 2
    \begin{subfigure}{0.495\linewidth}
        \centering
        \includegraphics[width=\linewidth]{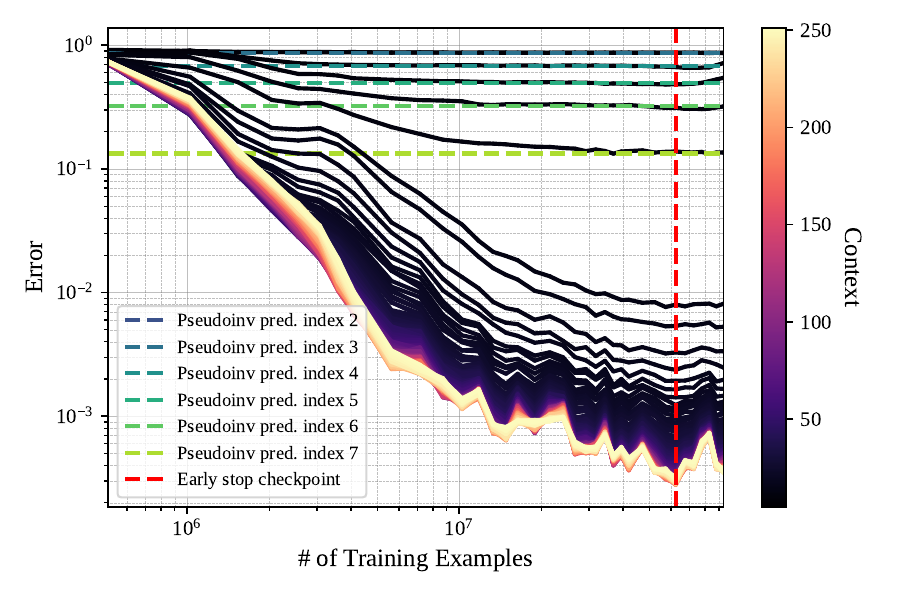}
        \caption{} \label{fig:zero_cut_context_train_conv}
    \end{subfigure}
    \caption{Performance on a long uninterleaved trace --- \ref{fig:zero_cut_context} and \ref{fig:zero_cut_context_train_conv} depict the in-context learning performance of the model on long uninterleaved test examples. The median squared-error is plotted against the context index in Fig.~\ref{fig:zero_cut_context}, and against the number of training examples seen in Fig.~\ref{fig:zero_cut_context_train_conv}. The red line in Fig.~\ref{fig:zero_cut_context_train_conv} at $6.25 \times 10^7$ training examples, and the blue curve in Fig.~\ref{fig:zero_cut_context} denote the checkpoint that we use for early stopping. in Fig.~\ref{fig:zero_cut_context} the optimal pseudoinverse predictor is the brown dashed curve, while in Fig.~\ref{fig:zero_cut_context_train_conv} it is denoted by the blue and green horizontal lines for each of the early context indices. in Fig.~\ref{fig:zero_cut_context_train_conv}, notice that the model gradually learns to make optimal predictions as opposed to the sudden emergence of associative recall ability seen in Section \ref{sec:assoc_recall_results}, since the prediction errors for early indices slowly converge towards the pseudoinverse predictor's performance. }\label{fig:combined_zero_cut}
\end{figure}

We first confirm that our trained model is able to in-context learn the first system seen in its context. We evaluate the model on the uninterleaved traces from the testing library specified in Section \ref{sec:basictest_setup}. In Fig.~\ref{fig:zero_cut_context}, we plot the median squared-error over these test traces vs the context index, and the color of each curve represents how far along the model is in training. The dotted line in this figure is the median squared-error of the pseudoinverse predictor in \eqref{eqn:pseudoinv_pred} over the same test traces. In Fig.~\ref{fig:zero_cut_context_train_conv} we plot the median squared-error over these test traces as training proceeds (measured by the number of training examples seen so far), the color of each solid curve represents the context index, the blue and green dotted horizontal lines are the optimal pseudoinverse predictor's median squared error for the specific early context indices.

Notice in Fig.~\ref{fig:zero_cut_context} that the early model checkpoints saturate out and cannot continue to make better predictions with more context. Nevertheless, after seeing $6.25 \times 10^7$ training examples, the model's prediction error on indices 2 through 7 closely match those of the pseudoinverse baseline from \eqref{eqn:pseudoinv_pred}. This is also seen in Fig.~\ref{fig:zero_cut_context_train_conv} where, as training proceeds, the dark curves representing the model's prediction errors on early indices converge to the prediction error of the pseudoinverse predictor for the corresponding index given by the blue and green curves. The best performance of the model is at the end of the context window with a median squared-error of $\approx 2\times10^{-4}$. According to \cite{liu2024can}, this is near the practical precision threshold for transformer models.

Notice in Fig.~\ref{fig:zero_cut_context_train_conv} the model is gradually learning to make better predictions as training continues. This ICL ability for the first system seen is the same ability that is studied by \cite{DuOymakTrans,pmlr-v235-sander24a} and using this evaluation metric, we see that sudden emergence is not present. Nonetheless, using the same evaluation metric, the ability to restart ICL for a new system (Section \ref{sec:restart_sys_main_paper}) and to recall a previously seen system (Section \ref{sec:assoc_recall_results}) both exhibit emergence.

Additionally, we notice in Fig.~\ref{fig:zero_cut_context_train_conv} that the squared-error bottoms out after $6.25 \times 10^7$ training examples, showing that the model suffers from overfitting late in training. Having seen this, we set our early stopping checkpoint at $6.25 \times 10^7$ training examples, as denoted by the red vertical line in Fig.~\ref{fig:zero_cut_context_train_conv} which corresponds to the blue curve in Fig.~\ref{fig:zero_cut_context}.

In summary, the model is able to use context to make better predictions of state observations from held-out test systems. The model's ability gradually develops during training, as opposed to how associative recall develops suddenly later in training as seen in Section \ref{sec:assoc_recall_results}.

\FloatBarrier
\subsubsection{Emergence of the ability to restart ICL on new systems}\label{sec:restart_sys_main_paper}

We now show the training dynamics for the ability to restart in-context learning for a new system that was not the first system seen in-context. We find that learning this restart ability is not gradual, as it was for learning the first system seen (Section~\ref{sec:first_sys_learn}). Instead, we see that the model begins to transition from poor restart performance to good restart performance early in training as compared to when associative recall emerges in training which we will see in Section~\ref{sec:assoc_recall_results}. We study the model's performance on the haystack segments of ``needle-in-a-haystack'' test examples (see the diagram in Fig.~\ref{fig:needle_in_haystack_diagram}). 

\begin{figure}[tbph]
    \centering
    \begin{subfigure}[b]{0.48\linewidth}
        \centering
        \includegraphics[width=\linewidth]{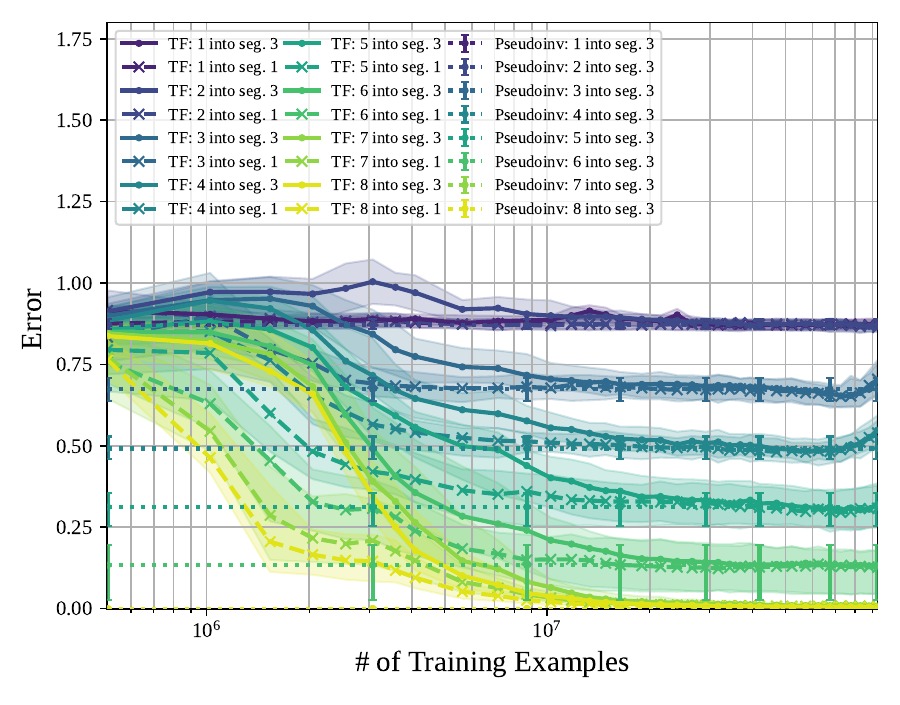}
        \caption{}
        \label{fig:new_system_train_conv}
    \end{subfigure}
    \hfill
    \begin{subfigure}[b]{0.48\linewidth}
        \centering
        \includegraphics[width=\linewidth]{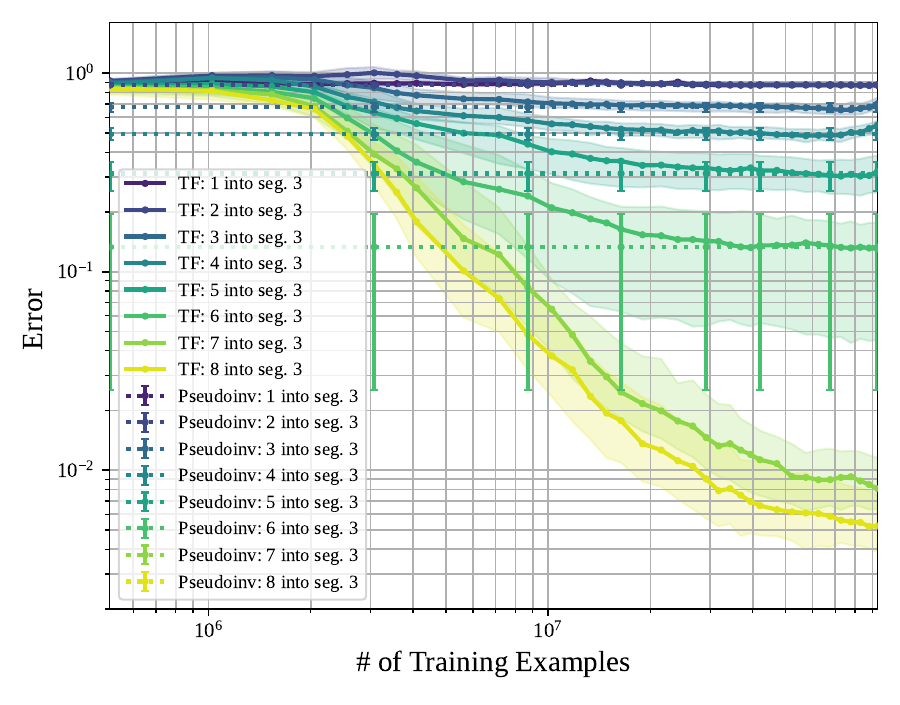}
        \caption{}
        \label{fig:restart_sys_train_conv_all_steps_log}
    \end{subfigure}
    \caption{Performance on new subsequent segments. \ref{fig:new_system_train_conv} is the squared-error of predictions on steps 1 through 8 into the first and third system segments, where each segment is seen for the first time in context. \ref{fig:restart_sys_train_conv_all_steps_log} is the squared-error for steps 1 through 8 into the third system segments on log scale. The error bars show the $25^{\text{th}}$ and $75^{\text{th}}$ percentiles across trace configurations of the model's prediction error across the medians over the 1000 initial states in each trace configuration. The horizontal dotted lines are the median squared-error of the optimal pseudoinverse predictor.}
    \label{fig:restart_sys_train_conv_both}
\end{figure}

In Fig.~\ref{fig:restart_sys_train_conv_both}, the median squared-error vs the number of training examples seen is plotted for steps 1 through 8 into the first system segment in the haystack and the third system segment in the haystack. Specifically, in Fig.~\ref{fig:new_system_train_conv} we see that at the beginning of training the model has not learned to restart its predictions for the third system segment, as its median squared-error in segment 3 is well above its counterpart predictions in segment 1 for all steps into each segment except for step 1. This shows that early on in training, there is clearly substantial interference from earlier segments when the model tries to learn to predict the behavior for a new sequence (explicitly labeled as such) that it is seeing in the third position in the haystack. As training proceeds, we see that the median squared-error for each step in segment 3 converges to the value of its segment 1 counterpart. Fig.~\ref{fig:restart_sys_train_conv_all_steps_log} shows that the model transitions towards restarting ICL correctly when training has processed $\approx 2\times 10^6$ training examples.

\begin{figure}[tbph]
    \centering
    \includegraphics[width=0.95\linewidth]{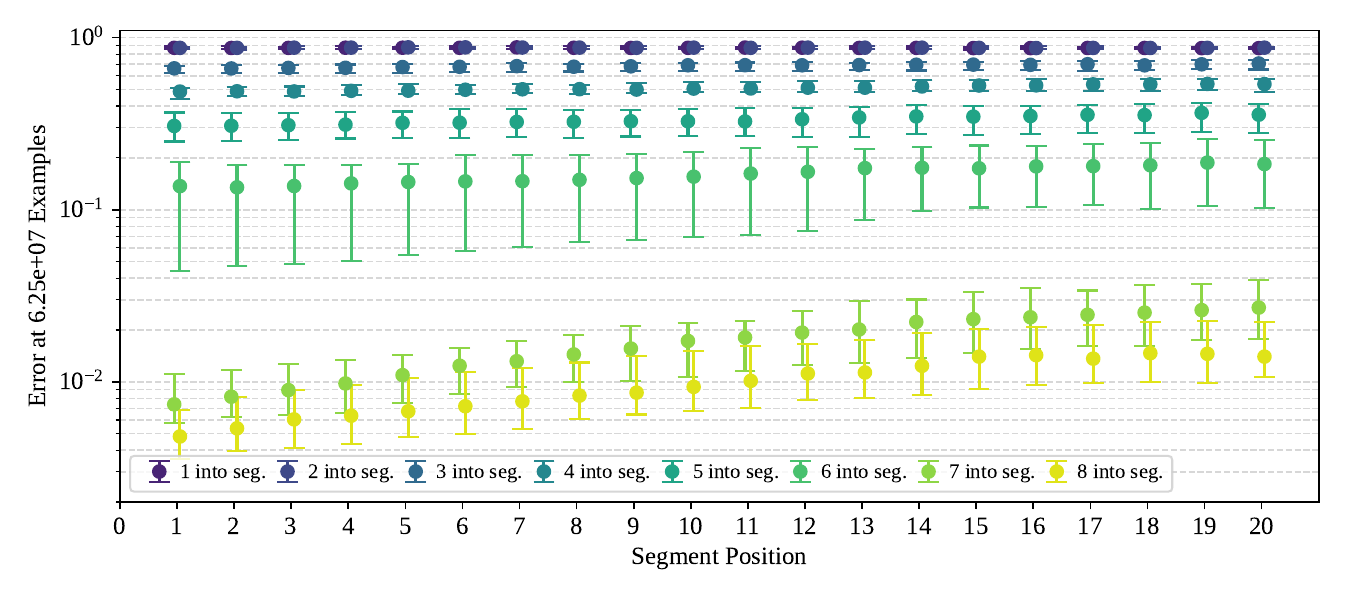}
    \caption{Restarting for a new system at the early-stopping checkpoint after seeing $6.25\times 10^{7}$ training examples.}
    \label{fig:restart_sys_6_25e7_log}
\end{figure}

In Fig.~\ref{fig:restart_sys_6_25e7_log}, we show the median squared-error of the model's predictions for up to 8 steps into each new segment at the early-stopping checkpoint of $6.25\times 10^{7}$ training examples. This log-scale plot shows that even at the end of training, the model's ability to restart ICL for new systems slowly degrades when predicting indices 6, 7 and 8 as more new systems are presented in the context window. This is seen as the upward trend in the green and yellow curves in Fig.~\ref{fig:restart_sys_6_25e7_log}.

% Again, in Section~\ref{sec:first_sys_learn}, we saw that the model gradually learns to do in-context learning of the orthogonal dynamics governing a particular sequence and to predict the next entry in the current sequence. There is no "emergence" behavior. 
% In Fig.~\ref{fig:new_system_train_conv}, we see that early on in training, there is clearly substantial interference from earlier segments when the model tries to learn to predict the behavior for a new sequence (explicitly labeled as such) that it is seeing in the third position. 

% \subsection{Learning associative recall} \label{sec:assoc_recall_results} 
\FloatBarrier
% \subsection{Training dynamics showing emergence}
\subsection{Emergence of associative recall}\label{sec:assoc_recall_results}
We now study the prediction performance of the model when queried for recall. In particular, how it depends on the index into the test segment. Furthermore, we study how the ability to predict different indices in the test segment develops differently during training. Particularly, the ability to predict the first index into the test segment develops much later in training and much more abruptly than the ability to predict the subsequent indices into the test segment. 

\begin{figure}[htbp]
    \centering
    \begin{subfigure}[b]{0.32\linewidth}
        \centering
        \includegraphics[width=\linewidth]{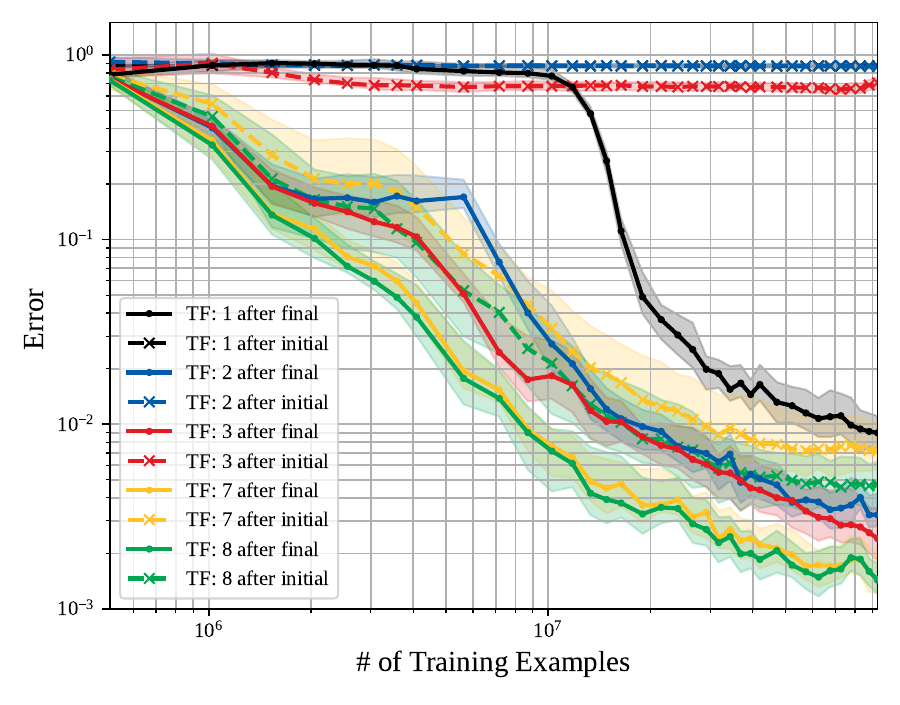}
        \caption{1 system haystack.}
        \label{fig:ortho_med_needle_train_conv_haystack_len_1_all_haystack_len_12_log_lin_log_main_paper}
    \end{subfigure}
    \hfill
    \begin{subfigure}[b]{0.32\linewidth}
        \centering
        \includegraphics[width=\linewidth]{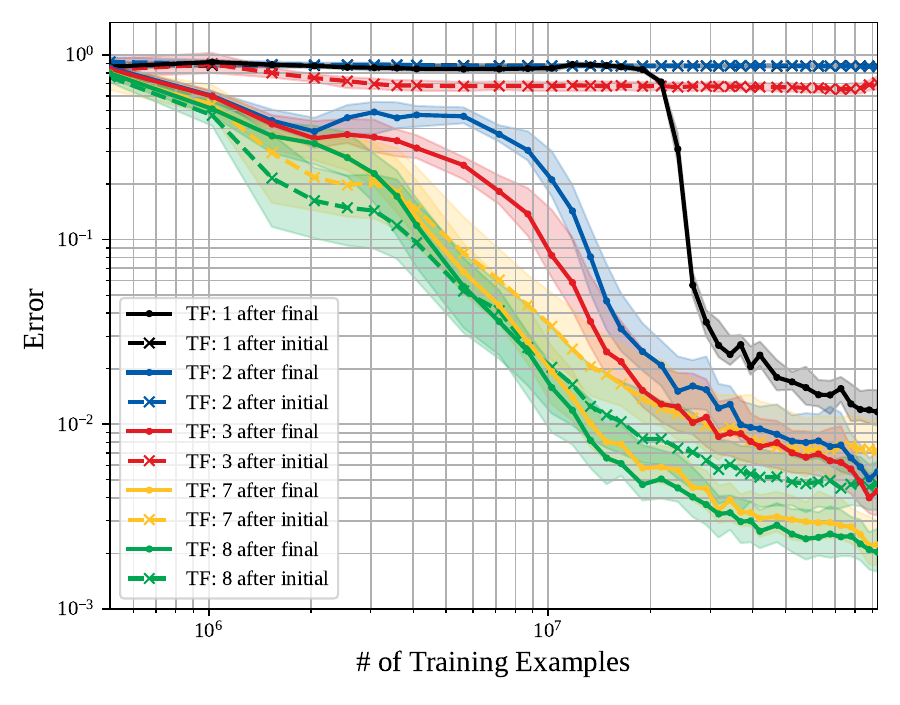}
        \caption{2 system haystack.}
        \label{fig:ortho_med_needle_train_conv_haystack_len_2_all_haystack_len_12_log_lin_log_main_paper}
    \end{subfigure}
    \hfill
    \begin{subfigure}[b]{0.32\linewidth}
        \centering
        \includegraphics[width=\linewidth]{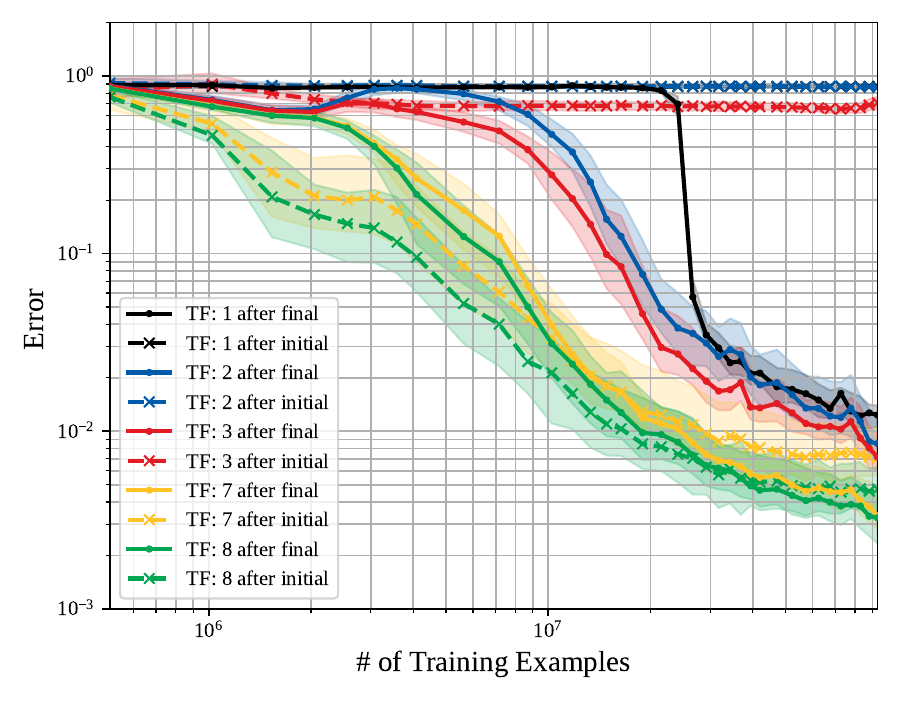}
        \caption{5 system haystack.}
        \label{fig:ortho_med_needle_train_conv_haystack_len_5_all_haystack_len_12_log_lin_log_main_paper}
    \end{subfigure} 
    \caption{Training dynamics for recall --- The $25^{\text{th}}$, $50^{\text{th}}$, and $75^{\text{th}}$ quartiles of the squared-error of the model's predictions vs the number of training examples seen during training so far are plotted on log-log plots for $N = 1$ in Fig.~\ref{fig:ortho_med_needle_train_conv_haystack_len_1_all_haystack_len_12_log_lin_log_main_paper}, $N=2$ in Fig.~\ref{fig:ortho_med_needle_train_conv_haystack_len_2_all_haystack_len_12_log_lin_log_main_paper}, and $N=5$ in Fig.~\ref{fig:ortho_med_needle_train_conv_haystack_len_5_all_haystack_len_12_log_lin_log_main_paper}. All haystack segments are of length 10 (excluding delimiting tokens). %The transformer model is shown the first 10 long segment (excluding open and close tokens) from a system trace, then is tasked to continue predicting the same trace after being interrupted by a segment from another system. 
    The test set consisted of 1,000 ``needle-in-a-haystack'' traces from each of 50 systems. The dashed curves marked with crosses show the performance for indices 1, 2, 3, 7, and 8 steps after the initial open symbol, while the solid curves marked with dots show the performance for the same indices after the final open symbol. Notice in all the above figures that the solid black curve, showing the model's ability to recall the correct system from just seeing its corresponding open symbol, very sharply improves after $10^{7}$ training examples.}
    \label{fig:combined_training_dynamics}
\end{figure}

To uncover this, we test the model on the ``needle-in-a-haystack'' test traces described in Section \ref{sec:first_sys_learn} with different values of $N$, for the number of systems in the haystack. Fig.~\ref{fig:combined_training_dynamics} shows results for $N = 1,2$, and $5$. Results for more $N$ values are given in Appendix~\ref{sec:more_train_dyn}. In Fig.~\ref{fig:combined_training_dynamics} the solid curves marked with dots show the median squared-error of the model's predictions on the first, second, third, seventh, and eighth indices into the test segment. These curves are labelled as ``after final'' in the legend, because their indices are after the final open symbol. To contrast, the dashed curves marked with crosses show the median squared-error of the model's predictions on the same indices into the first segment in the haystack. These curves are labelled as ``after initial'' in the legend, since their indices occur directly after the initial open symbol. These indices are depicted in Fig.~\ref{fig:needle_in_haystack_diagram}.

% We study the transformer's performance versus (a) the number of training examples seen so far %(measured by the number of training examples seen so far) 
% in Fig.~\ref{fig:combined_training_dynamics}; (b) the position of the needle in the haystack in Fig.~\ref{fig:ortho_needle_pos_error_ratios}; (c) the position in the final test segment in Fig.~\ref{fig:ortho_needle_context}; as well as (d) the number $N$ of systems in the haystack Fig.~\ref{fig:squared-error_vs_haystack_len}. 

To perform perfectly on the associative recall task, the model must implicitly learn and remember the correct $5\times 5$ orthogonal matrix corresponding to a particular system. In all subfigures in Fig.~\ref{fig:combined_training_dynamics}, the solid curves largely decrease as training proceeds.\footnote{For the solid blue and red curves showing the model's recall ability on the second and third indices after the query symbol, we notice double-descent behavior that is more pronounced for larger haystacks.} This shows that training is improving the model's ability to recall. More specifically, the solid black curve for the pure recall task of predicting the first observation in the test segment shows emergence-style behavior: first a steady high squared-loss until it begins to drop at some point in training. Interestingly, this ability begins to emerge {\em earlier} ($\approx 1.5\times 10^7$ training examples seen) for the simple case of $N=1$ (Fig.~\ref{fig:ortho_med_needle_train_conv_haystack_len_1_all_haystack_len_12_log_lin_log_main_paper}), than for case when $N=5$ where the transition happens closer to $2.5\times 10^7$ (Fig.~\ref{fig:ortho_med_needle_train_conv_haystack_len_5_all_haystack_len_12_log_lin_log_main_paper}). % Specifically, the ability emerges between 800K and 2.5M training example inputs processed during training for one system in the haystack, as compared to closer to 9M for 19.  %in Fig.~\ref{fig:ident_med_needle_train_conv_haystack_len_1_all_haystack_len_12_log_lin_log_main_paper}, as compared to closer to 9M in Fig.~\ref{fig:ident_med_needle_train_conv_haystack_len_19_all_haystack_len_12_log_lin_log_main_paper}.
One system in the haystack is an 
arguably-trivial recall test, since the model must recall the only system that it has seen so far.
%when there is only one system in the entire ``haystack'' than when we have 19 different systems in the context before asking it to recall one of them, as illustrated in Fig.~\ref{fig:ident_med_needle_train_conv_haystack_len_19_all_haystack_len_12_log_lin_log_main_paper} where the ability emerges closer to when 9M examples have been processed. 
While recall performance for $N=5$ is showing seemingly no improvement in Fig.~\ref{fig:ortho_med_needle_train_conv_haystack_len_5_all_haystack_len_12_log_lin_log_main_paper}, the recall for $N=1$ in Fig.~\ref{fig:ortho_med_needle_train_conv_haystack_len_1_all_haystack_len_12_log_lin_log_main_paper} has gotten substantially better. The whole time, we are also seeing improvements in recognizing the underlying state evolution of the time-series as evidenced by better progress in the other blue, red, yellow, and green solid curves --- all of which involve predicting the correct orthogonal transformation of the observed state.

% The solid black curves in Fig.~\ref{fig:combined_training_dynamics} represent the pure recall task show clear emergence of this ability to recall. As with identity systems, this emergence happens earlier for the easier problem where there is only one system in the haystack (After 12M examples in Fig.~\ref{fig:ortho_med_needle_train_conv_haystack_len_1_all_haystack_len_12_log_lin_log_main_paper}) than when there are four or nine systems in the haystack (25M examples in Fig.~\ref{fig:ortho_med_needle_train_conv_haystack_len_19_all_haystack_len_12_log_lin_log_main_paper}). Again, see Fig.~\ref{fig:phase_transition_haystack_len_conglomerate} for all haystack lengths.

\begin{figure}[tbph]
    \centering
    \includegraphics[width=0.7\linewidth]{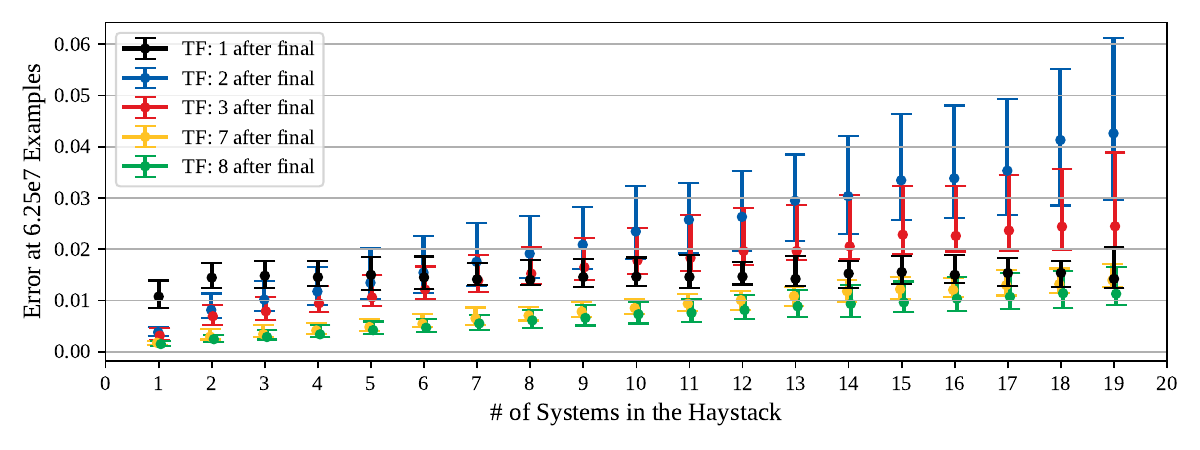}
    \caption{The $25^{\text{th}}$, $50^{\text{th}}$, and $75^{\text{th}}$ quartiles of the squared-error after $6.25\times 10^7$ training examples as the number of systems in the haystack $N$ increases. Notice that predicting 1-after the open symbol is largely unaffected by the value of $N$, as the black markers stay steady around $0.015$. Although the information encoded in the open symbolic label keeps the difficulty of the task constant as $N$ increases, nevertheless, the performance for 2,3,7, and 8-after the open symbol predictions steadily degrades as $N$ increases.}
    \label{fig:squared-error_vs_haystack_len}
\end{figure}
In Fig.~\ref{fig:squared-error_vs_haystack_len}, we plot the quartiles of the squared-error of the model's predictions as a function of the number of systems in the haystack.
% if the symbolic label had been used to perform the task, we would expect predictions in the final query sequence to be unaffected by the number of systems in the haystack. Contrarily, performing a symbol agnostic Bayesian prediction gets progressively more difficult with more systems in the haystack, since there are more candidate orthogonal matrices to average over. 
Observe that the 1-after final open label performance remains steady with more systems, while the predictions for the later indices get progressively worse. Information theoretically, making a prediction for 2-after the final open label is just as easy as making a prediction for 1-after the final open label, if not easier, since the open label provides the information required to make an optimal prediction. In light of this, Fig.~\ref{fig:squared-error_vs_haystack_len} suggests that the model has not learned to use symbolic labels well enough to maintain steady performance on the indices two or more after the final open symbol as the number of systems in the haystack increases.

\FloatBarrier
\section{Multiple Mechanisms for Prediction} \label{sec:multiple}

As discussed in Section \ref{sec:contributions}, we explore whether the trained model performs the associative recall task through a \textbf{label-based recall (H1)} mechanism or a \textbf{observation-based Bayesian recall (H2)} mechanism. If the mechanism for recall was label-based, the model would in-context learn the association of symbolic labels to their corresponding systems, and then perform inference based on recalling the queried system and continuing its evolution.  If the mechanism for recall was observation-based and Bayesian, the model would ignore the symbolic labels and instead would use the state observation to figure out which system the observation could have come from. The model then performs Bayesian prediction based on previous observations to make future predictions.

%insert hypotheses here 

\subsection{Out-of-Distribution Inference-Time Experiments}\label{sec:out_of_dist_experiments}

To decide if H1 or H2 is a valid hypothesis for the associative recall mechanism, we conducted three out-of-distribution experiments at inference-time: misdirecting the model towards the incorrect sequence in the haystack (Section \ref{sec:paren_swap}), synchronizing sequences in the haystack from different systems so that they would all have the same state after the final open symbol (Section \ref{sec:ortho_sync_experiment}), and misdirecting the model towards an unseen sequence not present in the haystack (Section \ref{sec:irrelevant_symbol}). Through these three out-of-distribution experiments, we found that neither H1 nor H2 fully describe the model's mechanism for associative recall. In fact, the evidence indicates that model uses H1 for predicting 1-after the final open symbol, and a suboptimal version of H2 for predicting two or more steps after the final open symbol.

To further explore the mechanism for restarting ICL on a new system, we conduct a fourth out-of-distribution experiment: misdirecting the model towards a seen sequence in the haystack (Section \ref{sec:irrelevant_hay_experiment}). This experiment's results provide evidence that the model ignores the open symbol when restarting ICL on a new system.

% rewrite hypotheses 
% reframe as misdirection
% first sentence as takeaway
% synchronizing rotations -> ambiguity around first token severely degrades second after mechanism
%change diagrams to reflect reframing of experiments as misdirection and move to each section

\subsubsection{Experiment 1: Misdirection towards the incorrect sequence in the haystack}\label{sec:paren_swap}
\begin{figure}[tbph]
        \centering
        % \vspace{-3mm}
        \includegraphics[width=0.7\linewidth]{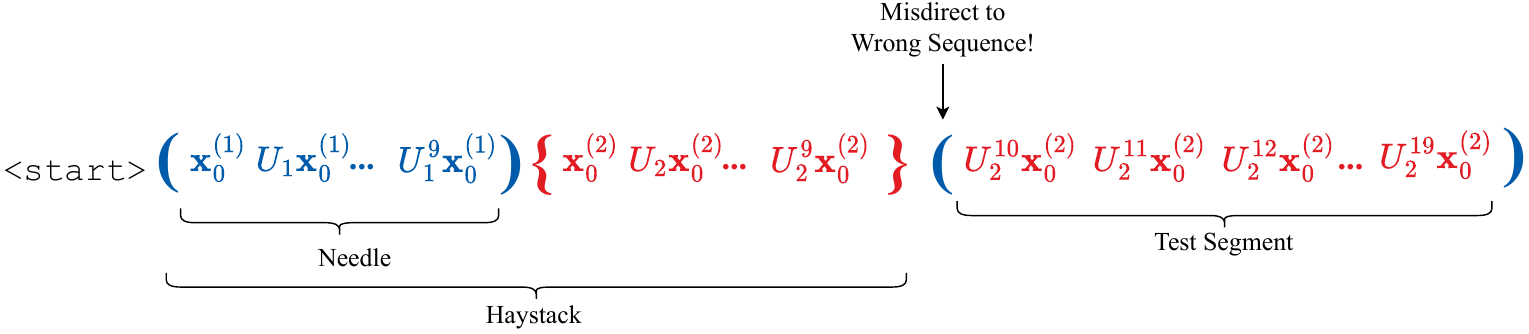}
        \caption{Misdirecting the model towards the incorrect sequence in the haystack.}
        \label{fig:paren_swap_diagram}
\end{figure}

For this out-of-distribution experiment, we test the model on ``needle-in-a-haystack'' test traces generated as specified in Section \ref{sec:needle_in_a_hay_test_setup}, except we swap the final open symbol with another open symbol that corresponds to a segment in the haystack that is not the ``needle'' as seen in Fig.~\ref{fig:paren_swap_diagram}. If H1 were true, the model would be using label-based recall and we would expect it to make predictions on the test segment for the wrong system. If H2 were true, the swapping of the label would not affect the prediction performance of the model, since the model would be using the seen states to make its predictions rather than the symbolic labels.

\begin{figure}[htbp]
    \centering
    % Subfigure 1
    \begin{subfigure}[b]{0.495\linewidth}
        \centering
        \includegraphics[width=\linewidth]{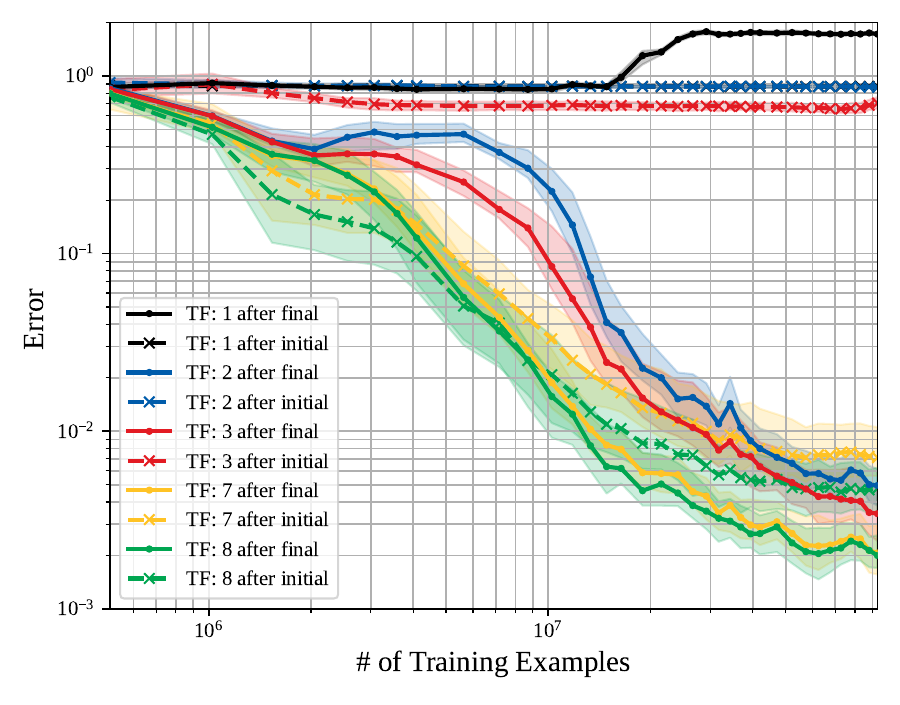}
        \caption{2 systems in the haystack.}
        \label{fig:paren_swap_len_2_med}
    \end{subfigure}
    \hfill
    % Subfigure 2
    \begin{subfigure}[b]{0.495\linewidth}
        \centering
        \includegraphics[width=\linewidth]{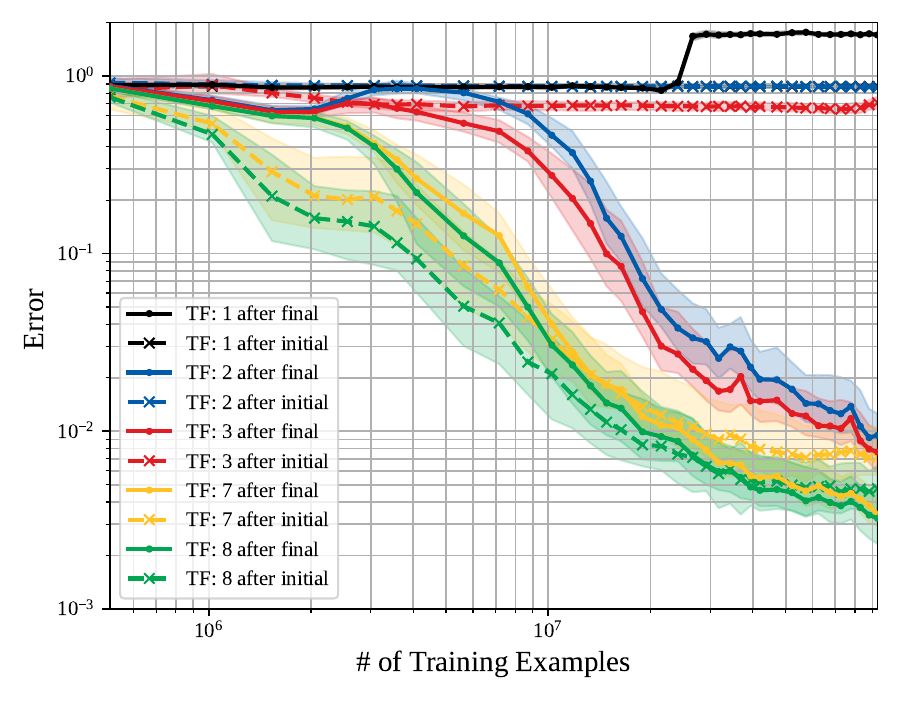}
        \caption{5 systems in the haystack.}
        \label{fig:paren_swap_len_5_med}
    \end{subfigure}

    \caption{Misdirection towards incorrect sequence --- The median squared-error of the model's predictions on the test segment after observing the incorrect final open label vs the number of training examples seen so far. The solid black curve sharply increases in these figures where it would have sharply decreased for a normal test trace as seen in Fig.~\ref{fig:combined_training_dynamics}, suggesting H1 is true for predicting the first index of the test segment. In contrast, we find that the model ignores the misdirection when predicting two or more indices into the test segment, since the solid blue, red, yellow, and green curves are almost identical to the corresponding curve in Fig.~\ref{fig:combined_training_dynamics}. This suggests that the model is using a Bayesian approach when predicting these indices.}
    \label{fig:paren_swap_experiment}
\end{figure}

\FloatBarrier

\textbf{Misdirecting the model towards the incorrect sequence in the haystack falsifies pure label-based recall and provides strong evidence that an observation-based Bayesian recall mechanism is present.} The results of this experiment are shown in Fig.~\ref{fig:paren_swap_experiment} for $N=2$ and $N=5$, where the median squared-error of the model's predictions are plotted against the number of training examples seen, mirroring the format of the Fig.~\ref{fig:combined_training_dynamics}. The first thing to notice in Fig.~\ref{fig:paren_swap_experiment} is that the solid black curve now sharply rises late in training, as opposed to its sharp fall for a normal ``needle-in-a-haystack'' test trace as shown in Fig.~\ref{fig:combined_training_dynamics}. Presumably, the model is using the label-based recall to predict the first index into the test segment. 

In contrast, the solid curves in Fig.~\ref{fig:paren_swap_experiment} for 2, 3, 7, and 8 after the final open symbol look almost identical to the corresponding solid curves in  Fig.~\ref{fig:combined_training_dynamics}. See the blue curve in Fig.~\ref{fig:paren_swap_len_2_med} and \ref{fig:ortho_med_needle_train_conv_haystack_len_2_all_haystack_len_12_log_lin_log_main_paper} at $2\times 10^7$ training examples. Their median squared-error both sit at around $2\times 10^{-2}$. The same correspondence can be seen for the red, yellow, and green curves as well. This shows that the model's predictions for 2, 3, 7, and 8 after the final open symbol are not very affected by the open symbol. Instead, they must be using the state observations after the final open symbol to decide which system to use for making predictions. This strengthens the case that the model uses an observation-based Bayesian recall mechanism for predicting the indices after the first index into the test segment.

% show that the model ignores the misdirection and leverages the observation to perform prediction. The baseline performance without misdirection for the 2+ after tasks shown in Fig.~\ref{fig:paren_swap_experiment} matches exactly what we see with misdirection in Fig.~\ref{fig:combined_training_dynamics}. This is strong evidence that the model applies a Bayesian approach for the 2+ after tasks when being shown a symbolic label it has seen in context. 

\subsubsection{Experiment 2: Synchronizing ``rotations'' in the haystack}\label{sec:ortho_sync_experiment}
\begin{figure}[tbph]
        \centering
        \includegraphics[width=0.7\linewidth]{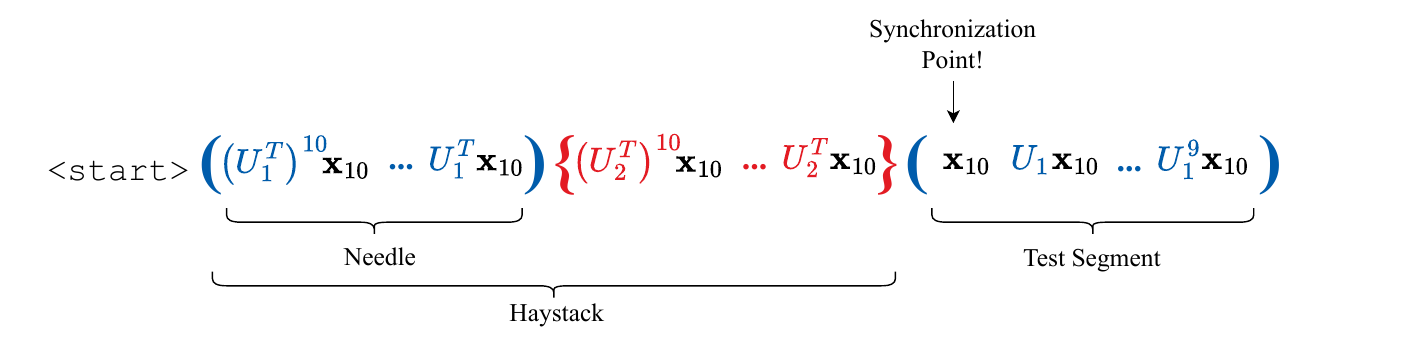}
        \caption{Synchronizing previous systems in the haystack, so the first observation in the test segment no longer reveals the system that is being continued.}
        \label{fig:ortho_sync_diagram}
\end{figure}

In Section \ref{sec:paren_swap}, the misdirection towards the incorrect sequence experiment showed that the model can make accurate predictions for indices 2 and further into the test segment without using the final open symbol. However, in that experiment the observation that is 1 index after the final open symbol can determine which system should be applied for predicting the subsequent indices, since if a predictor has observed $\xv_{9}^{(1)}$ from system 1 and $\xv_{9}^{(2)}$ from system 2, when it observes $\xv_{10}$ from either system 1 or system 2, it can check whether $\xv_{10} = U_1\xv_{9}^{(1)}$ or $\xv_{10} = U_2\xv_{9}^{(2)}$. Now, there is still the question of whether the model can use the final open symbol to predict the later indices in a situation where the 1 after final observation does not provide the necessary information. To answer this question, we conduct a synchronizing rotations experiment, where all of the sequences in the haystack from different systems all have the same state at the $10^{\text{th}}$ index, which corresponds to 1 index after the final open symbol. To do this, we first generate a single vector $x_{10} \sim \gauss{0}{\frac{1}{5}I}$ for all systems and generate the haystack by "rewinding" our systems back to their initial state $x_0$ by $x_{i-1} = U^Tx_i$ as is shown in Fig.~\ref{fig:ortho_sync_diagram}. The ambiguity of which system the first observation in the test segment comes from, means that the model must use the symbolic label to make an accurate prediction on the second index into the test segment.

\begin{figure}[htbp]
    \centering
    % Subfigure 1
    \begin{subfigure}[b]{0.495\linewidth}
        \centering
        \includegraphics[width=\linewidth]{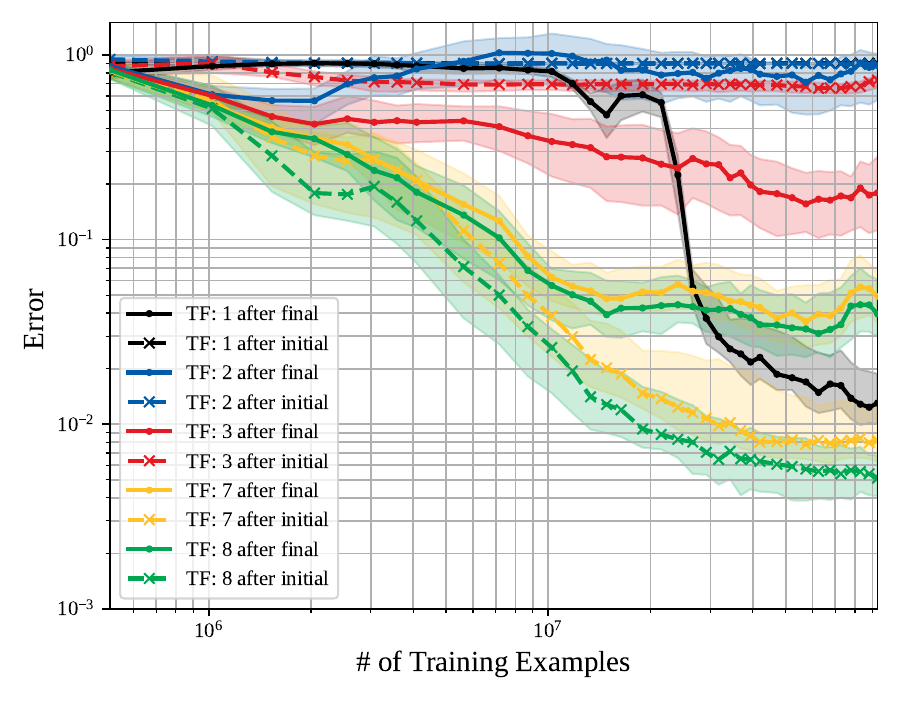}
        \caption{2 systems in the haystack.}
        \label{fig:ortho_sync_len_2_med}
    \end{subfigure}
    \hfill
    % Subfigure 2
    \begin{subfigure}[b]{0.495\linewidth}
        \centering
        \includegraphics[width=\linewidth]{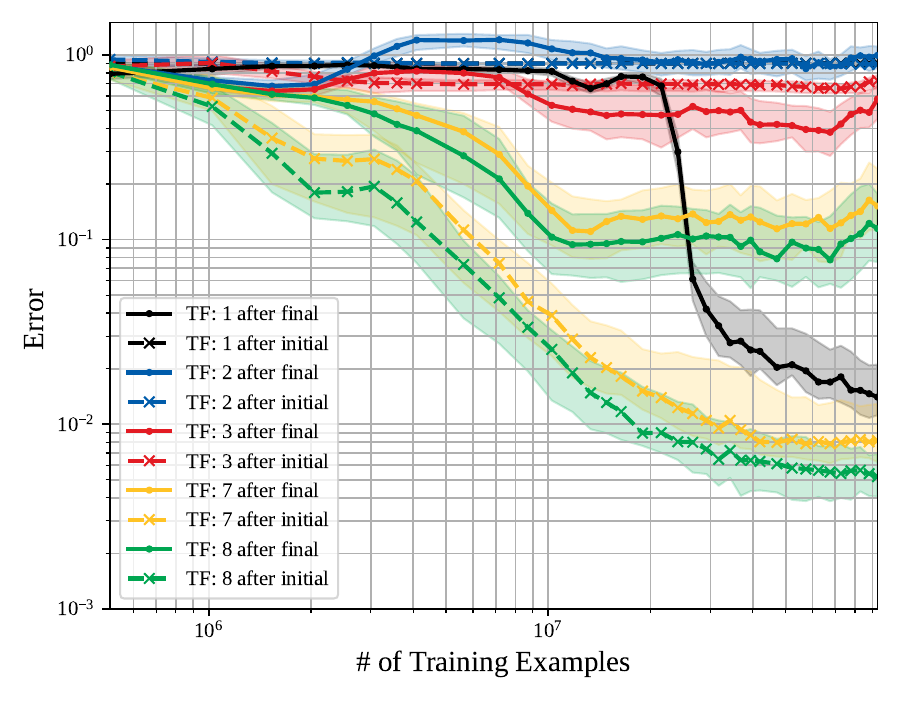}
        \caption{5 systems in the haystack.}
        \label{fig:ortho_sync_len_5_med}
    \end{subfigure}
    \caption{Synchronizing rotations --- The median squared-error of the model's predictions on the test segment when the first index into the test segment is a valid continuation for all haystack segments vs how many training examples have been processed during training. Synchronizing the rotations of the haystack segments means that after the first observation in the test segment, a predictor cannot determine which system is being continued. The solid black curve still shows emergence at the same point in training as it did in Fig.~\ref{fig:combined_training_dynamics}, supporting the validity of H1 for predicting the first index into the test segment. Otherwise, the model performance is significantly degraded for the rest of the indices into the test segment. For example, the solid blue curve in these plots for prediction errors on the second index into the test segment is near $1.0$ in Fig.~\ref{fig:ortho_sync_len_2_med} where it is near $1.0\times 10^{-2}$ in Fig.~\ref{fig:ortho_med_needle_train_conv_haystack_len_2_all_haystack_len_12_log_lin_log_main_paper}. This indicates that the model is unable to use the symbolic label well enough to make accurate predictions.}
    \label{fig:ortho_sync_experiment}
\end{figure}
\FloatBarrier
\textbf{Synchronizing the sequences in the haystack so the observation at the first index in the test segment is not informative of the continuing system shows that the model has not learned to use the symbolic label to make accurate predictions for two or more indices into the test segment.} In Fig.~\ref{fig:ortho_sync_experiment}, which plots the median squared-error of the predictions on the synchronizing test traces vs. the number of training examples seen, we see that the solid black curve still sharply decreases late in training\footnote{Due to the synchronization, the first observation in the test segment is a valid continuation for all systems in the haystack and therefore is always predictable.}, as it does in Fig.~\ref{fig:combined_training_dynamics}, showing that the model is able to use the final open symbol to predict $\xv_{10}$. On the other hand, the solid blue curves are almost horizontal throughout all of training, and the solid red, yellow, and green curves are have a significantly higher squared-error than their counterpart curves in Fig.~\ref{fig:combined_training_dynamics}. This means that the model is unable to make accurate predictions on the subsequent indices in the test segment after $\xv_{10}$, although the final open symbol provides all of the necessary information to do so. This strengthens the case for H2 being the right hypothesis for predicting the indices after $\xv_{10}$, although the model is clearly not a Bayes optimal predictor since seeing $\xv_{10}$ and the next observation $U\xv_{10}$ would disambiguate which system $U$ is being continued. Despite the system being disambiguated after seeing $U\xv_{10}$, the red solid curves in Fig.~\ref{fig:ortho_sync_experiment} are still much worse than their counterparts in Fig.~\ref{fig:combined_training_dynamics}.

% We construct an experiment where two different systems have equivalent payloads on the 1-after observation, effectively making the 2 after task ambiguous if the models leverage Bayes. To do this we generate a single payload at $x_{10} \sim \gauss{0}{\frac{1}{5}I}$ for all systems and generate the haystack by "rewinding" our systems back to $x_0$ by $x_{i-1} = U^Tx_i$ as is shown in Fig.~\ref{fig:ortho_sync_diagram}. In this case given the ambiguity of which system $x_{10}$ (the 1-after observation) corresponds to, the model must leverage the symbolic label to disambiguate the sequence. We find that not only does the model fail to do the 2-after task, \textit{but in fact the model continues to struggle even through 8-after} where it under performs the 8 after initial baseline as shown in Fig.~\ref{fig:ortho_sync_experiment}. This indicates that the model is not strictly performing Bayesian prediction either, as after seeing multiple examples following the symbolic label, it should be able to disambiguate between the systems.

\subsubsection{Experiment 3: Misdirection towards an unseen sequence}\label{sec:irrelevant_symbol}
\begin{figure}[tbph]
        \centering
        \includegraphics[width=0.7\linewidth]{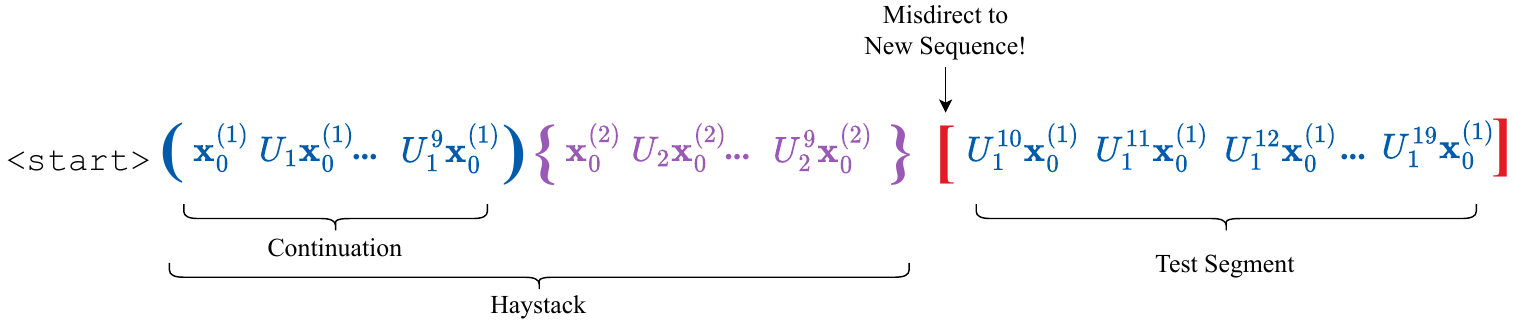}
        \caption{Misdirecting the model with an unseen symbolic label indicating a new sequence.}
        \label{fig:irrelevant_symbol_diagram}
\end{figure}

To further study whether the model uses the symbolic labels at all to continue its predictions on a recalled sequence, we misdirect the model to restart ICL for a sequence that has not appeared in the haystack so far. Fig.~\ref{fig:irrelevant_symbol_diagram} illustrates how we swap out the final open symbol of a normal ``needle-in-a-haystack'' test trace with an open symbol that does not correspond to any of the systems in the haystack. Given the results from the previous two sections, we expect the model to predict zero for the first index as that is optimal for restarting a new sequence, but we expect the model to make accurate predictions on the subsequent indices that are continuing a segment that is in the haystack. This is almost what happens, but late in training, the model suddenly transitions to (at least partially) restarting ICL on a new sequence.

\begin{figure}[htbp]
    \centering
    % Subfigure 1
    \begin{subfigure}[b]{0.495\linewidth}
        \centering
        \includegraphics[width=\linewidth]{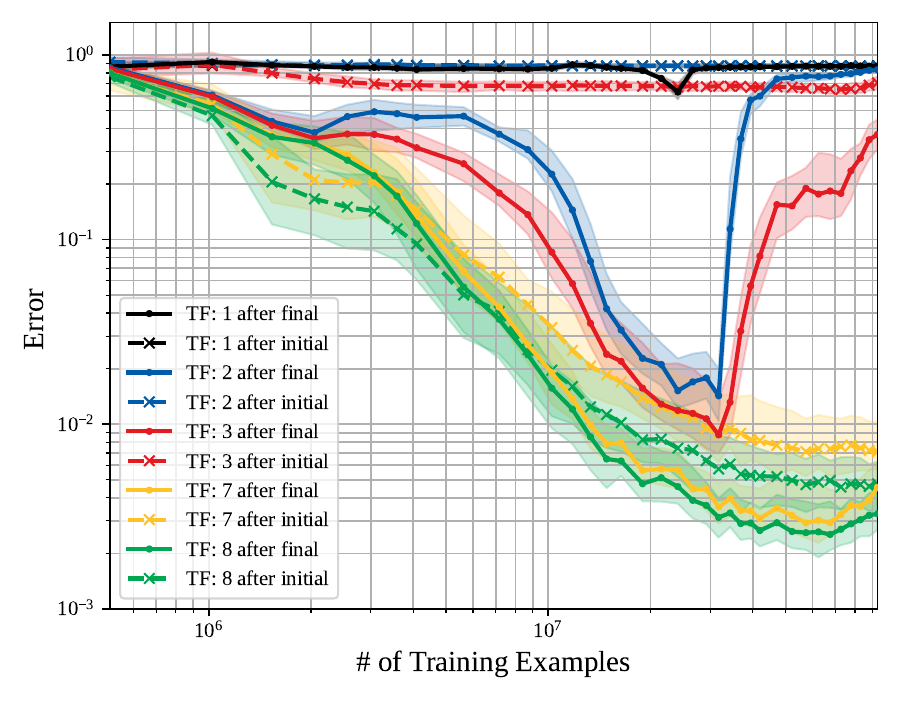}
        \caption{2 systems in the haystack.}
        \label{fig:irrelevant_symbol_len_2_med}
    \end{subfigure}
    \hfill
    % Subfigure 2
    \begin{subfigure}[b]{0.495\linewidth}
        \centering
        \includegraphics[width=\linewidth]{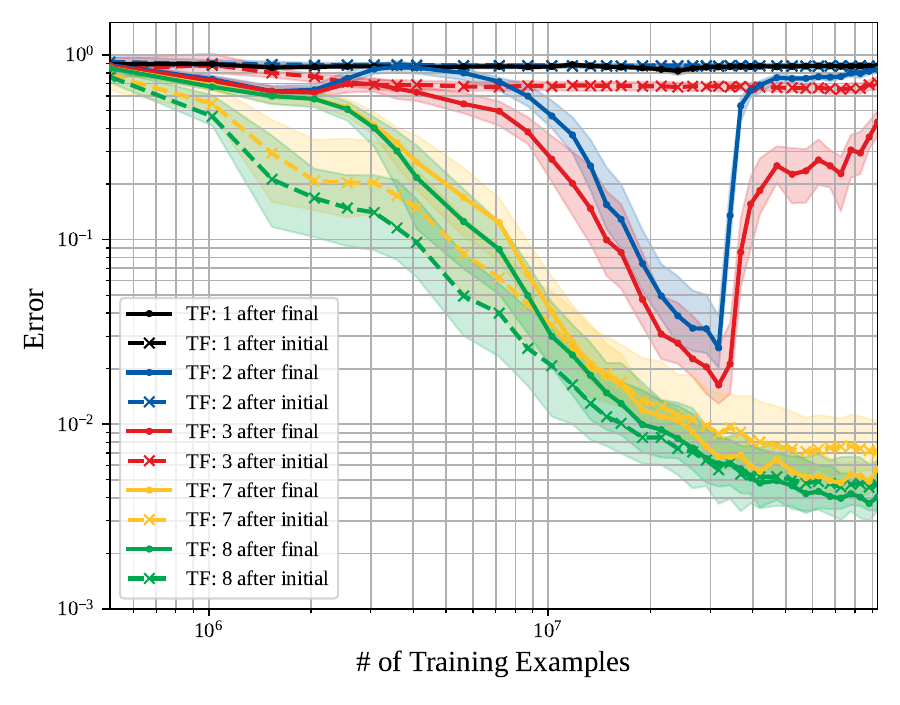}
        \caption{5 systems in the haystack.}
        \label{fig:irrelevant_symbol_len_5_med}
    \end{subfigure}
%%%%% Fix the description see Gireeja's comment %%%%%
    \caption{Misdirection towards an unseen system --- The median squared-error of the model's predictions on the test segment when the final open symbol does not correspond to any haystack systems vs the number of training examples seen so far. The solid blue and red curves match their counterparts in Fig.~\ref{fig:combined_training_dynamics} until $\approx 3\times 10^7$ training examples, at which point they sharply increase. This means the model continues predicting the existing system in early training then suddenly starts treating it more like an unseen system late in training. We note that this transition happens shortly after the emergence of associative recall, suggesting the model has learned that unseen labels also require ICL to be restarted.}
    \label{fig:irrelevant_symbol_experiment}
\end{figure}

\FloatBarrier

\textbf{Misdirecting with an unseen symbolic label highlights a new abrupt phase transition in model behavior  late in training from disregarding the symbolic label and continuing to predict the observed test segment to restarting ICL for a new system.} Fig.~\ref{fig:irrelevant_symbol_experiment} shows the median squared-error of the model predictions after being presented with a final open symbol that does not correspond to any system in the haystack while the test segment is actually a continuation of a haystack segment vs how far along the model is in training. In the same figure, the first after final predictions match the expected behavior of restarting predictions as seen by the solid black curve being a horizontal line near 1. For the rest of the indices, the model ignores the symbolic label through the initial stages of training and continues to correctly predict the test segment, since the solid curves in Fig.~\ref{fig:irrelevant_symbol_experiment} match their counterparts in Fig.~\ref{fig:combined_training_dynamics} up to around $3\times 10^7$ training examples. Once associative recall fully emerges later in training, and the model learns how to use the symbolic labels, the model transitions to treating a continuation of the old sequence as a brand new sequence corresponding to the new label, since the blue and red solid curves abruptly increase after $3\times 10^7$ training examples in Fig.~\ref{fig:irrelevant_symbol_experiment}. This shows that the model predictions for two or more indices into the test segment \textit{are} affected by the final open symbol, although Section \ref{sec:ortho_sync_experiment} showed that the model is unable to use it to make accurate predictions for recall on those indices.

\subsubsection{Experiment 4: Misdirection towards a seen sequence in the haystack}\label{sec:irrelevant_hay_experiment}
\begin{figure}[tbph]
        \centering
        \includegraphics[width=0.7\linewidth]{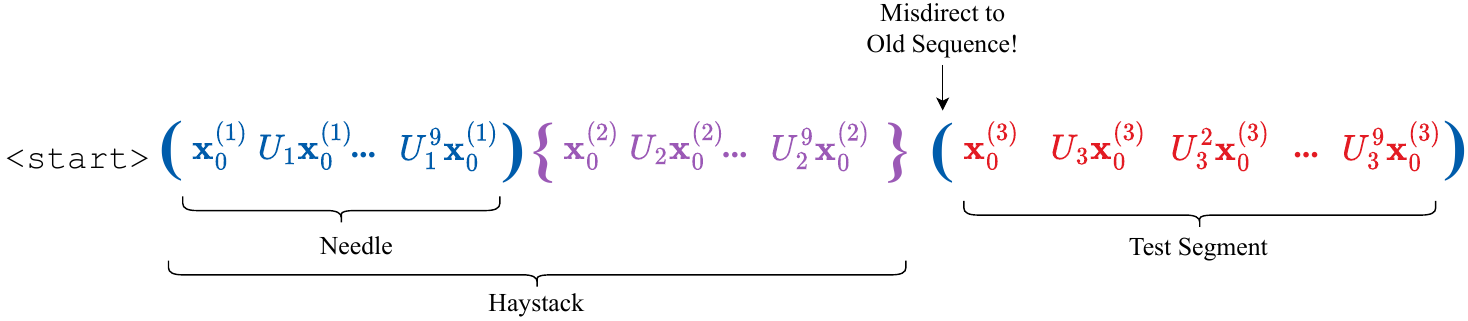}
        \caption{Misdirecting the model with a previously seen symbolic label.}
        \label{fig:irrelevant_hay_diagram}
        \vspace{-5mm}
\end{figure}

The first three out-of-distribution experiments studied the model's mechanisms for performing associative recall, but this section will study the mechanism for restarting its prediction on a new system. Again, we devise a misdirection experiment. This time we provide a final open symbol that points toward the ``needle'' in the haystack, but the test segment is a sequence from a system that is not in the haystack (Fig.~\ref{fig:irrelevant_hay_diagram}). 

\begin{figure}[htbp]
    \centering
    % Subfigure 1
    \begin{subfigure}[b]{0.495\linewidth}
        \centering
        \includegraphics[width=\linewidth]{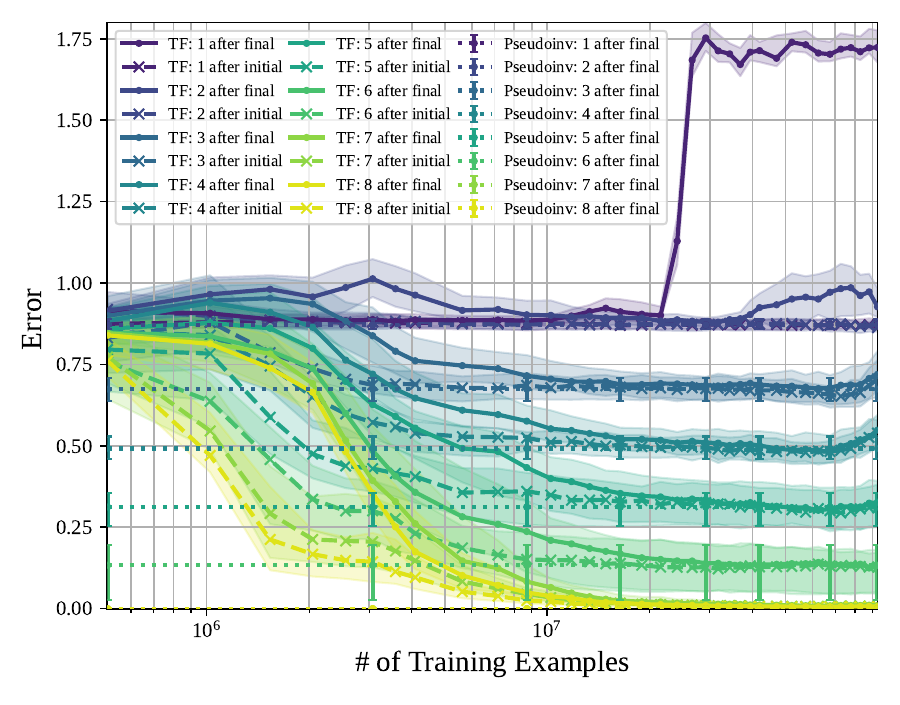}
        \caption{Linear scale.}
        \label{fig:new_hay_insert_len_2_med_lin}
    \end{subfigure}
    \hfill
    % Subfigure 2
    \begin{subfigure}[b]{0.495\linewidth}
        \centering
        \includegraphics[width=\linewidth]{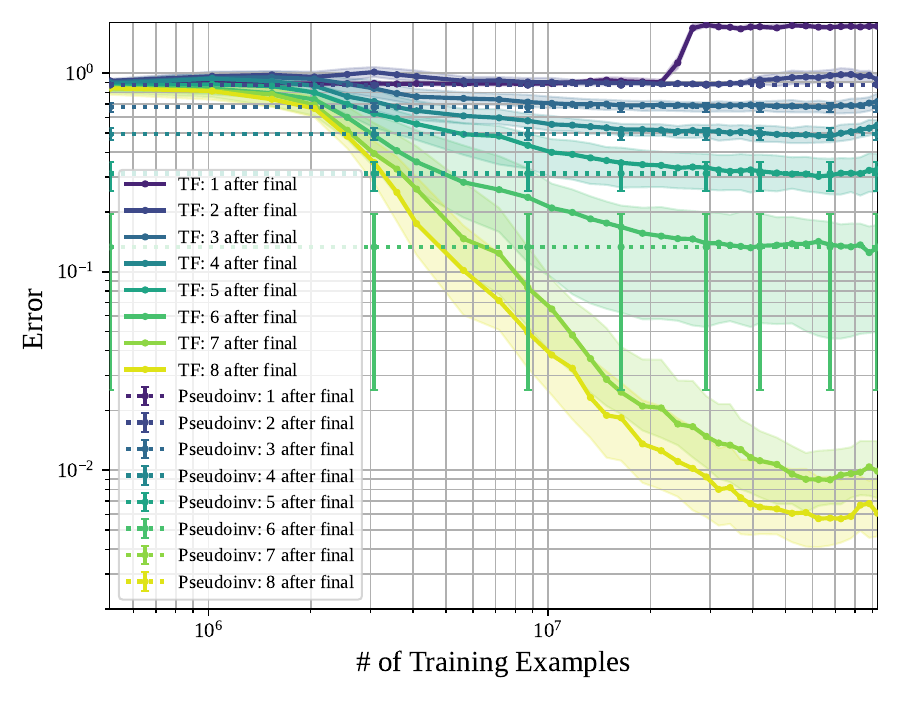}
        \caption{Log scale.}
        \label{fig:new_hay_insert_len_2_med_log}
    \end{subfigure}

    \caption{Misdirection towards a seen system --- The median squared-error of the model's predictions on the third segment that is an unseen sequence while the final open symbol corresponds to the first segment vs the number of training examples seen so far. The solid curves other than the 1-after final curve match the solid curves in Fig.~\ref{fig:restart_sys_train_conv_both}. This supports the hypothesis that the model uses the state observations to continue its prediction on a newly seen segment, rather than the open symbolic label.}
    \label{fig:new_hay_insert_experiment}
\end{figure}

    % We see the model performs Bayesian inference on the system and matches the performance when correctly prompted with a new symbolic label for a new sequence for the third position in the haystack (Fig.~\ref{fig:new_system_train_conv}). We note that at the same point in training that the model begins to leverage the symbolic token for the misdirection towards an unseen system (Fig.~\ref{fig:irrelevant_symbol_experiment}) the model starts to perform worse on the 2-after task in this setting.}

\FloatBarrier
%\paragraph{Inserting irrelevant hay as the query sequence}
\textbf{Misdirecting with a previously seen symbolic label does not stop the model from restarting its predictions on the new sequence for two or more indices into the test segment.} 
In Fig.~\ref{fig:new_hay_insert_experiment}, the median squared-error on this misdirection experiment is plotted against the number of training examples seen so far during training. We find that when given the correct unseen symbolic label (Fig.~\ref{fig:new_system_train_conv}), and when shown a symbolic label misdirecting to a sequence that has already been observed in context (Fig.~\ref{fig:new_hay_insert_experiment}), the model performs equivalently on predicting two or more indices into the newly seen segment, as the solid curves match except for the 1-after final curve that is using the symbolic open label. The model recognizes that the sequence it is seeing does not correspond to a sequence it has already seen. This supports a hypothesis that that model does not use the symbolic labels to restart ICL on a new system. Interestingly, misdirecting the model with a previously unseen symbolic label indicating a new system (Section \ref{sec:irrelevant_symbol}) shows that these unseen symbolic labels do affect the model's predictions in the test segment. This evidence shows that the mechanism for restarting ICL may not be purely label-based nor purely observation-based.

\FloatBarrier
\subsubsection{Summary of out-of-distribution experiment results}
After conducting the four out-of-distribution experiments, we find that neither H1 nor H2 can fully explain the model's mechanism for predicting interleaved time-series. The misdirection to the incorrect sequence experiment (Section \ref{sec:paren_swap}) supports H1 for predicting the first index into the test segment, but it also shows that the model can make accurately continue its predictions in the test segment without using the final open symbolic label. Therefore, H1 is insufficient. The synchronizing rotations in the haystack experiment (Section \ref{sec:ortho_sync_experiment}) further showed that the model is unable to use the final open symbolic label well enough to accurately continue its predictions in the test segment, providing strong evidence that the model performs a suboptimal version of H2 for this subtask. This evidence is further strengthened by the misdirection towards a seen sequence in Section \ref{sec:irrelevant_hay_experiment}. Finally, the misdirection to an unseen sequence experiment (Section \ref{sec:irrelevant_symbol}) showed that even for the later indices, the final open symbolic label can signal the model to restart its predictions for a new segment, invalidating H2 as the sole mechanism for continuing predictions. It is clear from these results that the conjecture C3, where the transformer model uses multiple mechanisms for the single task of interleaved time-series prediction holds true.

\FloatBarrier
\subsection{Transformer circuit analysis}\label{sec:edge_pruning}
% Using the base reference model and a continually updated pruned model, Edge Pruning optimizes over the KL divergence of the two models output distributions and computes an edge loss by enforcing a target sparsity on the edge masks using a Lagrangian term that penalizes deviation from a reference sparsity value. 

\begin{table}[tbph]
\centering
\begin{tabular}{lccc}
\toprule
\textbf{Circuit} & \textbf{\# Edges} & \textbf{1-After squared-error} & \textbf{2-After squared-error}\\ %\textcolor{red}{$\frac{\text{Full squared-error}}{\text{Pruned squared-error}}?$}
\midrule
% Identity System Full Model        & 32936 & 0.0002 & 0.00001\\
% Identity System Full Model 2-after        & 32936 & 0.00001 & 0.0002 \\
% Identity System 1-after Circuit         & 92    & 0.001 & 0.64\\
% Identity System 2-after Circuit          & 21    & 0.05 & 0.00001\\
Orthogonal Sys Full Model       & 32936 & 0.002 & 0.004\\
% Orthogonal System Full Model 2-after       & 32936 & 0.004 & 0.002\\
Orthogonal Sys 1-after Circuit        & 200   & 0.01 & 0.64\\
Orthogonal Sys 2-after Circuit        & 40    & 0.32 & 0.02\\
\bottomrule
\end{tabular}
\vspace{0.3em}
\caption{Edge pruning finds sparse transformer circuits with high evaluation accuracy in our GPT2-style model. We prune late checkpoints of a model using interleaved traces and data consisting of 5 systems in the haystack. We report the number of edges in the final circuit and the mean squared-error of the circuits' predictions for both the 1-after final and 2-after final tasks. We run an edge thresholding binary search to reach the target edge sparsity, set all pruned edges to have weights of 0, and then run inference. We visualize the 1-after final circuit in Fig.~\ref{fig:circuit}.}
\label{tab:edge_prune}
\end{table}

% \begin{wraptable}{r}{0.65\textwidth}
% \centering
% \scriptsize
% \vspace{-1em}
% \begin{tabular}{l@{\hspace{2pt}}c@{\hspace{2pt}}c@{\hspace{2pt}}c}
% \toprule
% \textbf{Circuit} & \textbf{\#Edges} & \textbf{1-after Squared Error} & \textbf{2-after Squared Error} \\
% \midrule
% Orthogonal Sys Full Model & 32936 & 0.002 & 0.004 \\
% Orthogonal Sys 1-after Circuit & 200 & 0.008 & 0.73 \\
% Orthogonal Sys 2-after Circuit & 40 & 0.35 & 0.02 \\
% \bottomrule
% \end{tabular}
% \caption{Edge pruning finds sparse transformer circuits with high evaluation accuracy in our GPT2 model. We prune late checkpoints of a model using interleaved traces and data consisting of 5 systems in the haystack. We report the number of edges in the final circuit and the squared-error of the circuits' predictions and the ground truth payloads for both the 1-after and 2-after tasks. We visualize the 1-after circuit in Fig.~\ref{fig:circuit}.}
% \label{tab:edge_prune}
% \vspace{-0.33em}
% \end{wraptable}
Now that it is clear the model uses multiple mechanisms for the single task of interleaved time-series prediction, we study if these different mechanisms have different computation graphs in the weights of the learned transformer model. To do this we use Edge Pruning, a transformer circuit-discovery method that optimizes over continuous masks over a disentangled transformer to find a sparse representation of a task \citep{bhaskar2024finding}. We run a modified version of Edge Pruning to distinguish the circuits being used by our model for the 1-after final and 2-after final tasks. As our model is solely trained with squared-error, we remove the KL objective in the original Edge Pruning method's loss function in \citep{bhaskar2024finding} and optimize on a scaled up squared-error added to the original edge loss. The loss function that we optimize to find a sparse computation graph is \( \mathcal{L}' = k \cdot \mathcal{L}_{\mathrm{squared-error}} + \mathcal{L}_{\mathrm{edge}, s} \). Our dataset for the edge pruning method consisted of one ``needle-in-a-haystack'' trace configuration generated in the same way as in Section \ref{sec:needle_in_a_hay_test_setup} for 5 systems in the haystack.\footnote{Further experiments should test more ``needle-in-a-haystack'' configurations, and would have a larger testing library of traces to create full train and test splits for training and evaluating the pruning method.} 

We report the size of the circuits and the squared-error of the predictions outputted from both the final checkpoint of the trained model and the pruned circuits in Table \ref{tab:edge_prune}.\footnote{The trained model that was used for these results is from an earlier training run than the ``orthogonal medium'' model that is throughout the rest of this paper. This earlier training run was trained on $5\times 5$ orthogonal matrices that were sampled from a non-uniform distribution over all $5\times 5$ orthogonal matrices \cite{mezzadri2006generate}.} Since the pruning method optimizes continuous gates for each node in the computation graph, these continuous gates must be quantized to 0 or 1 to get a pruned circuit.\cite{bhaskar2024finding} Our results in Table \ref{tab:edge_prune} show the accuracy of the pruned model \textit{after} this quantizing has been done, which differs from how \cite{bhaskar2024finding} reports their results. Post-quantization performance ensures that the edges that we believe to be irrelevant truly have no contribution to the output. Importantly, we find high accuracy and \textit{0\% edge overlap} between the 1-after final and 2-after final circuits, indicating that \textit{our model mostly leverages mechanistically different learned mechanisms for consecutive tokens}. See Appendix \ref{sec:edge_pruning_appendix} for further details on the pruned circuits.
\FloatBarrier
\section{Do Pretrained LLMs Also Display Multi-Mechanism Tendencies? Yes!}
\label{sec:nlp}

\begin{figure}[htbp]
    \centering
    \includegraphics[width=1\linewidth]{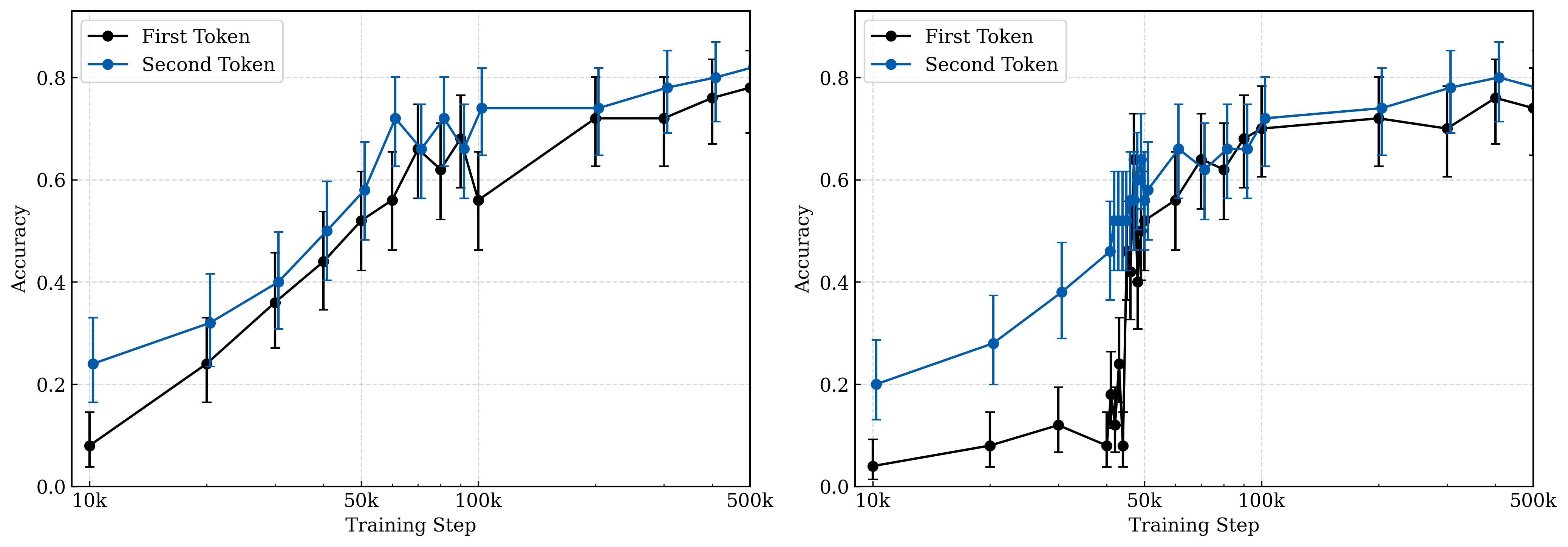}
    \caption{Comparative example of in-weights associative recall (left) and in-context associative recall (right) in a 2-shot prompting setting. Each point is a separate OLMo-2 7B model at different training steps. 
    % Similar to the toy model, the ability to apply an in-context recognized task is emergent, after which the performance of the first output token increases to the level of other output tokens. 
    We report 95\% credible intervals using Jeffreys prior (\({Beta}(0.5, 0.5))\) based on 100 samples per evaluation point.}    
    \label{fig:nlp}
\end{figure}

To see whether our conjecture C3 (multiple mechanisms for a single multi-token task) holds for natural language problems solvable by prompting LLMs, 
% We test whether the distinct learning dynamics, particularly the 1-after phase transition seen in the our toy models comes up in the training dynamics of natural language pretrained LLMs. We 
we leverage OLMo-2 7B checkpoints \citep{olmo20242olmo2furious} and a basic English to Spanish translation task that is inspired\footnote{We use Spanish instead of the International Phonetic Language (IPA) as IPA has tokens that are not compatible with the OLMo 2 tokenizer. Examples of the English to Spanish task are shown in Appendix \ref{app:nlp}. We also change the in-context labels to have no semantic meaning in light of \citep{wei2023symbol}.} by the IPA translation task used in previous works benchmarking and studying emergent behaviors \citep{wei2022emergent, srivastava2023beyond}. In Fig.~\ref{fig:nlp} (right), we see a similar phase transition in the first token prediction task, a parallel of the 1-after dynamics of the associative recall setup (Section \ref{sec:assoc_recall_results}). Meanwhile, the second-token performance is both better and more gradual in its improvement across training. This matches what we saw in our toy problem in Section \ref{sec:assoc_recall_results}.

One natural question is whether what we are seeing in Fig.~\ref{fig:nlp} is the emergence of true ICL recall or just the underlying ability to start a translation itself. This can be probed by replacing the purely symbolic task labels ``X:'' and ``Y:'' in the few-shot examples  with semantically informative ``Spanish:'' and ``English:'' labels. This replacement switches the problem from pure ICL for task recognition (\textit{in-context associative recall}) to leveraging a learned label (\textit{in-weights associative recall}). The resulting performance is seen in Fig.~\ref{fig:nlp} (left). Notice the marked improvement in the first-token performance that erases the entire gap to the second-token performance. This establishes that the model knows how to start a translation, it just can't in-context-learn well enough to know that is what it is supposed to do before the phase transition around 50k training steps.

\section{Pretraining Loss}\label{sec:pretrain_cong}

\begin{figure}[tbph]
    \centering
    \includegraphics[width=0.8\linewidth]{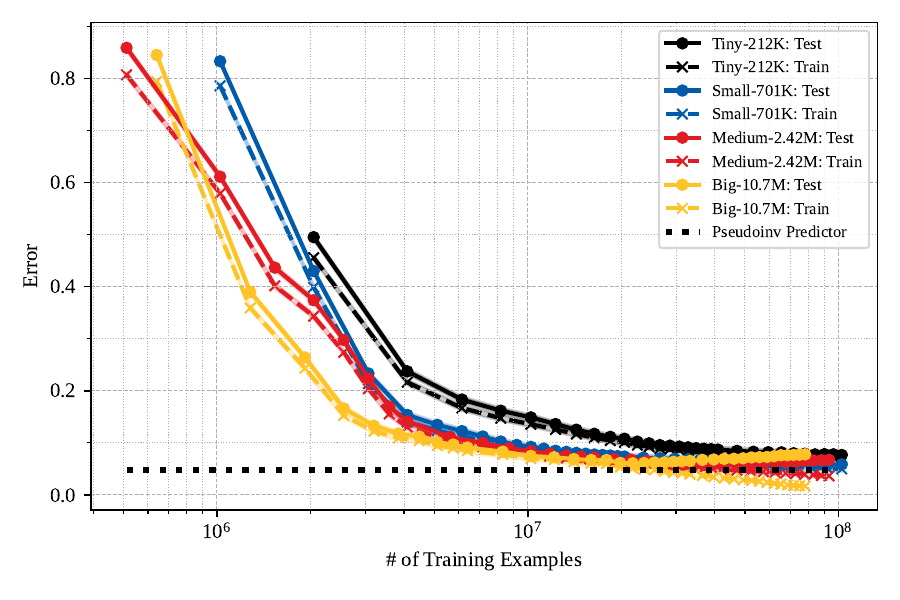}
    \caption{Pretraining loss --- The squared-error of each transformer model's predictions on traces interleaved in the style of the training data averaged over each time step of the trace. %The error bars are the standard deviation.
    For both the train and test data, averaging was done over 40,000 different interleaved traces, each of length 251. The horizontal black dotted line is the averaged squared-error of a predictor that computes an estimate of the underlying system dynamics by using the Moore-Penrose pseudoinverse of the observed data.}
    \label{fig:pretrain_loss_conglomerate}
\end{figure}
In \cite{du2025understandingemergentabilitieslanguage}, it was shown that many emergent abilities emerge for models of different sizes once a model's pretraining loss performance on a broad corpus of held-out data in the style of their pretraining data reaches a certain threshold. We want to see if this holds true in our toy setting of interleaved time-series prediction. 

We evaluate four different model sizes (see Table \ref{tab:model_size_training_params} in Appendix \ref{sec:model_size_effects} for model parameter details) and show how their pretraining loss dynamics differ throughout training. In Fig.~\ref{fig:pretrain_loss_conglomerate}, we see the performance of model training checkpoints on data that is in the style of what the model saw during training as specified in Section~\ref{sec:datagen}. The dotted curves are the models' performance on freshly drawn interleavings of traces from a held-out test library, while the crossed curves are on freshly drawn interleavings of traces from the training library. 
% The black dotted horizontal line gives a lower bound baseline for the prediction error of the model by computing the average squared-error of a predictor that computes $\widehat{\xv}_{i+1} = \widehat{U}\xv_i$, where 
% \begin{align}
%     \widehat{U} = \begin{bmatrix}\xv_1 & \dots & \xv_i \end{bmatrix}\begin{bmatrix}\xv_0 & \dots & \xv_{i-1} \end{bmatrix}^{\dagger},\label{eqn:pseudoinv_pred}
% \end{align}
% and $X^{\dagger}$ denotes the Moore-Penrose pseudoinverse of $X$. This perfect baseline also correctly unbraids the interleaved system so it knows exactly where it is in which sequence. Essentially, this baseline only makes non-zero errors on the first, second, third, fourth, fifth, and sixth entry in any sequence --- it gets everything else perfectly correct.

In Fig.~\ref{fig:pretrain_loss_conglomerate}, it is clear that larger models decrease their pretraining loss earlier in training than smaller models. Furthermore, classic overfitting behavior is evident, especially in the larger models. We see that the pretraining error on the held-out dataset deteriorates late in training, while the error on the training data continues to decrease. For the ``big'' model, its error on the training data even goes lower than the fundamental lower bound imposed by the error of the perfect pseudoinverse predictor.
\FloatBarrier

\subsection{Training dynamics with respect to the pretraining loss}

\begin{figure}[htbp]
    \centering
    \begin{subfigure}[b]{0.49\linewidth}
        \centering
        \includegraphics[width=\linewidth]{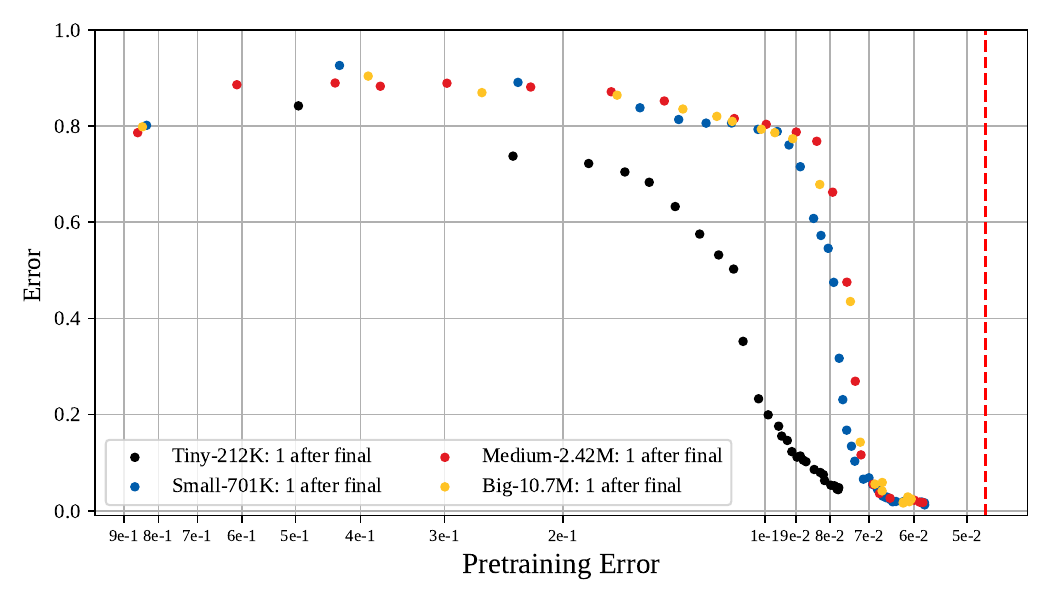}
        \caption{Recall 1 step after final --- 1 system haystack.}
        \label{fig:squared-error_vs_pretrain_loss_1_after_haystack_len_1}
    \end{subfigure}
    \hfill
    \begin{subfigure}[b]{0.49\linewidth}
        \centering
        \includegraphics[width=\linewidth]{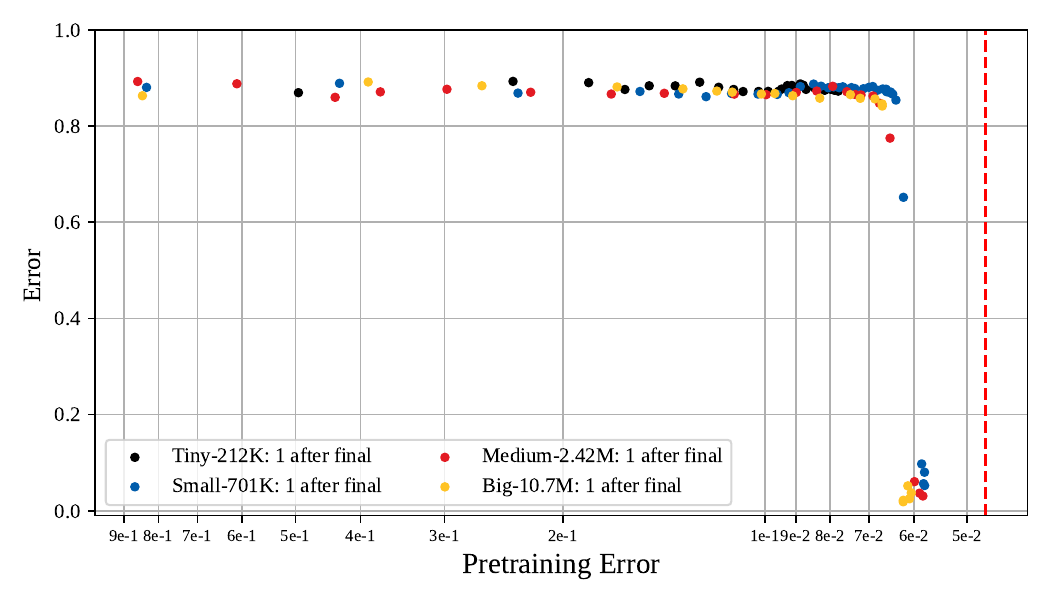}
        \caption{Recall 1 step after final --- 7 system haystack.}
        \label{fig:squared-error_vs_pretrain_loss_1_after_haystack_len_7}
    \end{subfigure}
    \begin{subfigure}[b]{0.49\linewidth}
        \centering
        \includegraphics[width=\linewidth]{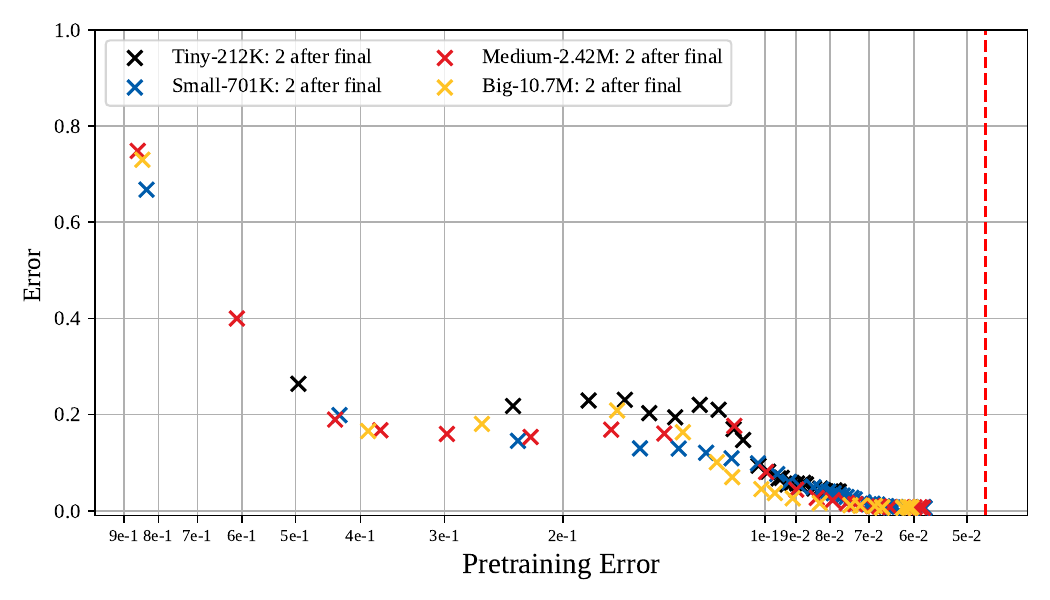}
        \caption{Recall 2 steps after final --- 1 system haystack.}
        \label{fig:squared-error_vs_pretrain_loss_2_after_haystack_len_1}
    \end{subfigure}
    \hfill
    \begin{subfigure}[b]{0.49\linewidth}
        \centering
        \includegraphics[width=\linewidth]{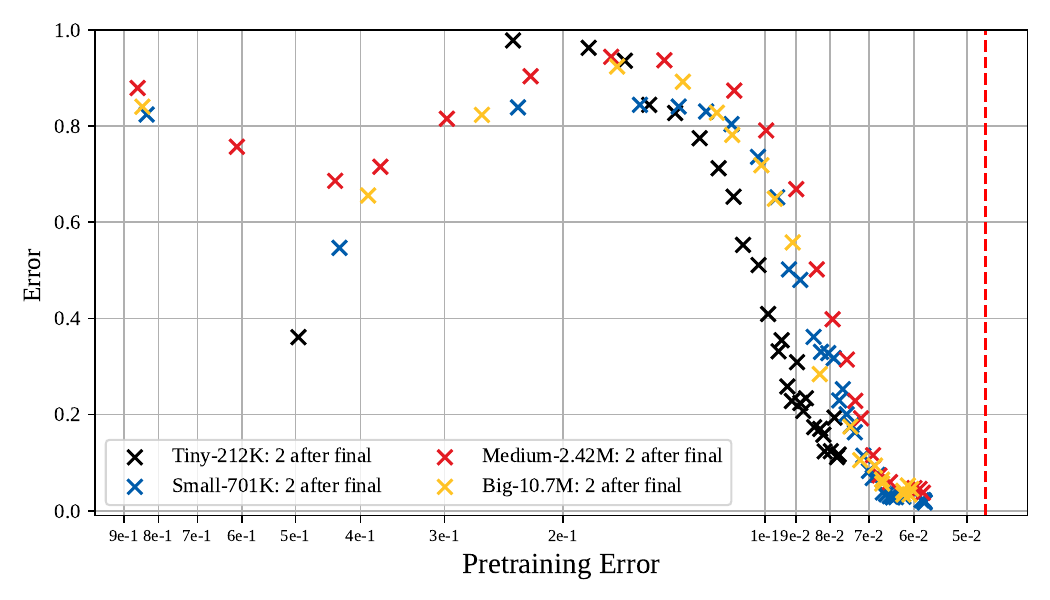}
        \caption{Recall 2 steps after final --- 7 system haystack.}
        \label{fig:squared-error_vs_pretrain_loss_2_after_haystack_len_7}
    \end{subfigure}
    \begin{subfigure}[b]{0.49\linewidth}
        \centering
        \includegraphics[width=\linewidth]{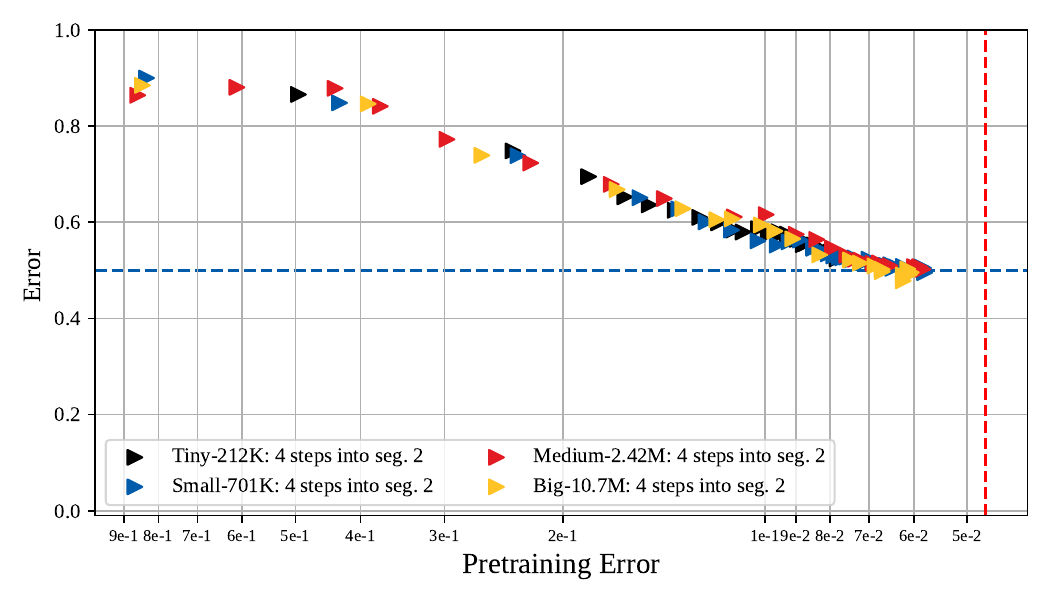}
        \caption{Restart 4 steps into segment 2.}
        \label{fig:squared-error_vs_pretrain_loss_4_init_haystack_len_1}
    \end{subfigure}
    \hfill
    \begin{subfigure}[b]{0.49\linewidth}
        \centering
        \includegraphics[width=\linewidth]{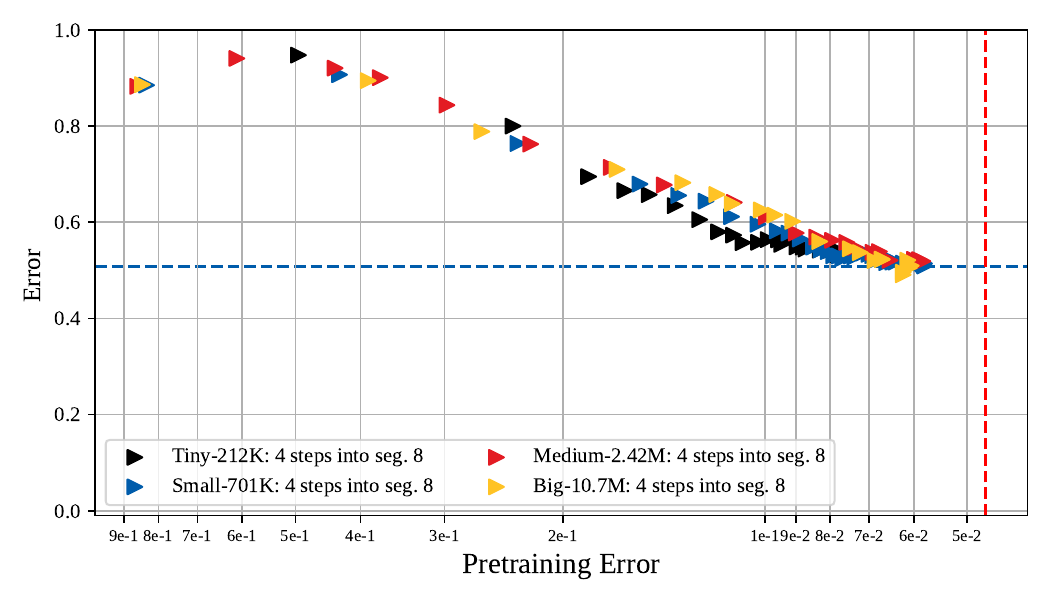}
        \caption{Restart 4 steps into segment 8.}
        \label{fig:squared-error_vs_pretrain_loss_4_init_haystack_len_7}
    \end{subfigure}
    \caption{Recall and restart performance vs pretraining loss on held-out data. The red vertical line is the fundamental lower bound pretraining loss achieved by the pseudoinverse predictor. The blue horizontal line in Figs.~\ref{fig:squared-error_vs_pretrain_loss_4_init_haystack_len_1} and \ref{fig:squared-error_vs_pretrain_loss_4_init_haystack_len_7} is the median error of the pseudoinverse predictor at the specific index plotted in each figure.}
    \label{fig:squared-error_vs_pretrain_loss}

\end{figure}

Taking inspiration from \cite{du2025understandingemergentabilitieslanguage}, in Fig.~\ref{fig:squared-error_vs_pretrain_loss} we look at when the ability to resume a recalled sequence, the ability to continue predicting a recalled sequence, and the ability to restart ICL on a previously unseen system emerge with respect to the pretraining loss as opposed to the number of training examples. This is to see if these abilities emerge at a specific pretraining loss threshold that is independent of the model size. This pretraining loss corresponds to the held-out loss specified in Section~\ref{sec:pretrain_cong}. 
% The dots represent the squared-error for predicting 1 index after the final query open token, the x's correspond to the squared-error for predicting 2 indices after the final query open token, and the triangles represent the squared-error for predicting 4 indices into the 2nd initial segment in Fig.~\ref{fig:squared-error_vs_pretrain_loss_haystack_len_1} and 4 steps into the 8th initial segment in Fig.~\ref{fig:squared-error_vs_pretrain_loss_haystack_len_7}. 
The red vertical line in the plots is the pretraining loss achieved by the pseudoinverse predictor specified in \eqref{eqn:pseudoinv_pred} averaged over all context indices. The blue horizontal line in Figs.~\ref{fig:squared-error_vs_pretrain_loss_4_init_haystack_len_1} and \ref{fig:squared-error_vs_pretrain_loss_4_init_haystack_len_7} is the median error of the pseudoinverse predictor at the specific index corresponding to the curve plotted.

Although figures in Appendices \ref{sec:when_emerge} and \ref{sec:more_train_dyn} show that larger models develop these emergent abilities before smaller models with respect to the number of training examples, in Fig.~\ref{fig:squared-error_vs_pretrain_loss_1_after_haystack_len_7} we see that all the models exhibit the ``1-after final'' phase transition at a pretraining loss of around $6.5 \times 10^{-2}$. Furthermore, in Fig.~\ref{fig:squared-error_vs_pretrain_loss_2_after_haystack_len_7}, the ``2-after final'' phase transition occurs for all models begin at a pretraining loss of a bit more than $1 \times 10^{-1}$. In Fig.~\ref{fig:squared-error_vs_pretrain_loss_1_after_haystack_len_1} the different models are again tightly linked, except for the ``tiny'' model whose emergence for ``1-after final'' occurs at a pretraining loss that is around $0.02$ larger than the rest. Overall, these results seems to qualitatively match the conclusions in \cite{du2025understandingemergentabilitieslanguage} that pretraining loss is a good indicator for when emergent abilities emerge.
% \section{Memorization of training traces}

% After training our models out sufficiently, we find that 

%% linear plots for 1,2,19 for normal overfitting and off by 1. Model clearly is memorizing but relies on seeing the first thing. Point back to synchronizing experiment where this indicates that getting the first thing wrong really messes you up
\FloatBarrier

\section{Discussion}

At this point, the community understands that ICL is rich and nuanced \cite{lin2024dualoperatingmodesincontext, wang2024investigatingpretrainingdynamicsincontext, min2022rethinkingroledemonstrationsmakes, park2025competitiondynamicsshapealgorithmic, lampinen2024broaderspectrumincontextlearning}. 
We contribute a new dimension of nuance by empirically pointing out that \textit{a single ICL-driven task can be performed, on different tokens, using multiple mechanisms that emerge separately}. 
When tasks have tight local coherency, there can often be approximate local underspecification --- there are multiple ways of knowing what the model is supposed to be doing here. 
The intrinsic Bayesian orientation of autoregressive next-token prediction \cite{xie2021explanation} means that this local underspecification can get picked up on, while the average-loss-over-tokens driving the training gradients means that an approximately correct mechanism that works most of the time will be rewarded and improved even if it can't solve the task completely. 
The very success of this approximately correct mechanism during training will further reduce the overall gradient pressure \cite{pezeshki2021gradient} for alternative and potentially better mechanisms, potentially forcing them to develop more slowly \cite{shah2020pitfalls}. 
However, structurally, there are certain aspects of tasks that are likely to require the use of the better mechanisms --- and it seems that starting an episode of a task might be one of them. 
These better mechanisms therefore can emerge later in training --- but their emergence does not mean that the better information they are acting on will automatically be incorporated by the mechanisms already favored for other parts of the task. 
This contrasts with situations where the earlier emerging mechanism foregrounds information that can be used by the later mechanism \cite{singh2025strategycoopetitionexplainsemergence}.

\section*{Acknowledgments}

This work was supported in part by the NSF CAREER grant ECCS-2240031.

\bibliography{neurips_2025}
\bibliographystyle{plain}

\newpage
\appendix

\section{Extended Related Work}
\label{app:related}

\paragraph{Emergence} 
%\subsubsection{Emergence}
%Since deep learning benefits from ever larger models trained with ever more data and compute, the research community has asked how capabilities emerge with scale.
Benchmark performance of large language models (LLMs) has been observed to improve abruptly at certain scales in a seemingly unpredictable manner~\citep{wei2022emergent, ganguli2022predictability}, leading to the conclusion that LLMs exhibit \emph{emergent} abilities, where different abilities may emerge at different scales or points during training. %  emergence. 
% How do capabilities in these deep networks \emph{emerge} with scale? %data and compute scale? 
% Understanding the emergence of capabilities with scale in deep networks 
%
% This discussion around emergence was largely initiated in the context of large language models by. %where performance on many benchmarks seems to jump/breakthrough quite abruptly at certain scales in a seemingly unpredictable manner. 
%Healthy skepticism emerged almost immediately:
A natural question was posed by \cite{schaeffer2024emergent}: is emergence a mirage due to the discrete nature of the evaluation metrics used? Would alternative continuous metrics (e.g. logits instead of generated responses) show continuous improvement and thus lead to more predictable performance? Or are these phase transitions real? At this point, the reality of phase-transitions in abilities during training is well established, with arguably the strongest evidence of this coming from the ``grokking'' literature \cite{power2022grokking,nanda2023progress, gromov2023grokking,zhong2024clock,mallinar2024emergence,humayun2024deep, mohamadi2023grokking, soudry2018implicit, liu2022omnigrok, prieto2025grokking, pmlr-v162-pezeshki22a, davies2023unifying} where there is a clear emergence of substantially improved generalization performance after what looks like a long period without improvement during training. Similar behavior has been observed in stylized regression-style examples that are very different from LLMs and ICL 
 \cite{lyu2023dichotomy, kumargrokking, nam2024exactlysolvablemodelemergence}. 
The toy problem in our paper is both regression-style and very much related to ICL and LLMs.

What exactly drives emergence is also an ongoing topic of investigation. While earlier work talked about model sizes and total compute, the story now is more nuanced. Powerful evidence connects the emergence of abilities to the pretraining losses attained \cite{du2025understandingemergentabilitieslanguage}, other information-oriented metrics \cite{chen2024quantifyingsemanticemergencelanguage}, and the idea that more complex or specialized abilities can only emerge after a model acquires prerequisite abilities during training \cite{chen2024sudden, lubana2024percolationmodelemergenceanalyzing}.

\paragraph{In-context learning }
% Since GPT3 \citep{brown2020language}, in-context learning has been understood as a key ability for pretrained transformers — and more recent work like \cite{lu2023emergent} deepens the case for its centrality by linking it to other seemingly emergent abilities as well. 

Our understanding of in-context learning (ICL) has greatly benefited from experiments in simple domains, with many works studying ICL through linear dynamics \citep{garg2022can, huang2025task, wu2024pretrainingtasksneededincontext}. To add a proxy for the order-dependence of language, works have used sequences from  Markov chains to study pretraining, revealing emergent behaviors related to the formation of induction heads \citep{edelman2024evolution, pmlr-v235-sander24a, DuOymakTrans, li2023transformers}. 

Recent works have analyzed the role of labels in ICL, giving additional perspective into information flow and the inherent limitations of ICL \citep{min2022rethinkingroledemonstrationsmakes, wang2023labelwordsanchorsinformation, huang2025task}. This is closely linked to the literature on in-weights learning (IWL) \citep{chan2025understandingincontextvsinweight, anand2025dualprocesslearningcontrolling}, which reveals structural contrasts in the learning mechanisms models can leverage.

Mechanistic explorations of ICL have advanced through both top-down classification of higher-order attention head behaviors \citep{elhage2021mathematical, olsson2022context, yin2025attentionheadsmatterincontext} and bottom-up analyses tracing the dynamics of these features \citep{singh2024needsrightinductionhead}. From analyzing the learning dynamics of ICL in various setups \citep{mainali2025exactlearningdynamicsincontext,zhang2025trainingdynamicsincontextlearning}, works have found that ICL enables a mixture of emergent behaviors \citep{reddy2023mechanisticbasisdatadependence, chan2022datadistributionalpropertiesdrive, nguyen2024differentiallearningkineticsgovern, lu2023emergent} and competes with other learning mechanisms \citep{park2025competitiondynamicsshapealgorithmic}. Contemporaneous work has also explored the transient dynamics of ICL \citep{singh2025strategycoopetitionexplainsemergence}, finding that a form of IWL both cooperates and competes with ICL in attention-only models. 

With learning dynamics displaying various surprising phenomena in different setups, works have also tackled discerning whether there are distinct modes of ICL (e.g. task learning (TL) vs. task recognition (TR)\footnote{From \cite{wang2024investigatingpretrainingdynamicsincontext}: "TR refers to the ability of an LLM to recognize the target task from demonstrations and only utilize its own knowledge obtained from pretraining to solve the task, while TL refers to the ability of an LLM to solve the target task solely based on demonstrations." The NLP example we choose in Section \ref{sec:nlp} of this paper falls into the "Task Recognition" subtype.}) through various lenses like ICL risk (premature convergence), competition dynamics, and differing data distributions \citep{pan2023incontextlearninglearnsincontext,wang2024investigatingpretrainingdynamicsincontext, lin2024dualoperatingmodesincontext, wies2023learnabilityincontextlearning}. Other work has argued for broader definitions than the "dual mode hypothesis", pushing for ICL to be understood as a spectrum of behaviors or mixture of algorithms \citep{lampinen2024broaderspectrumincontextlearning, park2025competitiondynamicsshapealgorithmic}. In a sense, our findings here add another nuance to this story since we provide evidence that different modes of ICL can be active at the same time for the same task, just on different tokens, and that these modes can emerge separately during training.

% We need to figure out how to incorporate these works into our analysis section. One way to say is: "Crucially, we display that multiple modes of ICL are intertwined during consecutive token predictions in what could seem as the same task." I think making the claim that it is "two" like the TR vs TL work is too strong a claim because our setups do not clearly align with those definitions it seems

\paragraph{Associative recall }
With the resurgence of state-space models, \cite{arora2023zoology} highlighted associative recall \citep{hopfield1982neural, zhang2017learning} as the key capability where attention-based models hold a distinct advantage. This was formalized by
introducing a discrete toy problem they called Multi-Query Associative Recall (MQAR). However, in basic MQAR, the solution involves the exact copying of the right token from context rather than recalling something conceptual that has to be applied locally. Meanwhile, in continuous Gauss-Markov models, next-observation prediction can still be performed effectively using state-space models, as least-squares algorithms can be adapted into a streaming form via standard Woodbury-matrix tricks. By using MQAR-style discrete symbolic labels to segment time-series drawn from Gauss-Markov models, we effectively create a naturally continuous toy problem where the information to be recalled is not an exact copying. 

% \todo{include paragraph that cites \cite{nichani2024understandingfactualrecalltransformers} and explains that associative recall should not be confused with associative memories (models that store the associations between certain input and output sets) and factual recall where models are prompted to recall atomic facts from their training data e.g.``UC Berkeley is a public institution.''}

% \textcolor{red}{From the HiLD reviewer: ``The paper is missing references to related work on using synthetic setups to study different aspects of ICL in transformers. For example, \cite{nichani2024understandingfactualrecalltransformers}.''}

% \textcolor{red}{\cite{nichani2024understandingfactualrecalltransformers} studies the capability of one-layer transformers to memorize facts in their weights using a toy task. The toy task creates a subject token sets, a relation token set, a noise token set, and an answer token set. The answers are mapped to a pair of subject and relation tokens. When there are many more subject tokens than relation tokens, it is shown that transformer models start training with random guessing, then predict based on the relation token while ignoring the subject token, and then finally learn the mapping from the subject-relation token pair to the answer token. This paper also shows that single layer transformers can memorize $N$ facts as long as the number of parameters scales with $N\mathrm{poly}\log{N}$.}

% \textcolor{red}{the first MOSS reviewer suggested that we add \cite{park2025competitiondynamicsshapealgorithmic} but we already cite it.}

\paragraph{Are deep neural networks Bayesian predictors?}

The Bayesian perspective on machine-learning generally is longstanding and consequently has been used to better understand the ICL capabilities of deep neural networks. \cite{xie2021explanation} frames in-context learning as implicit Bayesian inference. Using a simple, synthetic task setup, this work showed theoretically that if a model makes its predictions according to its pretraining distribution, it will asymptotically approach the performance of the Bayes optimal predictor as the number of in-context examples for this task increase to infinity. \cite{panwar2024incontextlearningbayesianprism} builds off of these ideas, and empirically investigates how close a transformer model's predictions are to the Bayes optimal predictor. They find that high-capacity transformers successfully perform Bayes-optimal prediction for Gaussian mixtures and other more complex mixtures. Interestingly, they hypothesize that low-capacity transformers attempt to approximate Bayesian inference through gradient descent. The facts that gradient descent reaches the global optimum for convex problems and \cite{von2023transformers} shows that multiple self-attention layers in a model can simulate multiple steps of gradient descent lead the authors in \cite{panwar2024incontextlearningbayesianprism} to believe that higher-capacity transformers perform Bayesian inference because they can take more steps of gradient descent.
% A distinguishability condition is when ICL occurs. (KL divergence between prompt task distribution and distribution for other distinct tasks seen in pretraining is sufficiently large. If a model predicts according to the pretraining distribution its predictions approach the Bayes optimal in 0-1 loss asymptotically in in-context examples. When distinguishability doesn't hold, KL between the test and train task is small and Bayes risk still decreases with in-context examples.)

Many works have shown that gradient descent can approximate a Bayes-optimal predictor. For example, \cite{MandtStephan2017SGDa} uses stochastic gradient descent with a constant learning rate to sample from an approximate posterior, by connecting stochastic optimization with the stochastic gradients used in Markov chain Monte Carlo. Additionally, \cite{mingardSGDBayesian} uses Gaussian processes to estimate a posterior and empirically computes the posterior probability of a deep neural network trained with stochastic gradient descent. Through comparing the two posterior distributions, they conclude that the Bayes posterior is a first-order correlate with the neural network's, while second-order differences can come from hyperparameter choices when training the model. 

Relating this work to the emergence that we observe for transformer models on our in-context recall task, singular learning theory predicts that Bayesian learners have phase transitions in their generalization error \cite{watanabe2009algebraicgeo}. Grounding their empirical investigation in this theory, \cite{hoogland2025losslandscapedegeneracydrives,chen2023dynamicalversusbayesianphase} study the loss landscape of transformer and autoencoder models, and are able to predict the occurrence of a phase transition by computing the local learning coefficient \cite{lau2024locallearningcoefficientsingularityaware}.

\newpage
\section{Edge Pruning Circuit Visualization}\label{sec:edge_pruning_appendix}
\begin{figure}[tbph]
    \centering
    \includegraphics[width=0.4\linewidth]{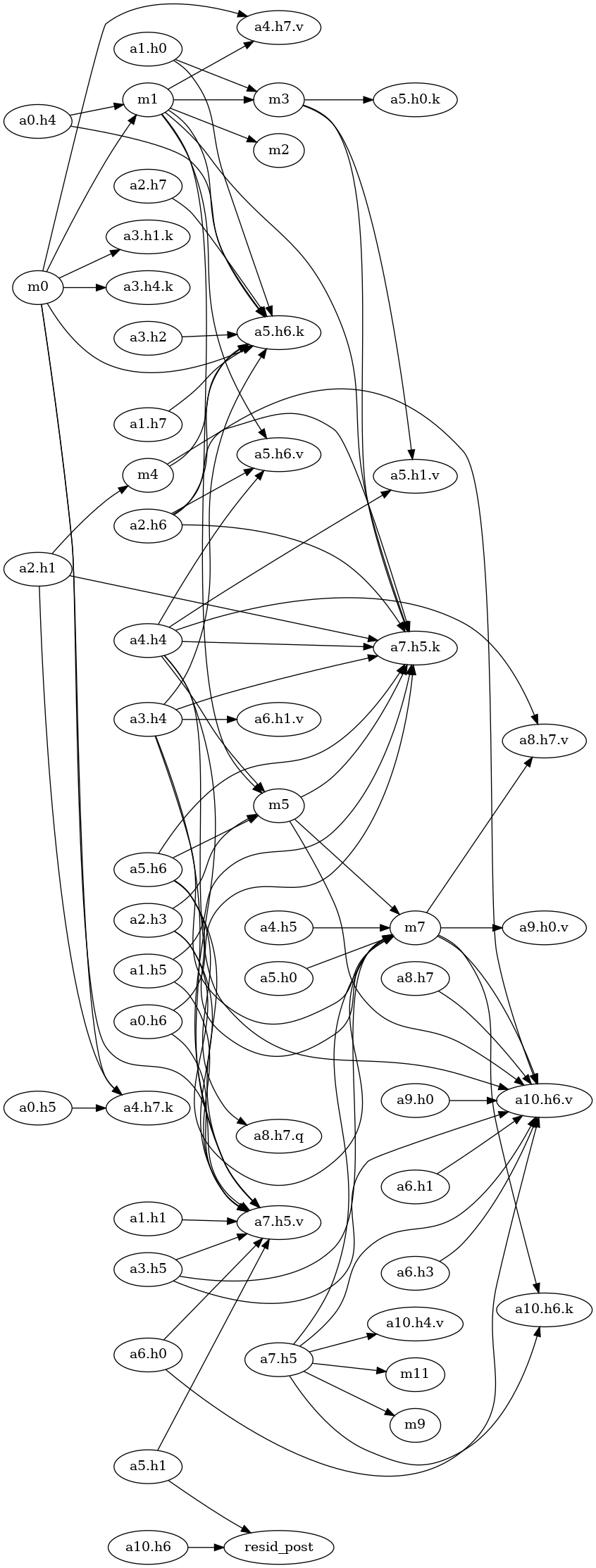}
    \caption{1-after final open symbol circuit in the orthogonal model. The output of the residual stream is labeled as \texttt{resid\_post}, MLPs are labeled as \texttt{m\{layer\_num\}}, attention heads are labeled as \texttt{a\{layer\_num\}.h\{head\_num\}}, and \texttt{.q}, \texttt{.k}, and \texttt{.v} stand for QKV matrices respectively.} 
    \label{fig:circuit}
\end{figure}

Section \ref{sec:edge_pruning} describes how optimizing over continuous masks on disentangled transformer nodes leads to a sparse computation graph after the continuous masks are quantized to 0 or 1. To perform this quantization, the quantization threshold must be set. We use binary search to find the smallest threshold (within 1e-5 precision) such that the pruned model’s edge sparsity is close to the target edge sparsity of 0.98. Importantly, this does not force a viable path from the start and end of the residual stream, it only captures significant contributions over edges with respect to the loss. In Fig.~\ref{fig:circuit}, the pruned circuit is visually presented in a directed computation graph. Nodes in this computation graph are specific QKV matrices of attention heads, entire attention heads, MLPs at a specific layer, or the output of the residual stream. The output of the residual stream is labeled as \texttt{resid\_post}, MLPs are labeled as \texttt{m\{layer\_num\}}, attention heads are labeled as \texttt{a\{layer\_num\}.h\{head\_num\}}, and \texttt{.q}, \texttt{.k}, and \texttt{.v} stand for QKV matrices respectively. 

% \newpage
\FloatBarrier
\section{NLP Prompts}
\label{app:nlp}

In this section, we provide sample prompts for the NLP experiments from Section \ref{sec:nlp}. We use 100 English prompts from the IPA transliterate benchmark task in BIG-bench \cite{srivastava2023beyond}, and pass these English prompts through Google Translate to obtain their Spanish translations.

Below are examples of three different task instances for English to Spanish for the in-weights associative recall task.

\texttt{"input": Spanish: Inició el ascenso del Imperio Británico en la India. English:} \\
\texttt{"target": It started the rise of the British Empire in India.} \\
\texttt{"input": Spanish: Hicieron acusaciones sobre la plataforma. English: } \\
\texttt{"target": They made accusations about the platform.} \\
\texttt{"input": Spanish: Che fue otro dictador que ascendió al poder mediante un levantamiento militar. English: } \\ 
\texttt{"target": Che was another dictator who rose to power by military uprising.}\\

Below are examples of three different task instances for English to Spanish for the in-context associative recall task. Notice that they are the same as the in-weights associative recall examples, except the semantically meaningful labels ``English'' and ``Spanish'' are swapped for the labels ``X'' and ``Y''.

\texttt{"input": X: Inició el ascenso del Imperio Británico en la India. Y:} \\
\texttt{"target": It started the rise of the British Empire in India.} \\
\texttt{"input": X: Hicieron acusaciones sobre la plataforma. Y: } \\
\texttt{"target": They made accusations about the platform.} \\
\texttt{"input": X: Che fue otro dictador que ascendió al poder mediante un levantamiento militar. Y: } \\ 
\texttt{"target": Che was another dictator who rose to power by military uprising.}\\

\section{Additional "Out-of-distribution" Experiment in OLMo}
%In this section, we explore a second  analogy of the "out-of-distribution" setups from our analysis on the toy models into the OLMo model to see if pretrained LLMs have clear regimes of label-based recall. 

In this section we explore a counterpart to the misdirecting-to-an-unseen-token task (Fig.~\ref{fig:irrelevant_symbol_diagram} and results shown in Fig.~\ref{fig:irrelevant_symbol_experiment}) using training checkpoints of the OLMo 2 model.
Here, we consider the task where the symbolic label that is used for directing the associative recall (e.g. the open SPL in the Markovian problem) is replaced with a different label that is so far unseen.
%where the label to direct associative recall is unseen in the "haystack". 
To adapt this task to a natural language setup, we keep our translation task with two few-shot examples exactly the same as in Section \ref{app:nlp}, but alter the final testing prompt to not use the `Y:' label tokens and to instead use a previously unseen `Z:' token. 

As such, at inference time, the first token prompt functionally looks like this. \\\\
\texttt{X: Inició el ascenso del Imperio Británico en la India.} \\ \texttt{Y: It started the rise of the British Empire in India.} \\
\texttt{X: Hicieron acusaciones sobre la plataforma.} \\
\texttt{Y: They made accusations about the platform.} \\
{\ttfamily X: Che fue otro dictador que ascendió al poder mediante un levantamiento militar.} \\
\texttt{Z:}

\begin{figure}[htbp]
    \centering
    \includegraphics[width=1\linewidth]{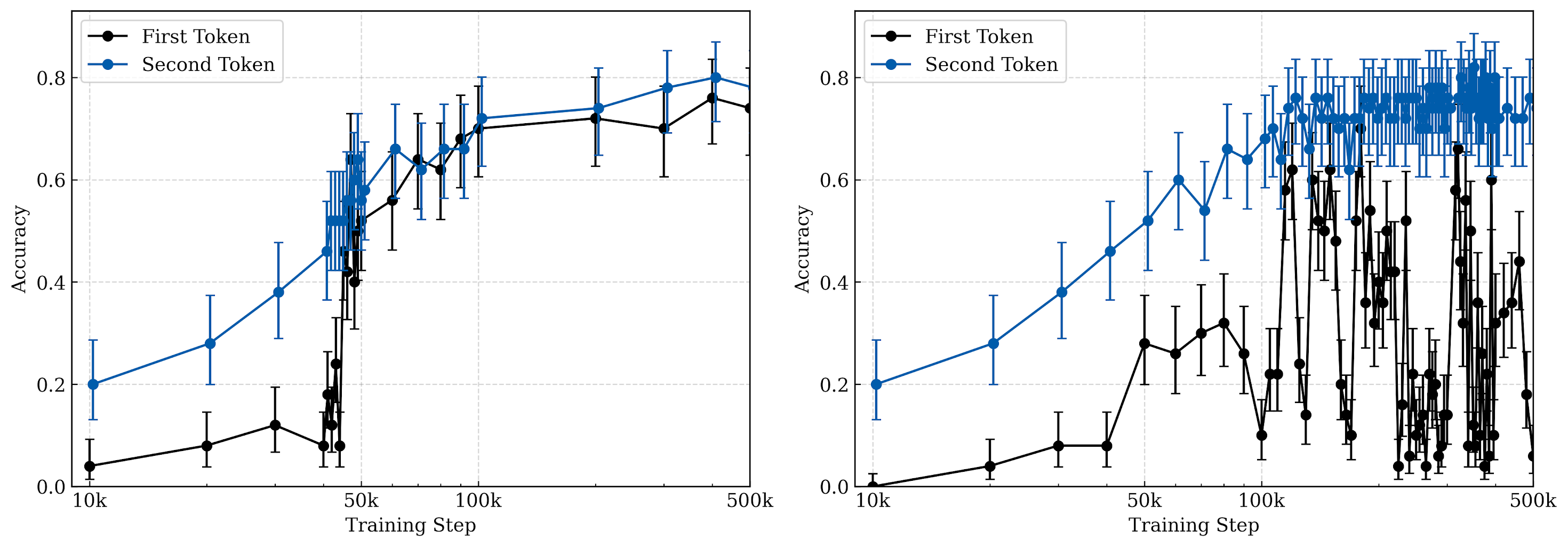}
    \caption{Comparison of the original in-context associative recall task (left) vs the in-context associative recall task that is misdirected to an unseen token (right). % We see that the first token goes through regimes of label based recall (low accuracy) and "lazy bayesian" style recall (higher accuracy). 
    Importantly, we see that the unseen token does not affect the second token accuracy when compared to the original task.}
    \label{fig:nlp_ood_final}
\end{figure}

A comparison of the two plots in Fig.~\ref{fig:nlp_ood_final} shows the results. Note that the figure on the left in Fig.~\ref{fig:nlp_ood_final} is identical to the figure on the right in Fig.~\ref{fig:nlp} and has been copied here for ease of visual comparisons. The key behavior to notice is that there is no qualitative difference between the two blue curves in the left and right subfigures of Fig.~\ref{fig:nlp_ood_final}. What this means is that the second-token-prediction task is functionally unaffected by the ``misdirection'' by the unseen token `Z:'. This gives further evidence that the second token prediction does not rely on the in-context label in order to complete the task once it has started. This matches what we saw in our toy model as well, lending further support for our multi-mechanism conjecture to hold in pretrained LLMs.

%The results illustrated in the right plot in Fig.~\ref{fig:enter-label} show that the second token prediction task is unaffected -- - meaning that if one looks at the case where we don't misdirect and just prompt with `Y:' that is copied as the left plot in Fig.~\ref{fig:enter-label}, we see that the blue curve for the second token prediction is qualitatively unchanged by the misdirection. This means that the model effectively does not leverage the in-context label in order to complete the task once it has started. This matches what we saw in our toy model as well, lending support for our multi-mechanism conjecture.

Meanwhile, for the first token task (black curve in Fig.~\ref{fig:nlp_ood_final}), in a sense, there is objectively no right answer for what to do when presented with `Z:'. Nonetheless, we measure accuracy against a correct ground-truth English translation. It seems that the model accuracy oscillates wildly as it trains. %Sometimes the first token is right quite frequently, but the model performance can degrade as it trains further, only to improve later, decline again, . 
%and so it is interesting to see how often the model chooses to start translating into English. What we see is that with further training, the model flip flops in what it does. Sometimes it seems inclined to start a translation into English, and sometimes not. 
This oscillation during training is an unexpected behavior for single-epoch training and deserves more investigation in future work. 

\FloatBarrier

\section{The Effects of Model Size}\label{sec:model_size_effects}

\begin{table}[htbp]
\centering
%\vspace{-1em}
\resizebox{\textwidth}{!}{
\begin{tabular}{lccccccc}
\toprule
\textbf{Model Name} & \textbf{$n_{params}$} & \textbf{$n_{layers}$} & \textbf{$d_{model}$} & \textbf{$n_{heads}$} & \textbf{$d_{head}$} & \textbf{Learning Rate} & \textbf{Batch Size} \\
\midrule
Orthogonal Tiny & 212K & 3 & 72 & 6 & 12 & $1.7\times10^{-4}$ & 2048 \\
Orthogonal Small & 701K & 6 & 96 & 6 & 16 & $4.5\times10^{-5}$ & 1024 \\
Orthogonal Medium & 2.42M & 12 & 128 & 8 & 16 & $1.6\times10^{-5}$ & 512 \\
Orthogonal Big & 10.7M & 24 & 192 & 12 & 16 & $1.5\times10^{-5}$ & 640 \\ % 2GPUS x 128

Identity Tiny & 212K  & 3 & 72 & 6 & 12 & $6.3\times10^{-5}$ & 8192 \\
Identity Small & 701K  & 6 & 96 & 6 & 16 & $3.2\times10^{-5}$ & 4096 \\
Identity Medium & 2.42M & 12 & 128 & 8 & 16 & $1.6\times10^{-5}$ & 1024 \\ %two gpus so effective batch size of 1024
Identity Big & 10.7M & 24 & 192 & 12 & 16 & $1.3\times10^{-5}$ & 512 \\

\bottomrule
\end{tabular}
}
\vspace{0.3em}
\caption{Model size and training hyperparameters.}
\label{tab:model_size_training_params}
%\vspace{-2em}
\end{table}

In order to test the effect of model size on our emergence results, we trained models across 4 different model sizes as shown in Table \ref{tab:model_size_training_params}. We originally tuned the learning rate for the medium model with a batch size of 512 on a single GPU. Following the model scaling that was done\footnote{Alternatively, one could follow the model scaling done in \cite{groeneveld2024olmo}. There, they also halve the number of layers when decreasing the model size, but decide to halve the model dimension and number of attention heads as well.} in \cite{gpt3paper}, when decreasing the size of our model from ``medium" to ``small" we halved the number of layers, multiplied the model dimension by $0.75$, maintained the same head dimension, and doubled the learning rate. To go from ``small'' to ``tiny'', we used the same process except we chose the head dimension to be $12$ to maintain an integer value for the number of heads. If we would have maintained the head dimension of $16$, the tiny model would have had $4.5$ heads. For this reason, $12$ was chosen as the head dimension since it is the largest integer less than $16$ that leads to an integer when dividing $72$. To go from ``medium" to ``big", we doubled the number of layers, multiplied the model dimension by $1.5$, maintained the same head dimension, and multiplied the learning rate by $\frac{5}{6}$. This scaling was used for the identity models and batch size was not taken into account. It was later brought to our attention that the learning rate should also scale with the batch size. For our later orthogonal runs, we additionally adopted the square-root learning-rate scaling as indicated in \cite{li2024surge}. Specifically, we took the learning rate we had scaled by model size and multiplied it by $\sqrt{\frac{batch\ \ size}{512}}$. However, we have not verified if this scaling is the best way to proceed with our training and further testing is still required. Furthermore, we trained the identity models on one Nvidia GH200 GPU with 80GB of RAM whereas we trained on one L40S GPU for the orthogonal ``tiny'', ``small'', and ``medium'' models and we trained on two L40S GPUs for the orthogonal ``big'' model. One L40S GPU has 48GB of RAM. This is the reason for the differing batch sizes across different model types.

\subsection{When does the associative recall ability emerge for predicting the first test segment observation?}\label{sec:when_emerge}

\begin{figure}[tbph]
    \centering
    \includegraphics[width=\linewidth]{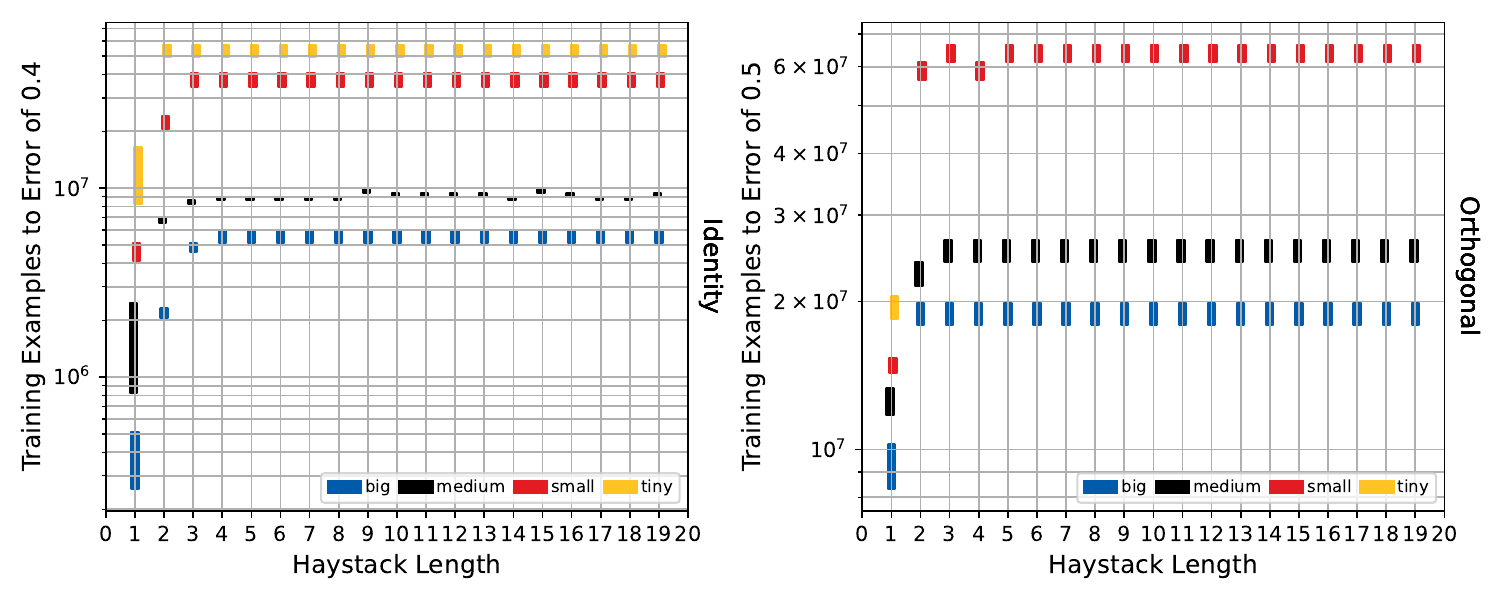}
    \caption{Emergence of associative recall on different haystack lengths across model sizes --- The plot above shows that associative recall emerges much earlier for larger model sizes both when trained on identity systems as well as when trained on orthogonal systems. We also see that associative recall emerges earlier when there is only one system in the haystack but levels out as the model learns to generalize associative recall regardless of the haystack size.}
    \label{fig:model_scaling_haystack_length}
\end{figure}

\begin{figure}[htbp]
    \centering
    \begin{subfigure}[b]{0.32\linewidth}
        \centering
        \includegraphics[width=\linewidth]{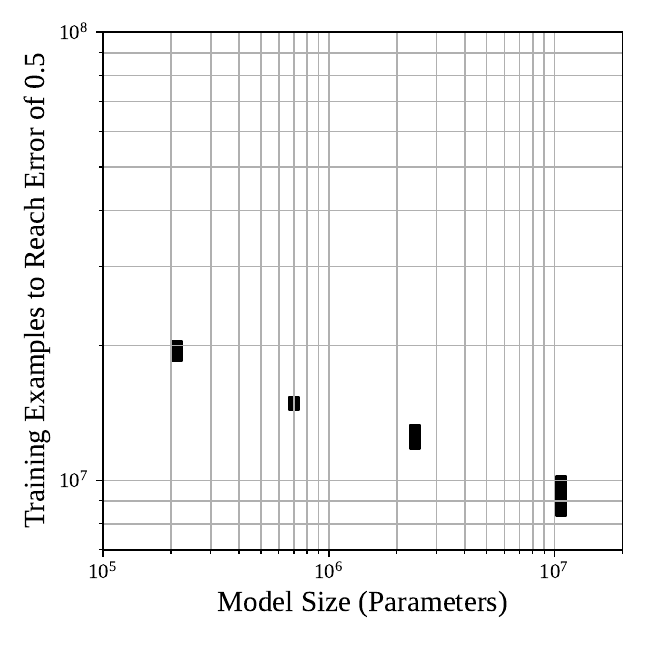}
        \caption{Orthogonal: 1 system haystack.}
        \label{fig:ortho_haystack_len_1_phase_transition}
    \end{subfigure}
    \hfill
    \begin{subfigure}[b]{0.32\linewidth}
        \centering
        \includegraphics[width=\linewidth]{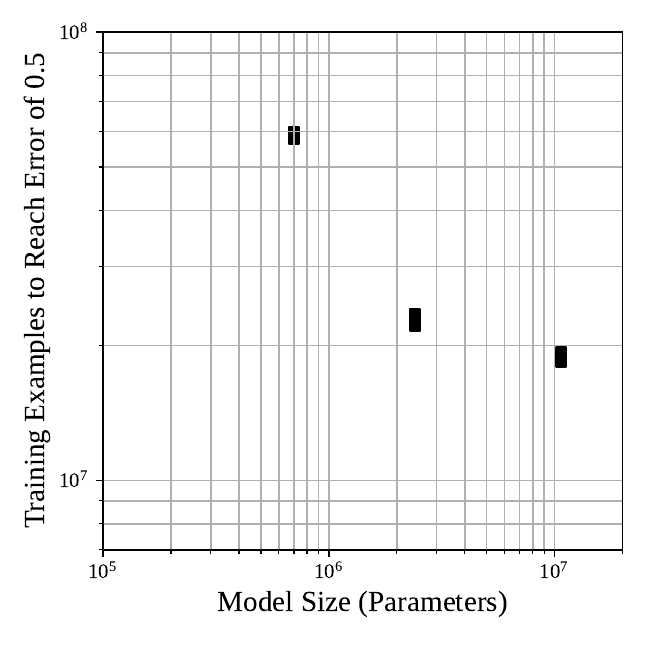}
        \caption{Orthogonal: 2 system haystack.}
        \label{fig:ortho_haystack_len_2_phase_transition}
    \end{subfigure}
    \hfill
    \begin{subfigure}[b]{0.32\linewidth}
        \centering
        \includegraphics[width=\linewidth]{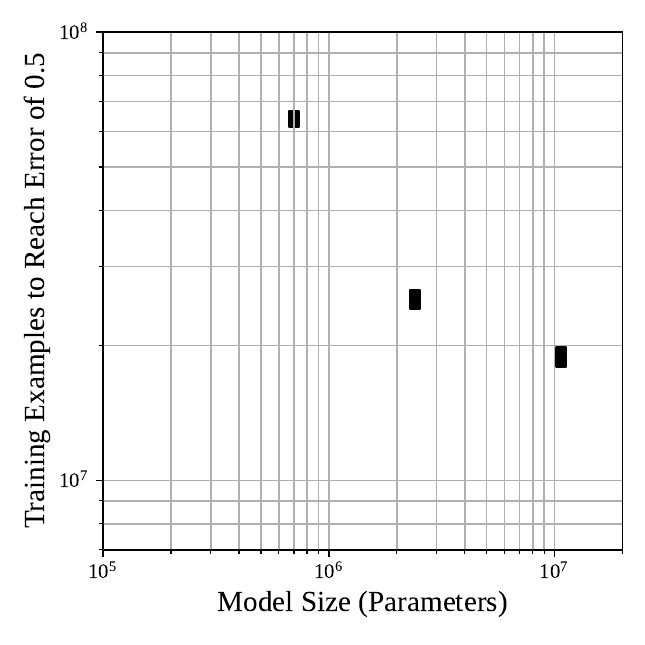}
        \caption{Orthogonal: 3 system haystack.}
        \label{fig:ortho_haystack_len_3_phase_transition}
    \end{subfigure}

    % \vspace{0.2cm}

    \centering
    \begin{subfigure}[b]{0.32\linewidth}
        \centering
        \includegraphics[width=\linewidth]{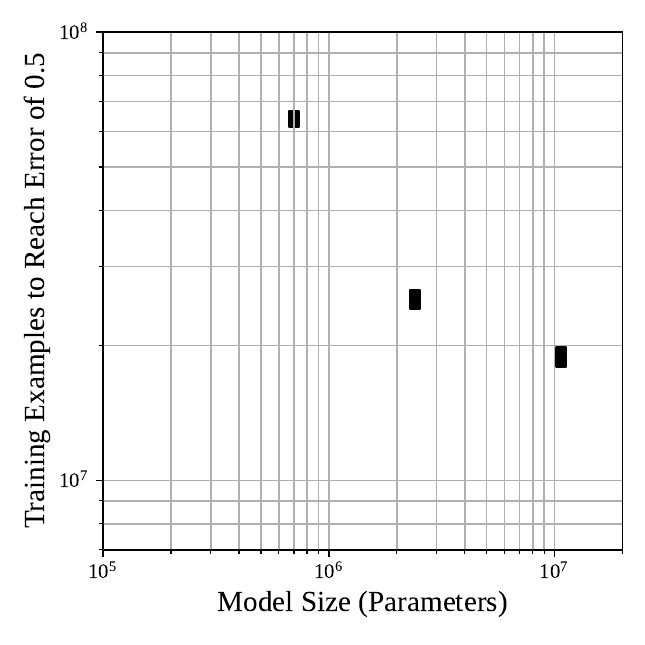}
        \caption{Orthogonal: 17 system haystack.}
        \label{fig:ortho_haystack_len_17_phase_transition}
    \end{subfigure}
    \hfill
    \begin{subfigure}[b]{0.32\linewidth}
        \centering
        \includegraphics[width=\linewidth]{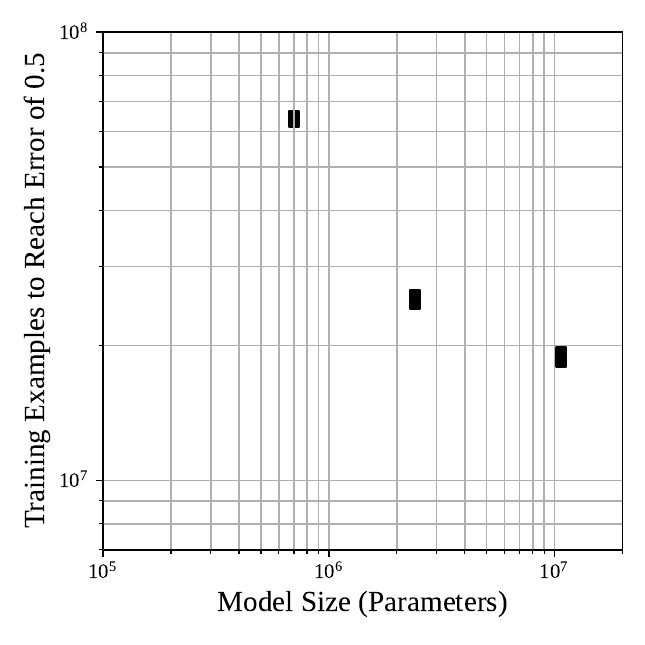}
        \caption{Orthogonal: 18 system haystack.}
        \label{fig:ortho_haystack_len_18_phase_transition}
    \end{subfigure}
    \hfill
    \begin{subfigure}[b]{0.32\linewidth}
        \centering
        \includegraphics[width=\linewidth]{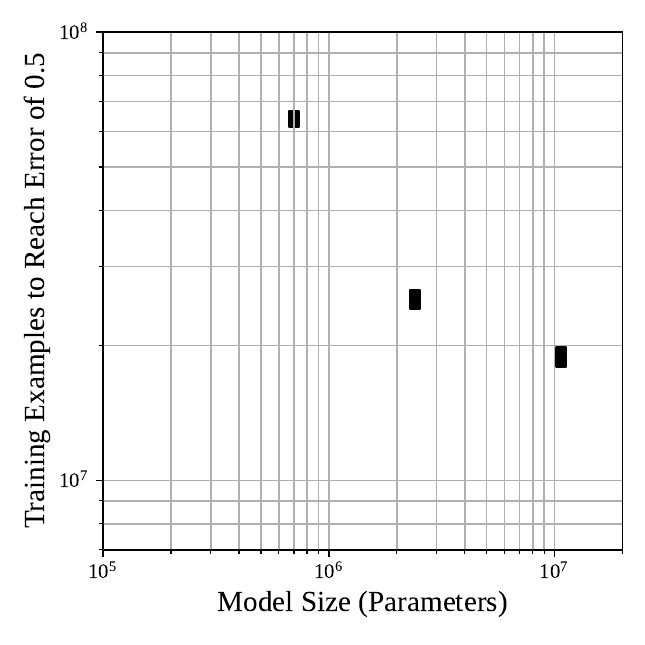}
        \caption{Orthogonal: 19 system haystack.}
        \label{fig:ortho_haystack_len_19_phase_transition}
    \end{subfigure}
    \caption{Emergence of associative recall in varying model sizes across haystack lengths for Orthogonal systems}
    \label{fig:ortho_haystack_phase_transitions}

\end{figure}

\begin{figure}[htbp]
    \centering
    \begin{subfigure}[b]{0.32\linewidth}
        \centering
        \includegraphics[width=\linewidth]{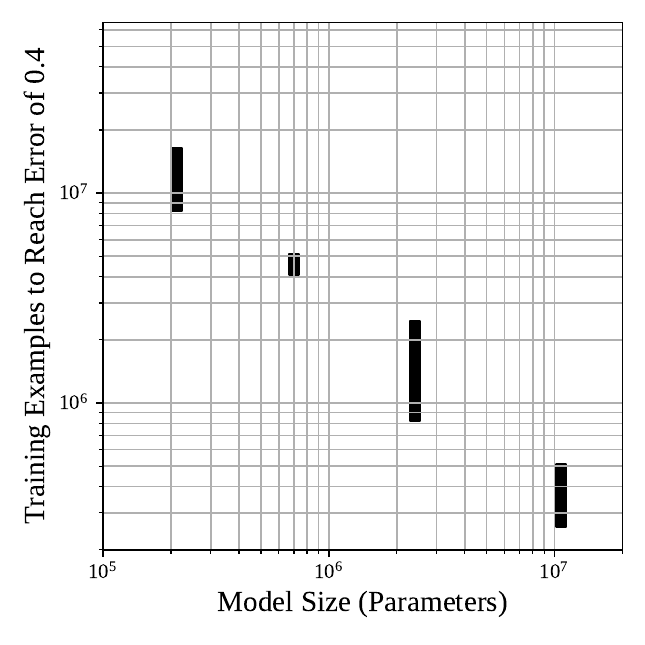}
        \caption{Identity: 1 system haystack.}
        \label{fig:ident_haystack_len_1_phase_transition}
    \end{subfigure}
    \hfill
    \begin{subfigure}[b]{0.32\linewidth}
        \centering
        \includegraphics[width=\linewidth]{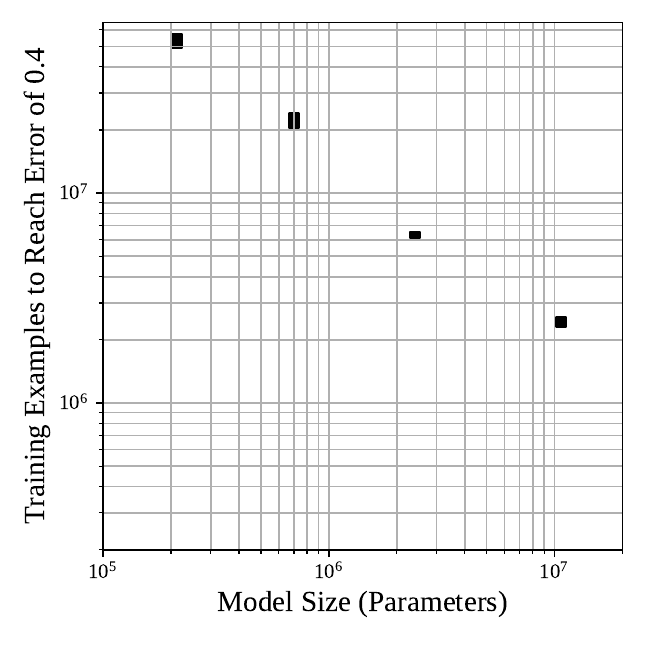}
        \caption{Identity: 2 system haystack.}
        \label{fig:ident_haystack_len_2_phase_transition}
    \end{subfigure}
    \hfill
    \begin{subfigure}[b]{0.32\linewidth}
        \centering
        \includegraphics[width=\linewidth]{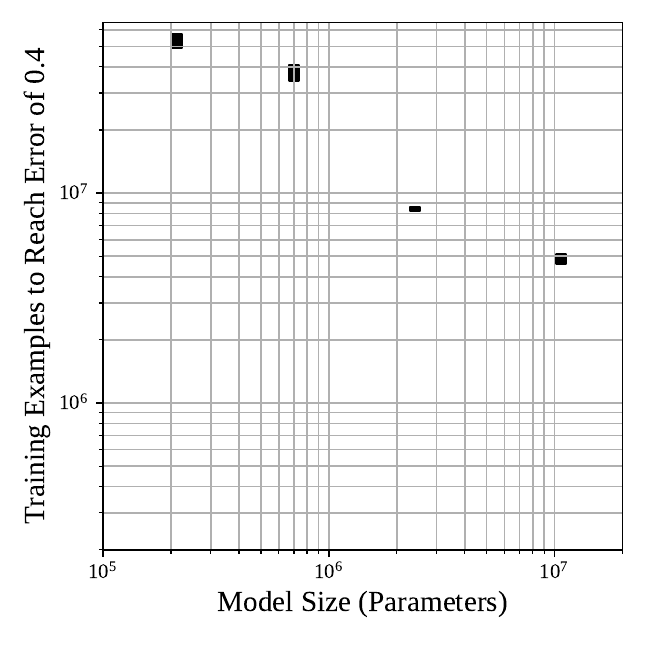}
        \caption{Identity: 3 system haystack.}
        \label{fig:ident_haystack_len_3_phase_transition}
    \end{subfigure}
    \vspace{0.2cm}
    \centering
    \begin{subfigure}[b]{0.32\linewidth}
        \centering
        \includegraphics[width=\linewidth]{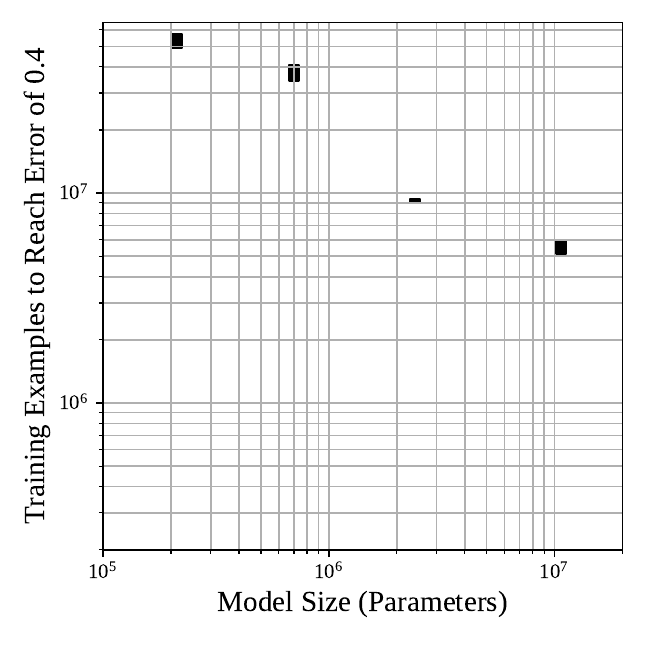}
        \caption{Identity: 17 system haystack.}
        \label{fig:ident_haystack_len_17_phase_transition}
    \end{subfigure}
    \hfill
    \begin{subfigure}[b]{0.32\linewidth}
        \centering
        \includegraphics[width=\linewidth]{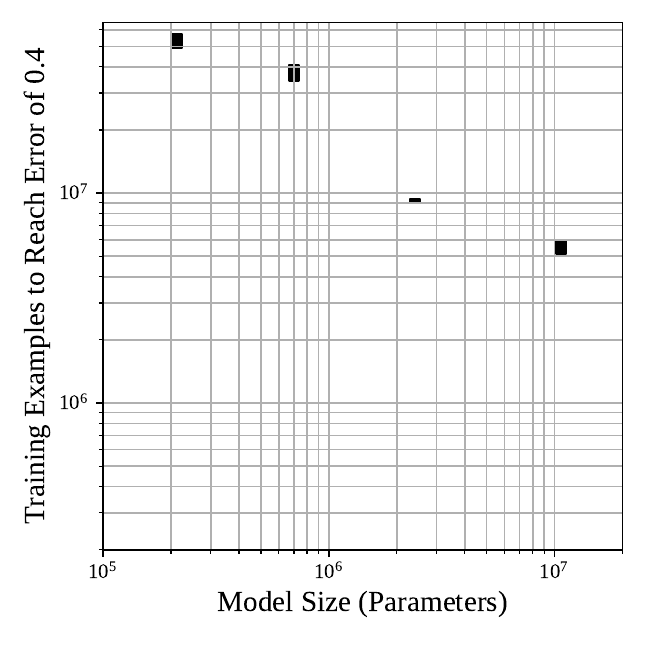}
        \caption{Identity: 18 system haystack.}
        \label{fig:ident_haystack_len_18_phase_transition}
    \end{subfigure}
    \hfill
    \begin{subfigure}[b]{0.32\linewidth}
        \centering
        \includegraphics[width=\linewidth]{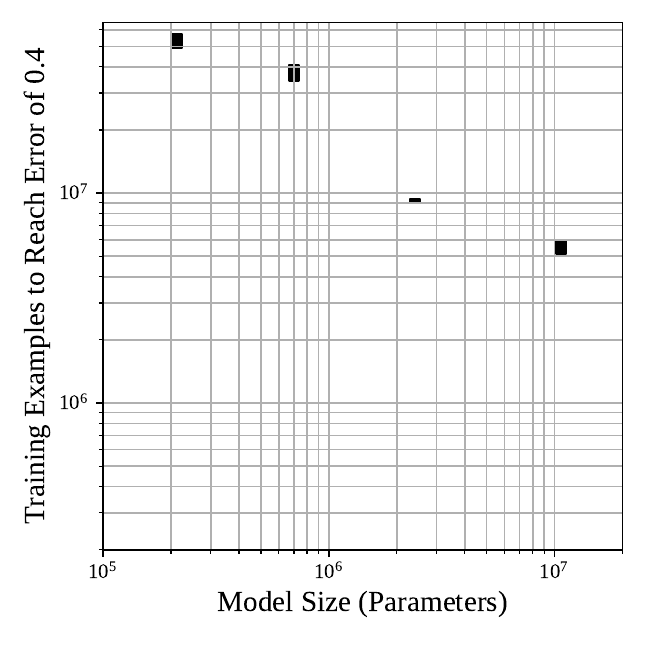}
        \caption{Identity: 19 system haystack.}
        \label{fig:ident_haystack_len_19_phase_transition}
    \end{subfigure}
    \caption{Emergence of associative recall in varying model sizes across haystack lengths for Identity systems}
    \label{fig:ident_haystack_phase_transitions}  
\end{figure}

Evidence that larger models see earlier emergence is shown in Figs.~\ref{fig:model_scaling_haystack_length}, \ref{fig:ortho_haystack_phase_transitions} and \ref{fig:ident_haystack_phase_transitions}, where the number of training examples until the phase transition is shown for each model size for the identity and orthogonal systems. This was done by recording the checkpoint before and after the mean-squared error for the pure recall task dropped below the cutoff values of $0.4$ for the identity systems and $0.5$ for the orthogonal systems. These cutoff values were chosen by visual inspection. 

% Given the questions around our learning rate we are unsure about how model size exactly affects the point of emergence. It is also interesting to note that the emergence of associative recall for a one system haystack emerges considerably earlier than it does for larger haystacks. We also found that while associative recall emerges for 1 system in the haystack on the tiny orthogonal model, it did not emerge for larger haystack sizes over the course of training that model.

\FloatBarrier

\section{More training dynamics plots}\label{sec:more_train_dyn}

\begin{figure}[htbp]
    \centering
    \begin{subfigure}[b]{0.29\linewidth}
        \centering
        \includegraphics[width=\linewidth]{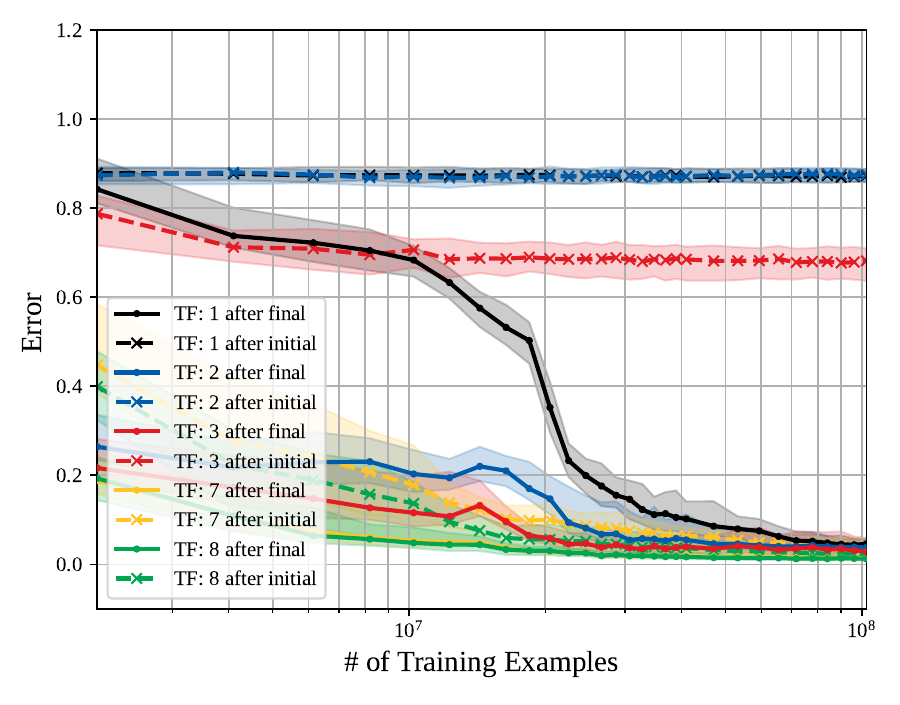}
        \caption{Orthogonal: 1 system haystack.}
        \label{fig:ortho_tiny_len_1_baseline_linear}
    \end{subfigure}
    \hfill
    \begin{subfigure}[b]{0.29\linewidth}
        \centering
        \includegraphics[width=\linewidth]{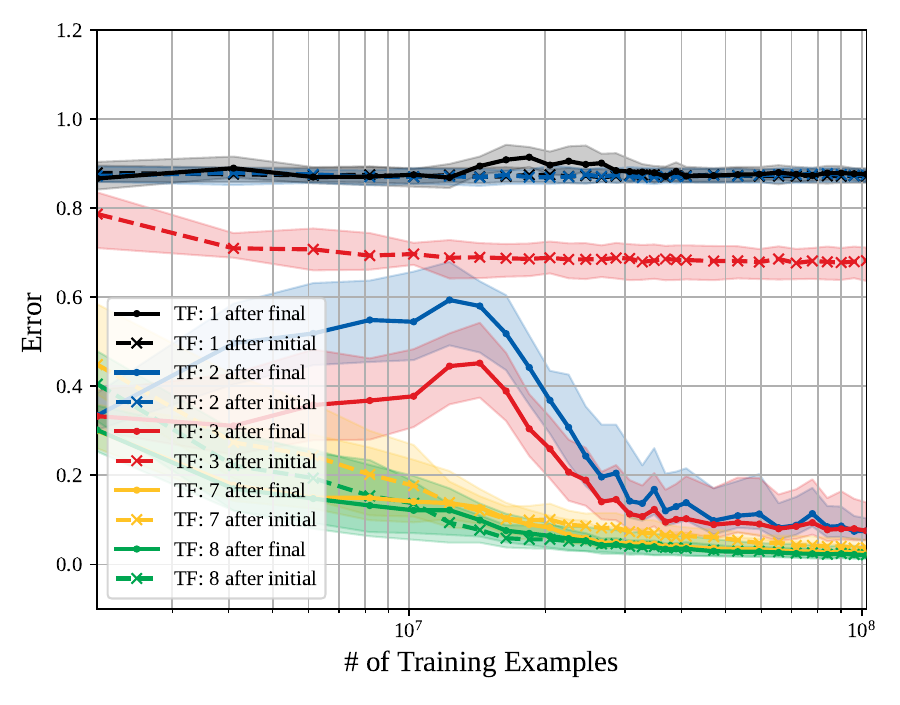}
        \caption{Orthogonal: 2 system haystack.}
        \label{fig:ortho_tiny_len_2_baseline_linear}
    \end{subfigure}
    \hfill
    \begin{subfigure}[b]{0.29\linewidth}
        \centering
        \includegraphics[width=\linewidth]{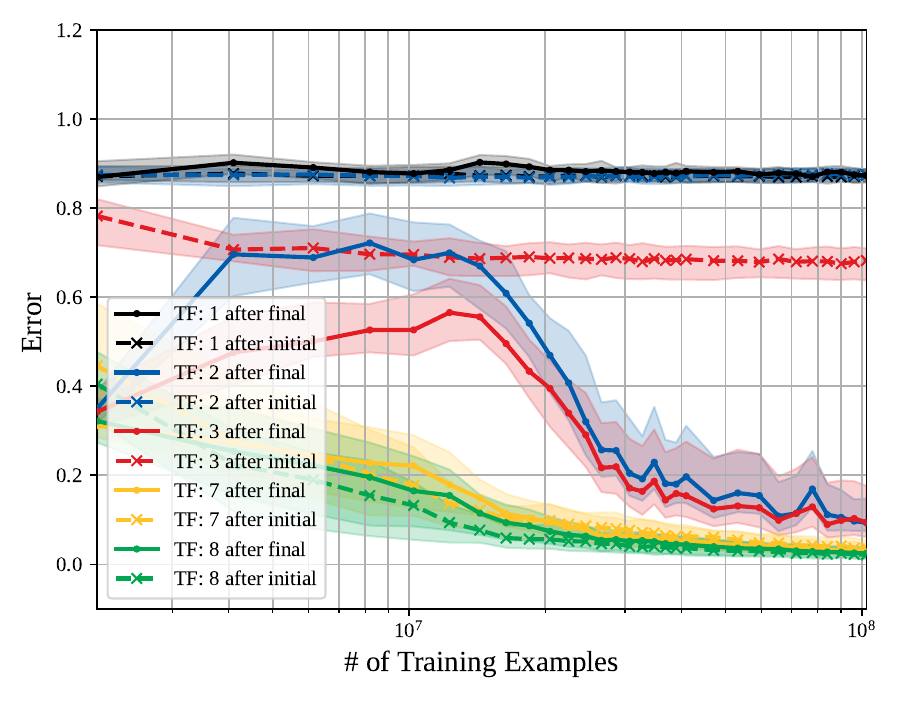}
        \caption{Orthogonal: 3 system haystack.}
        \label{fig:ortho_tiny_len_3_baseline_linear}
    \end{subfigure}

    % \vspace{0.2cm}

    \centering
    \begin{subfigure}[b]{0.29\linewidth}
        \centering
        \includegraphics[width=\linewidth]{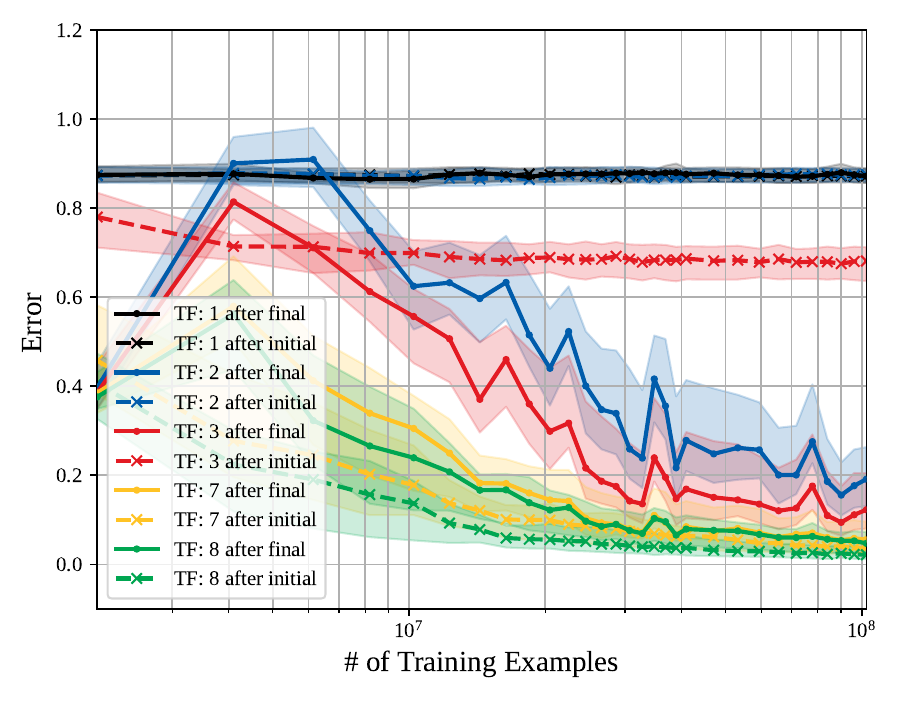}
        \caption{Orthogonal: 17 system haystack.}
        \label{fig:ortho_tiny_len_17_baseline_linear}
    \end{subfigure}
    \hfill
    \begin{subfigure}[b]{0.29\linewidth}
        \centering
        \includegraphics[width=\linewidth]{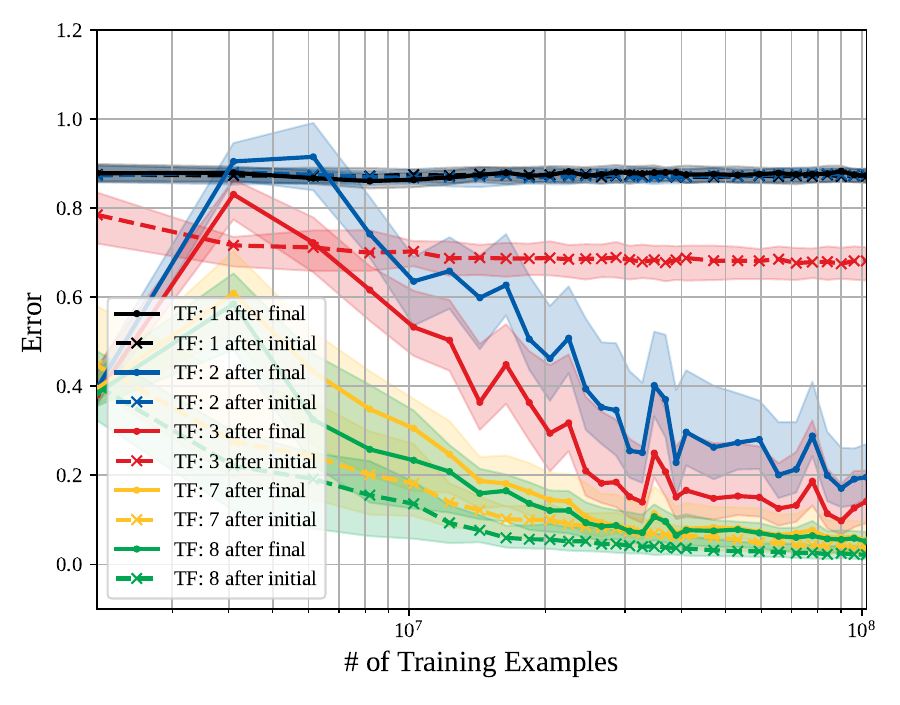}
        \caption{Orthogonal: 18 system haystack.}
        \label{fig:ortho_tiny_len_18_baseline_linear}
    \end{subfigure}
    \hfill
    \begin{subfigure}[b]{0.29\linewidth}
        \centering
        \includegraphics[width=\linewidth]{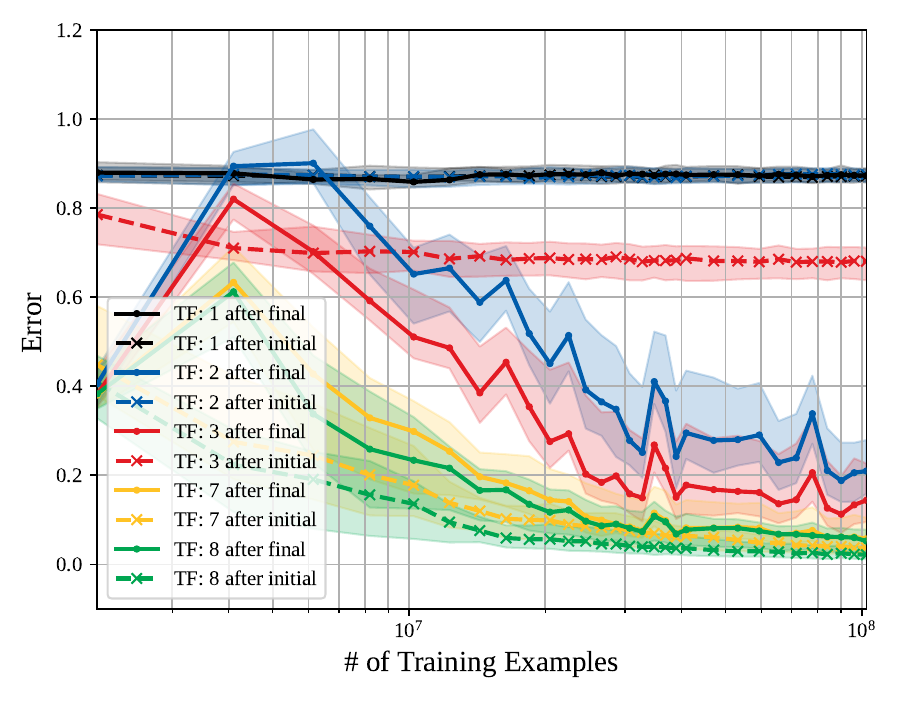}
        \caption{Orthogonal: 19 system haystack.}
        \label{fig:ortho_tiny_len_19_baseline_linear}
    \end{subfigure}
    \caption{Performance of tiny orthogonal model (212K params) across training --- linear-scale.}
    \label{fig:ortho_tiny_baseline_linear}

\end{figure}

\begin{figure}[htbp]
    \centering
    \begin{subfigure}[b]{0.29\linewidth}
        \centering
        \includegraphics[width=\linewidth]{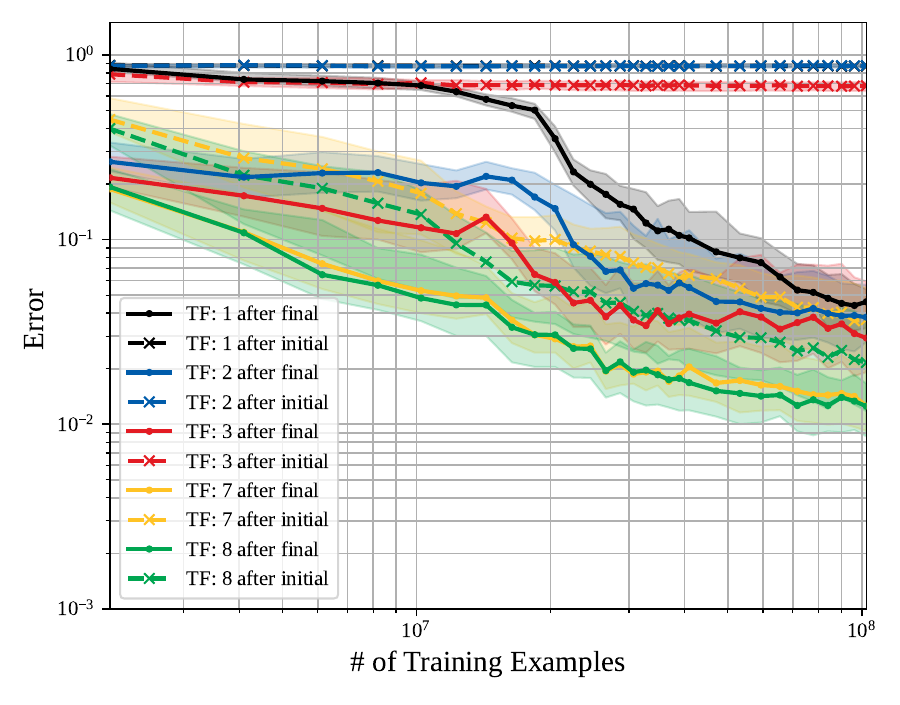}
        \caption{Orthogonal: 1 system haystack.}
        \label{fig:ortho_tiny_len_1_baseline_log}
    \end{subfigure}
    \hfill
    \begin{subfigure}[b]{0.29\linewidth}
        \centering
        \includegraphics[width=\linewidth]{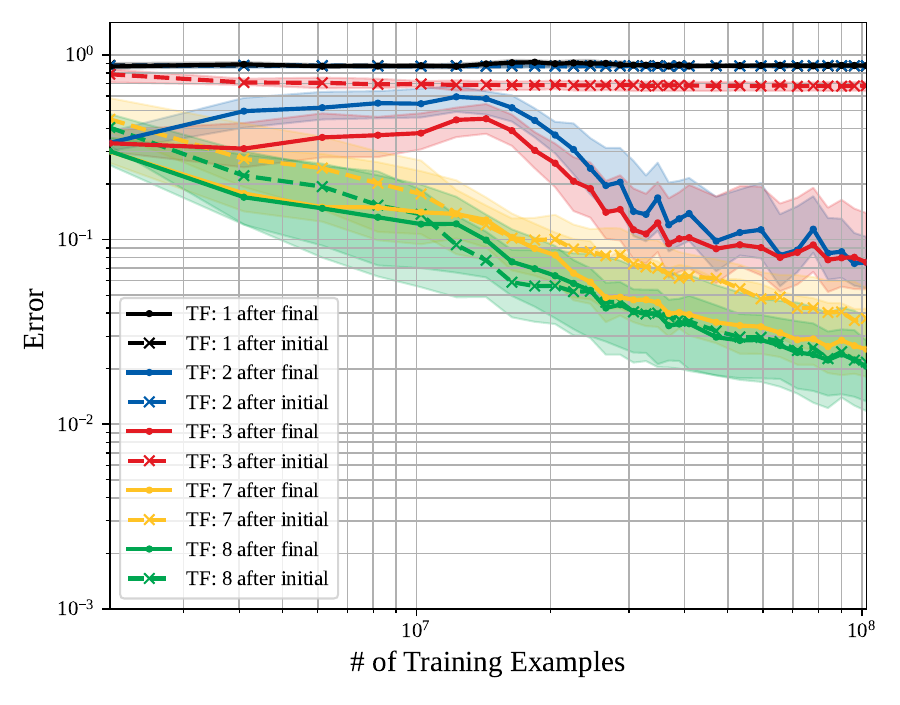}
        \caption{Orthogonal: 2 system haystack.}
        \label{fig:ortho_tiny_len_2_baseline_log}
    \end{subfigure}
    \hfill
    \begin{subfigure}[b]{0.29\linewidth}
        \centering
        \includegraphics[width=\linewidth]{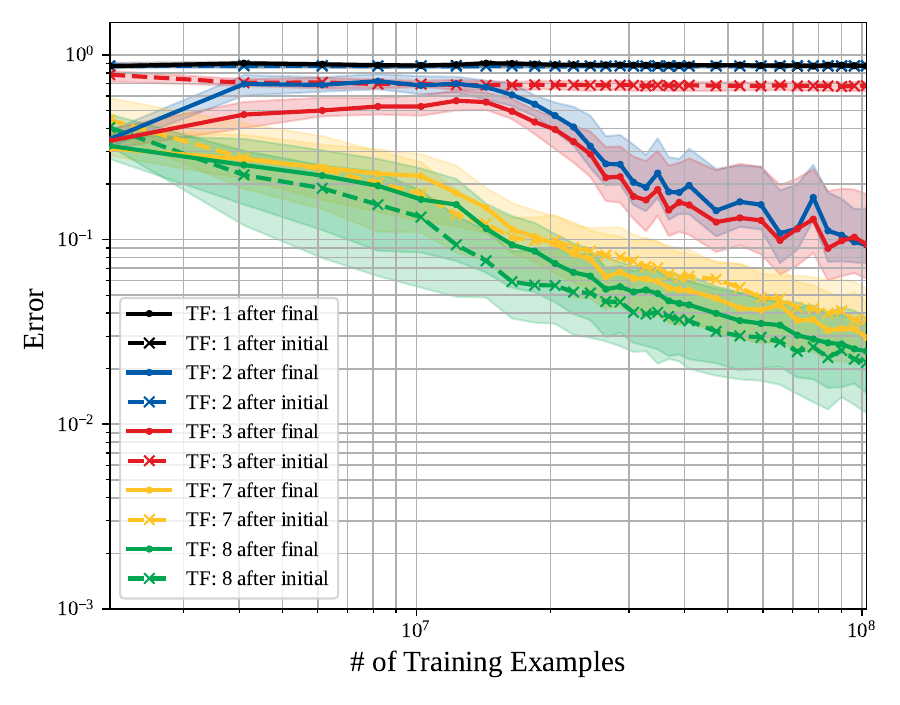}
        \caption{Orthogonal: 3 system haystack.}
        \label{fig:ortho_tiny_len_3_baseline_log}
    \end{subfigure}

    % \vspace{0.2cm}

    \centering
    \begin{subfigure}[b]{0.29\linewidth}
        \centering
        \includegraphics[width=\linewidth]{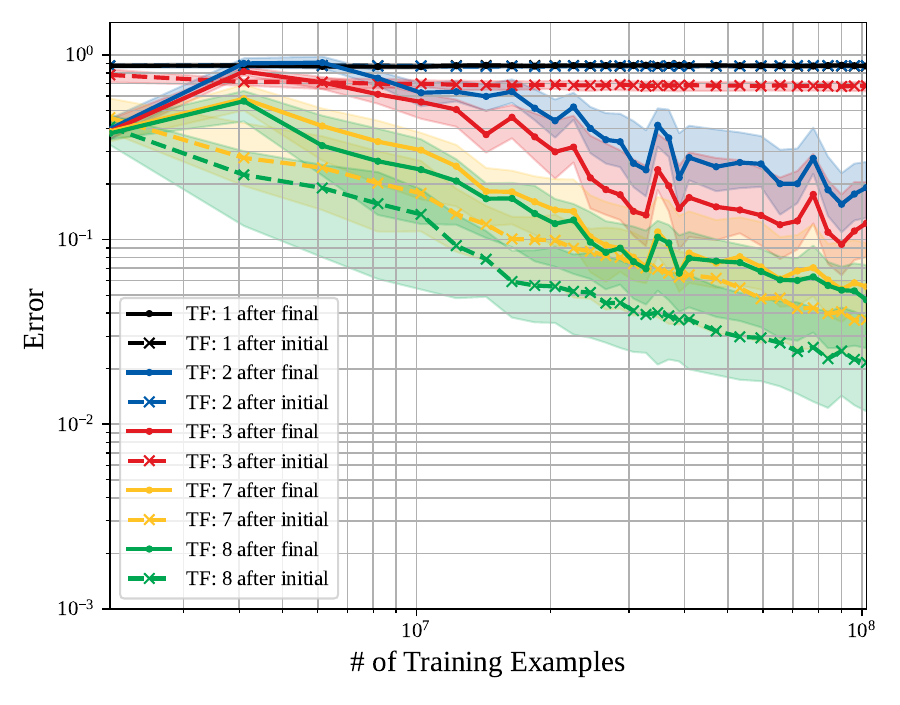}
        \caption{Orthogonal: 17 system haystack.}
        \label{fig:ortho_tiny_len_17_baseline_log}
    \end{subfigure}
    \hfill
    \begin{subfigure}[b]{0.29\linewidth}
        \centering
        \includegraphics[width=\linewidth]{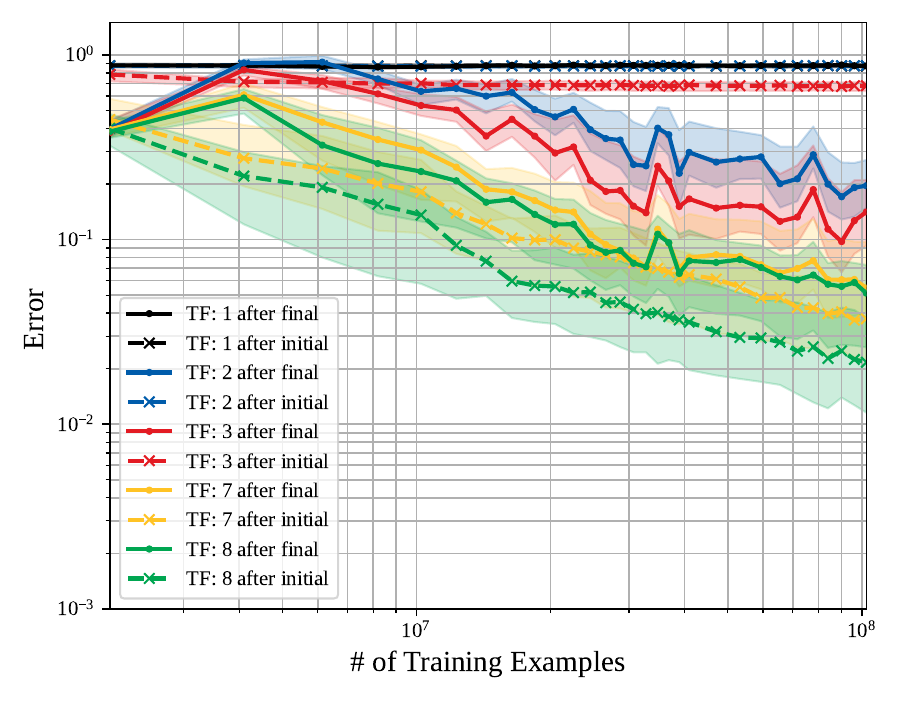}
        \caption{Orthogonal: 18 system haystack.}
        \label{fig:ortho_tiny_len_18_baseline_log}
    \end{subfigure}
    \hfill
    \begin{subfigure}[b]{0.29\linewidth}
        \centering
        \includegraphics[width=\linewidth]{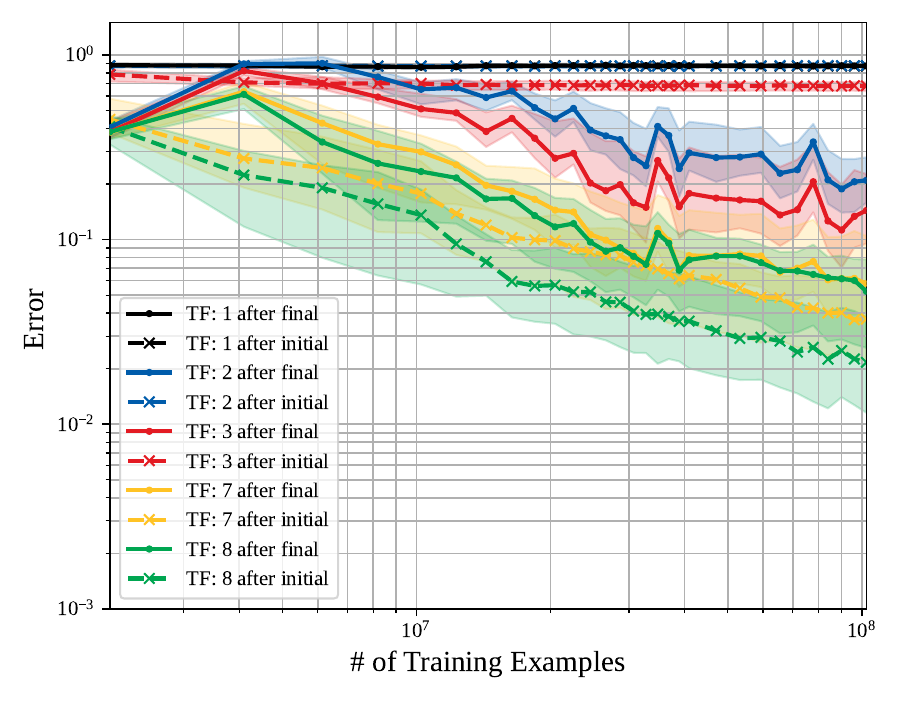}
        \caption{Orthogonal: 19 system haystack.}
        \label{fig:ortho_tiny_len_19_baseline_log}
    \end{subfigure}
    \caption{Performance of tiny orthogonal model (212K params) across training --- log-scale.}
    \label{fig:ortho_tiny_baseline_log}

\end{figure}

\begin{figure}[htbp]
    \centering
    \begin{subfigure}[b]{0.32\linewidth}
        \centering
        \includegraphics[width=\linewidth]{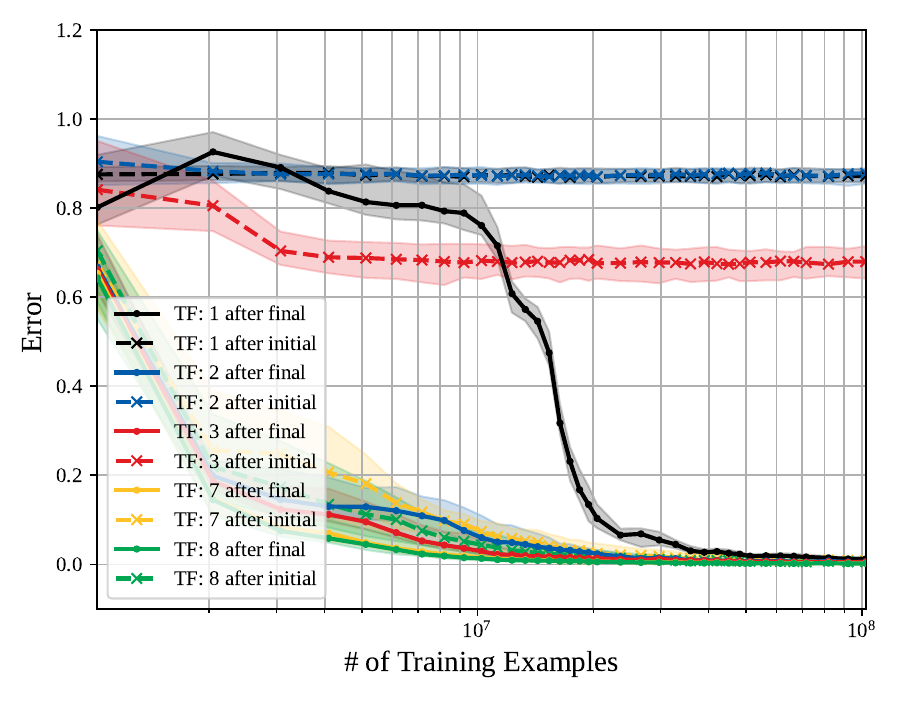}
        \caption{Orthogonal: 1 system haystack.}
        \label{fig:ortho_small_len_1_baseline_linear}
    \end{subfigure}
    \hfill
    \begin{subfigure}[b]{0.32\linewidth}
        \centering
        \includegraphics[width=\linewidth]{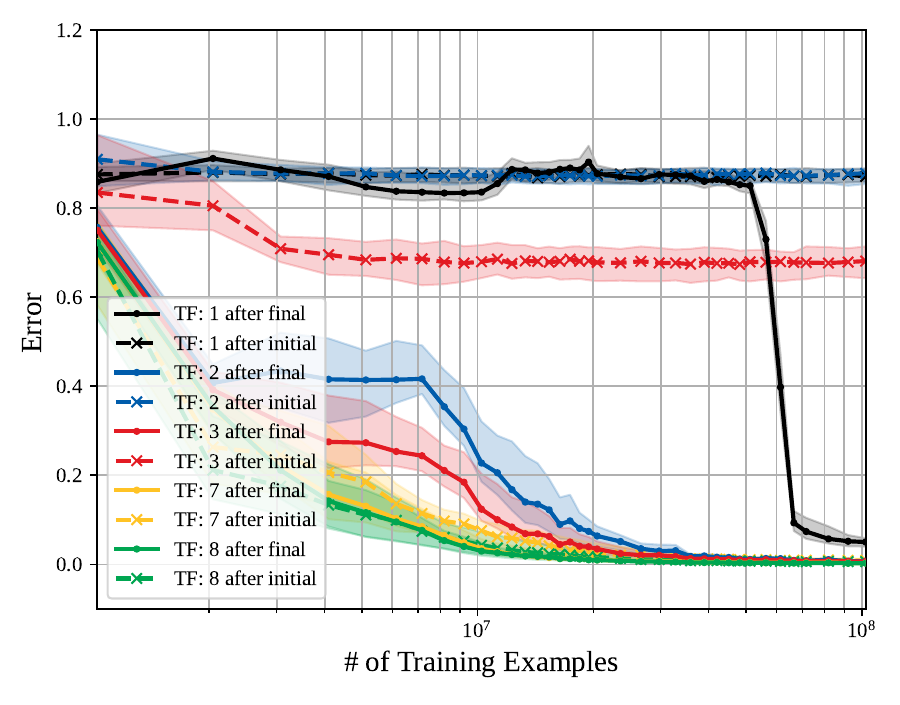}
        \caption{Orthogonal: 2 system haystack.}
        \label{fig:ortho_small_len_2_baseline_linear}
    \end{subfigure}
    \hfill
    \begin{subfigure}[b]{0.32\linewidth}
        \centering
        \includegraphics[width=\linewidth]{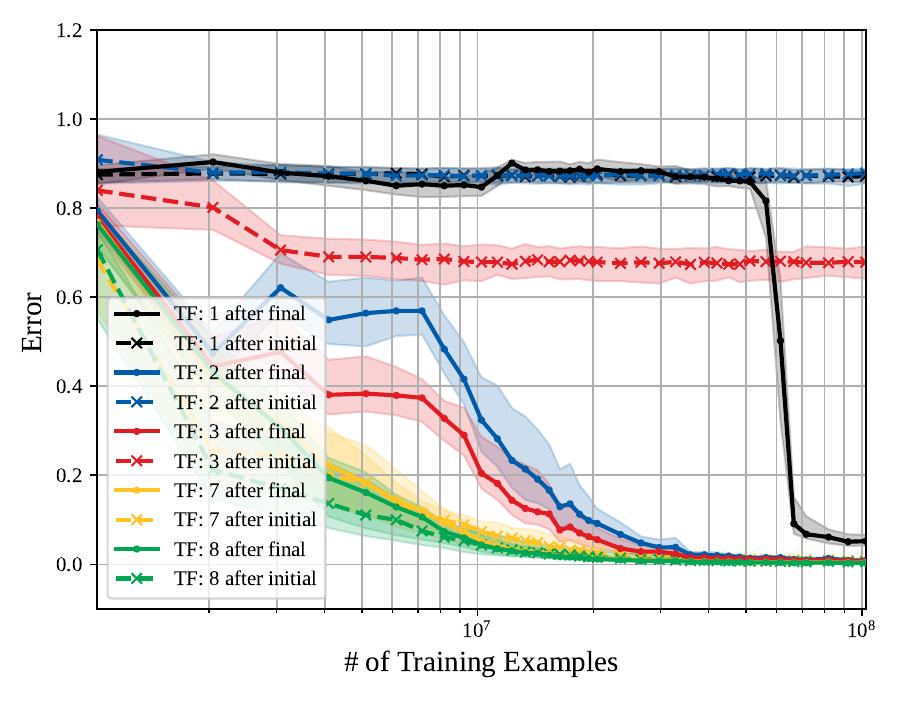}
        \caption{Orthogonal: 3 system haystack.}
        \label{fig:ortho_small_len_3_baseline_linear}
    \end{subfigure}

    \vspace{0.2cm}

    \centering
    \begin{subfigure}[b]{0.32\linewidth}
        \centering
        \includegraphics[width=\linewidth]{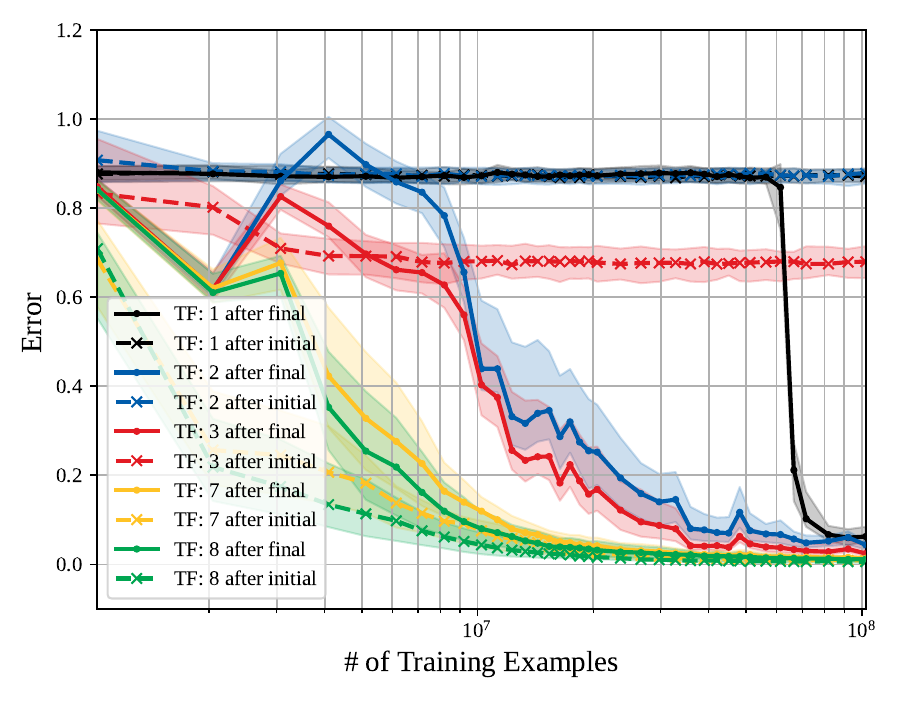}
        \caption{Orthogonal: 17 system haystack.}
        \label{fig:ortho_small_len_17_baseline_linear}
    \end{subfigure}
    \hfill
    \begin{subfigure}[b]{0.32\linewidth}
        \centering
        \includegraphics[width=\linewidth]{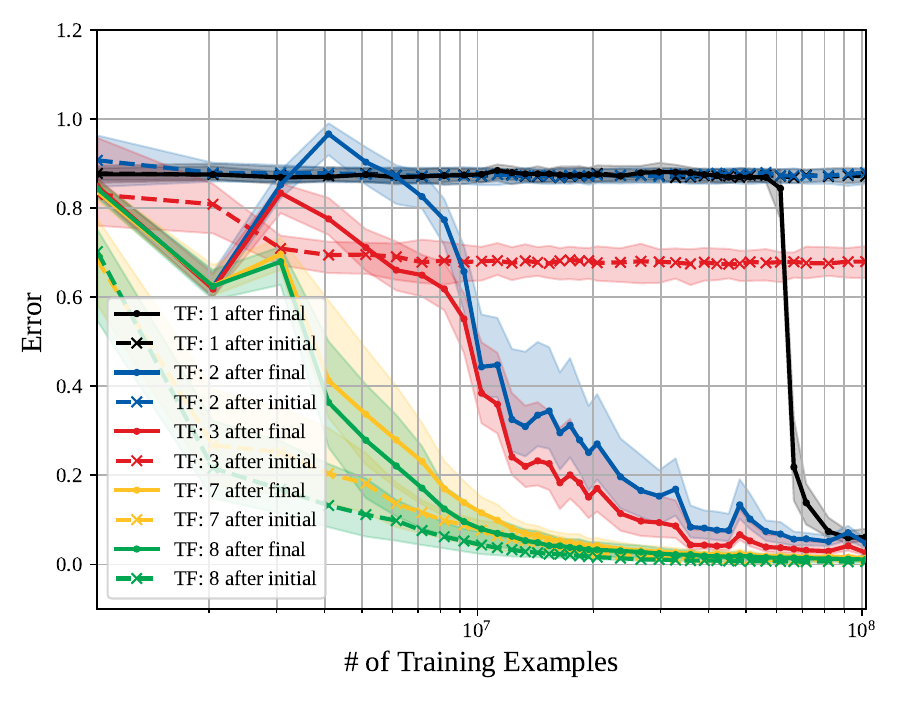}
        \caption{Orthogonal: 18 system haystack.}
        \label{fig:ortho_small_len_18_baseline_linear}
    \end{subfigure}
    \hfill
    \begin{subfigure}[b]{0.32\linewidth}
        \centering
        \includegraphics[width=\linewidth]{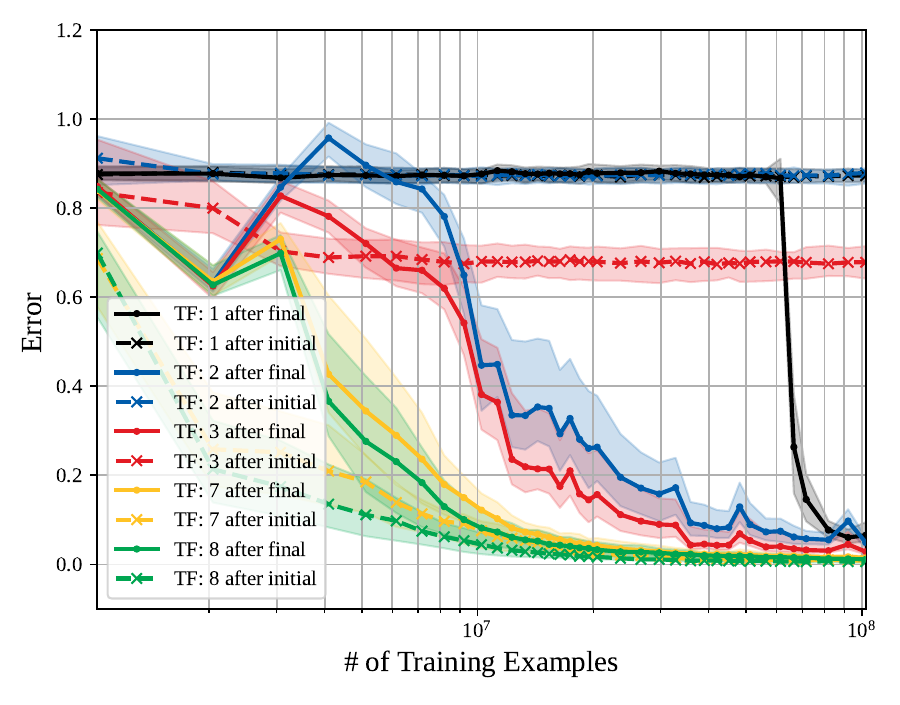}
        \caption{Orthogonal: 19 system haystack.}
        \label{fig:ortho_small_len_19_baseline_linear}
    \end{subfigure}
    \caption{Performance of small orthogonal model (701K params) across training --- linear-scale.}
    \label{fig:ortho_small_baseline_linear}

\end{figure}

\begin{figure}[htbp]
    \centering
    \begin{subfigure}[b]{0.32\linewidth}
        \centering
        \includegraphics[width=\linewidth]{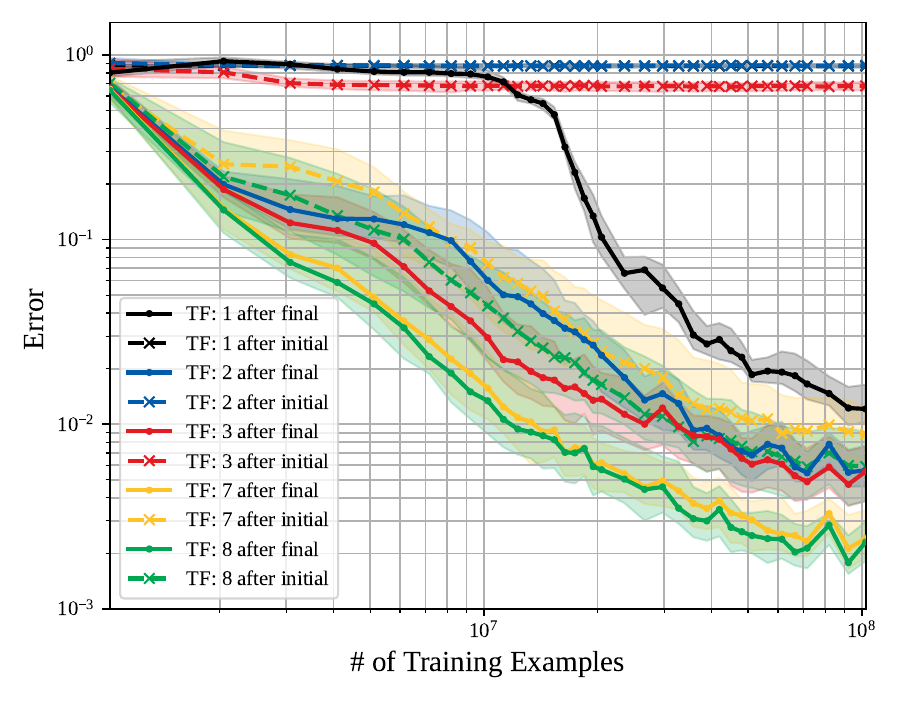}
        \caption{Orthogonal: 1 system haystack.}
        \label{fig:ortho_small_len_1_baseline_log}
    \end{subfigure}
    \hfill
    \begin{subfigure}[b]{0.32\linewidth}
        \centering
        \includegraphics[width=\linewidth]{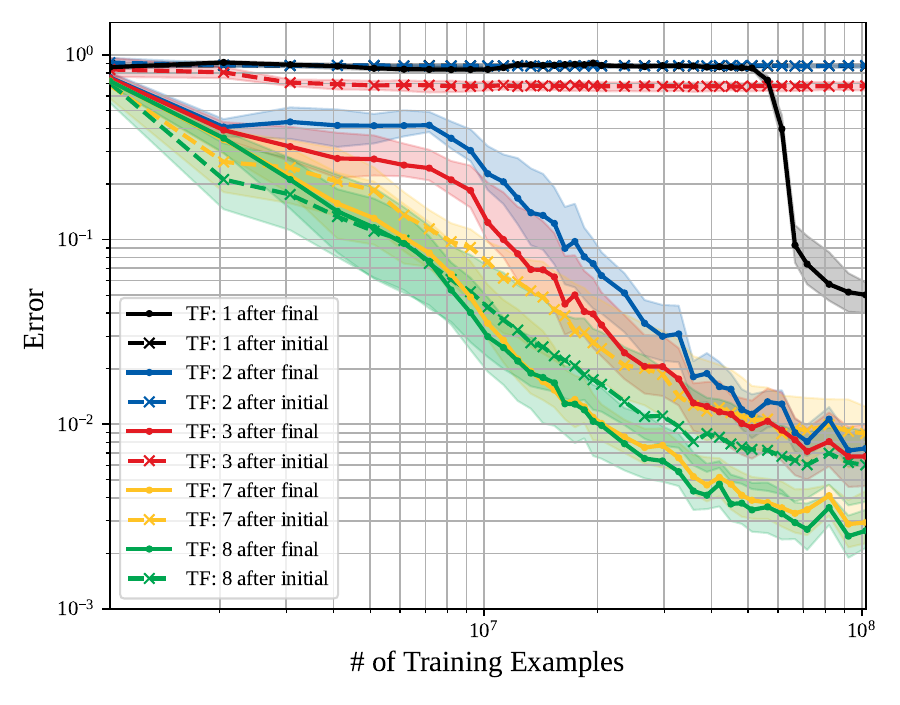}
        \caption{Orthogonal: 2 system haystack.}
        \label{fig:ortho_small_len_2_baseline_log}
    \end{subfigure}
    \hfill
    \begin{subfigure}[b]{0.32\linewidth}
        \centering
        \includegraphics[width=\linewidth]{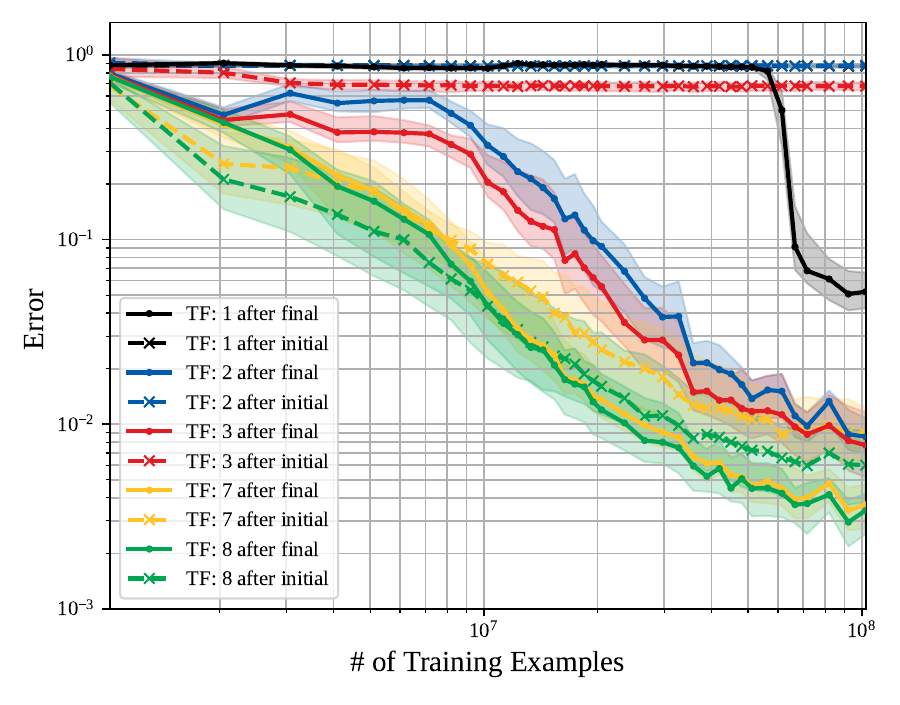}
        \caption{Orthogonal: 3 system haystack.}
        \label{fig:ortho_small_len_3_baseline_log}
    \end{subfigure}

    \vspace{0.2cm}

    \centering
    \begin{subfigure}[b]{0.32\linewidth}
        \centering
        \includegraphics[width=\linewidth]{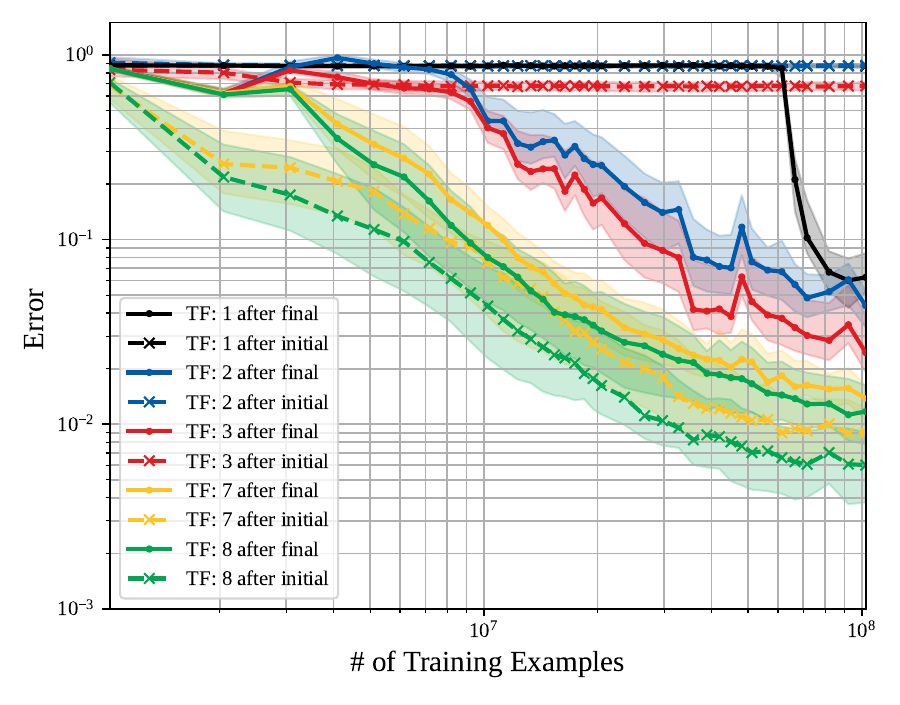}
        \caption{Orthogonal: 17 system haystack.}
        \label{fig:ortho_small_len_17_baseline_log}
    \end{subfigure}
    \hfill
    \begin{subfigure}[b]{0.32\linewidth}
        \centering
        \includegraphics[width=\linewidth]{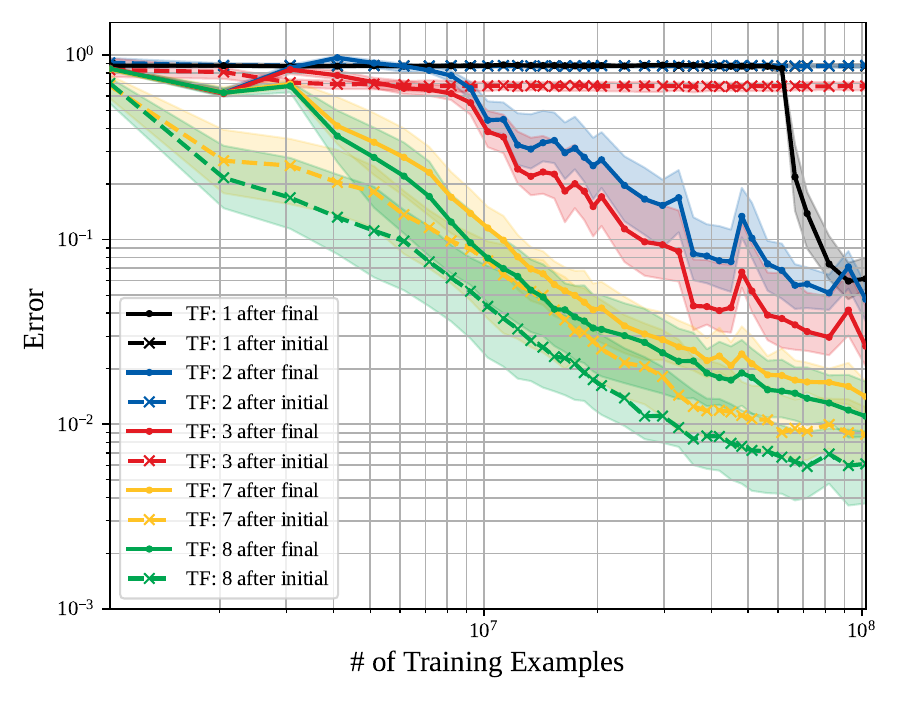}
        \caption{Orthogonal: 18 system haystack.}
        \label{fig:ortho_small_len_18_baseline_log}
    \end{subfigure}
    \hfill
    \begin{subfigure}[b]{0.32\linewidth}
        \centering
        \includegraphics[width=\linewidth]{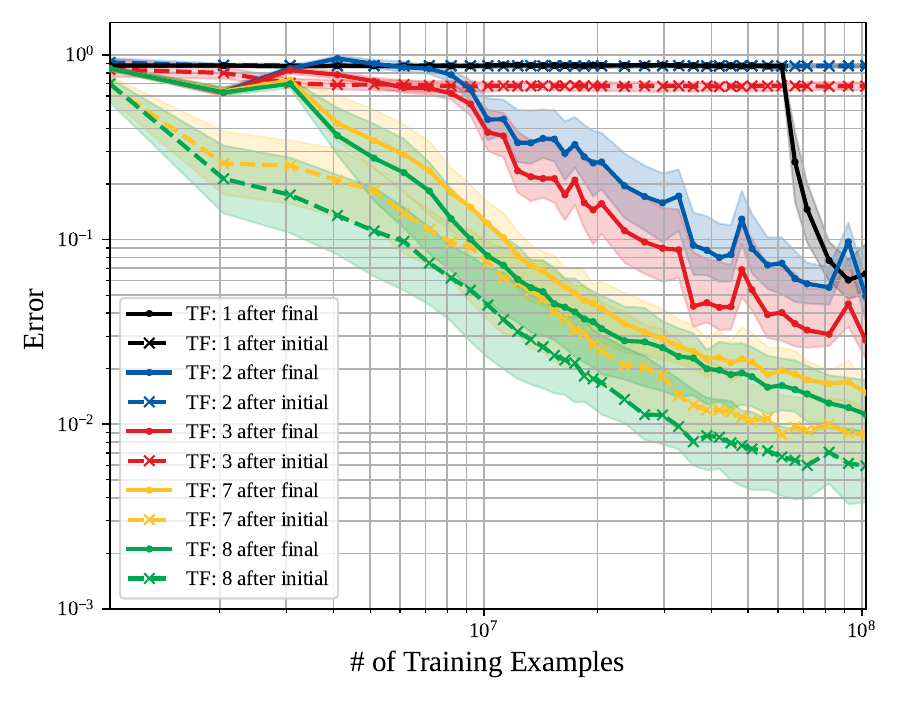}
        \caption{Orthogonal: 19 system haystack.}
        \label{fig:ortho_small_len_19_baseline_log}
    \end{subfigure}
    \caption{Performance of small orthogonal model (701K params) across training --- log-scale.}
    \label{fig:ortho_small_baseline_log}

\end{figure}

\begin{figure}[htbp]
    \centering
    \begin{subfigure}[b]{0.32\linewidth}
        \centering
        \includegraphics[width=\linewidth]{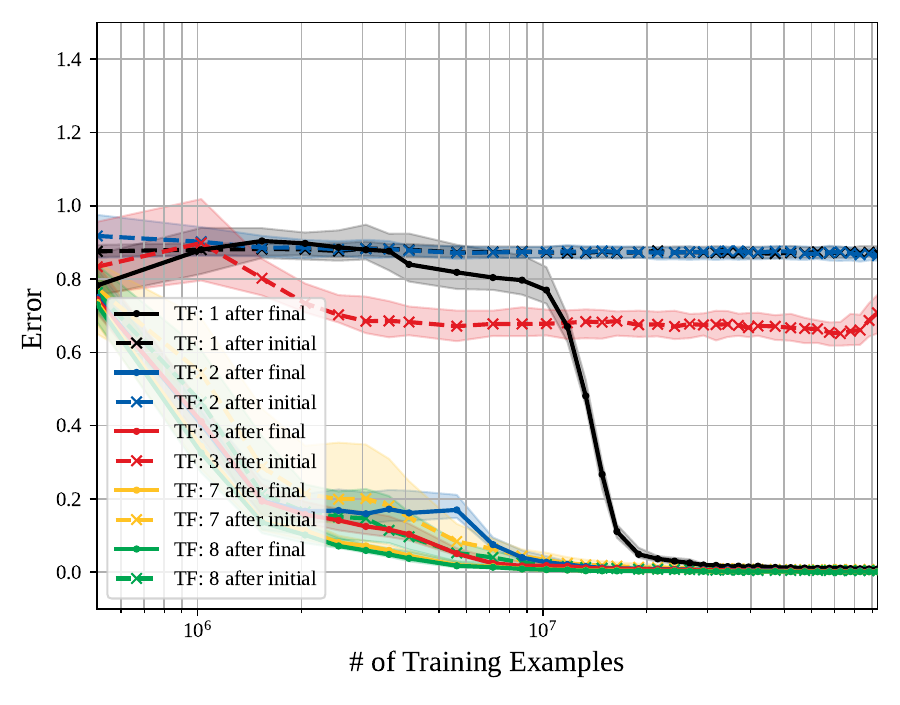}
        \caption{Orthogonal: 1 system haystack.}
        \label{fig:ortho_med_len_1_baseline_linear}
    \end{subfigure}
    \hfill
    \begin{subfigure}[b]{0.32\linewidth}
        \centering
        \includegraphics[width=\linewidth]{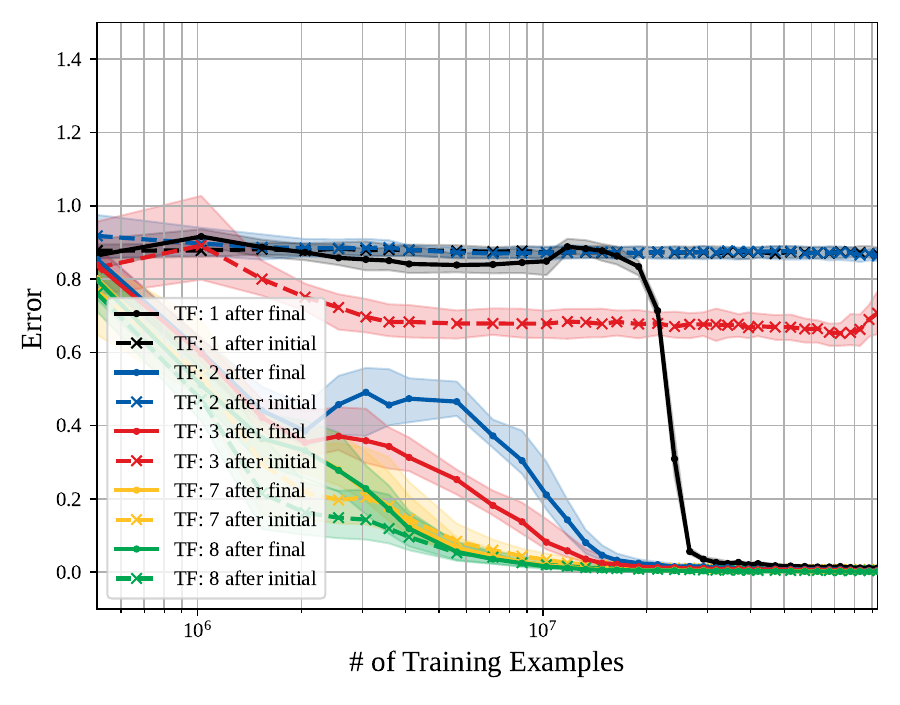}
        \caption{Orthogonal: 2 system haystack.}
        \label{fig:ortho_med_len_2_baseline_linear}
    \end{subfigure}
    \hfill
    \begin{subfigure}[b]{0.32\linewidth}
        \centering
        \includegraphics[width=\linewidth]{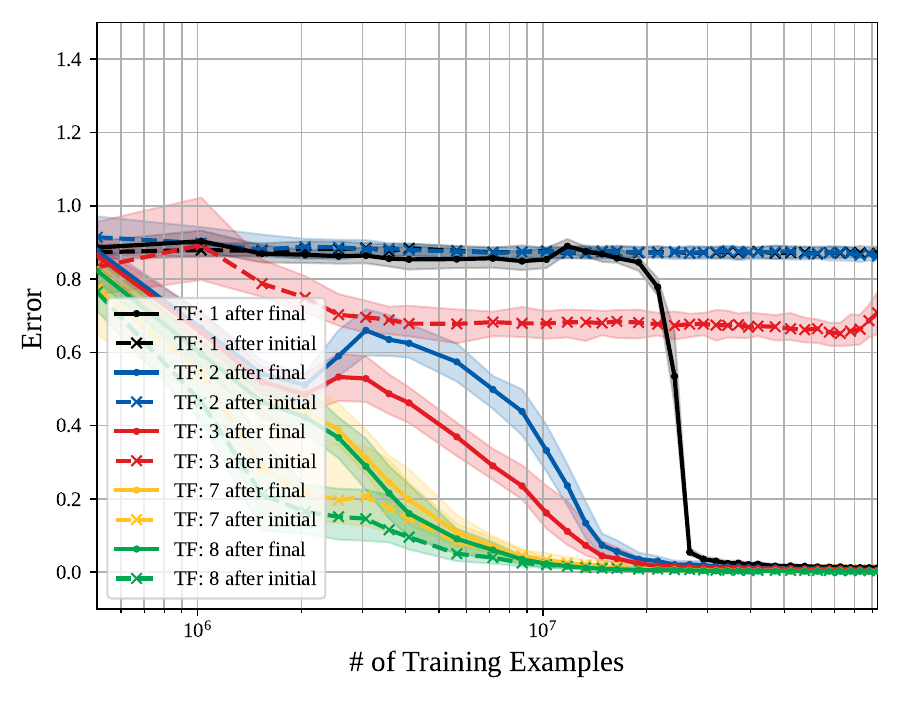}
        \caption{Orthogonal: 3 system haystack.}
        \label{fig:ortho_med_len_3_baseline_linear}
    \end{subfigure}

    \vspace{0.2cm}

    \centering
    \begin{subfigure}[b]{0.32\linewidth}
        \centering
        \includegraphics[width=\linewidth]{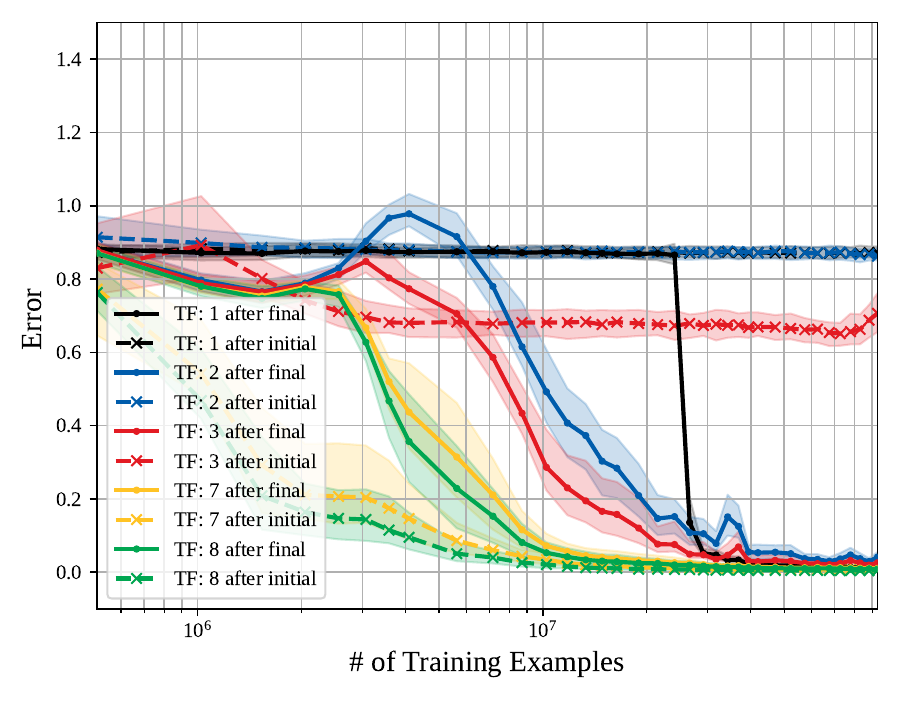}
        \caption{Orthogonal: 17 system haystack.}
        \label{fig:ortho_med_len_17_baseline_linear}
    \end{subfigure}
    \hfill
    \begin{subfigure}[b]{0.32\linewidth}
        \centering
        \includegraphics[width=\linewidth]{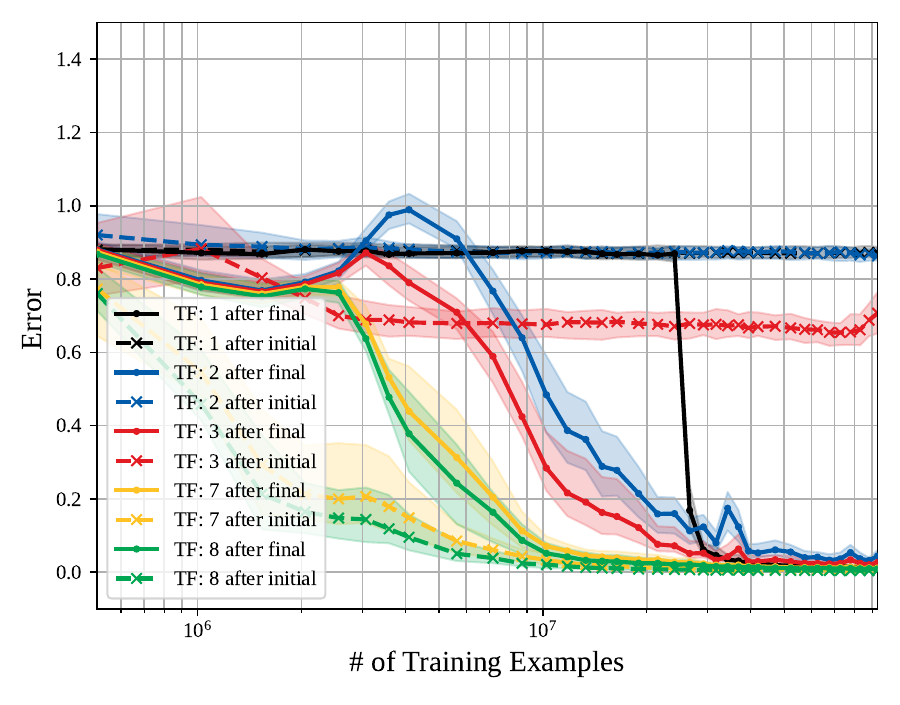}
        \caption{Orthogonal: 18 system haystack.}
        \label{fig:ortho_med_len_18_baseline_linear}
    \end{subfigure}
    \hfill
    \begin{subfigure}[b]{0.32\linewidth}
        \centering
        \includegraphics[width=\linewidth]{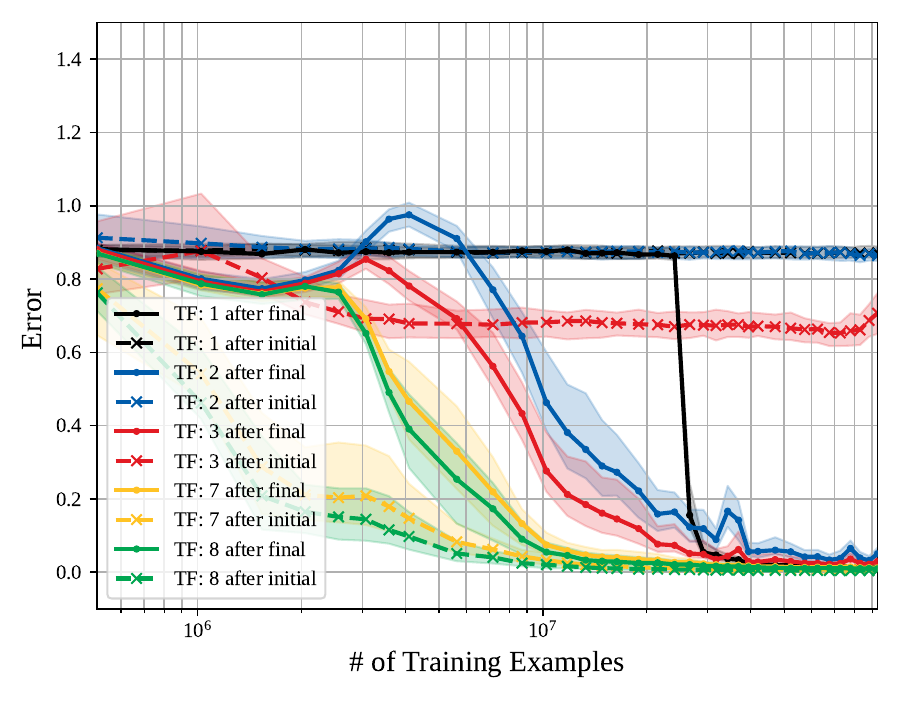}
        \caption{Orthogonal: 19 system haystack.}
        \label{fig:ortho_med_len_19_baseline_linear}
    \end{subfigure}
    \caption{Performance of medium orthogonal model (2.42M params) across training --- linear-scale.}
    \label{fig:ortho_med_baseline_linear}

\end{figure}

\begin{figure}[htbp]
    \centering
    \begin{subfigure}[b]{0.32\linewidth}
        \centering
        \includegraphics[width=\linewidth]{needle/train_conv/seg_len_10/ortho_haar_med/ortho_haar_embd_dim_128_train_conv_haystack_len_1_20250511_231940_logscale.pdf}
        \caption{Orthogonal: 1 system haystack.}
        \label{fig:ortho_med_len_1_baseline_log}
    \end{subfigure}
    \hfill
    \begin{subfigure}[b]{0.32\linewidth}
        \centering
        \includegraphics[width=\linewidth]{needle/train_conv/seg_len_10/ortho_haar_med/ortho_haar_embd_dim_128_train_conv_haystack_len_2_20250511_231941_logscale.pdf}
        \caption{Orthogonal: 2 system haystack.}
        \label{fig:ortho_med_len_2_baseline_log}
    \end{subfigure}
    \hfill
    \begin{subfigure}[b]{0.32\linewidth}
        \centering
        \includegraphics[width=\linewidth]{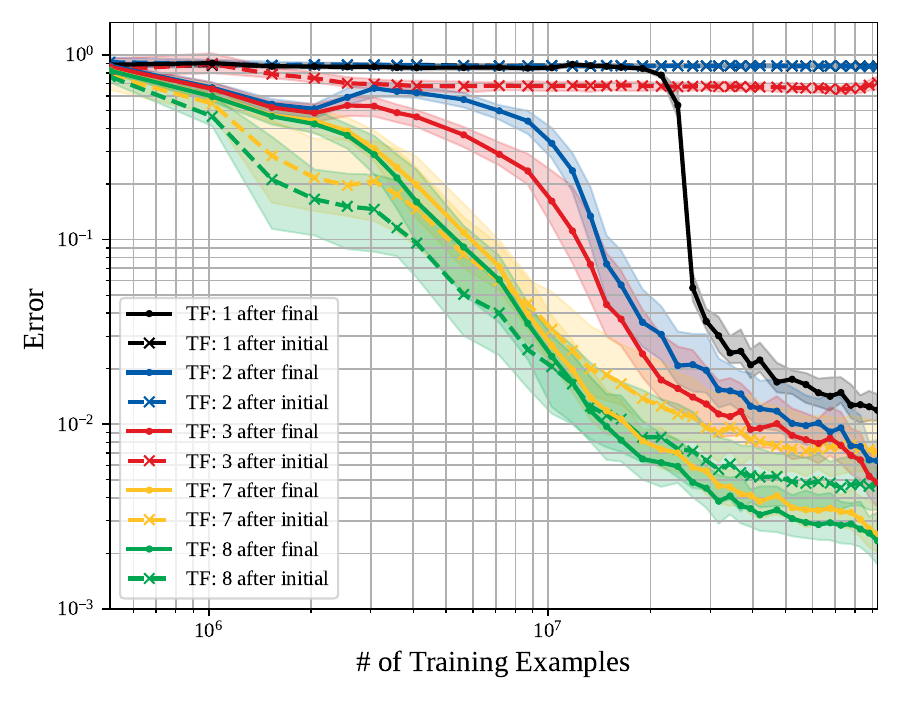}
        \caption{Orthogonal: 3 system haystack.}
        \label{fig:ortho_med_len_3_baseline_log}
    \end{subfigure}

    \vspace{0.2cm}

    \centering
    \begin{subfigure}[b]{0.32\linewidth}
        \centering
        \includegraphics[width=\linewidth]{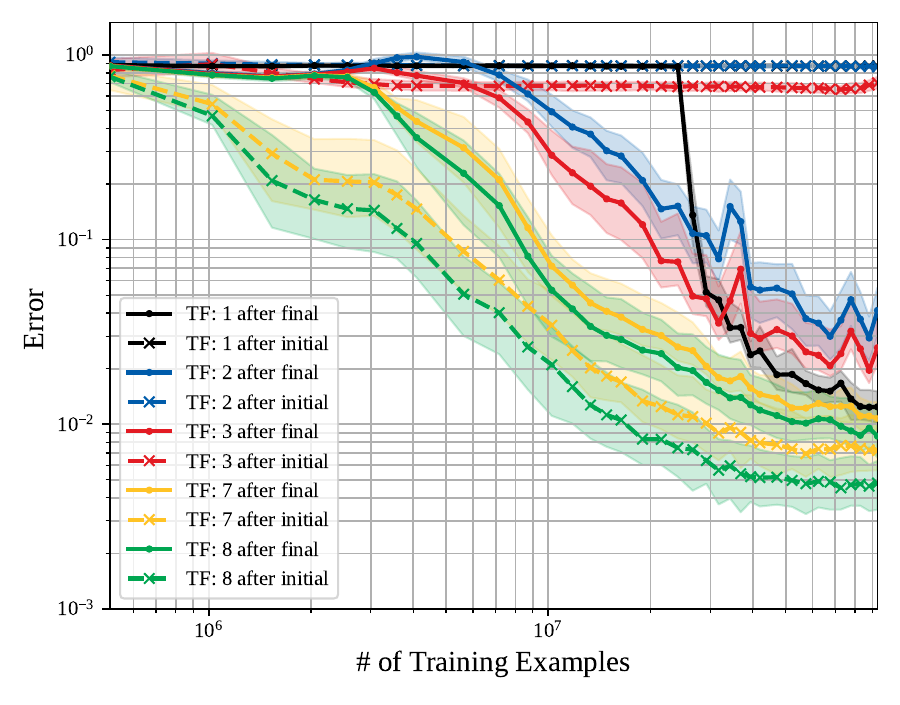}
        \caption{Orthogonal: 17 system haystack.}
        \label{fig:ortho_med_len_17_baseline_log}
    \end{subfigure}
    \hfill
    \begin{subfigure}[b]{0.32\linewidth}
        \centering
        \includegraphics[width=\linewidth]{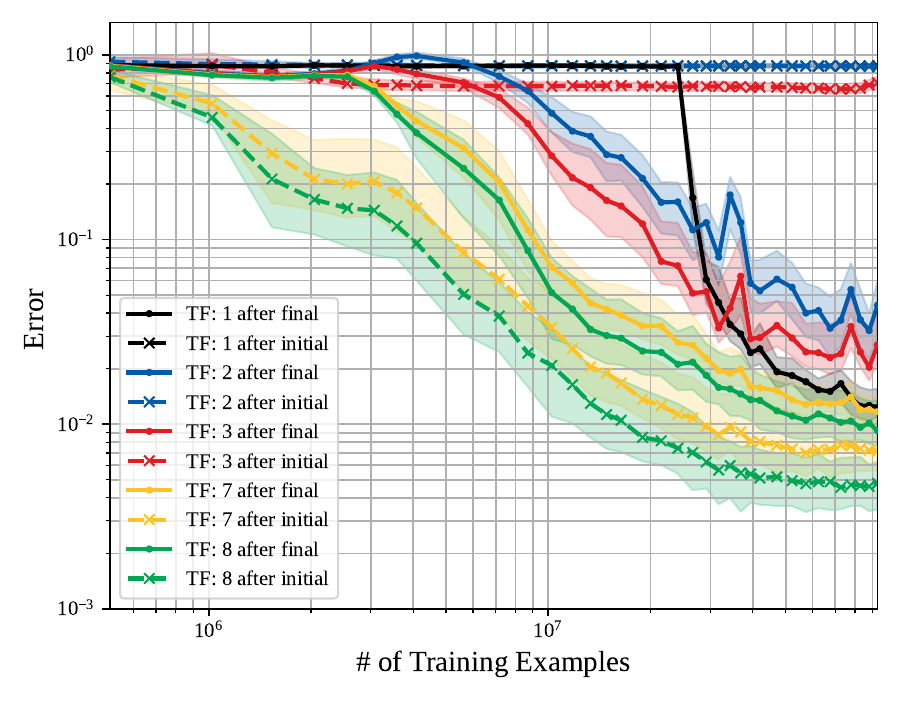}
        \caption{Orthogonal: 18 system haystack.}
        \label{fig:ortho_med_len_18_baseline_log}
    \end{subfigure}
    \hfill
    \begin{subfigure}[b]{0.32\linewidth}
        \centering
        \includegraphics[width=\linewidth]{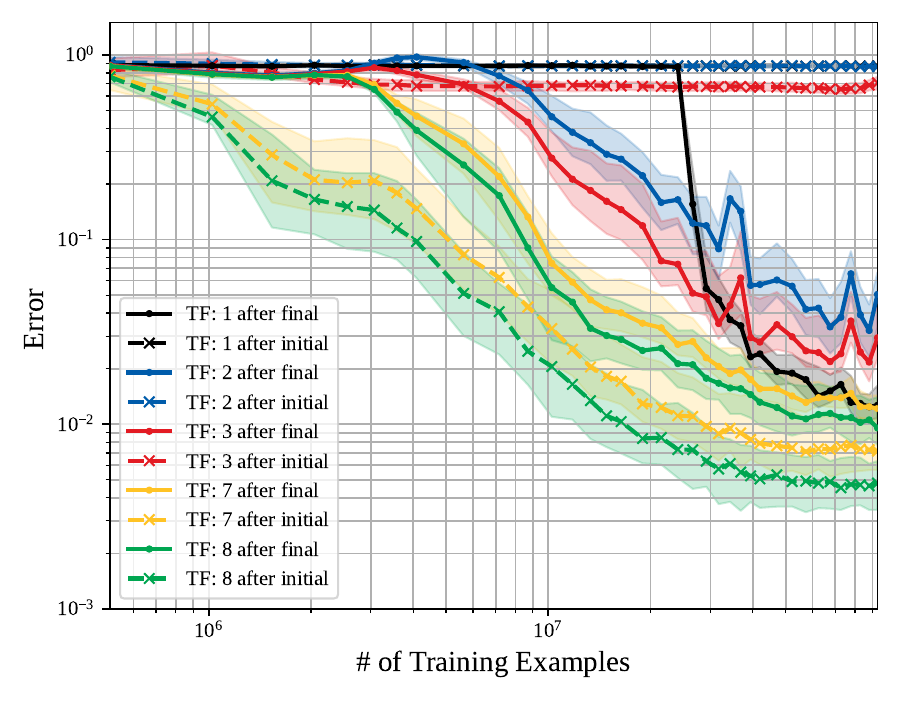}
        \caption{Orthogonal: 19 system haystack.}
        \label{fig:ortho_med_len_19_baseline_log}
    \end{subfigure}
    \caption{Performance of medium orthogonal model (2.42M params) across training --- log-scale.}
    \label{fig:ortho_med_baseline_log}

\end{figure}

\begin{figure}[htbp]
    \centering
    \begin{subfigure}[b]{0.32\linewidth}
        \centering
        \includegraphics[width=\linewidth]{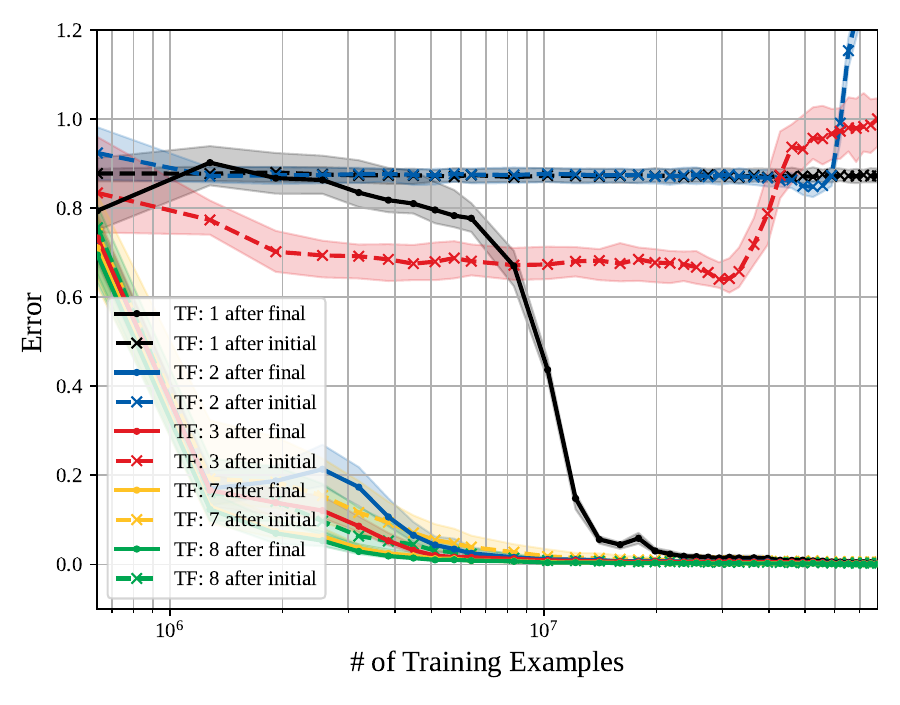}
        \caption{Orthogonal: 1 system haystack.}
        \label{fig:ortho_big_len_1_baseline_linear}
    \end{subfigure}
    \hfill
    \begin{subfigure}[b]{0.32\linewidth}
        \centering
        \includegraphics[width=\linewidth]{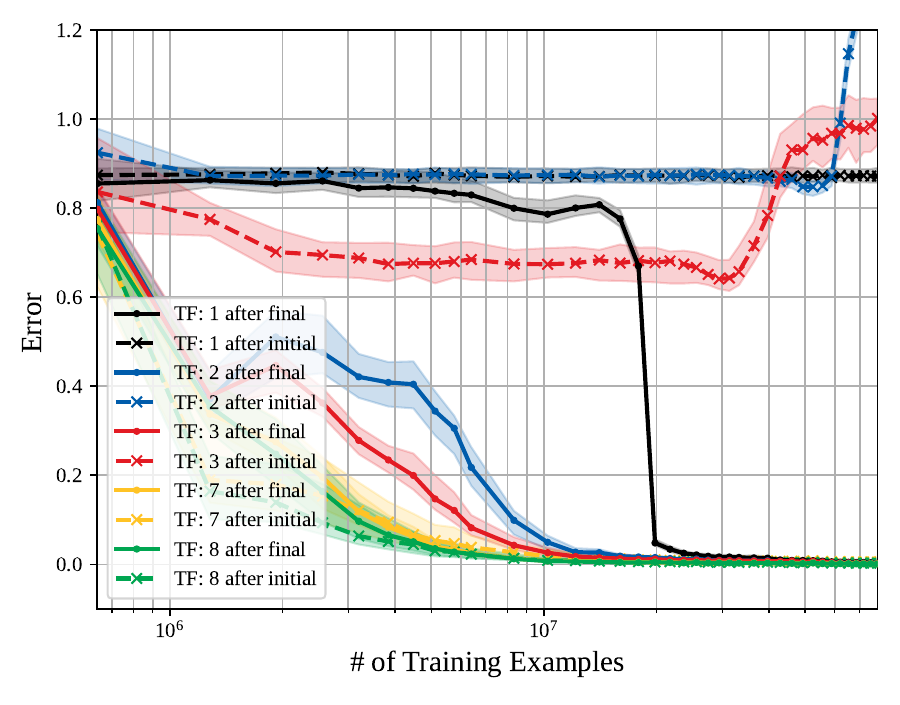}
        \caption{Orthogonal: 2 system haystack.}
        \label{fig:ortho_big_len_2_baseline_linear}
    \end{subfigure}
    \hfill
    \begin{subfigure}[b]{0.32\linewidth}
        \centering
        \includegraphics[width=\linewidth]{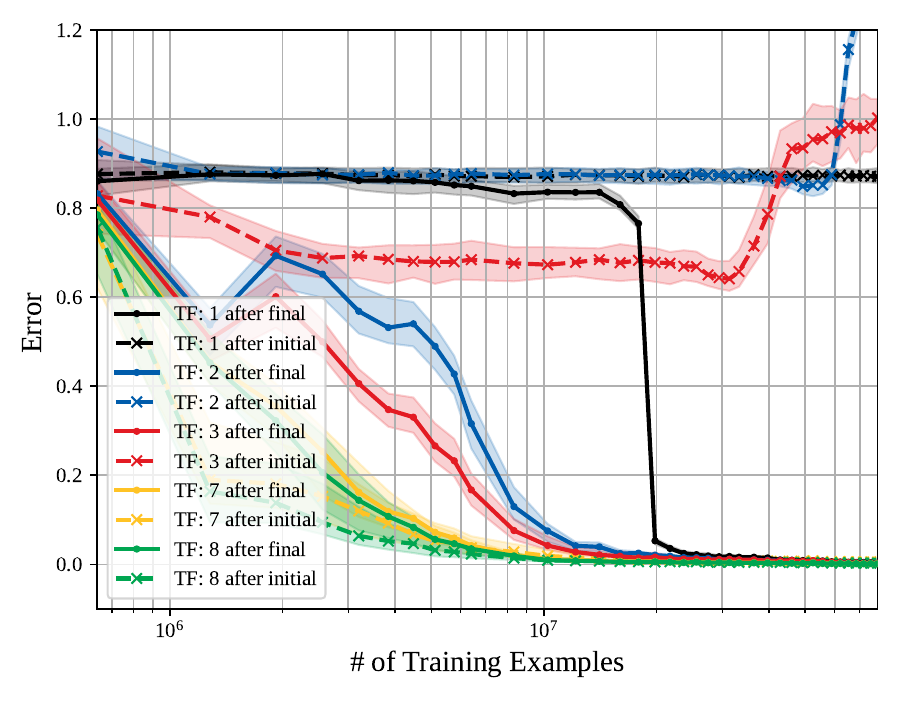}
        \caption{Orthogonal: 3 system haystack.}
        \label{fig:ortho_big_len_3_baseline_linear}
    \end{subfigure}

    \vspace{0.2cm}

    \centering
    \begin{subfigure}[b]{0.32\linewidth}
        \centering
        \includegraphics[width=\linewidth]{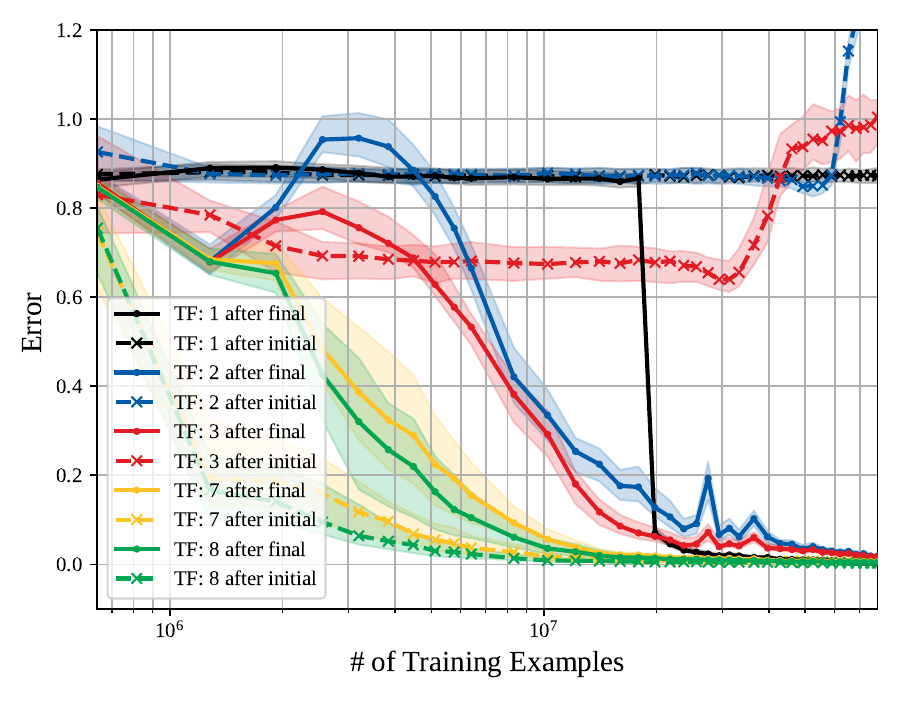}
        \caption{Orthogonal: 17 system haystack.}
        \label{fig:ortho_big_len_17_baseline_linear}
    \end{subfigure}
    \hfill
    \begin{subfigure}[b]{0.32\linewidth}
        \centering
        \includegraphics[width=\linewidth]{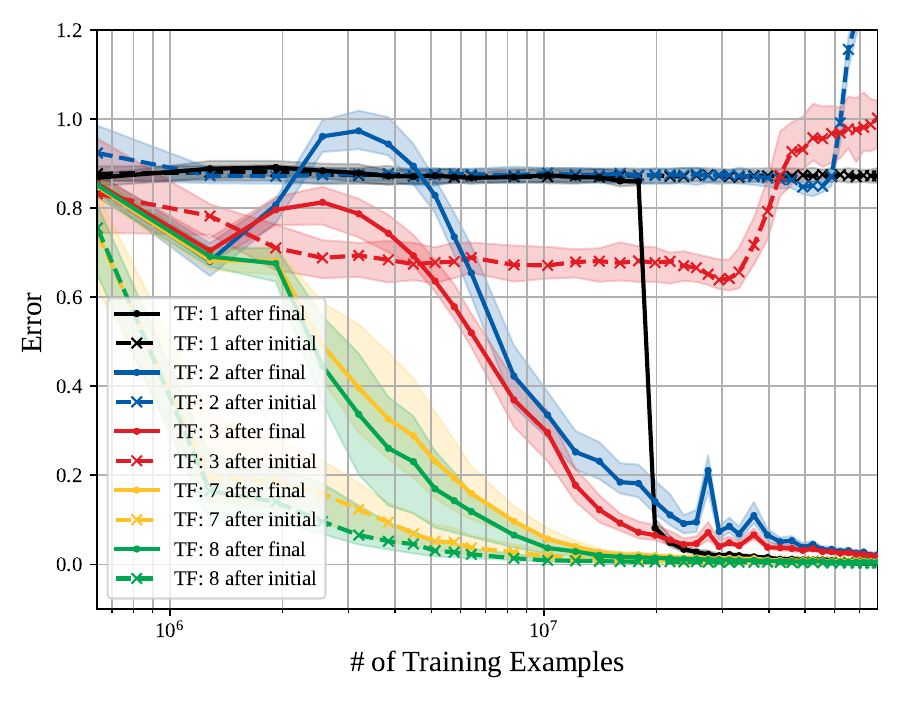}
        \caption{Orthogonal: 18 system haystack.}
        \label{fig:ortho_big_len_18_baseline_linear}
    \end{subfigure}
    \hfill
    \begin{subfigure}[b]{0.32\linewidth}
        \centering
        \includegraphics[width=\linewidth]{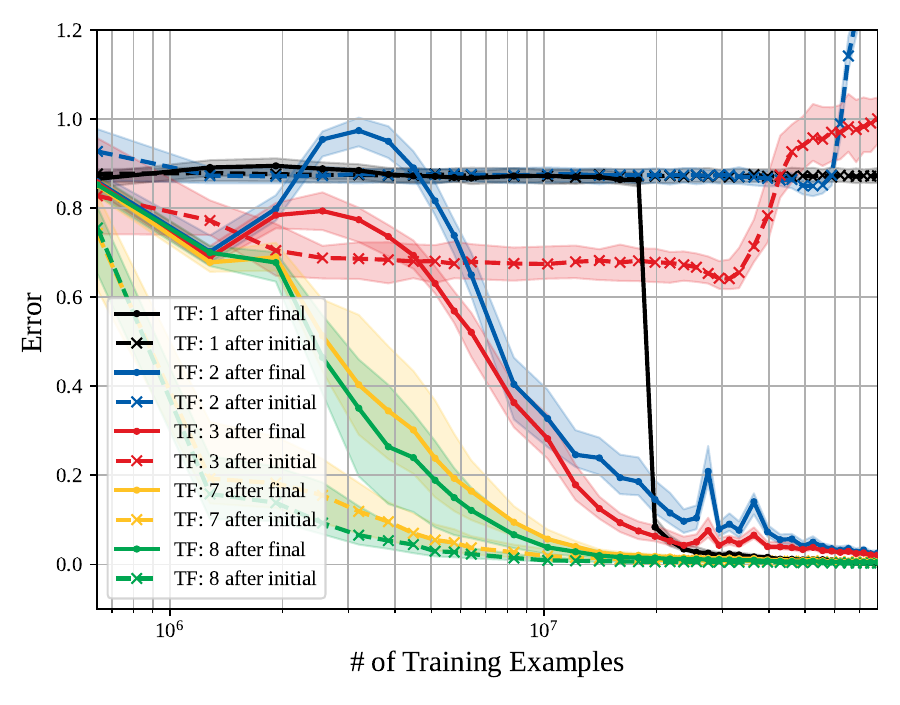}
        \caption{Orthogonal: 19 system haystack.}
        \label{fig:ortho_big_len_19_baseline_linear}
    \end{subfigure}
    \caption{Performance of big orthogonal model (10.7M params) across training --- linear-scale.}
    \label{fig:ortho_big_baseline_linear}

\end{figure}

\begin{figure}[htbp]
    \centering
    \begin{subfigure}[b]{0.32\linewidth}
        \centering
        \includegraphics[width=\linewidth]{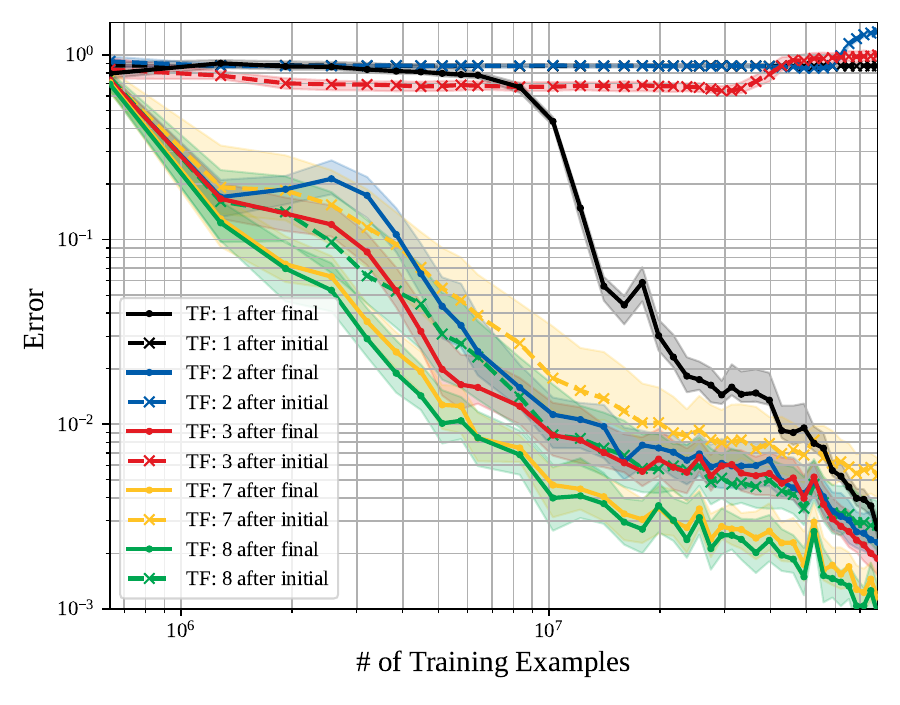}
        \caption{Orthogonal: 1 system haystack.}
        \label{fig:ortho_big_len_1_baseline_log}
    \end{subfigure}
    \hfill
    \begin{subfigure}[b]{0.32\linewidth}
        \centering
        \includegraphics[width=\linewidth]{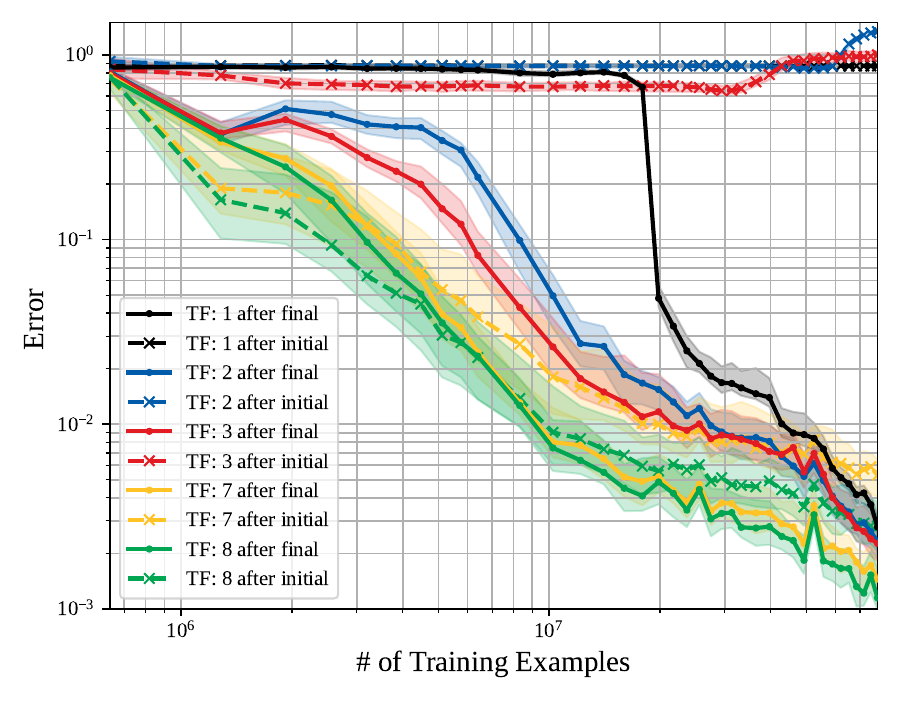}
        \caption{Orthogonal: 2 system haystack.}
        \label{fig:ortho_big_len_2_baseline_log}
    \end{subfigure}
    \hfill
    \begin{subfigure}[b]{0.32\linewidth}
        \centering
        \includegraphics[width=\linewidth]{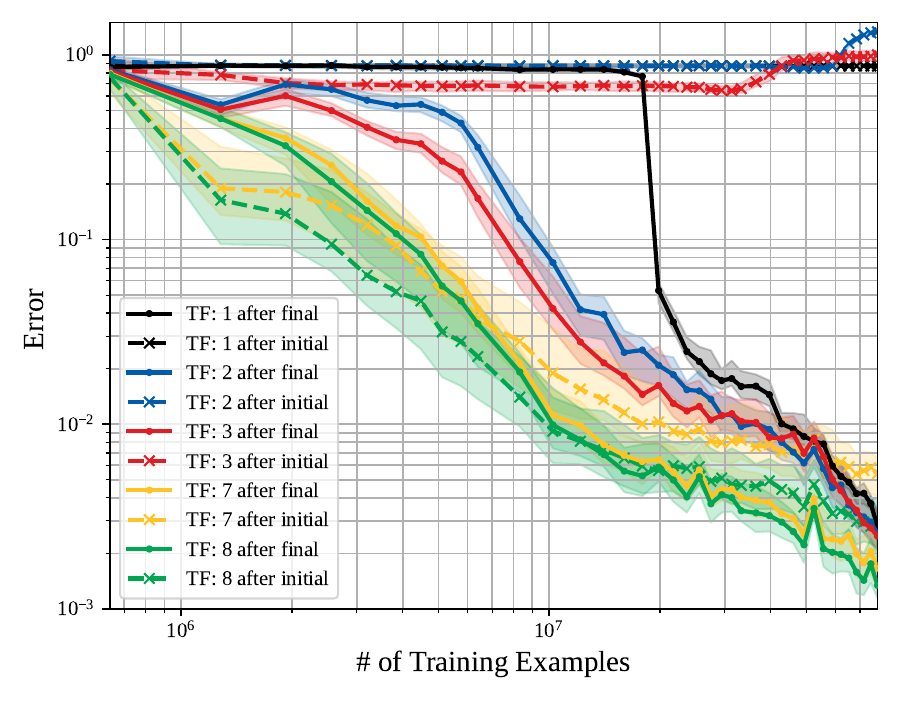}
        \caption{Orthogonal: 3 system haystack.}
        \label{fig:ortho_big_len_3_baseline_log}
    \end{subfigure}

    \vspace{0.2cm}

    \centering
    \begin{subfigure}[b]{0.32\linewidth}
        \centering
        \includegraphics[width=\linewidth]{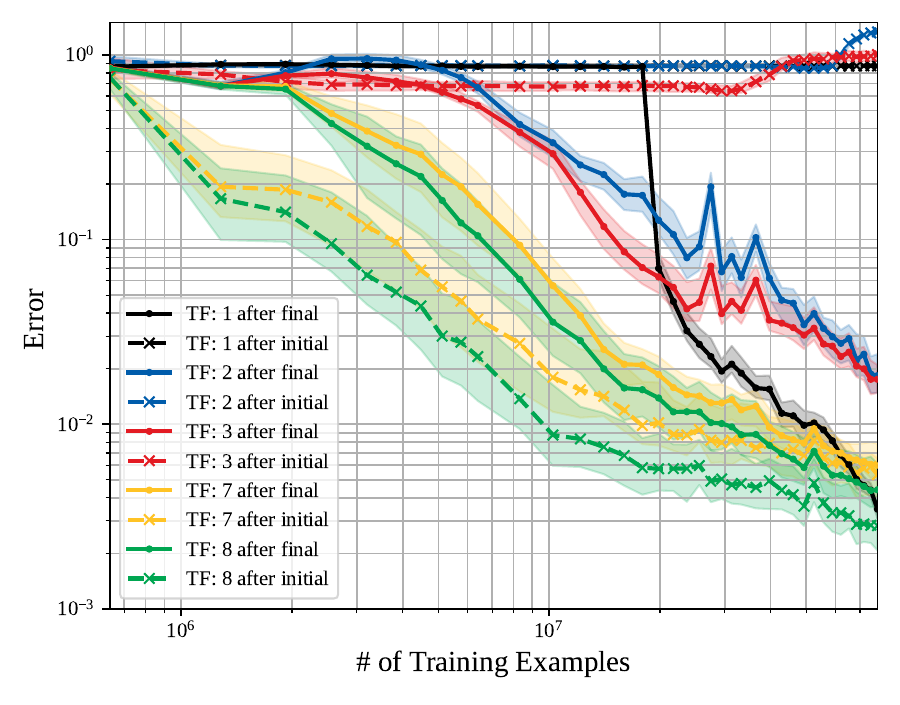}
        \caption{Orthogonal: 17 system haystack.}
        \label{fig:ortho_big_len_17_baseline_log}
    \end{subfigure}
    \hfill
    \begin{subfigure}[b]{0.32\linewidth}
        \centering
        \includegraphics[width=\linewidth]{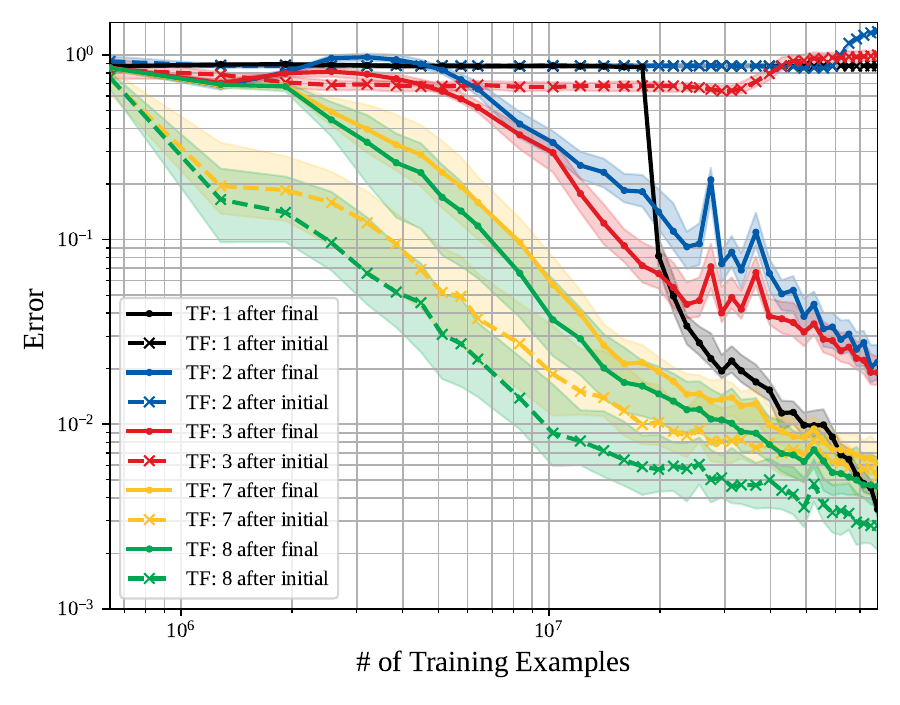}
        \caption{Orthogonal: 18 system haystack.}
        \label{fig:ortho_big_len_18_baseline_log}
    \end{subfigure}
    \hfill
    \begin{subfigure}[b]{0.32\linewidth}
        \centering
        \includegraphics[width=\linewidth]{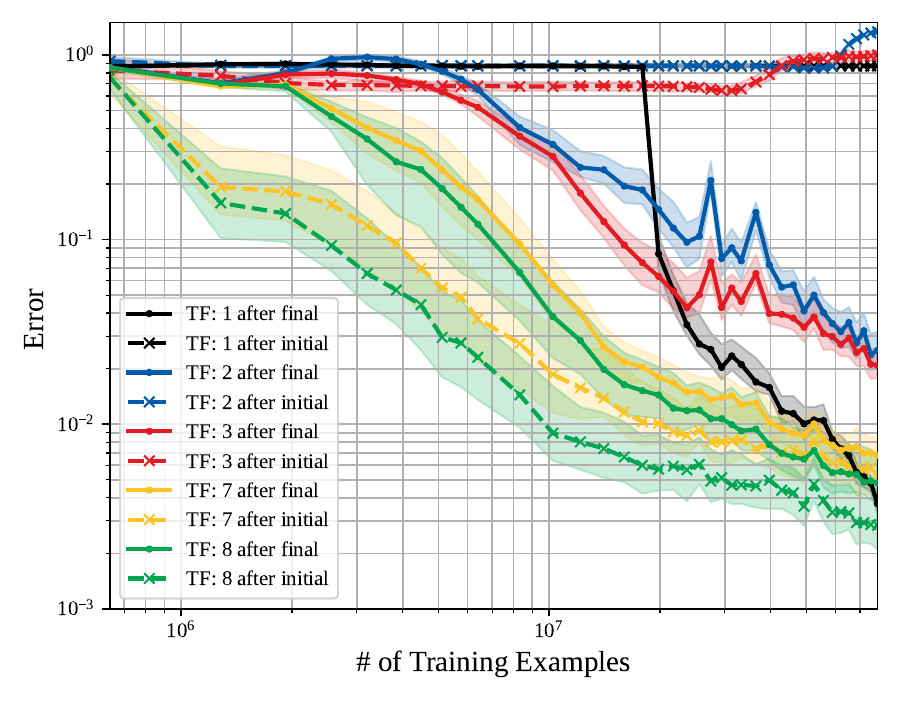}
        \caption{Orthogonal: 19 system haystack.}
        \label{fig:ortho_big_len_19_baseline_log}
    \end{subfigure}
    \caption{Performance of big orthogonal model (10.7M params) across training --- log-scale.}
    \label{fig:ortho_big_baseline_log}

\end{figure}

\begin{figure}[htbp]
    \centering
    \begin{subfigure}[b]{0.32\linewidth}
        \centering
        \includegraphics[width=\linewidth]{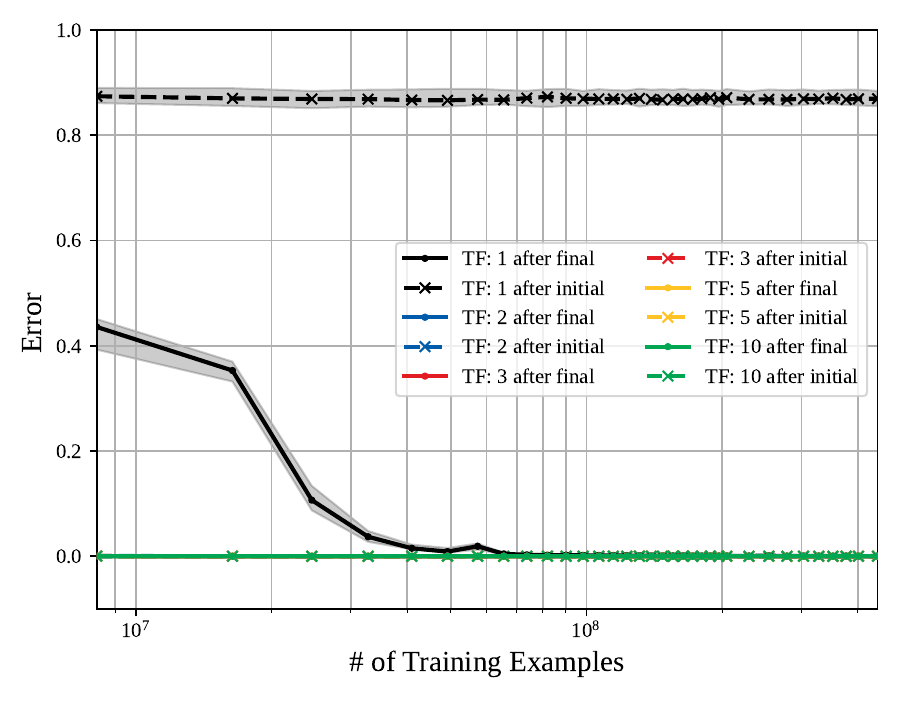}
        \caption{Identity: 1 system haystack.}
        \label{fig:ident_tiny_len_1_baseline_linear}
    \end{subfigure}
    \hfill
    \begin{subfigure}[b]{0.32\linewidth}
        \centering
        \includegraphics[width=\linewidth]{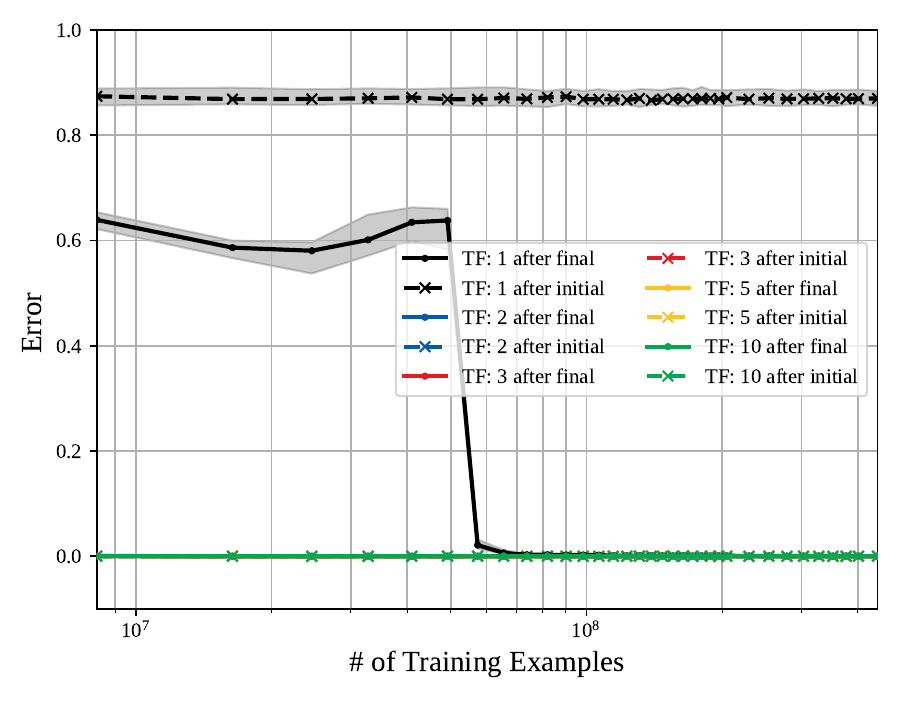}
        \caption{Identity: 2 system haystack.}
        \label{fig:ident_tiny_len_2_baseline_linear}
    \end{subfigure}
    \hfill
    \begin{subfigure}[b]{0.32\linewidth}
        \centering
        \includegraphics[width=\linewidth]{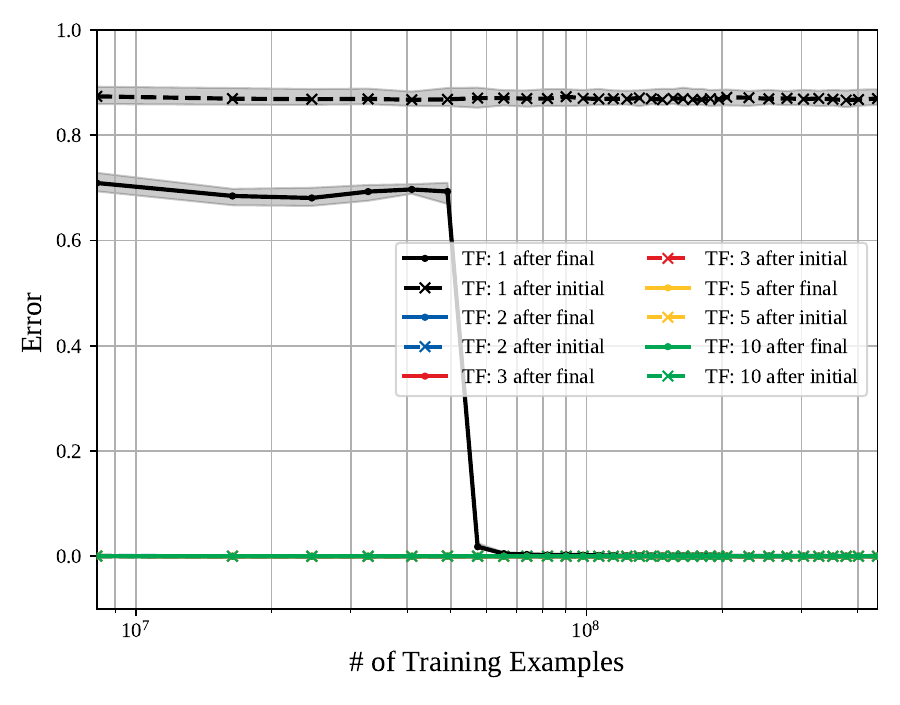}
        \caption{Identity: 3 system haystack.}
        \label{fig:ident_tiny_len_3_baseline_linear}
    \end{subfigure}

    \vspace{0.2cm}

    \centering
    \begin{subfigure}[b]{0.32\linewidth}
        \centering
        \includegraphics[width=\linewidth]{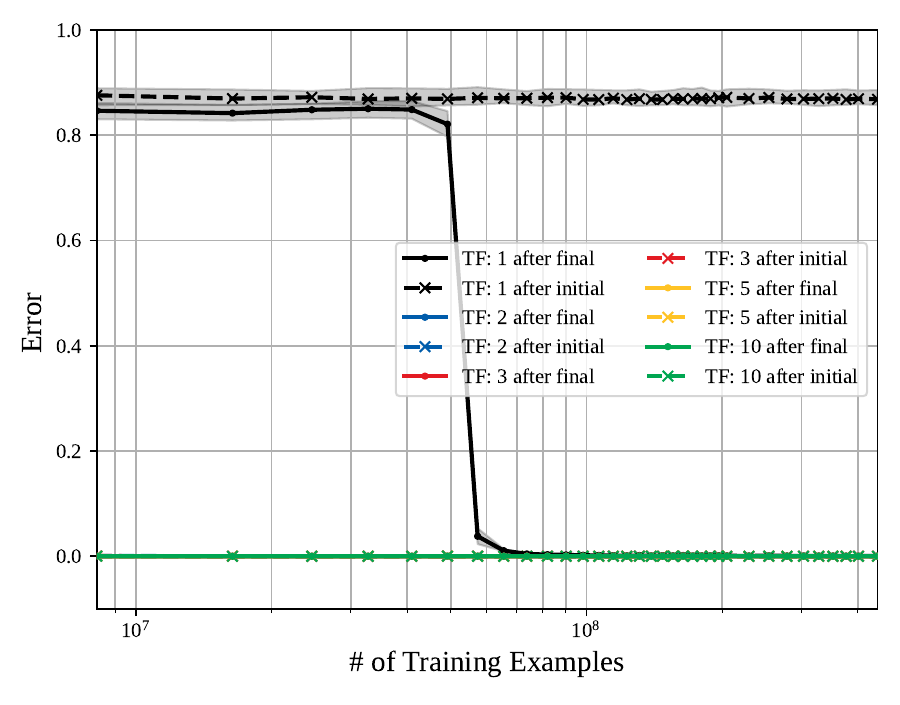}
        \caption{Identity: 17 system haystack.}
        \label{fig:ident_tiny_len_17_baseline_linear}
    \end{subfigure}
    \hfill
    \begin{subfigure}[b]{0.32\linewidth}
        \centering
        \includegraphics[width=\linewidth]{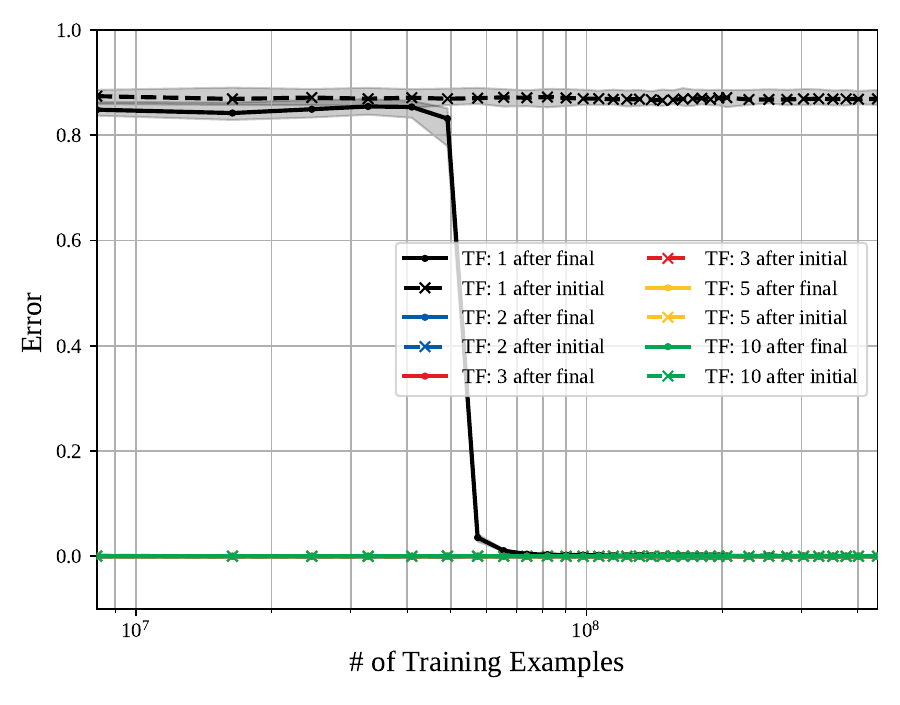}
        \caption{Identity: 18 system haystack.}
        \label{fig:ident_tiny_len_18_baseline_linear}
    \end{subfigure}
    \hfill
    \begin{subfigure}[b]{0.32\linewidth}
        \centering
        \includegraphics[width=\linewidth]{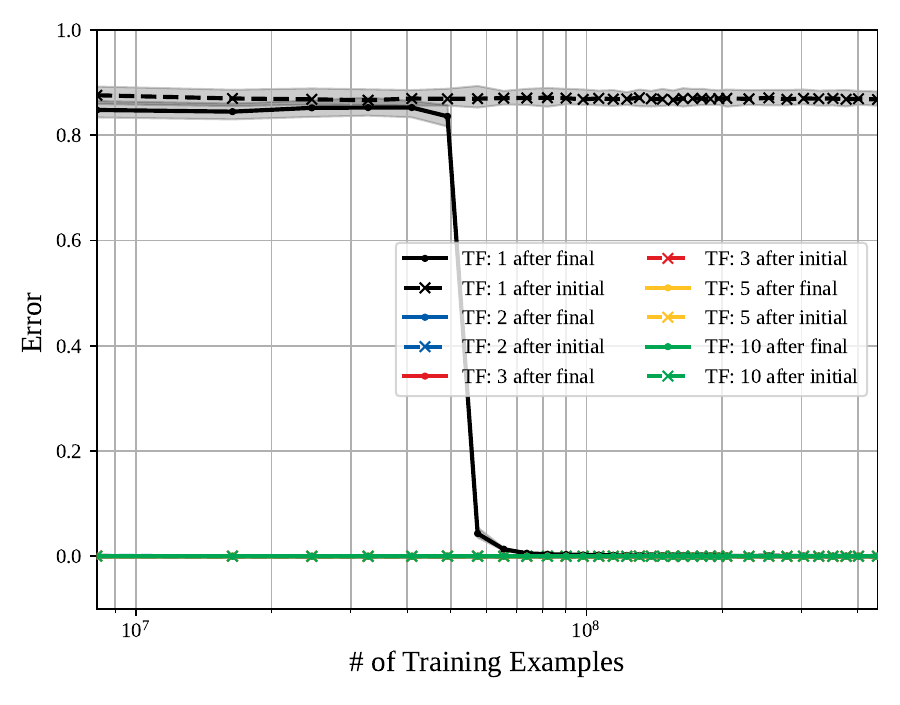}
        \caption{Identity: 19 system haystack.}
        \label{fig:ident_tiny_len_19_baseline_linear}
    \end{subfigure}
    \caption{Performance of tiny identity model (212K params) across training --- linear-scale.}
    \label{fig:ident_tiny_baseline_linear}

\end{figure}

\begin{figure}[htbp]
    \centering
    \begin{subfigure}[b]{0.32\linewidth}
        \centering
        \includegraphics[width=\linewidth]{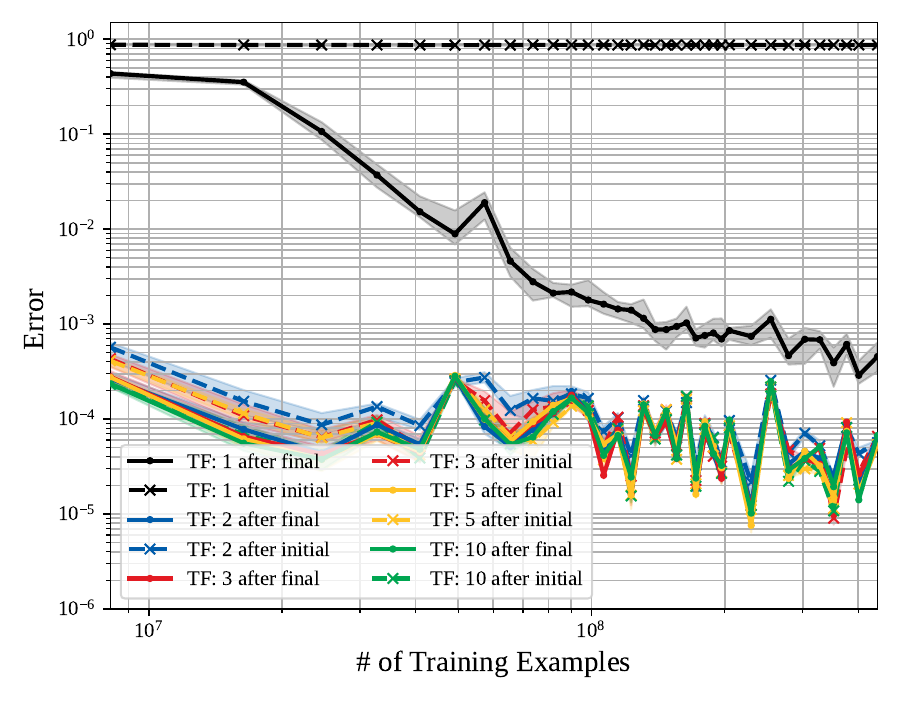}
        \caption{Identity: 1 system haystack.}
        \label{fig:ident_tiny_len_1_baseline_log}
    \end{subfigure}
    \hfill
    \begin{subfigure}[b]{0.32\linewidth}
        \centering
        \includegraphics[width=\linewidth]{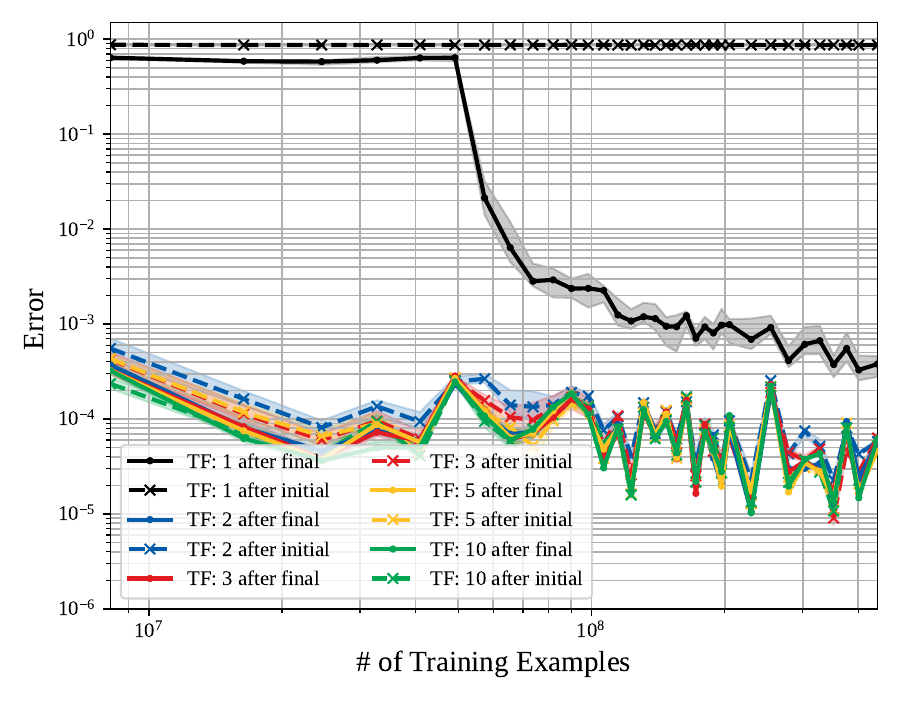}
        \caption{Identity: 2 system haystack.}
        \label{fig:ident_tiny_len_2_baseline_log}
    \end{subfigure}
    \hfill
    \begin{subfigure}[b]{0.32\linewidth}
        \centering
        \includegraphics[width=\linewidth]{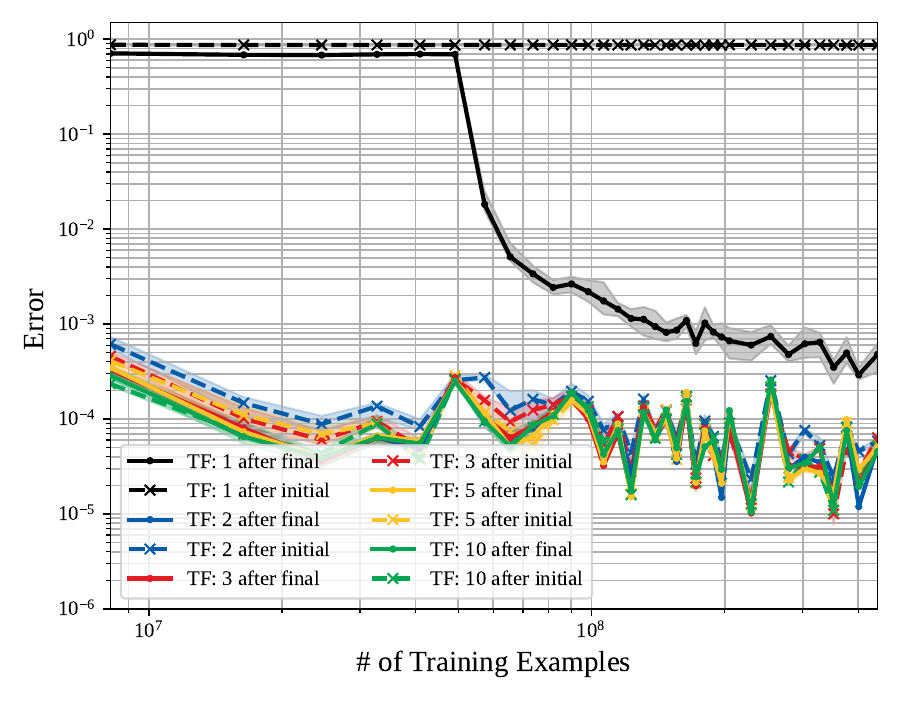}
        \caption{Identity: 3 system haystack.}
        \label{fig:ident_tiny_len_3_baseline_log}
    \end{subfigure}

    \vspace{0.2cm}

    \centering
    \begin{subfigure}[b]{0.32\linewidth}
        \centering
        \includegraphics[width=\linewidth]{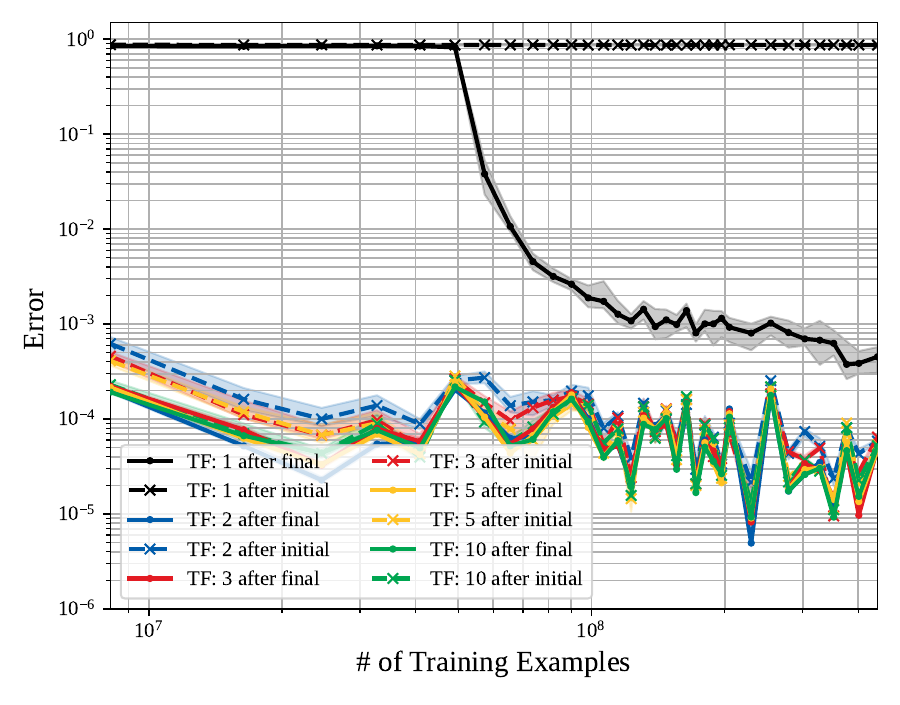}
        \caption{Identity: 17 system haystack.}
        \label{fig:ident_tiny_len_17_baseline_log}
    \end{subfigure}
    \hfill
    \begin{subfigure}[b]{0.32\linewidth}
        \centering
        \includegraphics[width=\linewidth]{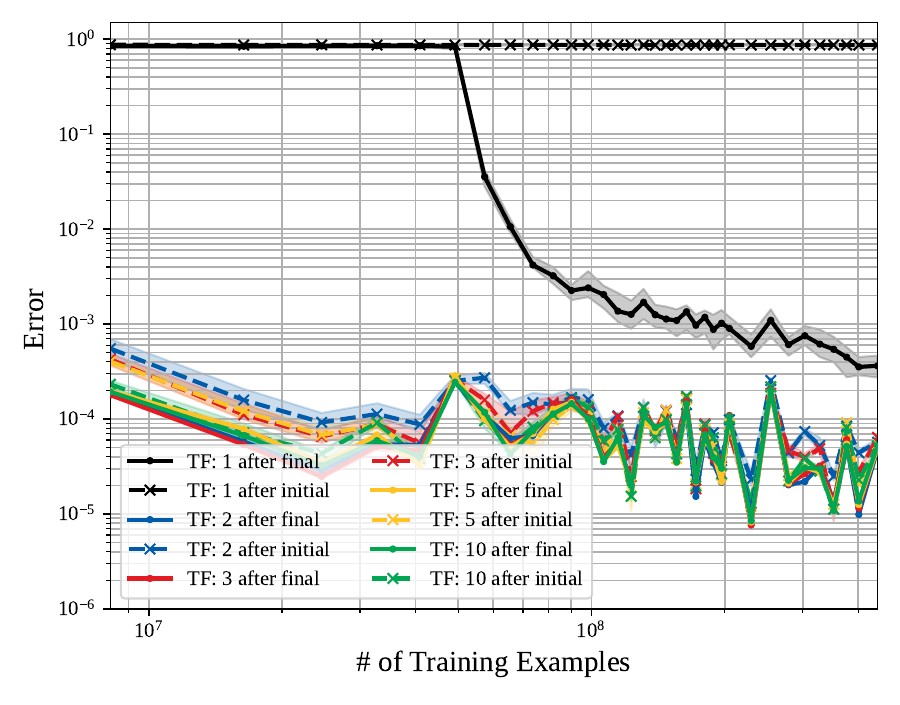}
        \caption{Identity: 18 system haystack.}
        \label{fig:ident_tiny_len_18_baseline_log}
    \end{subfigure}
    \hfill
    \begin{subfigure}[b]{0.32\linewidth}
        \centering
        \includegraphics[width=\linewidth]{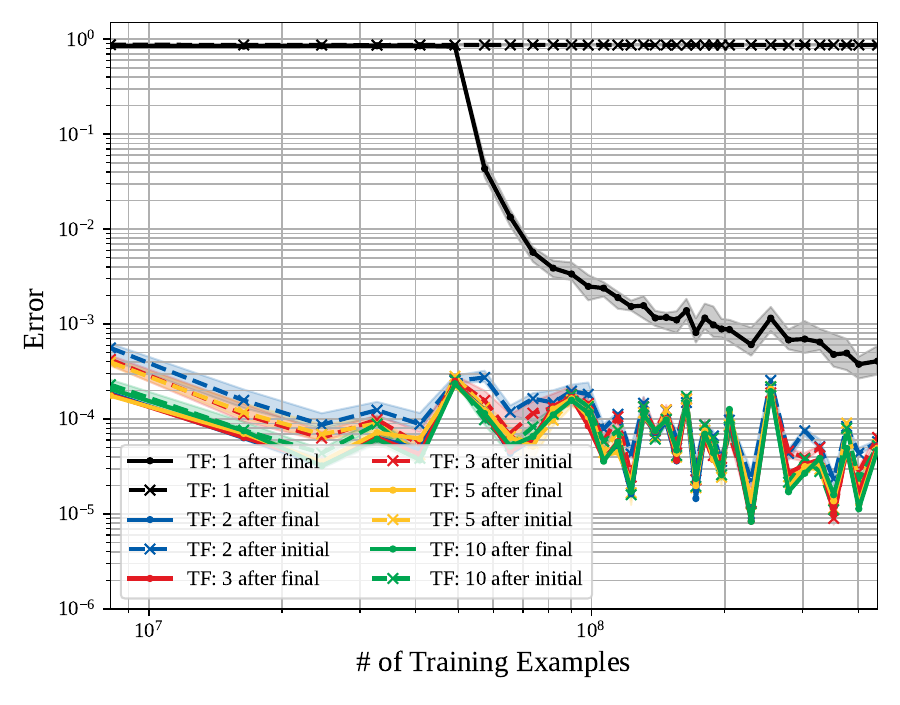}
        \caption{Identity: 19 system haystack.}
        \label{fig:ident_tiny_len_19_baseline_log}
    \end{subfigure}
    \caption{Performance of tiny identity model (212K params) across training --- log-scale.}
    \label{fig:ident_tiny_baseline_log}

\end{figure}

\begin{figure}[htbp]
    \centering
    \begin{subfigure}[b]{0.32\linewidth}
        \centering
        \includegraphics[width=\linewidth]{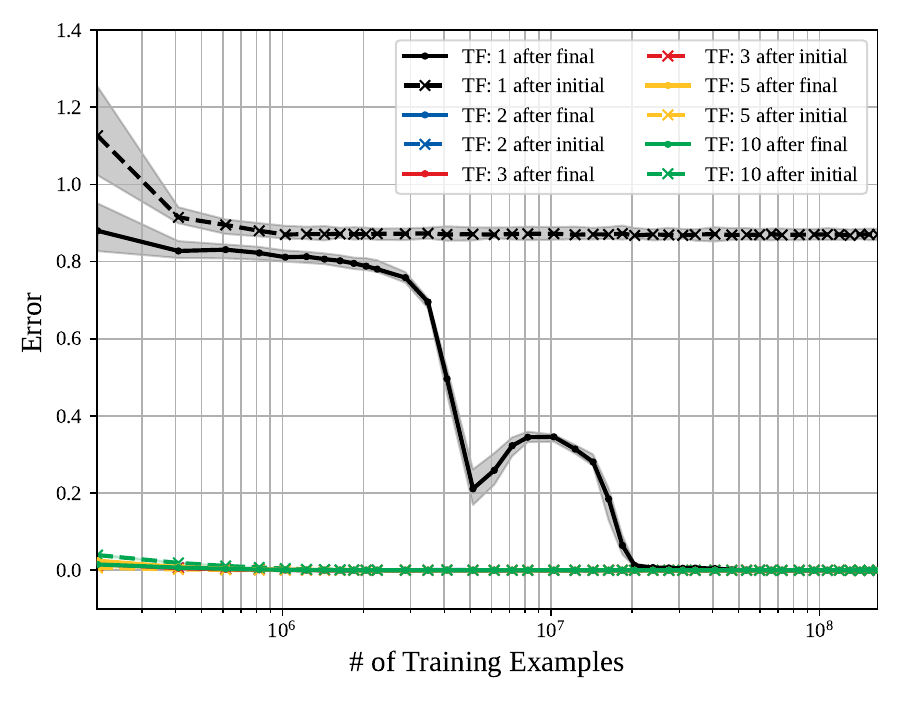}
        \caption{Identity: 1 system haystack.}
        \label{fig:ident_small_len_1_baseline_linear}
    \end{subfigure}
    \hfill
    \begin{subfigure}[b]{0.32\linewidth}
        \centering
        \includegraphics[width=\linewidth]{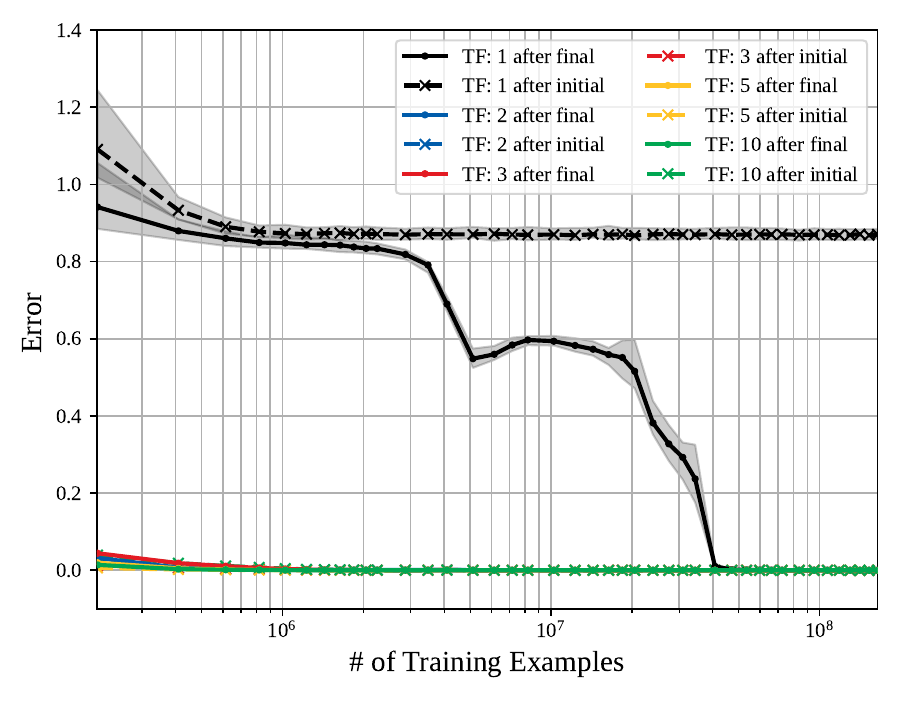}
        \caption{Identity: 2 system haystack.}
        \label{fig:ident_small_len_2_baseline_linear}
    \end{subfigure}
    \hfill
    \begin{subfigure}[b]{0.32\linewidth}
        \centering
        \includegraphics[width=\linewidth]{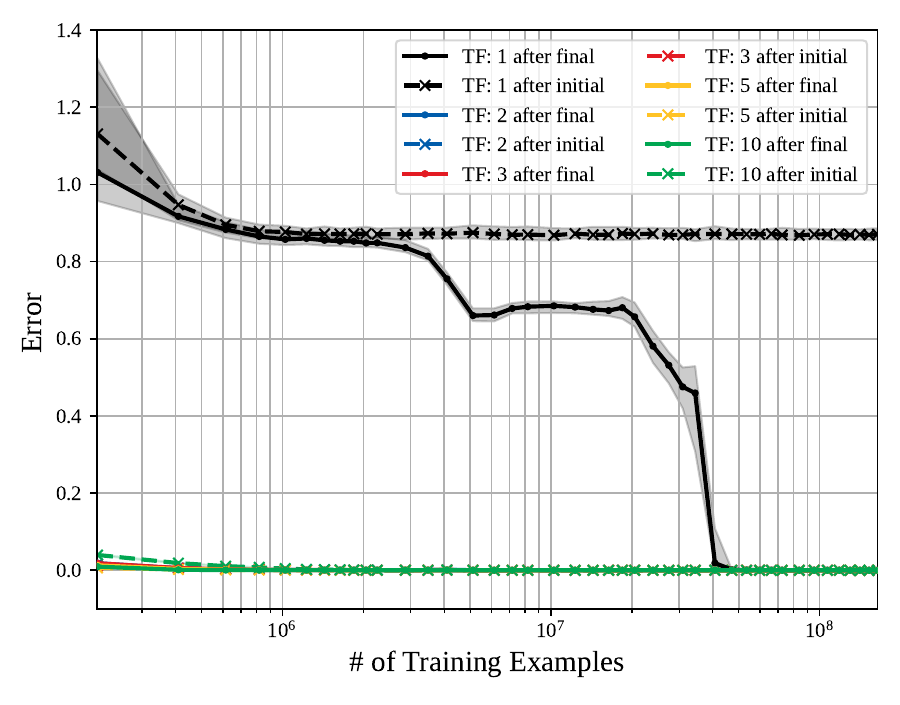}
        \caption{Identity: 3 system haystack.}
        \label{fig:ident_small_len_3_baseline_linear}
    \end{subfigure}

    \vspace{0.2cm}

    \centering
    \begin{subfigure}[b]{0.32\linewidth}
        \centering
        \includegraphics[width=\linewidth]{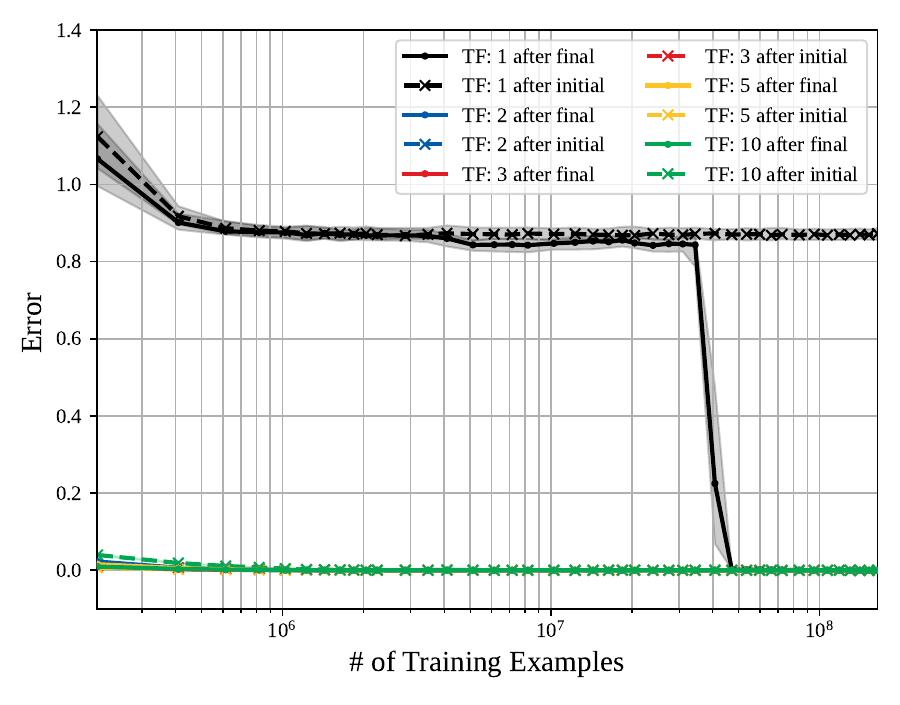}
        \caption{Identity: 17 system haystack.}
        \label{fig:ident_small_len_17_baseline_linear}
    \end{subfigure}
    \hfill
    \begin{subfigure}[b]{0.32\linewidth}
        \centering
        \includegraphics[width=\linewidth]{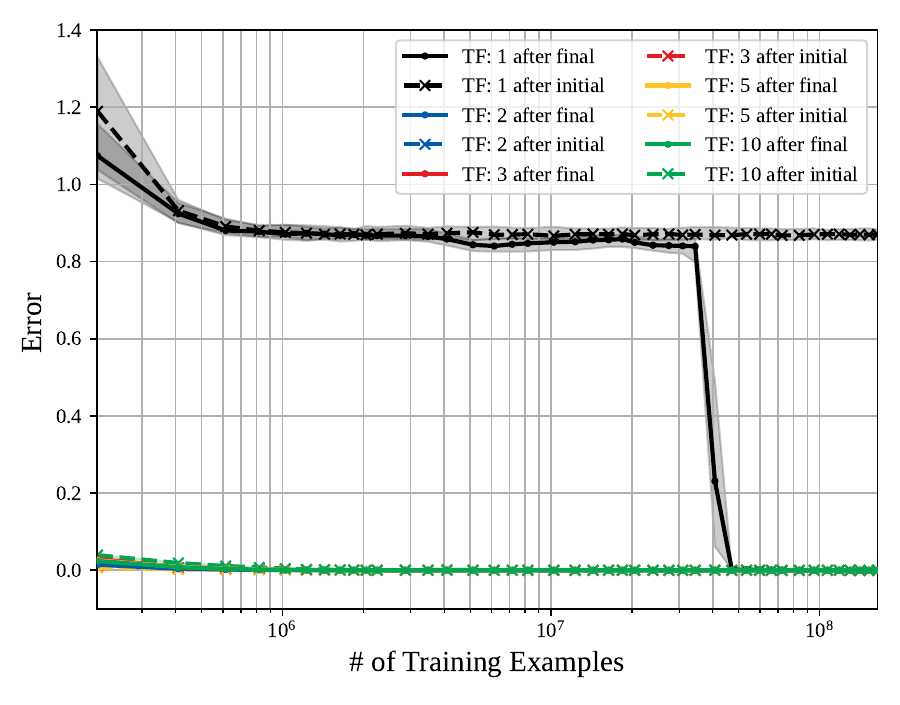}
        \caption{Identity: 18 system haystack.}
        \label{fig:ident_small_len_18_baseline_linear}
    \end{subfigure}
    \hfill
    \begin{subfigure}[b]{0.32\linewidth}
        \centering
        \includegraphics[width=\linewidth]{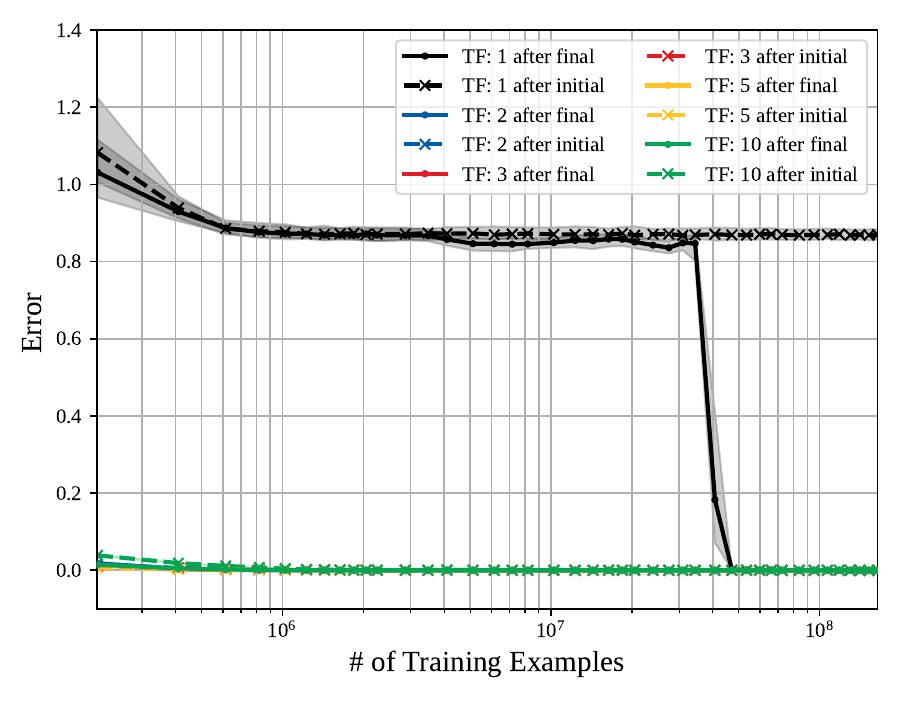}
        \caption{Identity: 19 system haystack.}
        \label{fig:ident_small_len_19_baseline_linear}
    \end{subfigure}
    \caption{Performance of small identity model (701K params) across training --- linear-scale.}
    \label{fig:ident_small_baseline_linear}

\end{figure}

\begin{figure}[htbp]
    \centering
    \begin{subfigure}[b]{0.32\linewidth}
        \centering
        \includegraphics[width=\linewidth]{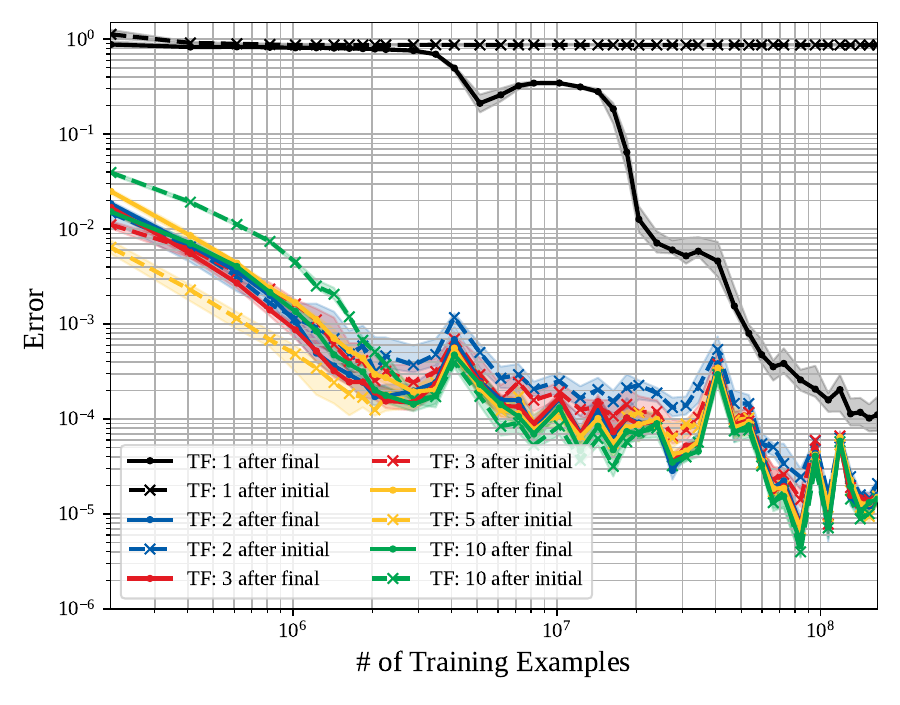}
        \caption{Identity: 1 system haystack.}
        \label{fig:ident_small_len_1_baseline_log}
    \end{subfigure}
    \hfill
    \begin{subfigure}[b]{0.32\linewidth}
        \centering
        \includegraphics[width=\linewidth]{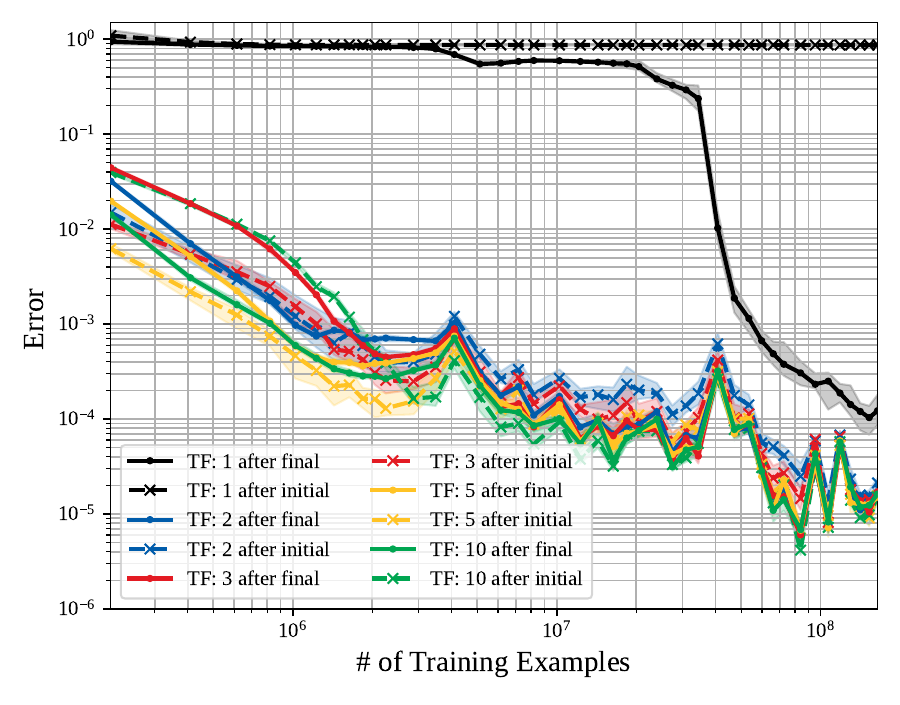}
        \caption{Identity: 2 system haystack.}
        \label{fig:ident_small_len_2_baseline_log}
    \end{subfigure}
    \hfill
    \begin{subfigure}[b]{0.32\linewidth}
        \centering
        \includegraphics[width=\linewidth]{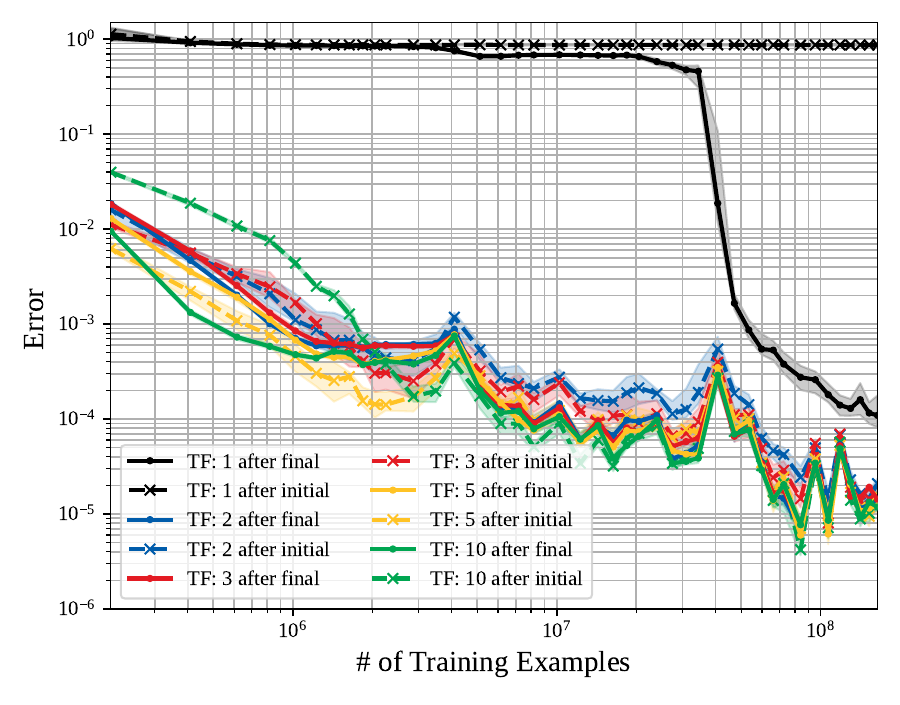}
        \caption{Identity: 3 system haystack.}
        \label{fig:ident_small_len_3_baseline_log}
    \end{subfigure}

    \vspace{0.2cm}

    \centering
    \begin{subfigure}[b]{0.32\linewidth}
        \centering
        \includegraphics[width=\linewidth]{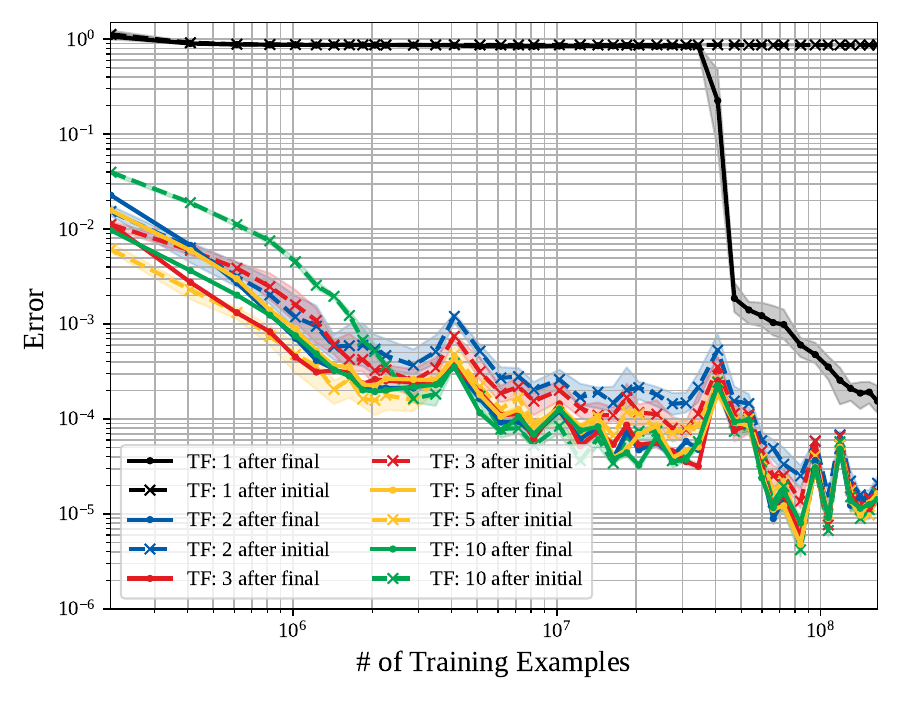}
        \caption{Identity: 17 system haystack.}
        \label{fig:ident_small_len_17_baseline_log}
    \end{subfigure}
    \hfill
    \begin{subfigure}[b]{0.32\linewidth}
        \centering
        \includegraphics[width=\linewidth]{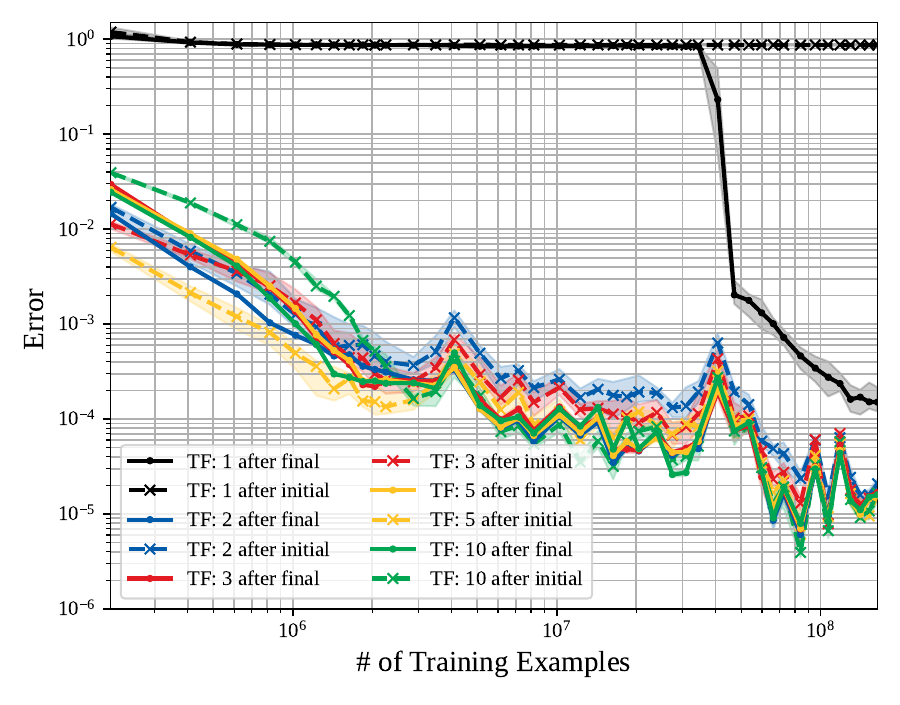}
        \caption{Identity: 18 system haystack.}
        \label{fig:ident_small_len_18_baseline_log}
    \end{subfigure}
    \hfill
    \begin{subfigure}[b]{0.32\linewidth}
        \centering
        \includegraphics[width=\linewidth]{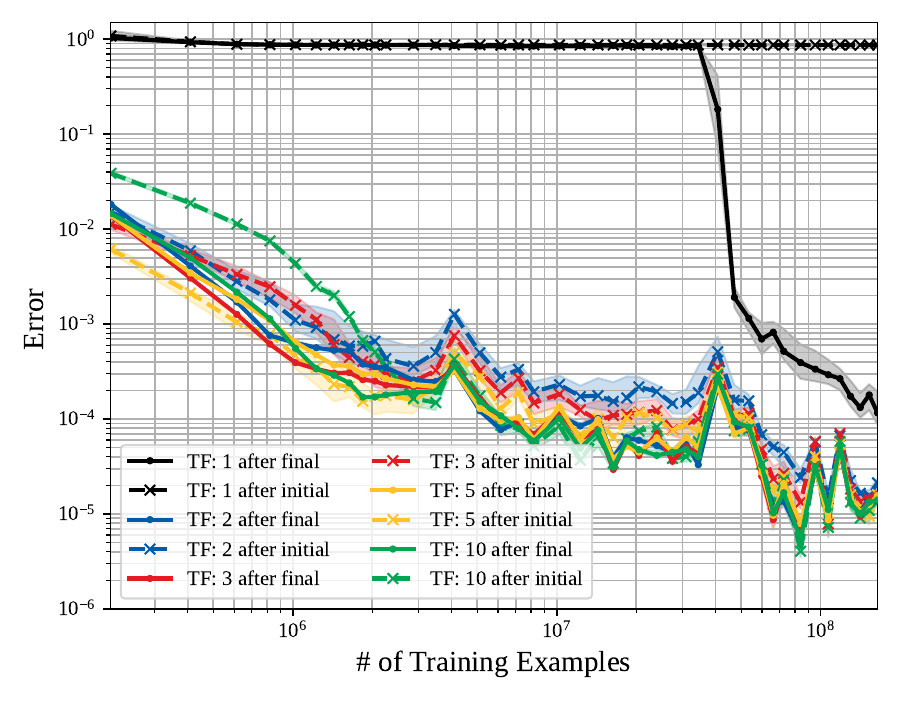}
        \caption{Identity: 19 system haystack.}
        \label{fig:ident_small_len_19_baseline_log}
    \end{subfigure}
    \caption{Performance of small identity model (701K params) across training --- log-scale.}
    \label{fig:ident_small_baseline_log}

\end{figure}

\begin{figure}[htbp]
    \centering
    \begin{subfigure}[b]{0.32\linewidth}
        \centering
        \includegraphics[width=\linewidth]{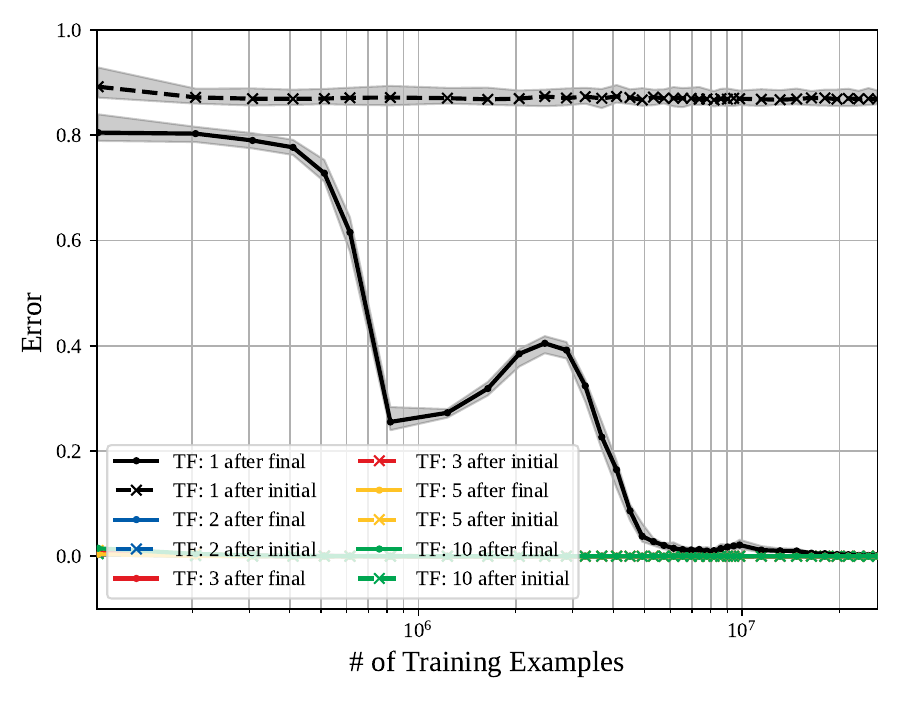}
        \caption{Identity: 1 system haystack.}
        \label{fig:ident_med_len_1_baseline_linear}
    \end{subfigure}
    \hfill
    \begin{subfigure}[b]{0.32\linewidth}
        \centering
        \includegraphics[width=\linewidth]{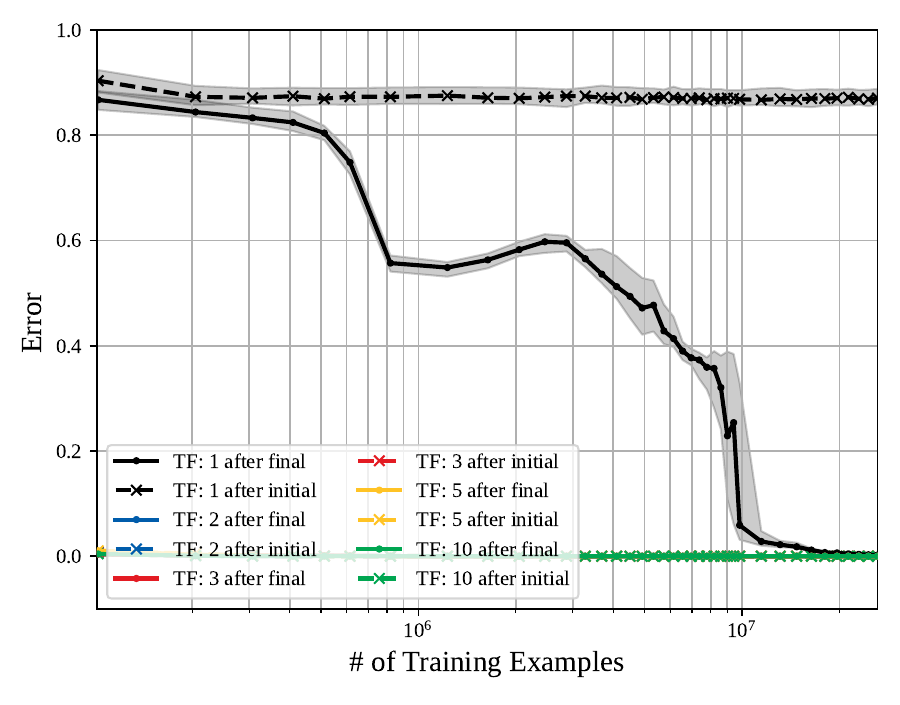}
        \caption{Identity: 2 system haystack.}
        \label{fig:ident_med_len_2_baseline_linear}
    \end{subfigure}
    \hfill
    \begin{subfigure}[b]{0.32\linewidth}
        \centering
        \includegraphics[width=\linewidth]{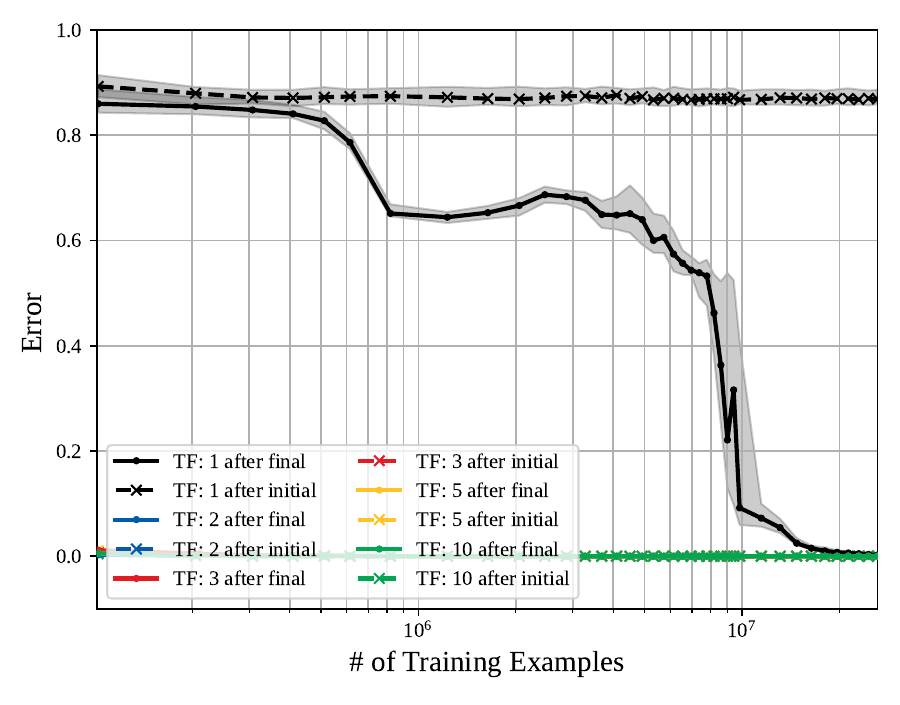}
        \caption{Identity: 3 system haystack.}
        \label{fig:ident_med_len_3_baseline_linear}
    \end{subfigure}

    \vspace{0.2cm}

    \centering
    \begin{subfigure}[b]{0.32\linewidth}
        \centering
        \includegraphics[width=\linewidth]{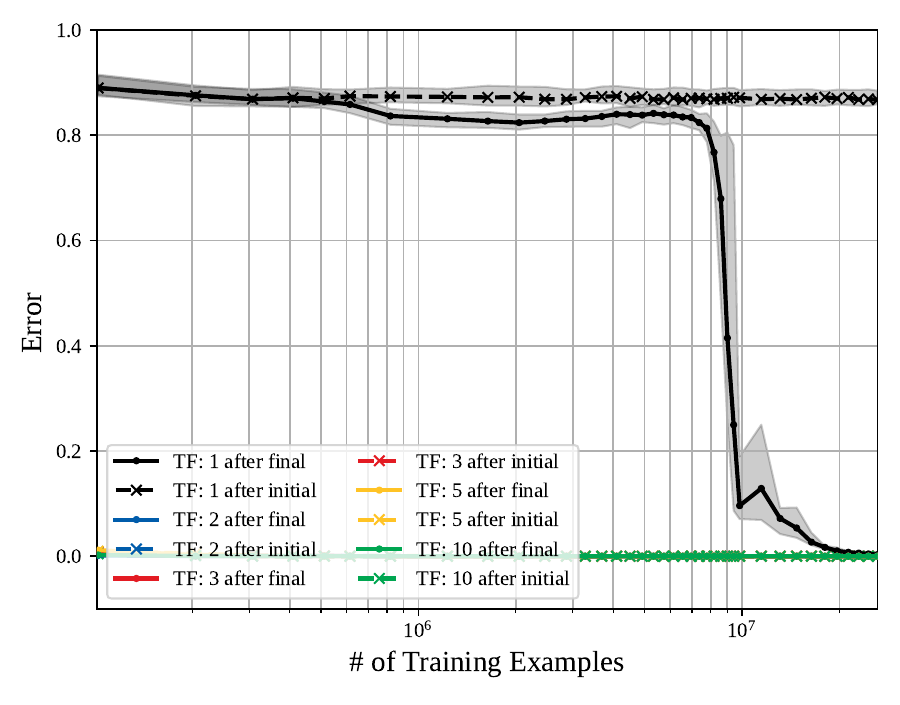}
        \caption{Identity: 17 system haystack.}
        \label{fig:ident_med_len_17_baseline_linear}
    \end{subfigure}
    \hfill
    \begin{subfigure}[b]{0.32\linewidth}
        \centering
        \includegraphics[width=\linewidth]{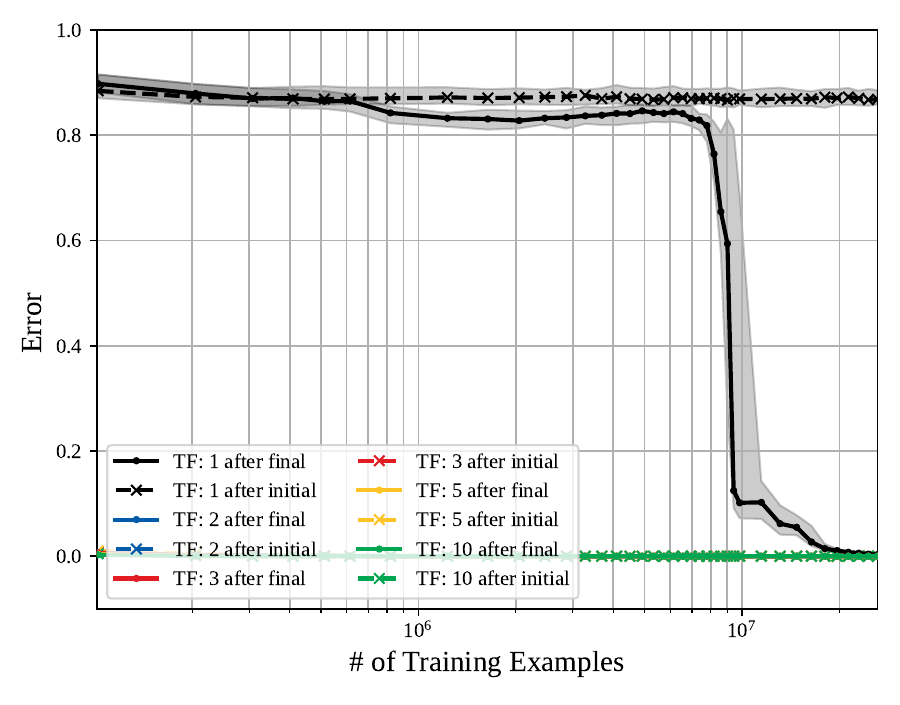}
        \caption{Identity: 18 system haystack.}
        \label{fig:ident_med_len_18_baseline_linear}
    \end{subfigure}
    \hfill
    \begin{subfigure}[b]{0.32\linewidth}
        \centering
        \includegraphics[width=\linewidth]{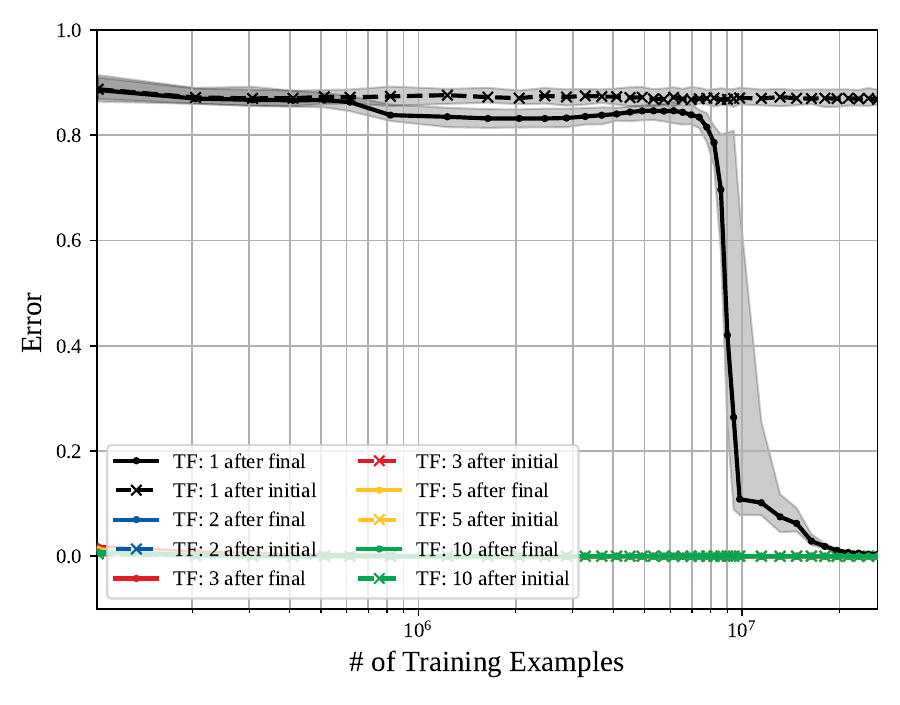}
        \caption{Identity: 19 system haystack.}
        \label{fig:ident_med_len_19_baseline_linear}
    \end{subfigure}
    \caption{Performance of medium identity model (2.42M params) across training --- linear-scale.}
    \label{fig:ident_med_baseline_linear}

\end{figure}

\begin{figure}[htbp]
    \centering
    \begin{subfigure}[b]{0.32\linewidth}
        \centering
        \includegraphics[width=\linewidth]{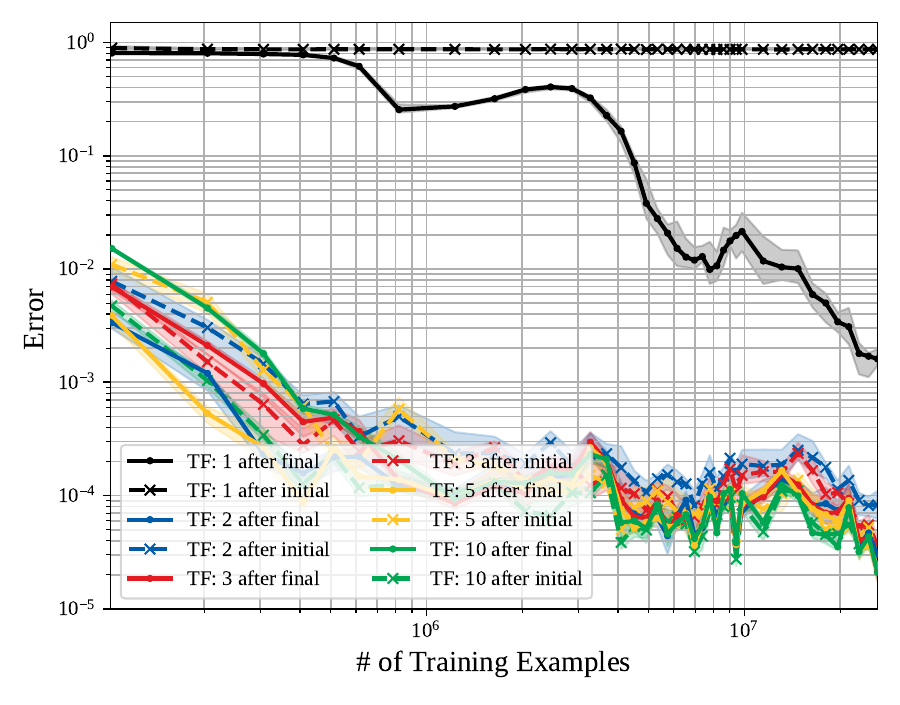}
        \caption{Identity: 1 system haystack.}
        \label{fig:ident_med_len_1_baseline_log}
    \end{subfigure}
    \hfill
    \begin{subfigure}[b]{0.32\linewidth}
        \centering
        \includegraphics[width=\linewidth]{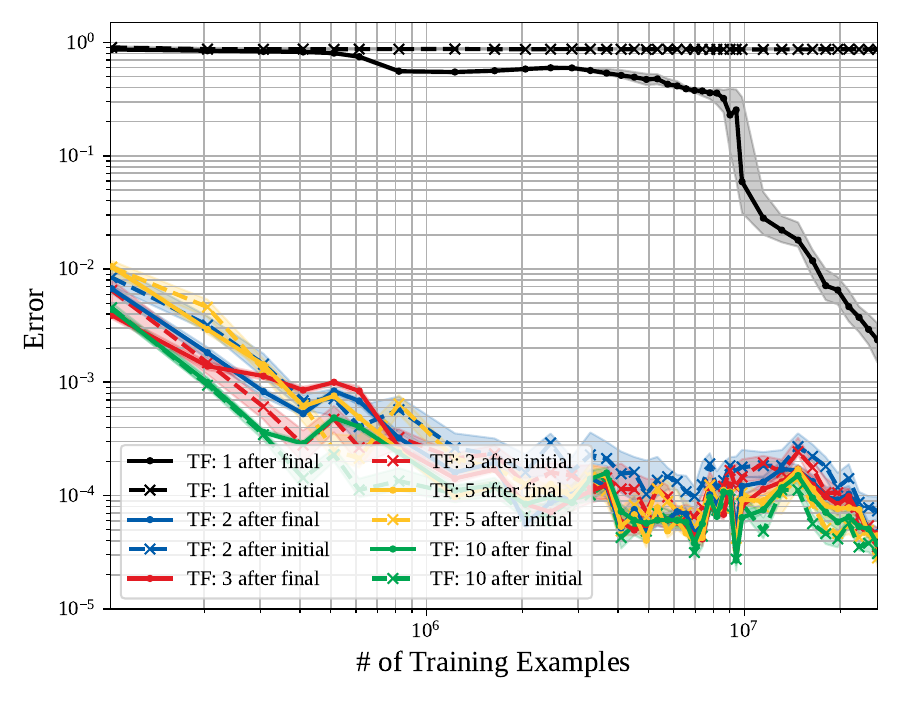}
        \caption{Identity: 2 system haystack.}
        \label{fig:ident_med_len_2_baseline_log}
    \end{subfigure}
    \hfill
    \begin{subfigure}[b]{0.32\linewidth}
        \centering
        \includegraphics[width=\linewidth]{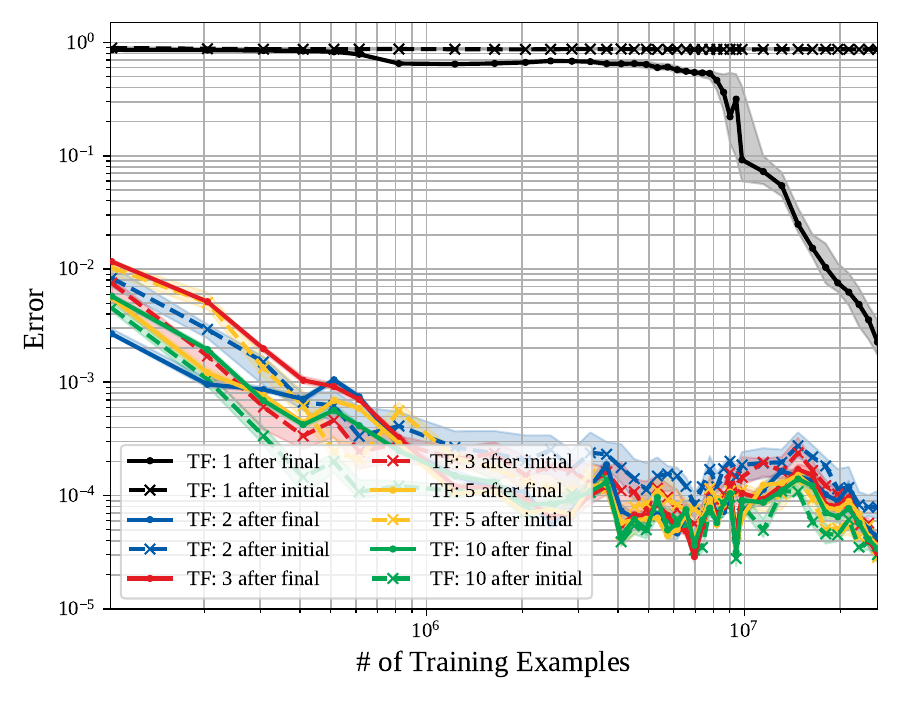}
        \caption{Identity: 3 system haystack.}
        \label{fig:ident_med_len_3_baseline_log}
    \end{subfigure}

    \vspace{0.2cm}

    \centering
    \begin{subfigure}[b]{0.32\linewidth}
        \centering
        \includegraphics[width=\linewidth]{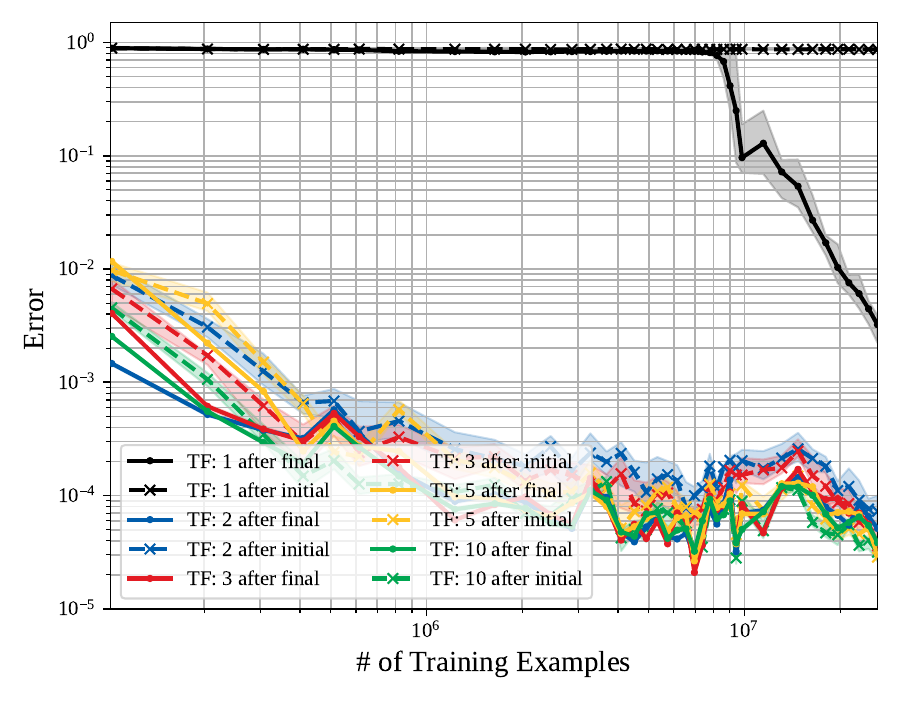}
        \caption{Identity: 17 system haystack.}
        \label{fig:ident_med_len_17_baseline_log}
    \end{subfigure}
    \hfill
    \begin{subfigure}[b]{0.32\linewidth}
        \centering
        \includegraphics[width=\linewidth]{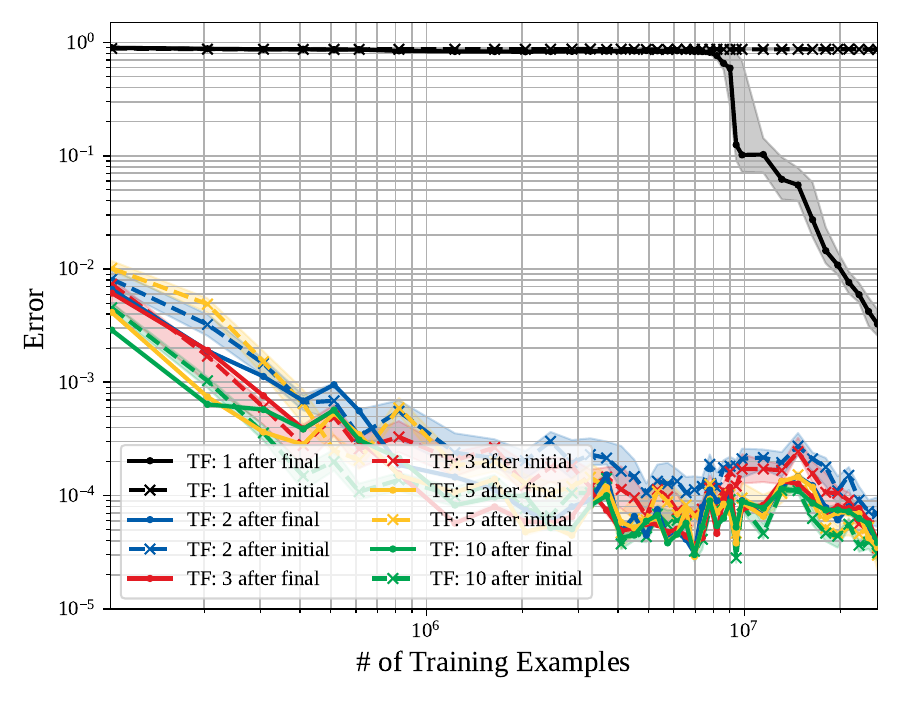}
        \caption{Identity: 18 system haystack.}
        \label{fig:ident_med_len_18_baseline_log}
    \end{subfigure}
    \hfill
    \begin{subfigure}[b]{0.32\linewidth}
        \centering
        \includegraphics[width=\linewidth]{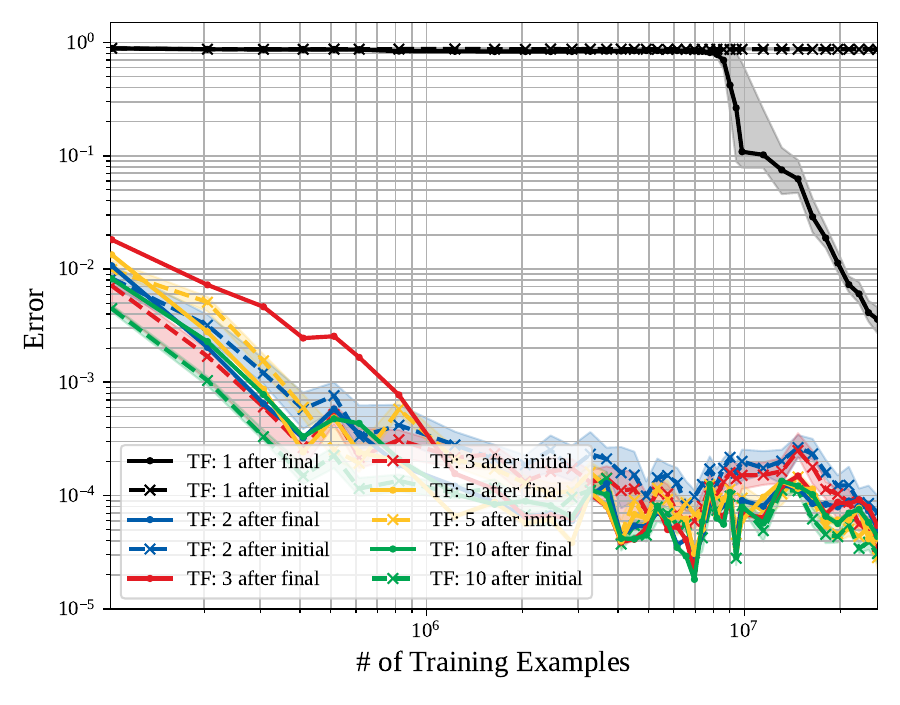}
        \caption{Identity: 19 system haystack.}
        \label{fig:ident_med_len_19_baseline_log}
    \end{subfigure}
    \caption{Performance of medium identity model (2.42M params) across training --- log-scale.}
    \label{fig:ident_med_baseline_log}

\end{figure}

\begin{figure}[htbp]
    \centering
    \begin{subfigure}[b]{0.32\linewidth}
        \centering
        \includegraphics[width=\linewidth]{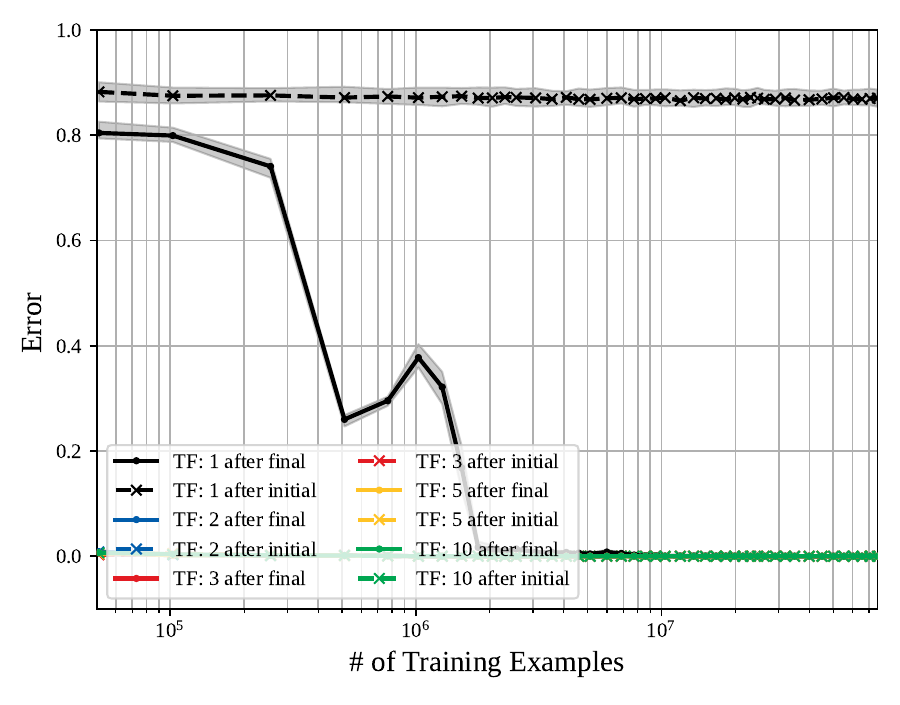}
        \caption{Identity: 1 system haystack.}
        \label{fig:ident_big_len_1_baseline_linear}
    \end{subfigure}
    \hfill
    \begin{subfigure}[b]{0.32\linewidth}
        \centering
        \includegraphics[width=\linewidth]{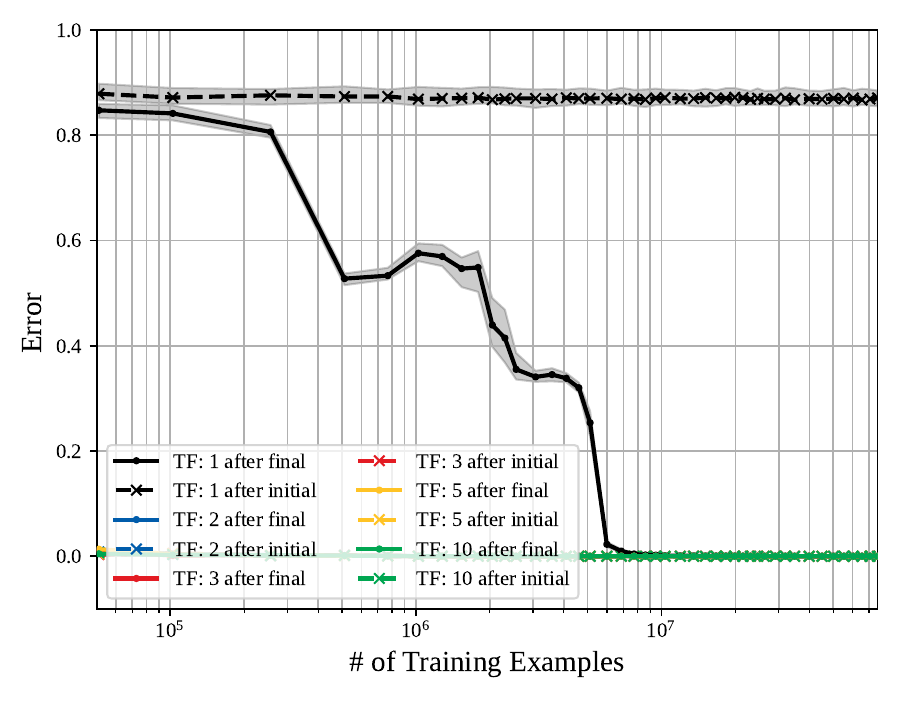}
        \caption{Identity: 2 system haystack.}
        \label{fig:ident_big_len_2_baseline_linear}
    \end{subfigure}
    \hfill
    \begin{subfigure}[b]{0.32\linewidth}
        \centering
        \includegraphics[width=\linewidth]{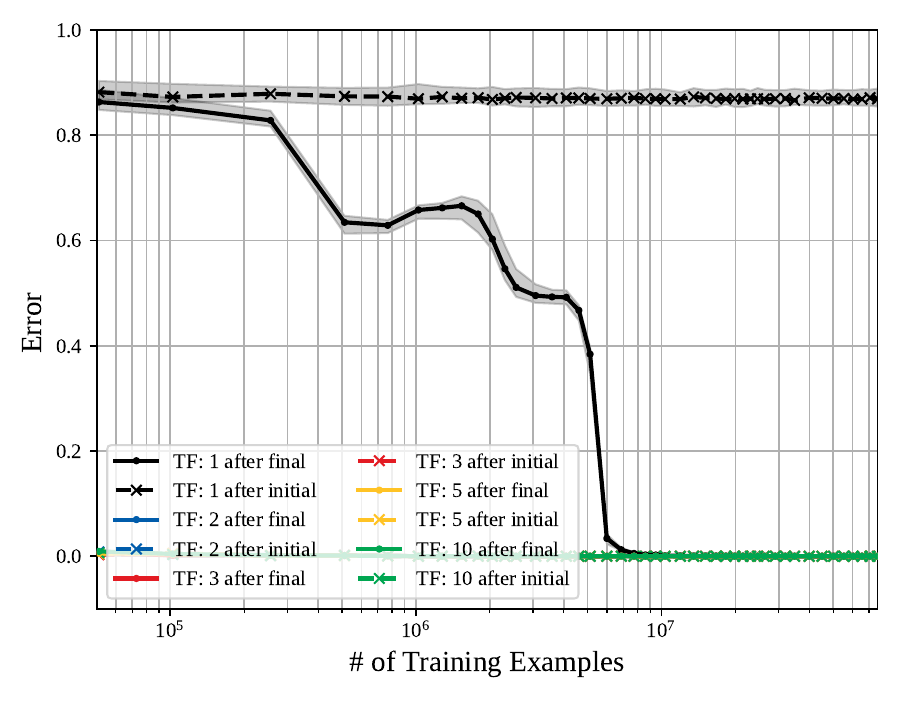}
        \caption{Identity: 3 system haystack.}
        \label{fig:ident_big_len_3_baseline_linear}
    \end{subfigure}

    \vspace{0.2cm}

    \centering
    \begin{subfigure}[b]{0.32\linewidth}
        \centering
        \includegraphics[width=\linewidth]{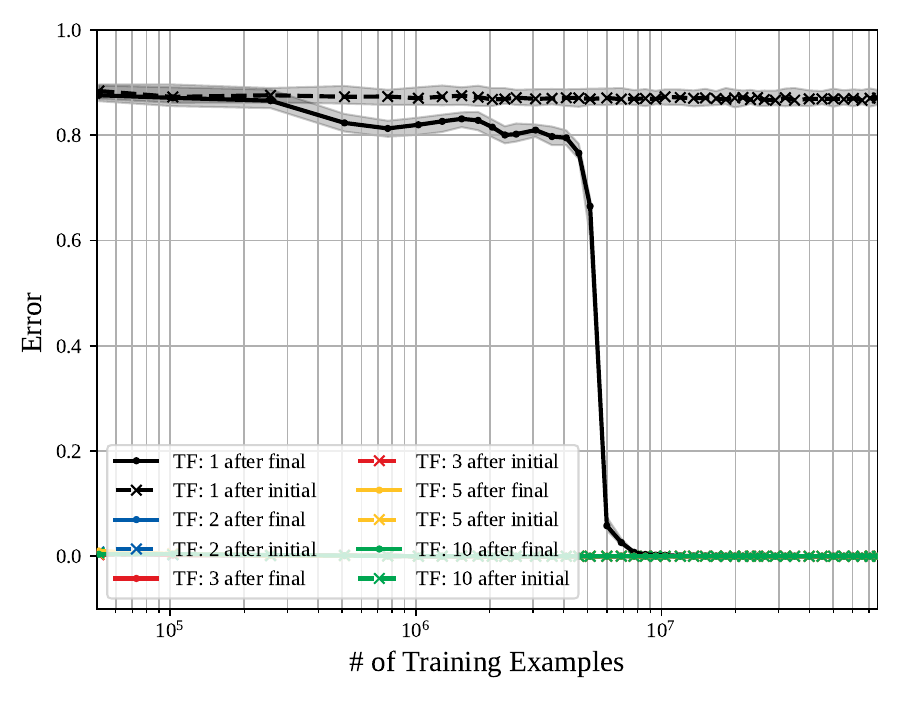}
        \caption{Identity: 17 system haystack.}
        \label{fig:ident_big_len_17_baseline_linear}
    \end{subfigure}
    \hfill
    \begin{subfigure}[b]{0.32\linewidth}
        \centering
        \includegraphics[width=\linewidth]{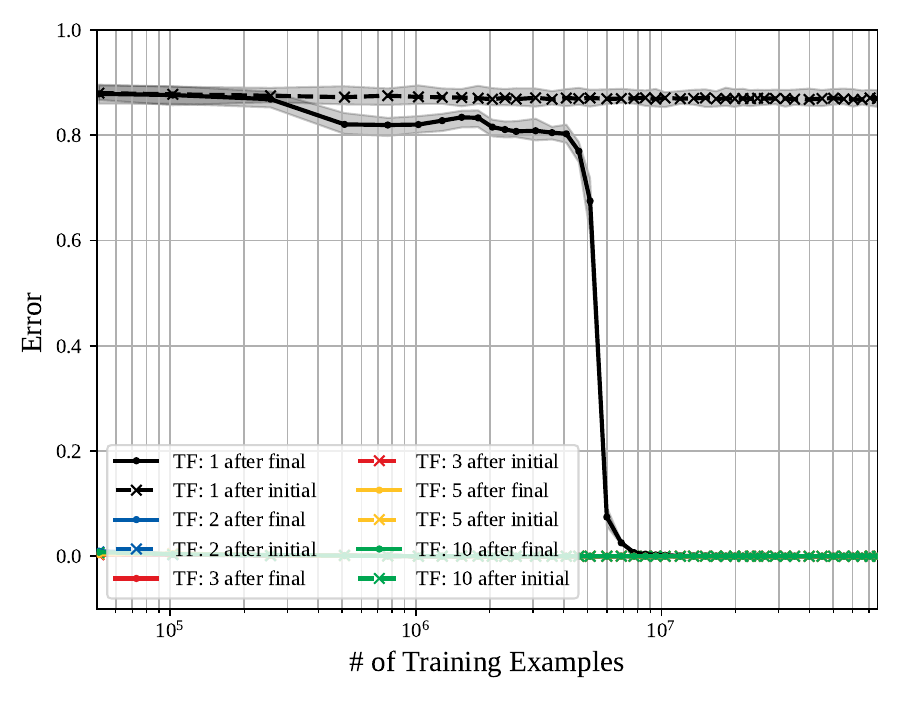}
        \caption{Identity: 18 system haystack.}
        \label{fig:ident_big_len_18_baseline_linear}
    \end{subfigure}
    \hfill
    \begin{subfigure}[b]{0.32\linewidth}
        \centering
        \includegraphics[width=\linewidth]{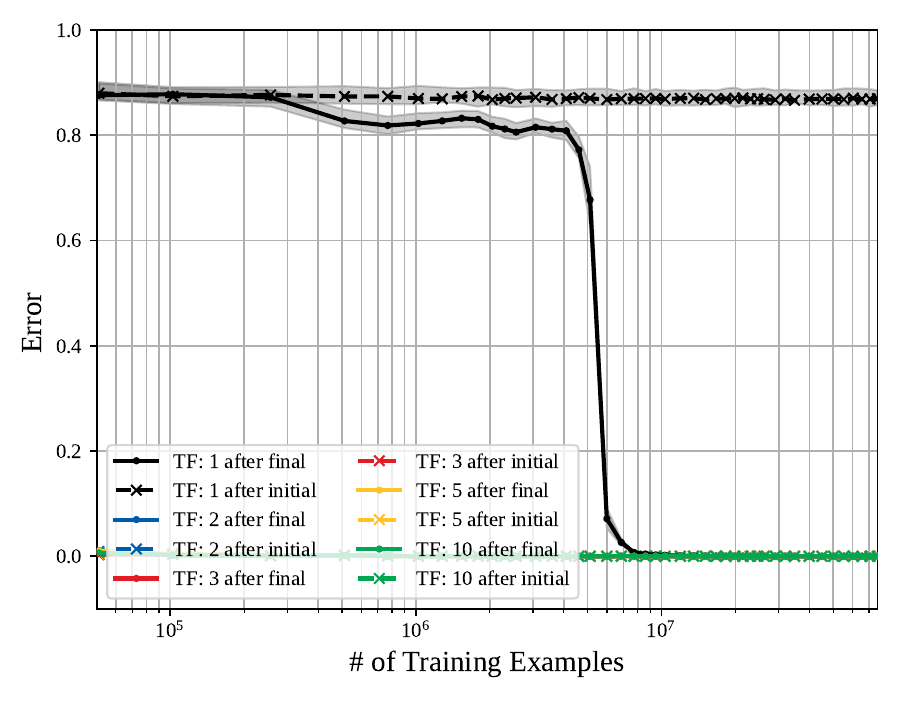}
        \caption{Identity: 19 system haystack.}
        \label{fig:ident_big_len_19_baseline_linear}
    \end{subfigure}
    \caption{Performance of big identity model (10.7M params) across training --- linear-scale.}
    \label{fig:ident_big_baseline_linear}

\end{figure}

\begin{figure}[htbp]
    \centering
    \begin{subfigure}[b]{0.32\linewidth}
        \centering
        \includegraphics[width=\linewidth]{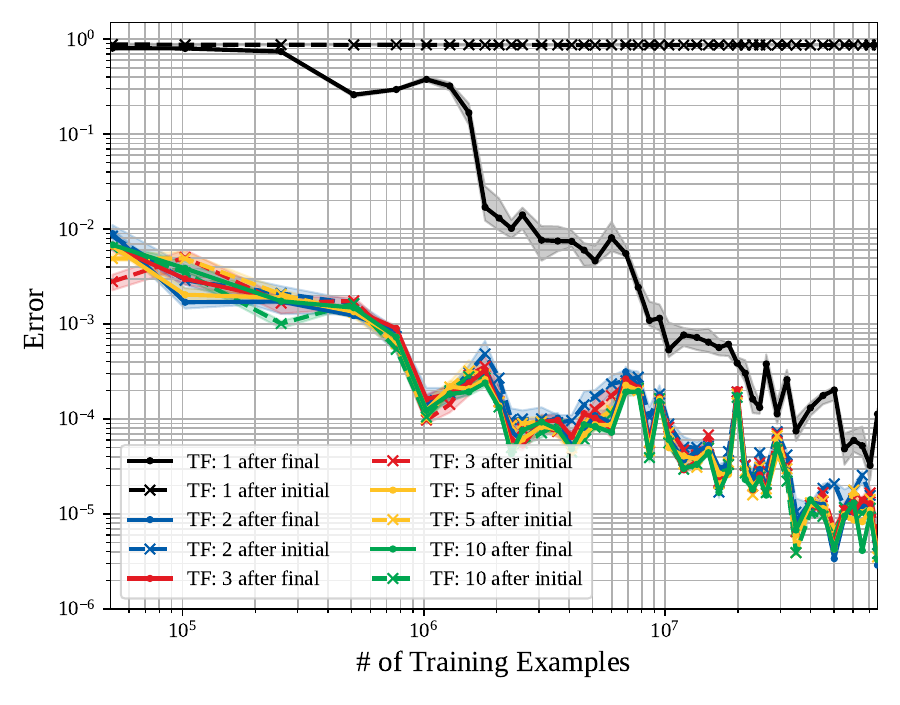}
        \caption{Identity: 1 system haystack.}
        \label{fig:ident_big_len_1_baseline_log}
    \end{subfigure}
    \hfill
    \begin{subfigure}[b]{0.32\linewidth}
        \centering
        \includegraphics[width=\linewidth]{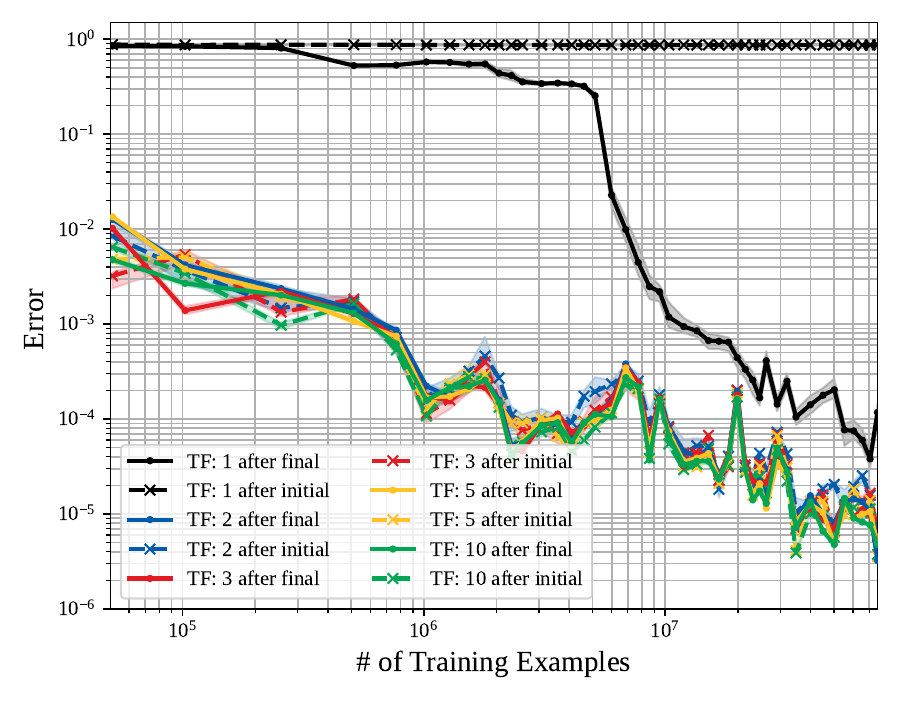}
        \caption{Identity: 2 system haystack.}
        \label{fig:ident_big_len_2_baseline_log}
    \end{subfigure}
    \hfill
    \begin{subfigure}[b]{0.32\linewidth}
        \centering
        \includegraphics[width=\linewidth]{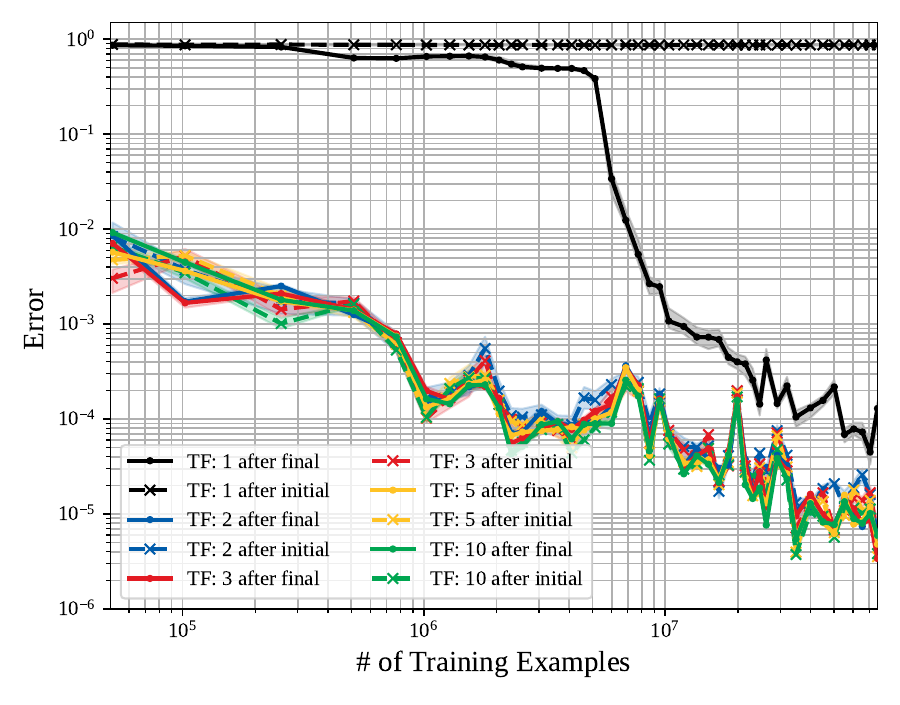}
        \caption{Identity: 3 system haystack.}
        \label{fig:ident_big_len_3_baseline_log}
    \end{subfigure}

    \vspace{0.2cm}

    \centering
    \begin{subfigure}[b]{0.32\linewidth}
        \centering
        \includegraphics[width=\linewidth]{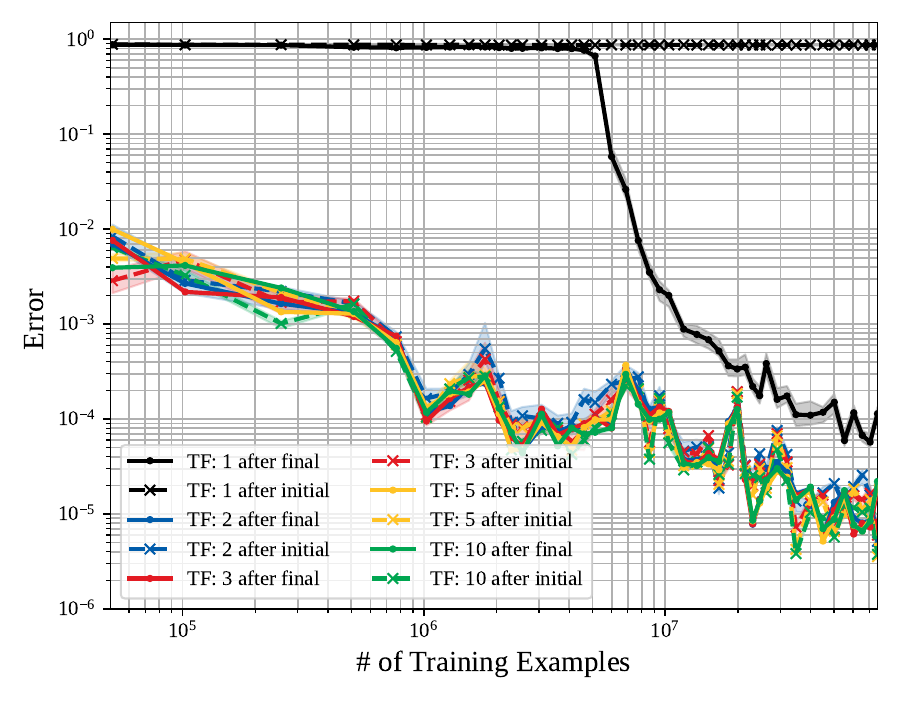}
        \caption{Identity: 17 system haystack.}
        \label{fig:ident_big_len_17_baseline_log}
    \end{subfigure}
    \hfill
    \begin{subfigure}[b]{0.32\linewidth}
        \centering
        \includegraphics[width=\linewidth]{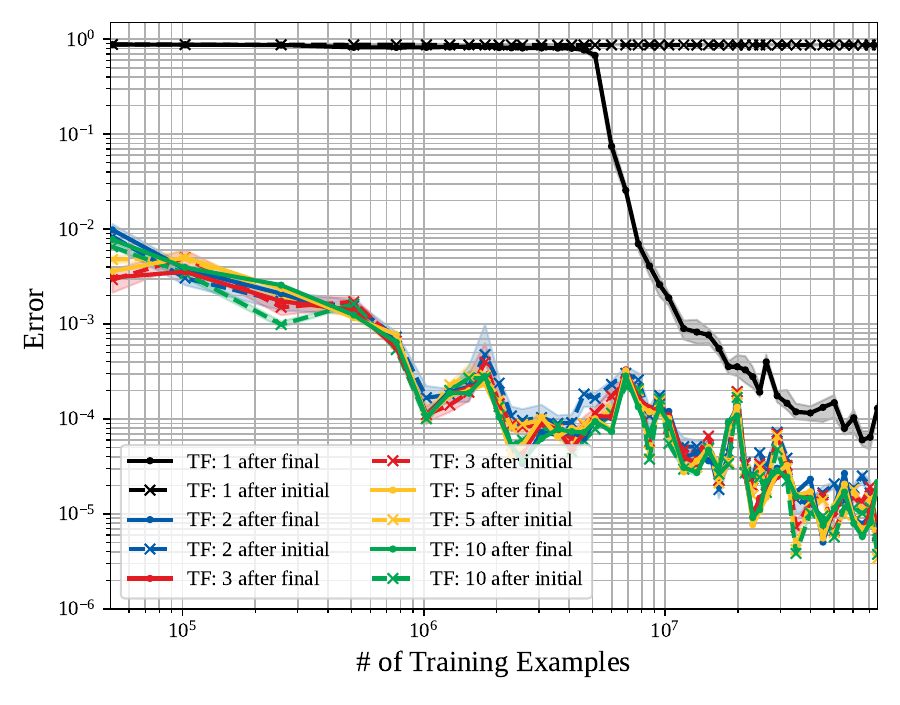}
        \caption{Identity: 18 system haystack.}
        \label{fig:ident_big_len_18_baseline_log}
    \end{subfigure}
    \hfill
    \begin{subfigure}[b]{0.32\linewidth}
        \centering
        \includegraphics[width=\linewidth]{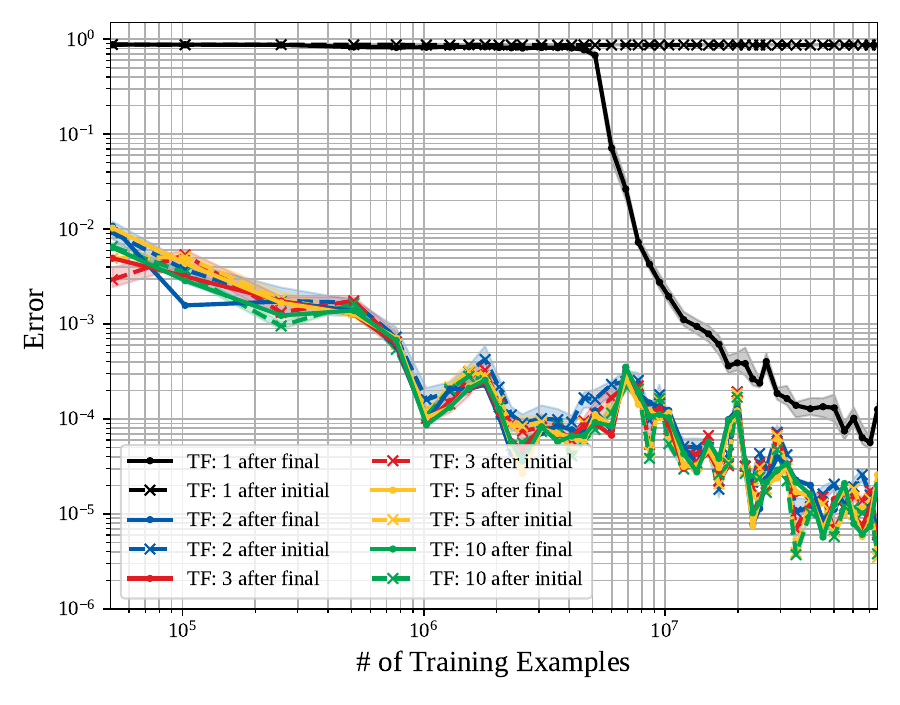}
        \caption{Identity: 19 system haystack.}
        \label{fig:ident_big_len_19_baseline_log}
    \end{subfigure}
    \caption{Performance of big identity model (10.7M params) across training --- log-scale.}
    \label{fig:ident_big_baseline_log}

\end{figure}

\vspace{-0.75em}
Following the summary results from above, we provide full training dynamics plots for our different model sizes and show the results when tested on haystack lengths of 1, 2, 3, 17, 18 and 19. For your convenience here are the different figures: Orthogonal tiny (linear scale) Fig.~\ref{fig:ortho_tiny_baseline_linear}, Orthogonal tiny (log scale) Fig.~\ref{fig:ortho_tiny_baseline_log}, Orthogonal small (linear scale) Fig.~\ref{fig:ortho_small_baseline_linear}, Orthogonal small (log scale) Fig.~\ref{fig:ortho_small_baseline_log}, Orthogonal medium (linear scale) Fig.~\ref{fig:ortho_med_baseline_linear}, Orthogonal medium (log scale) Fig.~\ref{fig:ortho_med_baseline_log}, Orthogonal big (linear scale) Fig.~\ref{fig:ortho_big_baseline_linear}, Orthogonal big (log scale) Fig.~\ref{fig:ortho_big_baseline_log}, Identity tiny (linear scale) Fig.~\ref{fig:ident_tiny_baseline_linear}, Identity tiny (log scale) Fig.~\ref{fig:ident_tiny_baseline_log}, Identity small (linear scale) Fig.~\ref{fig:ident_small_baseline_linear}, Identity small (log scale) Fig.~\ref{fig:ident_small_baseline_log}, Identity medium (linear scale) Fig.~\ref{fig:ident_med_baseline_linear}, Identity medium (log scale) Fig.~\ref{fig:ident_med_baseline_log}, Identity big (linear scale) Fig.~\ref{fig:ident_big_baseline_linear}, Identity big (log scale) Fig.~\ref{fig:ident_big_baseline_log}

%look at training dynamics for different model sizes following the results from before. They are included here for convienience

%ortho tiny linear

%ortho tiny log

%ortho small linear

%ortho small log

%ortho med linear

%ortho med log

%ortho big linear

%ortho big log

%ident tiny log

%ident small linear

%ident small log

%ident med linear

%ident med log

%ident big linear

%ident big log
\FloatBarrier

\section{When does the model learn to restart ICL for a new system?}\label{sec:restart_new_supplemental}

\begin{figure}[htbp]
    \centering
    \setcounter{subfigure}{0}
    \begin{subfigure}[b]{0.49\linewidth}
        \centering
        \includegraphics[width=\linewidth]{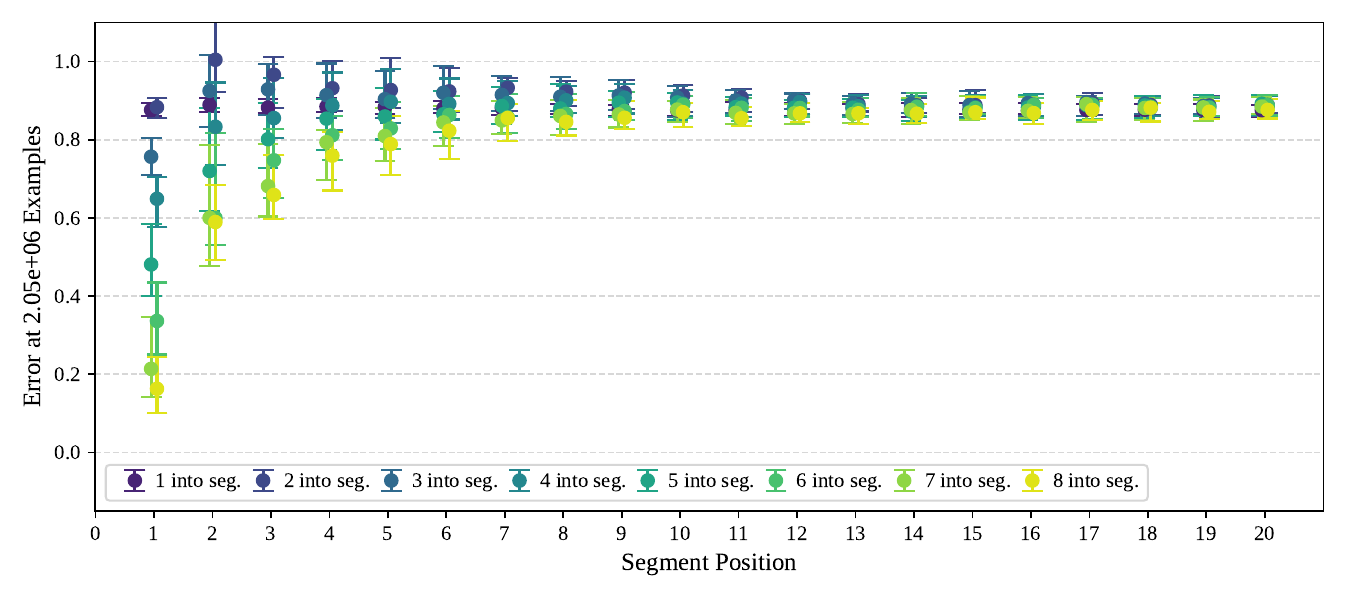}
        \caption{$2.05\times 10^{6}$ training examples.}
        \label{fig:restart_sys_2_05e6}
    \end{subfigure}
    \hfill
    \begin{subfigure}[b]{0.49\linewidth}
        \centering
        \setcounter{subfigure}{5}
        \includegraphics[width=\linewidth]{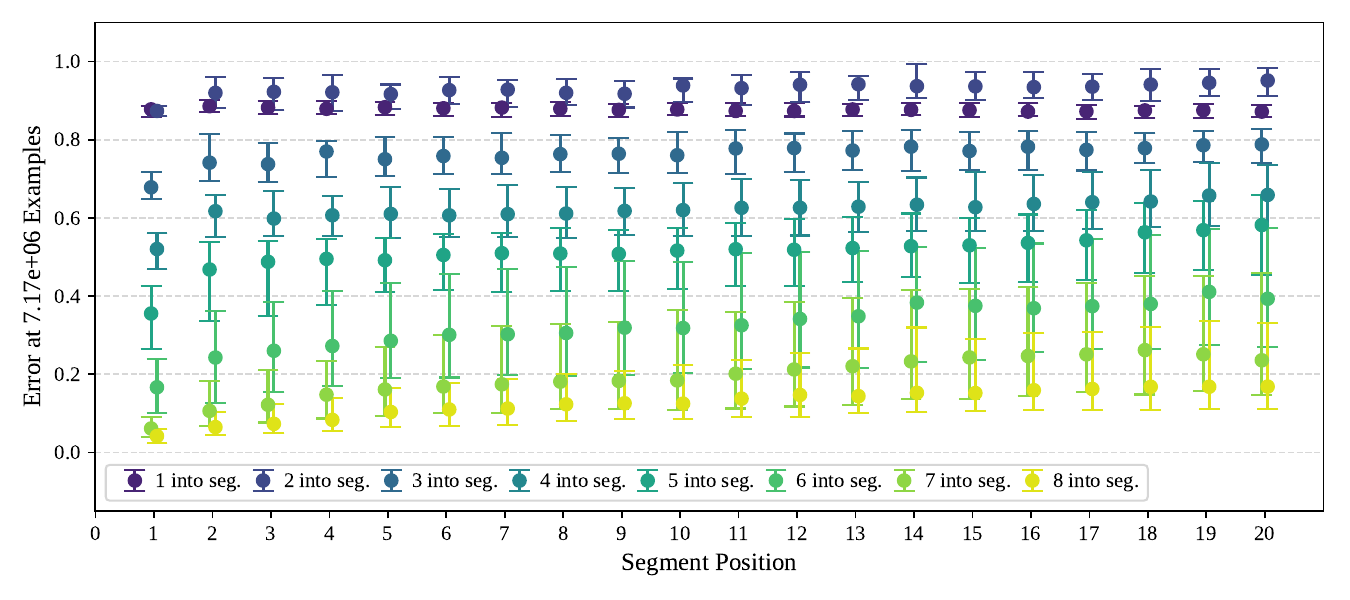}
        \caption{$7.17\times 10^{6}$ training examples.}
        \label{fig:restart_sys_7_17e6}
    \end{subfigure}

    \begin{subfigure}[b]{0.49\linewidth}
        \centering
        \setcounter{subfigure}{1}
        \includegraphics[width=\linewidth]{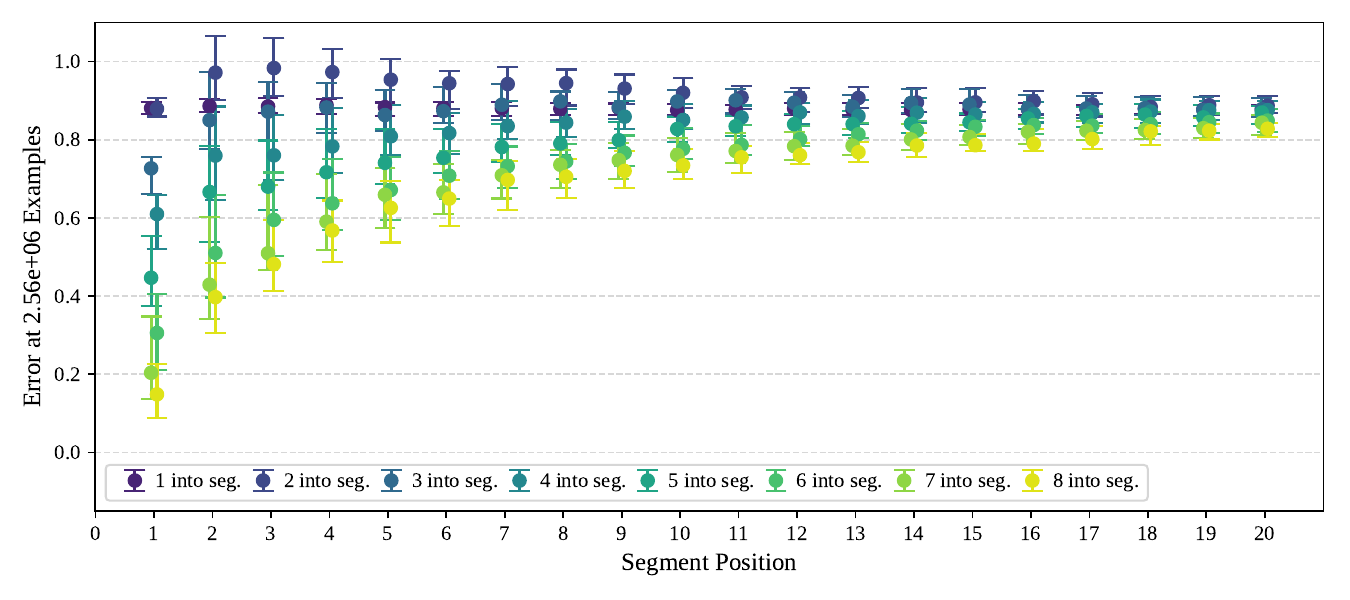}
        \caption{$2.56\times 10^{6}$ training examples.}
        \label{fig:restart_sys_2_56e6}
    \end{subfigure}
    \hfill
    \begin{subfigure}[b]{0.49\linewidth}
        \centering
        \setcounter{subfigure}{6}
        \includegraphics[width=\linewidth]{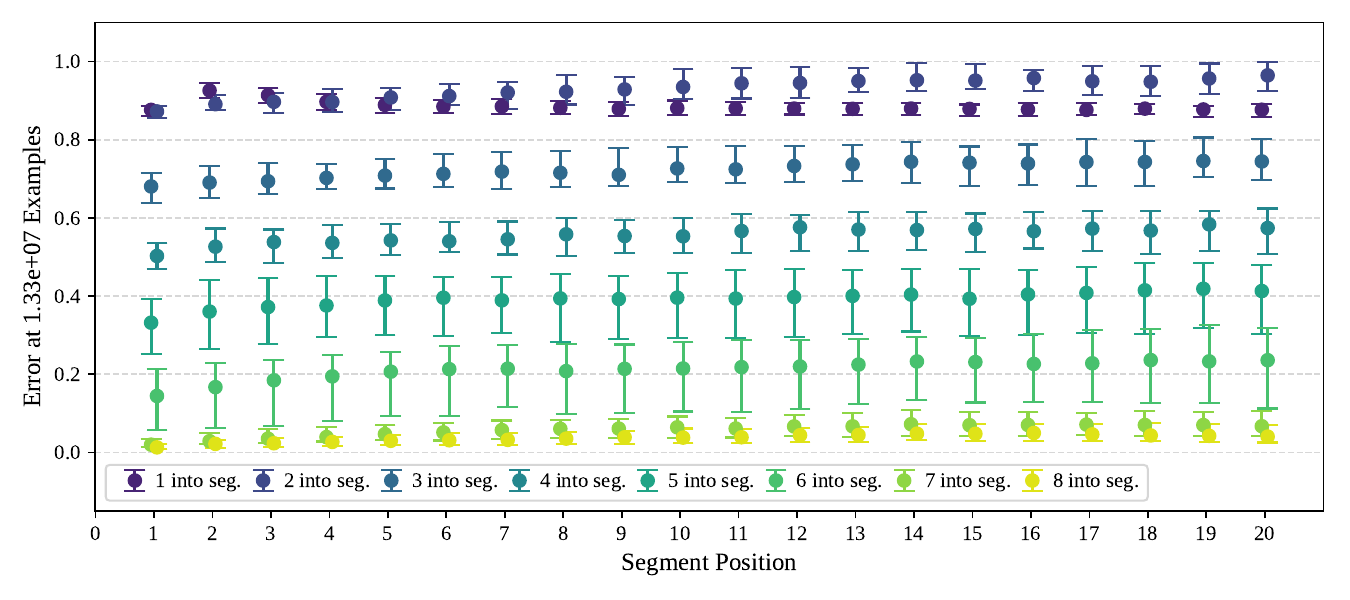}
        \caption{$1.33\times 10^{7}$ training examples.}
        \label{fig:restart_sys_1_33e7}
    \end{subfigure}

    \begin{subfigure}[b]{0.49\linewidth}
        \centering
        \setcounter{subfigure}{2}
        \includegraphics[width=\linewidth]{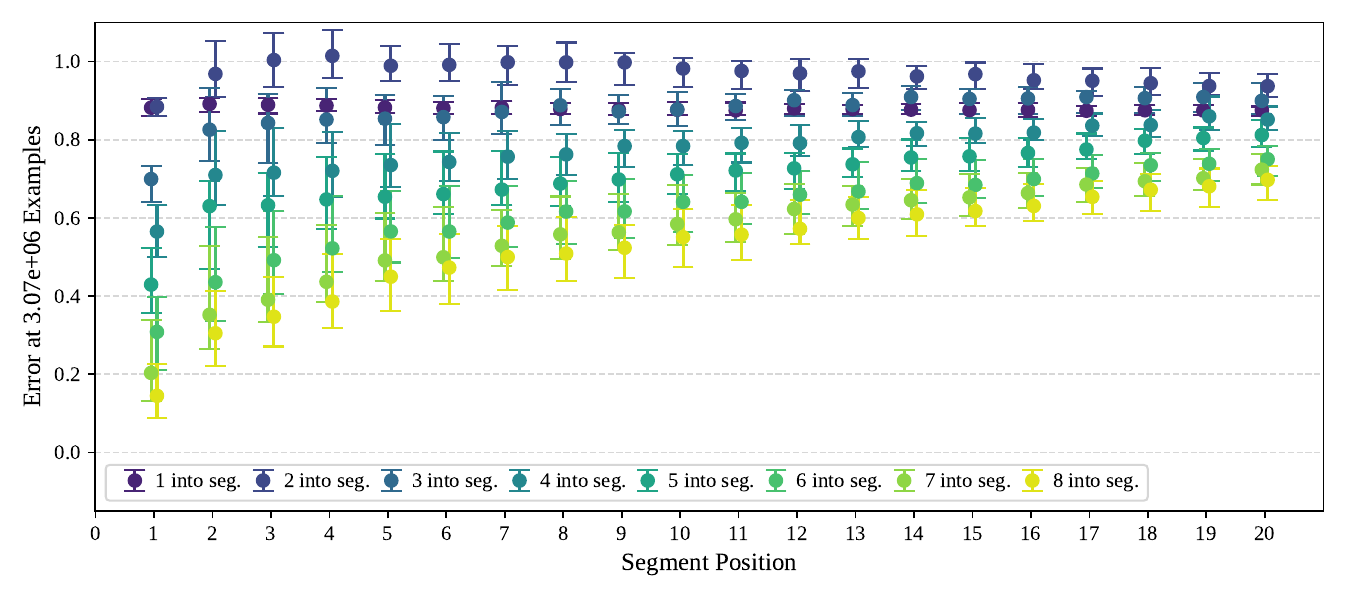}
        \caption{$3.07\times 10^{6}$ training examples.}
        \label{fig:restart_sys_3_07e6}
    \end{subfigure}
    \hfill    
    \begin{subfigure}[b]{0.49\linewidth}
        \centering
        \setcounter{subfigure}{7}
        \includegraphics[width=\linewidth]{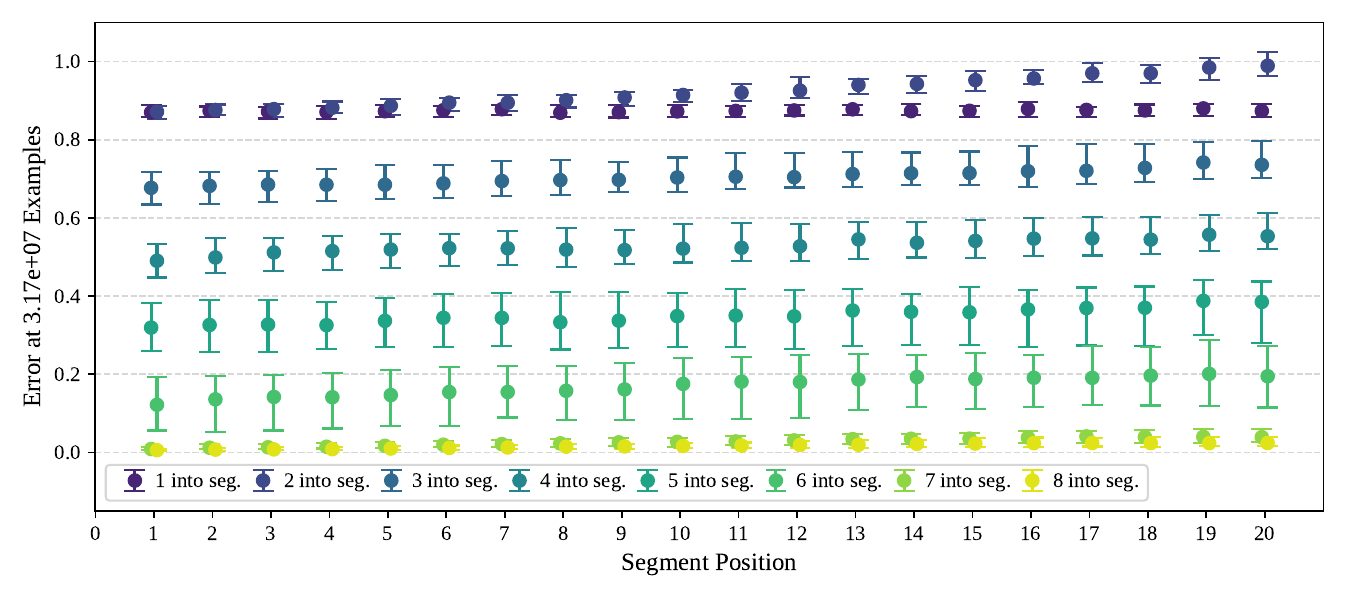}
        \caption{$3.17\times 10^{7}$ training examples.}
        \label{fig:restart_sys_3_17e7}
    \end{subfigure}

    \begin{subfigure}[b]{0.49\linewidth}
        \centering
        \setcounter{subfigure}{3}
        \includegraphics[width=\linewidth]{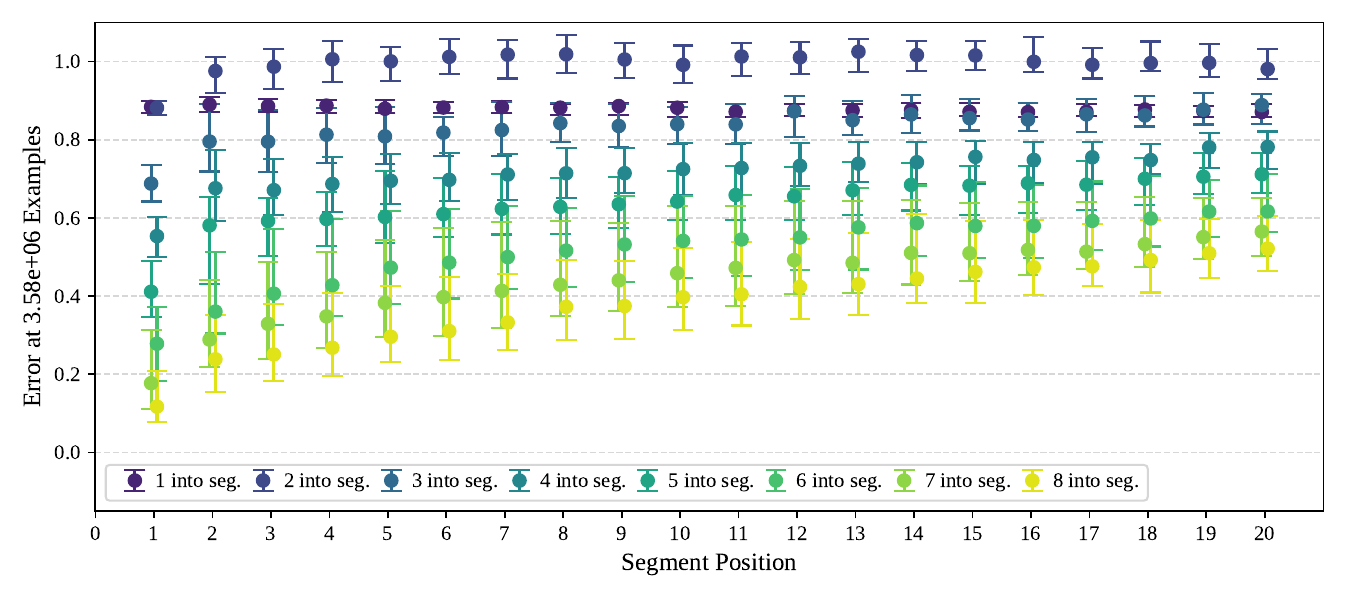}
        \caption{$3.58\times 10^{6}$ training examples.}
        \label{fig:restart_sys_3_58e6}
    \end{subfigure}
    \hfill
    \begin{subfigure}[b]{0.49\linewidth}
        \centering
        \setcounter{subfigure}{8}
        \includegraphics[width=\linewidth]{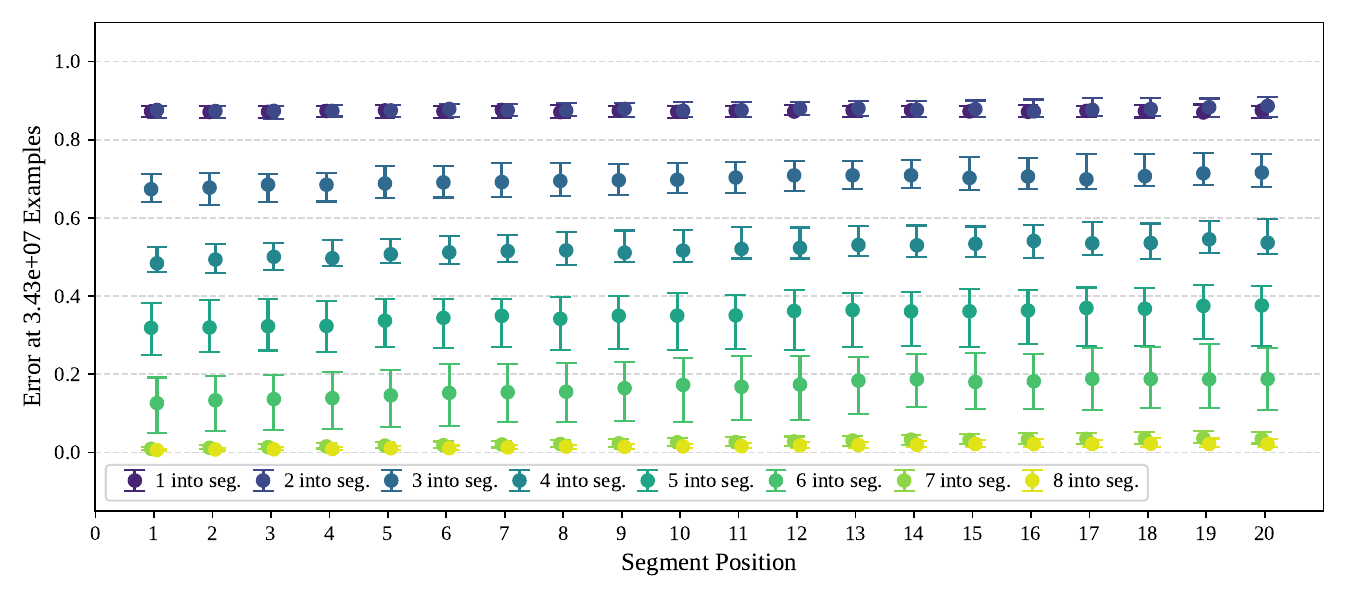}
        \caption{$3.43\times 10^{7}$ training examples.}
        \label{fig:restart_sys_3_43e7}
    \end{subfigure}
    % Add more subfigures here as needed
    \begin{subfigure}[b]{0.49\linewidth}
        \centering
        \setcounter{subfigure}{4}
        \includegraphics[width=\linewidth]{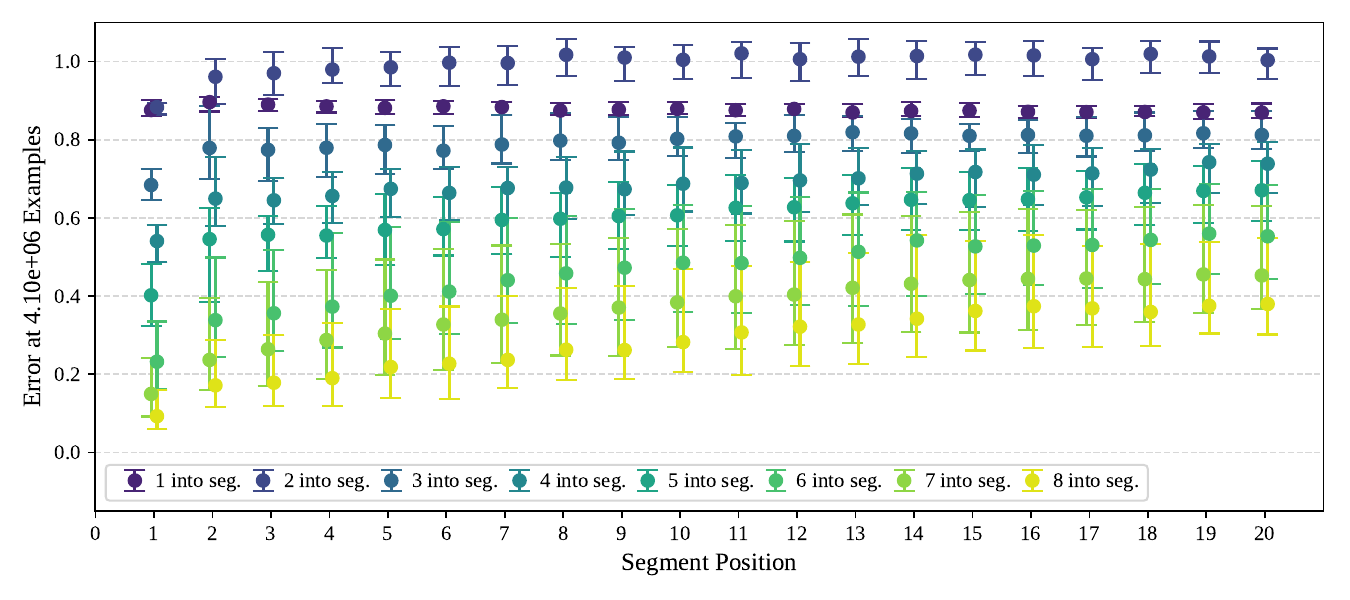}
        \caption{$4.10\times 10^{6}$ training examples.}
        \label{fig:restart_sys_4_10e6}
    \end{subfigure}
    \hfill
    \begin{subfigure}[b]{0.49\linewidth}
        \centering
        \setcounter{subfigure}{9}
        \includegraphics[width=\linewidth]{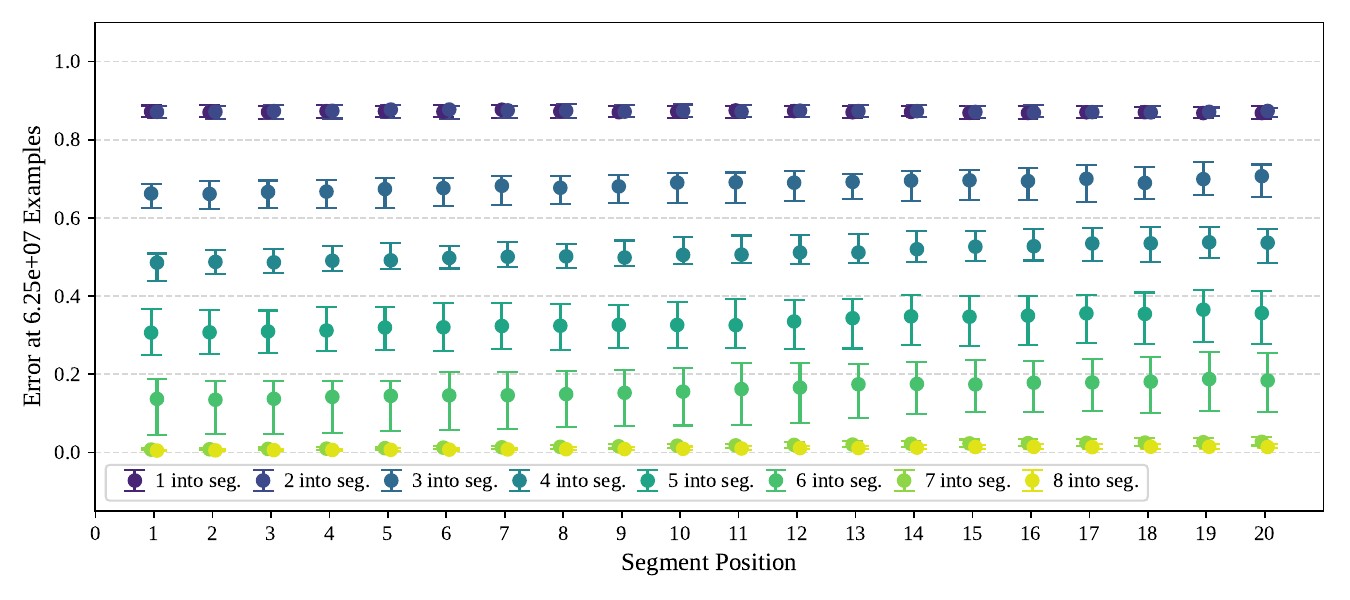}
        \caption{$6.25\times 10^{7}$ training examples.}
        \label{fig:restart_sys_6_25e7}
    \end{subfigure}

    \caption{Restarting for a new system --- Prediction error vs. the position of a previously unseen system segment in the haystack. Position 1 is the beginning and position 20 is the end. The colored curves correspond to different indices into each system segment.}
    \label{fig:restart_sys_all}
\end{figure}

Continuing on from Section~\ref{sec:restart_sys_main_paper}, this supplemental appendix takes a more comprehensive look at how the ability emerges for the model to restart its ICL predictions for a new system.

Fig.~\ref{fig:restart_sys_all} depicts the prediction error of different model checkpoints on initial segments from previously unseen systems when these segments are at different positions in the haystack. The first segment in the haystack is denoted position 1 and the last segment is denoted position 20. Furthermore, we look at the prediction error of the model for 1 through 8 indices after the initial open symbol initiating the segment. Subfigure (a) clearly shows that at this early checkpoint $2.05\times10^6$, while the model has learned to in-context learn the dynamics for the first segment, there is substantial interference with later segments where performance is poor. Rapidly across the checkpoints illustrated in (b),(c),(d),(e) the model learns how to interpret the symbolic tokens well enough to properly restart its predictions for each new segment and by checkpoint $4.1\times10^6$, it is quite good --- except at predicting the fundamentally unpredictable second position in a new segment. 

Subfigures (f),(g),(h),(i) of Fig.~\ref{fig:restart_sys_all} show the more gradual improvements in this ability during much later checkpoints --- where the focus is largely on improvements to the performance in predicting the second position in a new segment. Finally, Fig.~\ref{fig:restart_sys_6_25e7} shows the model's good ability to restart ICL at the end of its training.

% Looking first at Figs.~\ref{fig:restart_sys_2_05e6}--\ref{fig:restart_sys_4_10e6}, at $2.05\times 10^{6}$ training examples the model has the expected performance for the segment at position 1, but the rest of the segment positions have poor performance. By $4.10\times 10^{6}$ training examples, the performance at later segment positions for 8 indices into the segment has significantly improved, but its prediction for 2 indices into the segment has actually deteriorated from its performance at $2.05\times 10^{6}$ training examples. In Figs.~\ref{fig:restart_sys_3_17e7}--\ref{fig:restart_sys_3_43e7}, we fast-forward to $3.17\times 10^{7}$ training examples and see that its performance for 2 indices into the segment has improved to a near-optimal error by $3.43\times 10^{7}$ training examples. Finally, Figs.~\ref{fig:restart_sys_6_25e7} and \ref{fig:restart_sys_6_25e7_log} show the model's performance at our early-stopping checkpoint of $6.25\times 10^{7}$ training examples.

% Consequently, although this paper is primarily focused on the emergence of successful recall for initiating a resumed sequence (which happens around 25M training examples for our medium model), and the emergence of successful continuation of a resumed sequence (which happens around 10M training examples for our medium model), there is another ability that emerges at around 3M training examples for our medium model --- namely being able to restart fresh ICL predictions when seeing a brand new system. 
\FloatBarrier

\section{Final performance with position in segment and needle position within haystack}\label{sec:needle_last_seg_context}

% \begin{figure*}[tbph]
%     \centering
%     % Subfigure 2
%     \begin{subfigure}[b]{0.495\linewidth}
%         \centering
%         \includegraphics[width=\linewidth]{needle/context/ortho_haar/last_seg_context_ortho_haar_embd_dim_128_step_122000_haystack_len_19_20250515_123225.pdf}
%         \caption{Orthogonal systems.}
%         \label{fig:ortho_needle_context}
%     \end{subfigure}
%     \hfill
%     % Subfigure 4
%     \begin{subfigure}[b]{0.495\linewidth}
%         \centering
%         \includegraphics[width=\linewidth]{needle/needle_pos/ortho_haar/error_ratios_ortho_haar_embd_dim_128_state_dim_5_ident_C_step_122000_haystack_len_19_20250515_123225.pdf}
%         \caption{Orthogonal systems.}
%         \label{fig:ortho_needle_pos_error_ratios}
%     \end{subfigure}
%     \caption{\ref{fig:ortho_needle_context} shows a comparison between predictions made in the final test segment (black) and predictions made from a continuing segment (red) as well as the initial segment (yellow) for the identity and orthogonal model respectively. This is the performance of the identity model after $\approx 1.8\times 10^7$ training examples and the orthogonal model after $\approx 6.25 \times 10^7$ training examples.}
%     \label{fig:needle_summary}
% \end{figure*}

\begin{figure}[tbph]
    \centering
    % Orthogonal systems: context
    \begin{subfigure}[b]{0.49\linewidth}
        \centering
        \includegraphics[width=\linewidth]{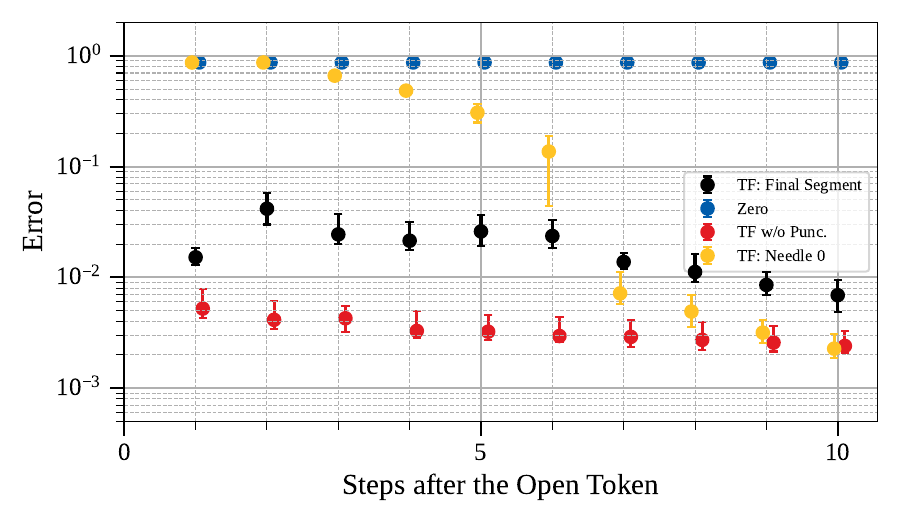}
        \caption{Orthogonal}
        \label{fig:ortho_needle_context}
    \end{subfigure}
    \hfill
    % Identity systems: context
    \begin{subfigure}[b]{0.49\linewidth}
        \centering
        \includegraphics[width=\linewidth]{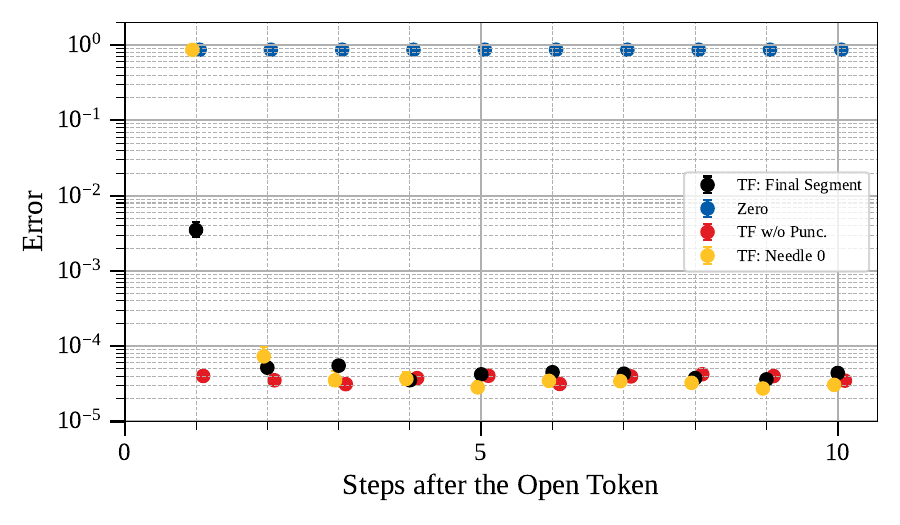}
        \caption{Identity}
        \label{fig:ident_needle_context}
    \end{subfigure}

    % Orthogonal systems: needle pos
    \begin{subfigure}[b]{0.49\linewidth}
        \centering
        \includegraphics[width=\linewidth]{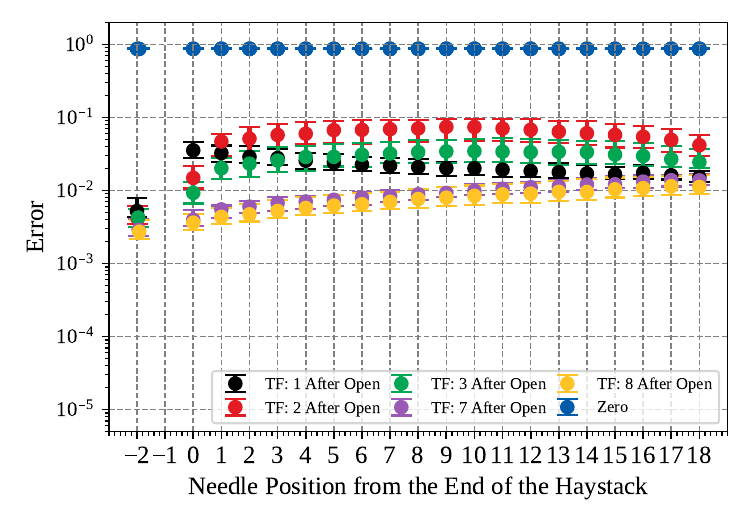}
        \caption{Orthogonal}
        \label{fig:ortho_needle_pos_error_ratios}
    \end{subfigure}
    \hfill
    % Identity systems: needle pos
    \begin{subfigure}[b]{0.49\linewidth}
        \centering
        \includegraphics[width=\linewidth]{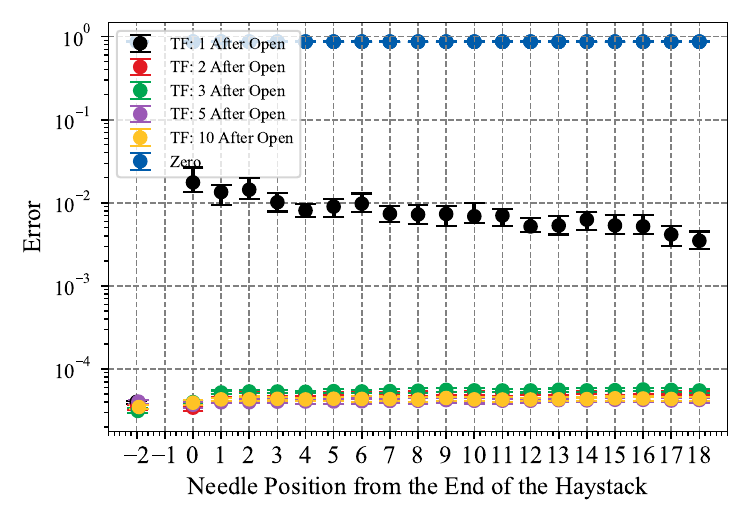}
        \caption{Identity}
        \label{fig:ident_needle_pos_error_ratios}
    \end{subfigure}
    \caption{Comparison of recall performance for orthogonal and identity systems. Figs.~\ref{fig:ortho_needle_context} and \ref{fig:ident_needle_context} show predictions made in the final test segment (black), from a continuing segment (red), and from the initial segment (yellow). Figs.~\ref{fig:ortho_needle_pos_error_ratios} and \ref{fig:ident_needle_pos_error_ratios} show median-squared error vs. the position of the needle in the haystack. For these plots, the x-axis value of $-2$ corresponds to a final segment that is the continuation of the segment before with no interruption by open and close symbols. The x-axis value of 0 is the closest needle position to
the final segment while the value 18 is the furthest away. The identity model is after $\approx 1.8\times 10^7$ training examples and the orthogonal model after $\approx 6.25 \times 10^7$ training examples. ``Zero'' in the legend corresponds to a predictor that always predicts 0.}
    \label{fig:needle_summary_combined}
\end{figure}

%make a subfigure block
\begin{figure}
    \centering
    \begin{subfigure}[b]{0.49\linewidth}
        \centering
        \includegraphics[width=\linewidth, trim=0mm 5mm 0mm 0mm, clip]{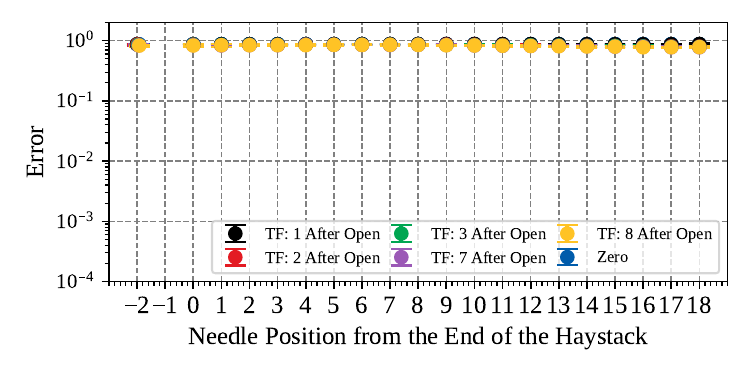}
        \caption{$2.05\times 10^{6}$ training examples.}
        \label{fig:ortho_needle_pos_4000}
    \end{subfigure}
    \hfill
    \begin{subfigure}[b]{0.49\linewidth}
        \centering
        \includegraphics[width=\linewidth, trim=0mm 5mm 0mm 0mm, clip]{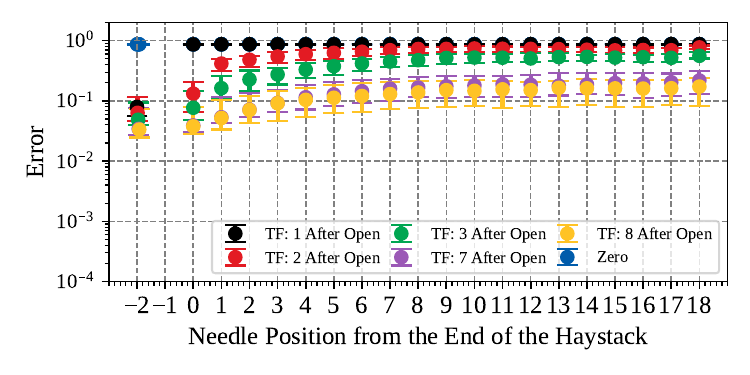}
        \caption{$7.17\times 10^{6}$ training examples.}
        \label{fig:ortho_needle_pos_14000}
    \end{subfigure}

    \centering
    \begin{subfigure}[b]{0.49\linewidth}
        \centering
        \includegraphics[width=\linewidth, trim=0mm 5mm 0mm 0mm, clip]{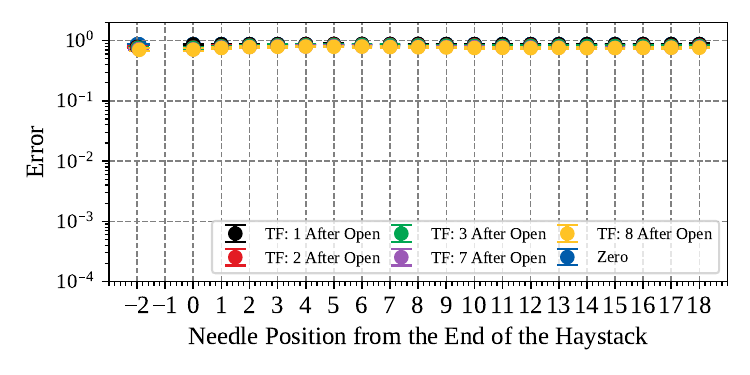}
        \caption{$2.56\times 10^{6}$ training examples.}
        \label{fig:ortho_needle_pos_5000}
    \end{subfigure}
    \hfill
    \begin{subfigure}[b]{0.49\linewidth}
        \centering
        \includegraphics[width=\linewidth, trim=0mm 5mm 0mm 0mm, clip]{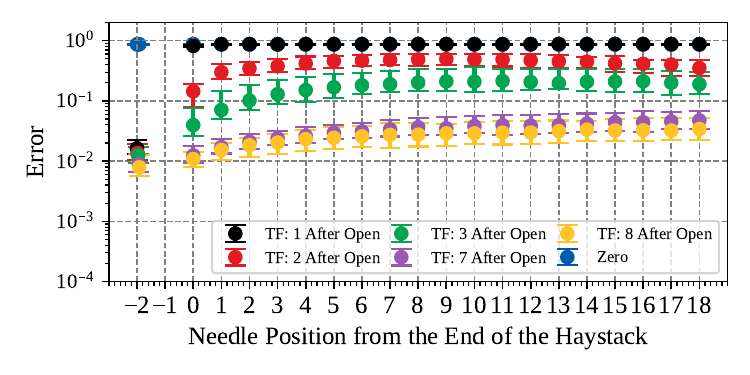}
        \caption{$1.33\times 10^{7}$ training examples.}
        \label{fig:ortho_needle_pos_26000}
    \end{subfigure}

    \centering
    \begin{subfigure}[b]{0.49\linewidth}
        \centering
        \includegraphics[width=\linewidth, trim=0mm 5mm 0mm 0mm, clip]{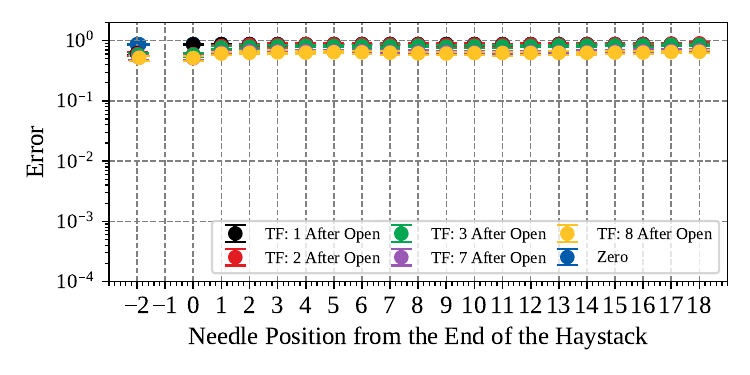}
        \caption{$3.07\times 10^{6}$ training examples.}
        \label{fig:ortho_needle_pos_6000}
    \end{subfigure}
    \hfill
    \begin{subfigure}[b]{0.49\linewidth}
        \centering
        \includegraphics[width=\linewidth, trim=0mm 5mm 0mm 0mm, clip]{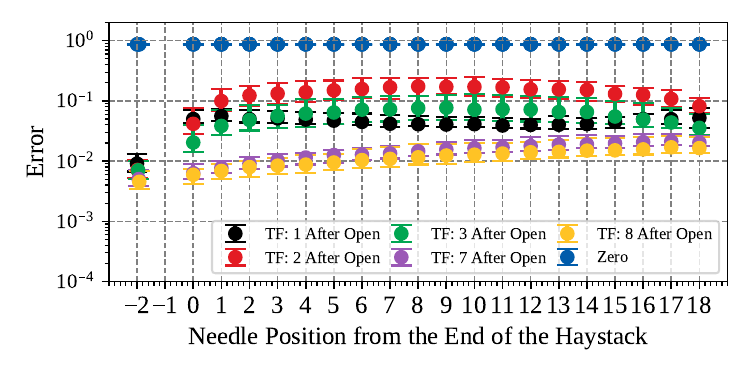}
        \caption{$3.17\times 10^{7}$ training examples.}
        \label{fig:ortho_needle_pos_62000}
    \end{subfigure}

    \centering
    \begin{subfigure}[b]{0.49\linewidth}
        \centering
        \includegraphics[width=\linewidth, trim=0mm 5mm 0mm 0mm, clip]{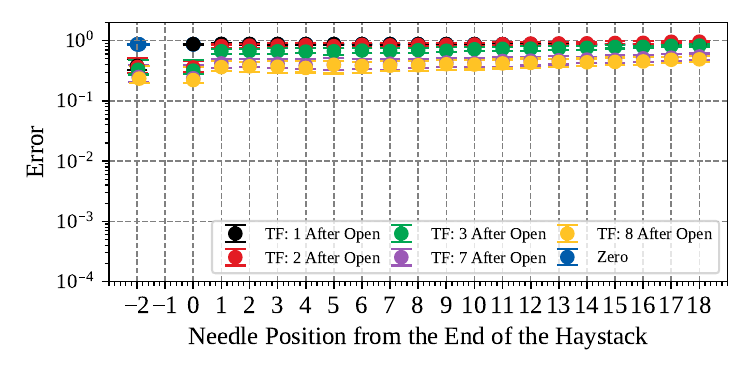}
        \caption{$3.58\times 10^{6}$ training examples.}
        \label{fig:ortho_needle_pos_7000}
    \end{subfigure}
    \hfill
    \begin{subfigure}[b]{0.49\linewidth}
        \centering
        \includegraphics[width=\linewidth, trim=0mm 5mm 0mm 0mm, clip]{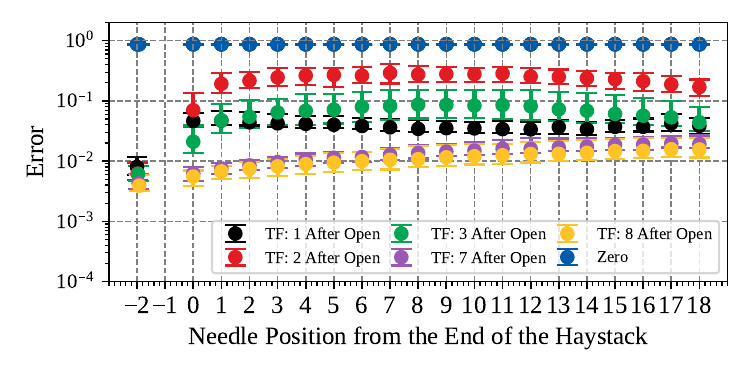}
        \caption{$3.43\times 10^{7}$ training examples.}
        \label{fig:ortho_needle_pos_67000}
    \end{subfigure}

    \centering
    \begin{subfigure}[b]{0.49\linewidth}
        \centering
        \includegraphics[width=\linewidth, trim=0mm 5mm 0mm 0mm, clip]{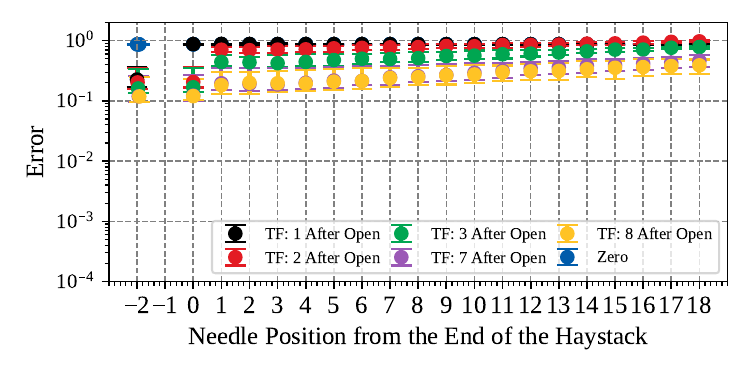}
        \caption{$4.10\times 10^{6}$ training examples.}
        \label{fig:ortho_needle_pos_8000}
    \end{subfigure}
    \hfill
    \begin{subfigure}[b]{0.49\linewidth}
        \centering
        \includegraphics[width=\linewidth, trim=0mm 5mm 0mm 0mm, clip]{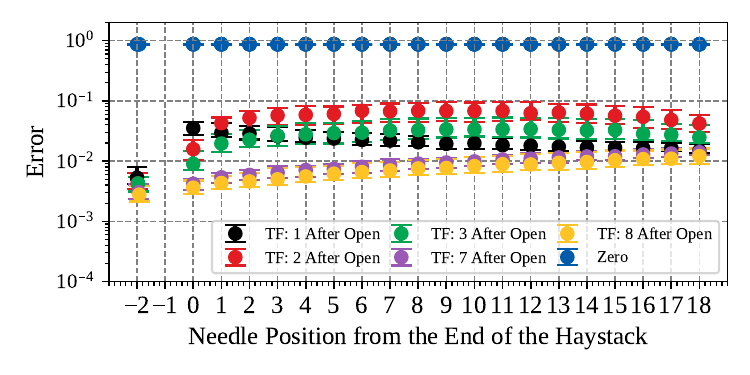}
        \caption{$6.25\times 10^{7}$ training examples.}
        \label{fig:ortho_needle_pos_122000}
    \end{subfigure}

    \caption{The effect of the needle position throughout training --- The median-squared error vs. the position of the needle in the haystack for different training checkpoints of the orthogonal model. For these plots, the x-axis value of $-2$ corresponds to a test segment that is the continuation of the segment before with no interruption by open and close symbols. The x-axis value of 0 is the closest needle position to
the test segment while the value 18 is the furthest away. ``Zero'' in the legend corresponds to a predictor that always predicts 0.}
    \label{fig:needle_pos_combined}
\end{figure}

% \begin{figure}[tbph]
%     \centering
%     \includegraphics[width=0.9\linewidth]{needle/context/ident/last_seg_context_ident_embd_dim_128_step_17600_haystack_len_19_20250308_153908.pdf}
%     \caption{Identity systems. Medium model. A comparison between predictions made in the final test segment (black) and predictions made from a continuing segment (red) as well as the initial segment (yellow) after being trained with $\approx 1.8\times 10^7$ training examples.}
%     \label{fig:ident_needle_context}
% \end{figure} 

Using a haystack of length $N=19$ for the results in this section, in Fig.~\ref{fig:ortho_needle_context}, the black dots show the performance of the orthogonal 128M parameter model on the final test segment of a needle-in-haystack sequence.  The yellow points represent how well it does on the first system it saw --- for which the first point is completely unpredictable and the rest do a bit better. To properly understand recall, we add one more possibility represented by the red dots. This represents the alternative of not terminating the final segment in the haystack at all and simply letting it continue on as the test sequence. We see that the black dots perform worse than the red points for merely continuing the last system in the haystack, but better than the yellow points showing what happens the first time we see this system. This clearly shows the ability to do recall, but there is worse prediction performance for the second observation in the test segment than the first (see Fig.~\ref{fig:squared-error_vs_haystack_len}). The conceptual difference represented by the red dots is that performing well here involves no recall at all, merely being able to continue predicting within one sequence.

Figure ~\ref{fig:ident_needle_context} shows what this looks like for the identity systems using the black dots. 
The recall of the constant is a bit spotty in the first position but we get more than 90\% of the way there with a squared-error of only 0.01 instead of the 1 we would have if we just guessed the all-zero vector. But after that initial observation in the test segment, the quality of understanding the constant-nature of these segments is quite good with all squared-losses around $10^{-4}$ or below.

We can also see how well the model can use recalled information as a function of the position we are trying to predict within the test segment. This parallels the natural language questions in \cite{liu-etal-2024-lost} that spawned widespread ``needle in haystack'' testing for LLMs generally \citep{kamradt2023needle}. Figs.~\ref{fig:ortho_needle_pos_error_ratios} and~\ref{fig:ident_needle_pos_error_ratios} show the effect of the needle position on the prediction performance of the orthogonal and identity models respectively. Here, we can see the first observation prediction quality in black showing higher quality when the needle is in the earlier positions within the context (larger position values in the figure). This is somewhat different from the ``U-shaped'' curve found for language-based LLMs although both have better performance when asking about the first thing seen. Meanwhile, the observation prediction quality for the second position is actually slightly better if the needle is closest to the test segment. 

In Fig.~\ref{fig:needle_pos_combined}, we look at how the effect of the needle position develops throughout the training of the orthogonal model. Again, the median squared error of its predictions on different indices in the test segment are shown for 19 needle positions for 10 different training checkpoints. For indices 7 and 8 in the test segment, the trend of better predictions for closer positions to the test segment is the first trend to develop in training and is sustained throughout. This same trend is also seen in indices two and three after $4.10 \times 10^6$ training examples have been processed, but by $3.17 \times 10^7$ training examples the model's predictions for the needle positions furthest away from the test segment start to improve for indices one, two, and three.

\FloatBarrier

\end{document}